\documentclass[runningheads]{llncs}

\usepackage{eccv}

\usepackage{eccvabbrv}

\usepackage{graphicx}
\usepackage{booktabs}

\usepackage[accsupp]{axessibility}  

\usepackage{hyperref}

\usepackage{orcidlink}

\usepackage[table]{xcolor}
\usepackage{booktabs}
\usepackage{multirow}

\usepackage[accsupp]{axessibility}
\usepackage{pgf}
\usepackage{multirow}
\usepackage{pifont}
\usepackage{pgfplots}
\pgfplotsset{compat=1.17}
\usepackage{pgfplotstable}
\usepackage{tikz}
\usepackage{csvsimple}
\usepackage{graphicx}
\usepackage{xifthen}
\usepackage{arydshln}
\usepackage{adjustbox}
\usepackage{balance}
\usepackage{tcolorbox}
\usepackage{etoolbox}
\usetikzlibrary{shapes,calc,positioning, decorations.text, arrows.meta, calc, shadows.blur, shadings, fit}
\usepackage{graphicx}
\usepackage{xspace}
\definecolor{cvprblue}{rgb}{0.21,0.49,0.74}

\title{MV$^2$: Multi-View Multi-Vehicle Driving Dataset for Novel View Synthesis}
\titlerunning{MV$^2$: Multi-Vehicle Driving Dataset}

\author{Sanjay Bhargav Dharavath\inst{1}\orcidlink{0009-0009-0994-7050} \and
Hanvitha Saraswathi Mukkamala\inst{1}\orcidlink{0009-0001-2541-300X
} \and
Faizan Farooq Khan\inst{3}\orcidlink{0000-0002-3645-603X} \and Ioannis Kakogeorgiou\inst{4}\orcidlink{0000-0001-5200-2620} \and Aditya Arun\inst{5}\orcidlink{0000-0002-1187-9108
} \and C.V. Jawahar\inst{1}\orcidlink{0000-0001-6767-7057} \and Zakaria Laskar\inst{2}\orcidlink{0000-0000-0000-0000}\thanks{Project Guide}}

\authorrunning{S.~B.~Dharavath et al.}

\institute{International Institute of Information Technology, Hyderabad \and Indian Institute of Science Education and Research, Thiruvananthapuram \and
King Abdullah University of Science and Technology (KAUST) \and IIT, National Centre for Scientific Research “Demokritos” \and Adobe MDSR, India}

\begin{document}
\maketitle
\begin{center}
    \vspace{-1.5em}
    \centering
    \captionsetup{type=figure}
    \setlength{\fboxrule}{1.5pt}
    \fcolorbox{black}{white}{\includegraphics[width=0.92\textwidth]{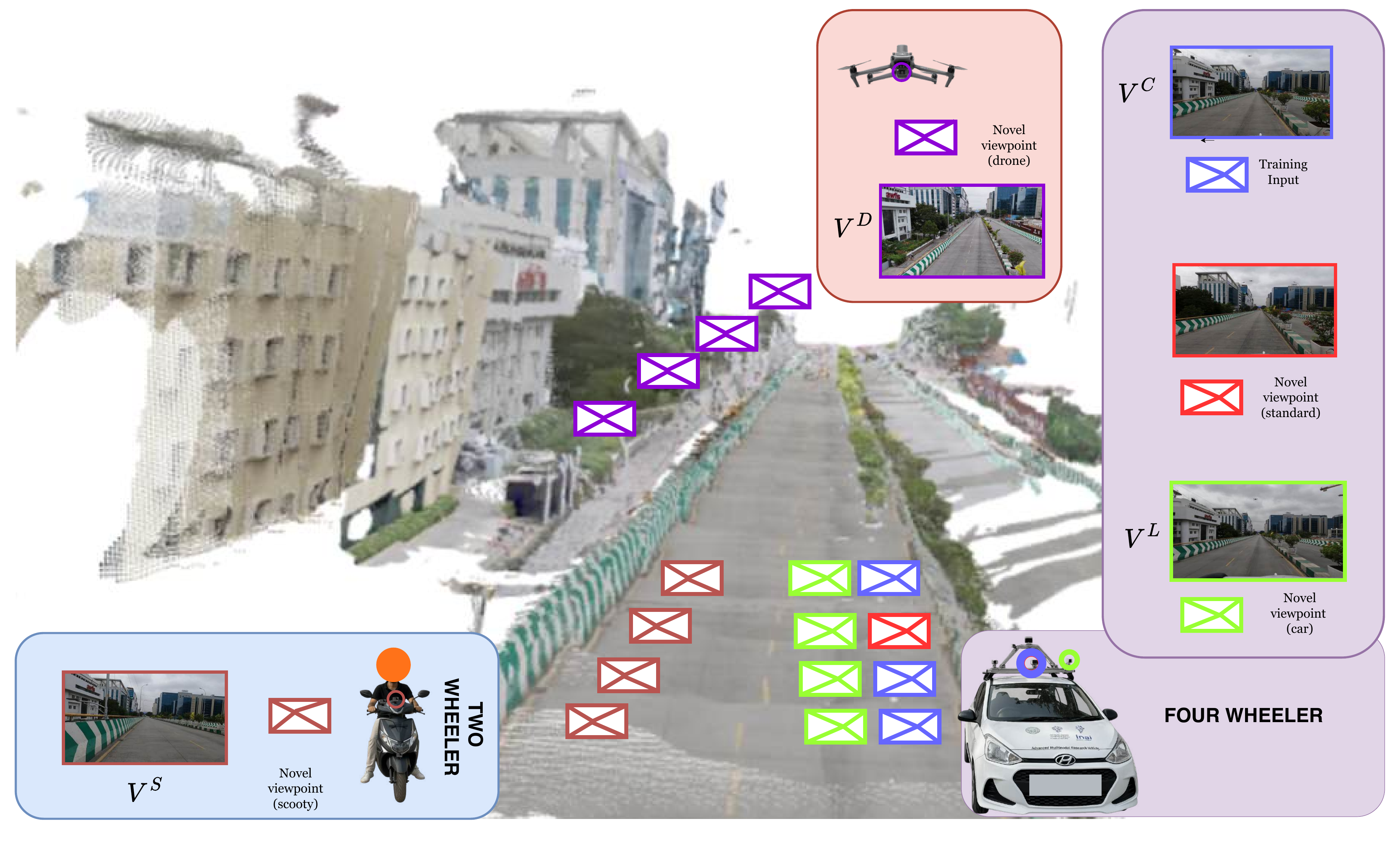} }
    \caption{\textbf{Example images and camera trajectories from the proposed MV$^2$ dataset and benchmark}. The dataset consists of camera trajectories from sensors placed on different vehicles - a four-wheeler, a two-wheeler and drone, capturing the same outdoor scene from diverse viewpoints. The diverse novel viewpoints (\textcolor{red}{red}, \textcolor{green}{green}, \textcolor{brown}{brown}, \textcolor{violet}{violet} cameras) sampled from different camera trajectories makes MV$^2$ a challenging Novel View Synthesis (NVS) benchmark, in contrast to existing works where novel views (\textcolor{red}{red} camera) are sampled from train trajectory (\textcolor{blue}{blue} cameras), and only consider ground-ground or aerial-aerial NVS evaluation.}   
    \label{fig:car-scooty}
\end{center}%

\newcommand{\red}[1]{{\color{red}#1}}
\newcommand{\cyan}[1]{{\color{cyan}#1}}
\newcommand{\todo}[1]{{\color{red}#1}}
\newcommand{\TODO}[1]{\textbf{\color{red}[TODO: #1]}}
\newcommand{\zl}[1]{\emph{\color{Blue}#1}}
\newcommand{\cvj}[1]{\emph{\color{Red}[CVJ: #1]}}
\newcommand{\sm}[1]{\tiny{#1}}
\newcommand{\tb}[1]{\textbf{#1}}




\definecolor{red4}{RGB}{255,199,199}
\definecolor{red3}{RGB}{255,214,214}
\definecolor{red2}{RGB}{255,230,230}
\definecolor{red1}{RGB}{255,242,242}
\definecolor{green1}{RGB}{240,255,240}
\definecolor{green2}{RGB}{225,255,225}
\definecolor{green3}{RGB}{200,255,200}
\definecolor{green4}{RGB}{170,255,170}
\definecolor{cvprblue}{rgb}{0.21,0.49,0.74}

\newcommand{\MV}{MV$^2$\xspace}%
\newcommand{\vggt}{VggT\xspace}%
\newcommand{\cartrain}{\textbf{Eval-Car-Train}\xspace}%
\newcommand{\dronetrain}{\textbf{Eval-Drone-Train}\xspace}%
\newcommand{\E}[2]{\textbf{\textit{test-E^[#1]_[#2]}}\xspace}%
\newcommand{\M}[2]{\textbf{\textit{test-M^[#1]_[#2]}}\xspace}%
\renewcommand{\H}[2]{\textbf{\textit{test-H^[#1]_[#2]}}\xspace}%

\makeatletter
\DeclareRobustCommand\onedot{\futurelet\@let@token\@onedot}
\def\@onedot{\ifx\@let@token.\else.\null\fi\xspace}
\def\eg{\emph{e.g}\onedot} \def\Eg{\emph{E.g}\onedot}
\def\ie{\emph{i.e}\onedot} \def\Ie{\emph{I.e}\onedot}
\def\vs{\emph{vs\onedot}}
\def\cf{\emph{cf}\onedot} \def\Cf{\emph{C.f}\onedot}
\def\etc{\emph{etc}\onedot} \def\vs{\emph{vs}\onedot}
\def\wrt{w.r.t\onedot} \def\dof{d.o.f\onedot}
\def\etal{\emph{et al}\onedot}
\makeatother

\newcommand{\tmark}{\ding{45}}%
\newcommand{\cmark}{\ding{51}}%
\newcommand{\xmark}{\ding{55}}%

\newcommand{\ph}{\phantom{0}}

\newcommand{\figref}[1]{Fig.~\ref{#1}}
\newcommand{\secref}[1]{Section~\ref{#1}}
\newcommand{\algref}[1]{Algorithm~\ref{#1}}
\newcommand{\eqnref}[1]{Eq.~\eqref{#1}}
\newcommand{\tabref}[1]{Table~\ref{#1}}

\newcommand{\boldparagraph}[1]{\noindent{\bf #1}}
\newcommand{\scatterwidthstats}{0.246\textwidth}
\newcommand{\scatterheightstats}{0.25\textwidth}

\pgfplotsset{compat = 1.3}

\newenvironment{customlegend}[1][]{%
    \begingroup
    \csname pgfplots@init@cleared@structures\endcsname
    \pgfplotsset{#1}%
}{%
    \csname pgfplots@createlegend\endcsname
    \endgroup
}%
\def\addlegendimage{\csname pgfplots@addlegendimage\endcsname}

\begin{abstract}
Differentiable rendering has advanced novel view synthesis (NVS), yet applying it to real-world driving remains difficult due to sparse capture viewpoints, dynamic objects, and limited multi-trajectory data. We introduce the \textbf{Multi-View Multi-Vehicle (\MV)} dataset and benchmark for evaluating NVS models under large viewpoint changes in dynamic urban scenes. \MV\ features synchronized captures from a car, scooter, and drone, each following distinct yet synchronized trajectories. Training NVS methods on one vehicle’s camera stream and testing on another enables evaluation under substantially larger viewpoint variations than existing single-trajectory datasets. All sequences are registered via Structure-from-Motion and camera poses verified using manual pixel-level correspondence annotations, yielding 50 high-quality scenes with 12000 images. Benchmarking recent NVS and camera pose estimation methods shows that NVS performance degrades with increasing viewpoint disparity, and that feed-forward pose estimators notably lag behind optimization-based approaches, highlighting \MV\ as a rigorous testbed for NVS in driving. The dataset, benchmark protocol, and project resources are available at
\url{https://mv2-dataset.github.io/}.


\end{abstract}

\section{Introduction}


Differentiable rendering has driven major progress in novel view synthesis (NVS) - the task of generating unseen views of a scene from a set of posed input images. NVS holds strong potential for autonomous driving, enabling photorealistic simulations for training and evaluating perception and planning systems across diverse viewpoints. However, applying NVS to driving scenarios is challenging due to sparse camera coverage, dynamic objects, and large untextured regions like the sky. These issues are amplified when multi-sensor setups (e.g., LiDAR or multi-camera rigs~\cite{geiger2012we, caesar2020nuscenes}) are unavailable. In this work, we evaluate state-of-the-art NVS methods~\cite{mildenhall2021nerf, kerbl20233d, depthsplat, mvsplat} in the demanding setting of monocular captures with large viewpoint variations in novel viewpoints.

Evaluating NVS methods in real-world driving scenarios is limited by the lack of purpose-built datasets. Most existing benchmarks \cite{yan2024street, tonderski2024neurad, fischer2024dynamic, yang2023emernerf} repurpose standard driving datasets \cite{geiger2012we, cabon2020virtual, caesar2020nuscenes, waymo2020}, which capture a single camera trajectory through traffic scenes. Training and test frames are sampled from the same path, typically by taking every $k$-th frame, so these setups only test interpolation, not true generalization to unseen viewpoints. Concurrent efforts remain constrained: some depend on synthetic data \cite{li2024xld}, others assume static scenes by masking dynamic objects \cite{ni2025lane, li2025mtgs}, and Ma et al.~\cite{ma2025novel} use FID as a proxy for cross-view evaluation, which measures perceptual realism rather than geometric consistency. Practically, recording identical dynamic scenes from multiple viewpoints with a single vehicle is infeasible due to traffic constraints.

To address this gap, we present the Multi-View Multi-Vehicle (\MV) dataset for benchmarking NVS methods in dynamic driving environments. The dataset is collected using a multi-platform configuration: a four-wheeler (car), a two-wheeler (scooty) and an aerial vehicle (drone), each equipped with one or multiple cameras - recording the same scenes from independent, real-world trajectories. The car and scooter follow separate lanes to provide lateral viewpoint variation, while the drone captures the same scene from above, offering occlusion-free coverage. Our setup enables a unique evaluation protocol: models trained on one vehicle’s camera stream are tested on their ability to synthesize views from another vehicle’s perspective. The substantial viewpoint differences between the vehicle trajectories allow us to directly evaluate extrapolation quality, a key capability missing from existing benchmarks that only measure interpolation~\cite{yan2024street, tonderski2024neurad, fischer2024dynamic, yang2023emernerf} .


\MV defines two evaluation setups based on the camera platform used for training NVS models. In the first setup \cartrain, training images come from one of the car's onboard camera, while the images from scooty, drone and the car's alternate cameras form the test set. In the second setup \dronetrain, we consider an aerial-to-ground NVS task, where training uses drone imagery, and testing is performed on images from car and scooty cameras. In both setups, we also consider an additional test set consisting of every $k^{th}$ image from the respective training sequence in line with existing works~\cite{ljungbergh2024neuroncap,wu2025towards,xie2025vid2sim}. This diversity in test images from multiple vehicles enables a comprehensive evaluation of NVS methods in synthesizing images from novel viewpoints - a critical requirement in driving simulation~\cite{ljungbergh2024neuroncap,wu2025towards,xie2025vid2sim}. 


For each scene, images from all camera sensors are registered to a unified coordinate frame using Structure-from-Motion (SfM). After rigorous annotation and filtering to remove pose errors, the final dataset comprises 50 scenes with 12,000 images at 1080 $\times$ 1920 resolution. We evaluate per-scene optimization–based and feed-forward NVS methods, including static~\cite{kerbl20233d, mildenhall2021nerf}, dynamic~\cite{desiregs, pvg}, and learning-based 3D Gaussian Splatting (3DGS) approaches~\cite{depthsplat, mvsplat, monosplat}. Key findings include:

\begin{itemize}
\item NVS performance consistently drops with an increase in camera baseline between train and test images, demonstrating the effectiveness of \MV’s viewpoint diversity.
\item Aerial-to-ground NVS is a challenging setting, where both aerial-specific~\cite{dronesplat} and general NVS models struggle.
\item Optimization-based methods consistently outperform feed-forward 3DGS methods for novel view synthesis (NVS) under both small and large viewpoint changes between training and test views.

\item Optimization-based methods benefit from depth priors obtained using off-the-shelf depth estimation models~\cite{lin2025depth3recoveringvisual} with appropriate filtering, providing an effective alternative to LiDAR. Additionally, accurate dynamic object segmentation masks further improve performance
\item Recent feed-forward camera pose estimators~\cite{wang2025vggt,mapanything,reloc3r,mast3r} lag behind optimization-based methods~\cite{colmap}, highlighting the value of \MV as a camera pose estimation benchmark.
\end{itemize}

\section{Related Work}

In this section, we review existing NVS datasets and methods in the context of driving scenes.

\setlength{\fboxrule}{1.5pt}
\begin{figure*}[ht!]
\centering
\begin{tabular}{ccccc}
  \textbf{Scene 1} & \textbf{Scene 2} & \textbf{Scene 3} & \textbf{Scene 4} \\

\fcolorbox{blue}{white}{\includegraphics[width=0.20\textwidth]{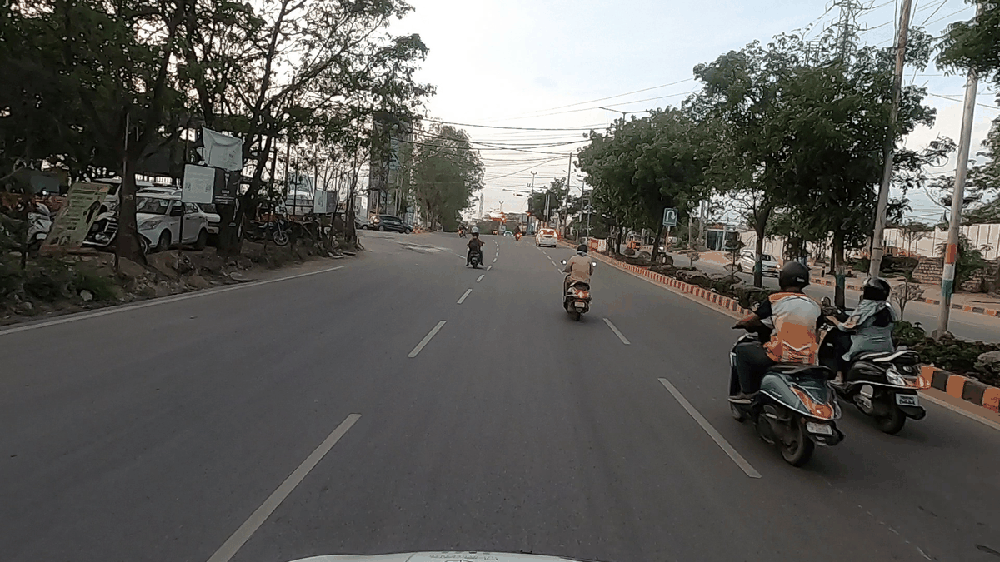}} &
\fcolorbox{blue}{white}{\includegraphics[width=0.20\textwidth]{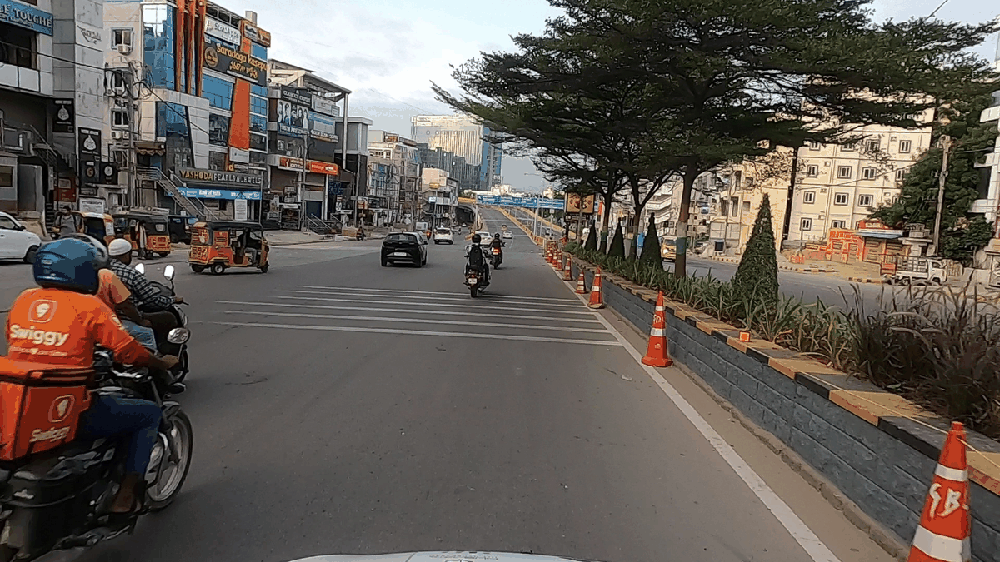}} &
\fcolorbox{blue}{white}{\includegraphics[width=0.20\textwidth]{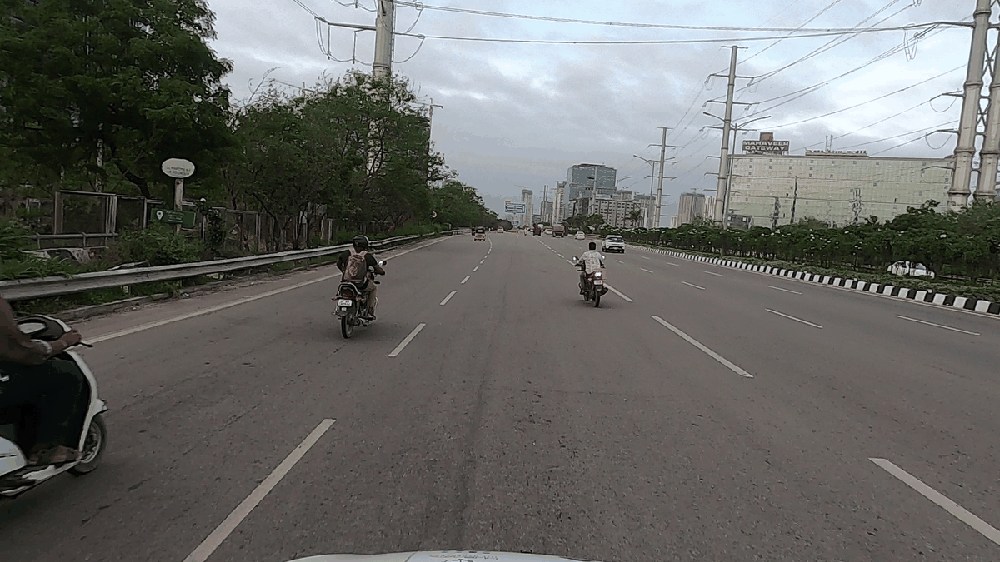}} &
\fcolorbox{blue}{white}{\includegraphics[width=0.20\textwidth]{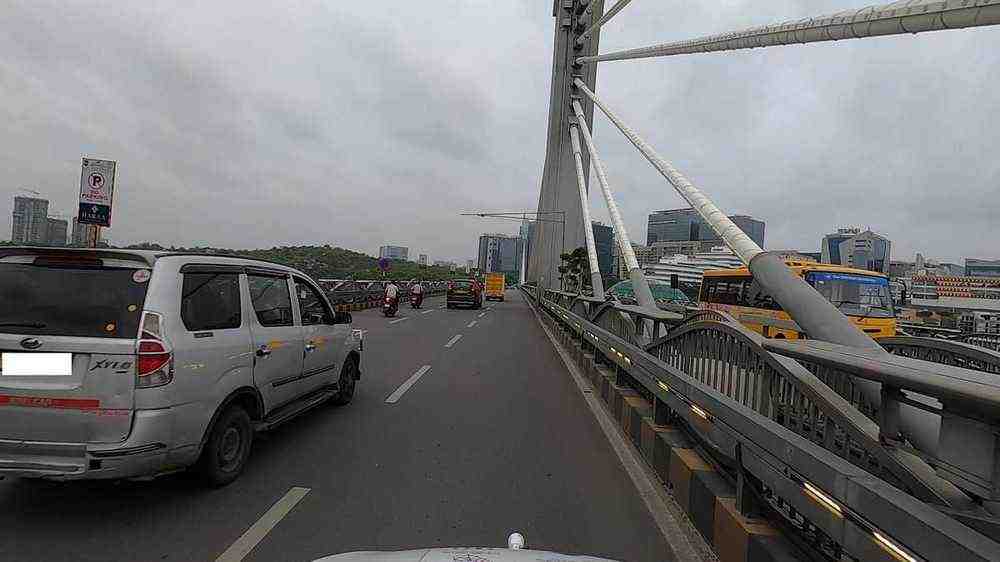}} \\[3pt]

\fcolorbox{violet}{white}{\includegraphics[width=0.20\textwidth]{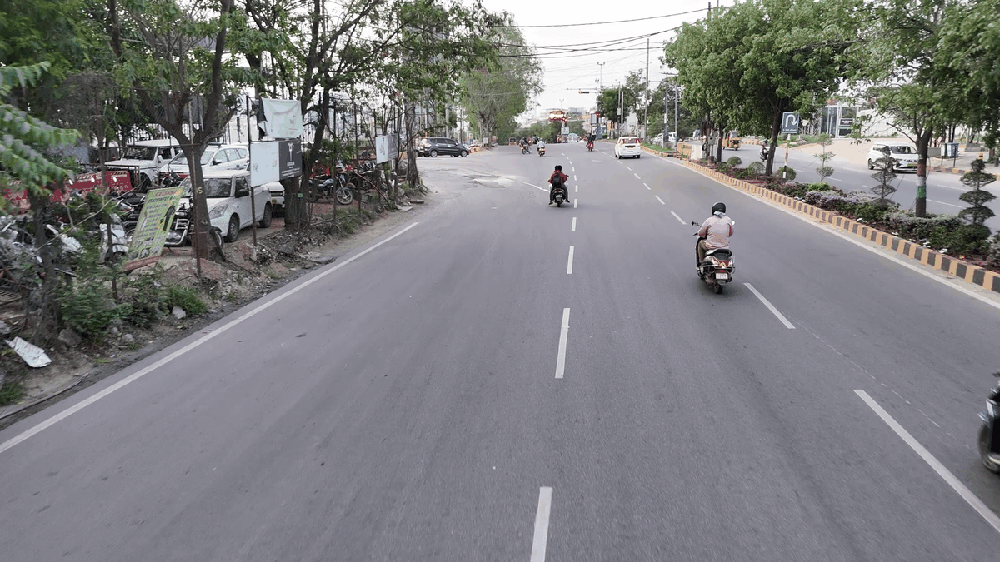}} &
\fcolorbox{violet}{white}{\includegraphics[width=0.20\textwidth]{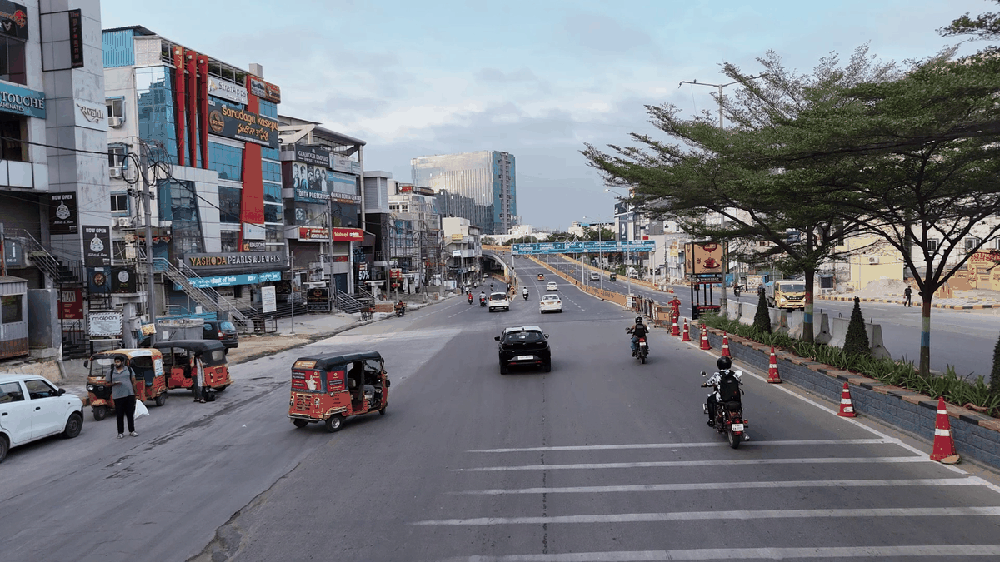}} &
\fcolorbox{violet}{white}{\includegraphics[width=0.20\textwidth]{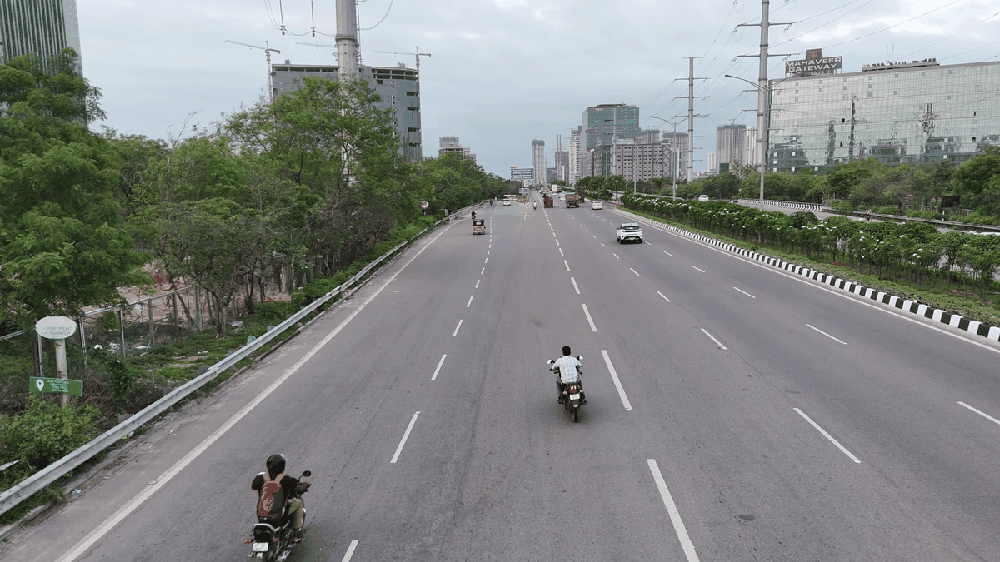}} &
\fcolorbox{violet}{white}{\includegraphics[width=0.20\textwidth]{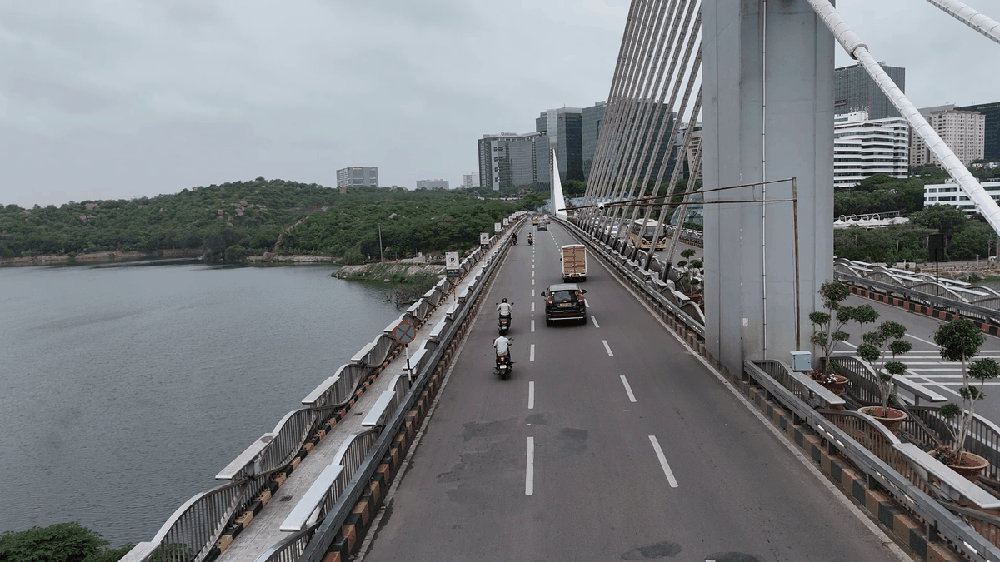}} \\[3pt]

\fcolorbox{brown}{white}{\includegraphics[width=0.20\textwidth]{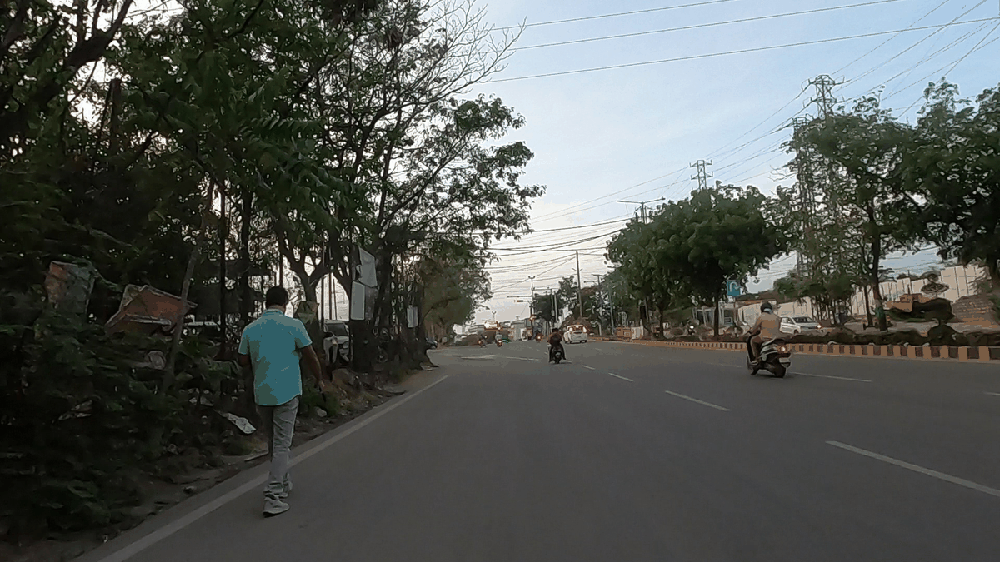}} &
\fcolorbox{brown}{white}{\includegraphics[width=0.20\textwidth]{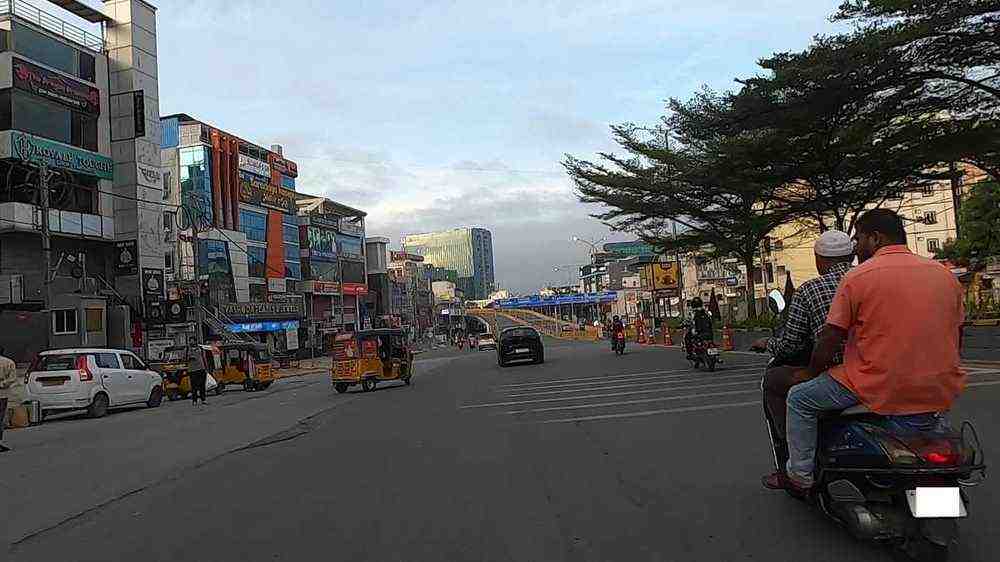}} &
\fcolorbox{brown}{white}{\includegraphics[width=0.20\textwidth]{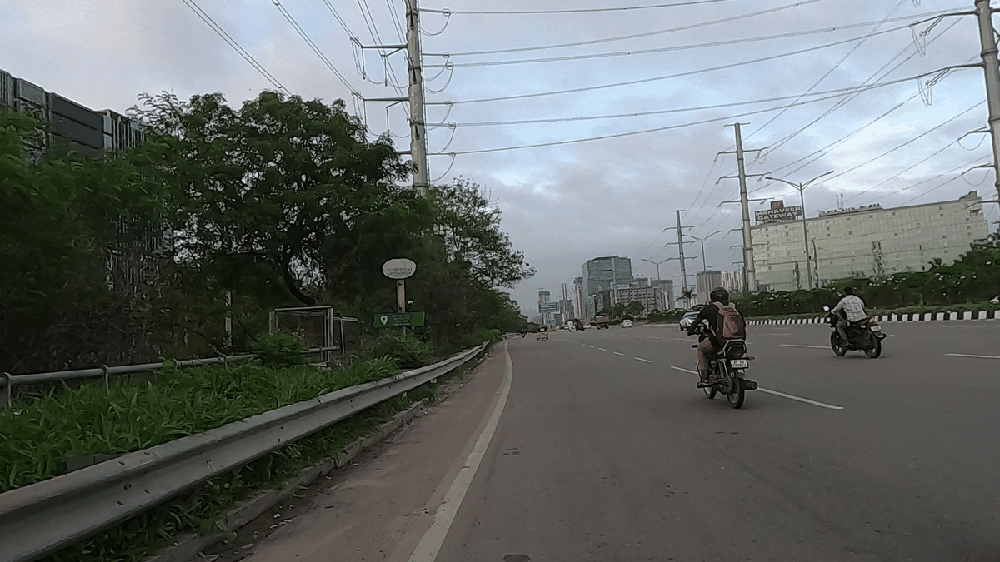}} &
\fcolorbox{brown}{white}{\includegraphics[width=0.20\textwidth]{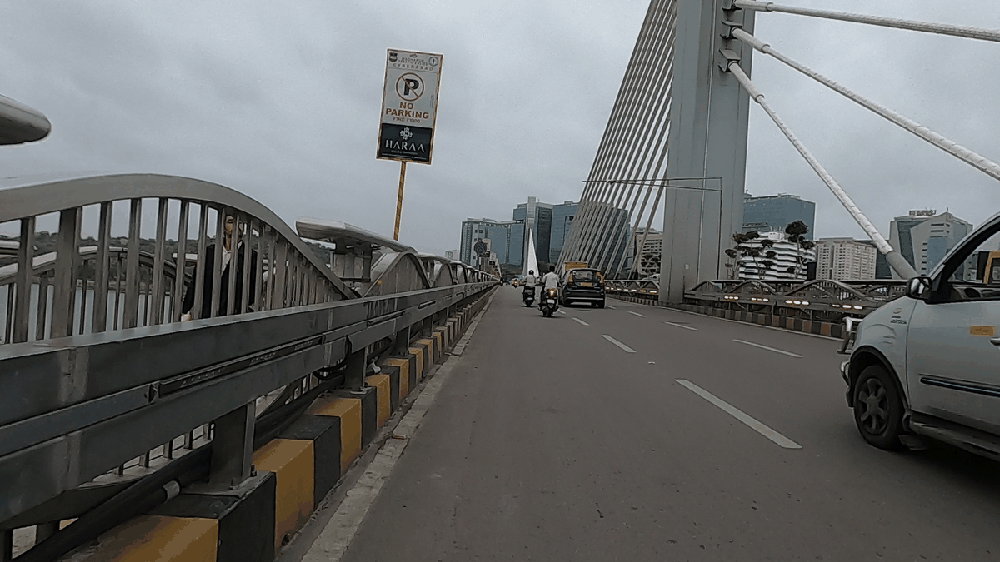}} 
\end{tabular}

\caption{Sample multi-view scenes captured across diverse environments. Each column represents a distinct scene, while each row corresponds to a different acquisition viewpoint \textcolor{blue}{$V^C$}, \textcolor{violet}{$V^D$}, and \textcolor{brown}{$V^S$}. The colored borders highlight the viewpoint source of each image, helping visualize cross-view consistency across sensors.}
\label{fig:sample_images}
\vspace{-15pt}
\end{figure*}

\noindent\textbf{Driving Datasets for NVS.}
 Datasets for novel view synthesis primarily consist of densely collected multi-view images of small-scale scenes~\cite{barron2022mip, mildenhall2021nerf}. These datasets consist of small-scale in-bounded scenes captured with a circular camera trajectory around a given object or part of the scene. In contrast, outdoor scenes face challenges such as unbounded large-scale scenes, the presence of distant objects such as sky, and dynamic objects such as vehicles and pedestrians. Several driving datasets, \eg KITTI~\cite{geiger2012we},
KITTI-360, vKITTI~\cite{cabon2020virtual}, Waymo Open Dataset~\cite{waymo2020}, nuScenes~\cite{caesar2020nuscenes} and Argoverse 2~\cite{fischer2024multi, wilson2023argoverse} have been repurposed for evaluating novel view synthesis of driving scenes. These datasets consist of multi-view images of the scene captured from a forward moving camera. Additional sensors such as GPS, IMU and importantly LiDAR are also used to estimate camera trajectory (pose) and an initial 3D model of the scene. Unlike existing NVS datasets with dynamic objects~\cite{ni2025paralanemultilanedatasetregistering} captured from diverse viewpoints, existing driving datasets are captured from a single vehicle trajectory. This limitation is also addressed in recent concurrent works~\cite{li2024xld,li2025mtgs}. While XLD~\cite{li2024xld} uses asynthetic dataset based on CARLA, MTGS~\cite{li2025mtgs} and Para-lane~\cite{ni2025lane} collect multiple traversals through the same scene, failing to \textit{capture dynamic objects from multiple viewpoints}. In contrast, the proposed \MV dataset consists of real-world images capturing the \textit{dynamic} driving scene from the viewpoints of on-road and aerial vehicles. Statistics and example images of dynamic objects are presented in \secref{suppsec:dynamic_objects} in the Supplementary and \figref{fig:sample_images}.

Aerial-to-ground datasets of outdoor driving scenes are limited. Existing aerial datasets do not provide pose information. Recently, the proposed co-registered aerial-to-ground image dataset~\cite{vuong2025aerialmegadepth} is restricted to scenes from MegaDepth~\cite{vuong2025aerialmegadepth} and mainly functions as a training set. Similarly, the dataset introduced in DroneSplat~\cite{dronesplat} only considers the aerial-to-aerial NVS problem. In contrast to existing aerial-to-ground datasets for NVS, proposed \MV provides complementarity by covering a large number of outdoor driving scenes, and provides co-registered aerial-to-ground camera poses. 

\noindent\textbf{NVS Methods for Autonomous Driving.}
Recent advances in neural rendering have been driven by NeRF-based volumetric methods and point-based differentiable rendering frameworks~\cite{mildenhall2021nerf, barron2022mip, xu2022point}. Building on these ideas, 3D Gaussian Splatting (3DGS)~\cite{kerbl20233d} introduced an explicit, continuous scene representation that enables real-time radiance field rendering through tile-based rasterization.
Following its success, numerous approaches have adapted 3DGS for large-scale driving scenes and dynamic urban environments~\cite{pvg, yan2024street, zhou2024drivinggaussian, desiregs, s3gaussian}. For example, PVG~\cite{pvg} models temporal dynamics via periodic vibration fields; Street Gaussians~\cite{yan2024street} jointly represent static and dynamic components using semantic logits; and DrivingGaussian~\cite{zhou2024drivinggaussian} introduces composite dynamic graphs for multi-object motion modeling.
More recent self-supervised extensions, such as PVG~\cite{pvg}, DeSiRe-GS~\cite{desiregs} and S$^3$Gaussians~\cite{s3gaussian}, further improve static–dynamic decomposition and surface fidelity without dynamic object mask supervision.

Beyond scene-specific optimization, feed-forward generalizable 3DGS networks aim to predict Gaussian parameters directly from multi-view inputs.
Methods such as PixelSplat~\cite{pixelsplat}, MVSplat~\cite{mvsplat}, MonoSplat~\cite{monosplat}, and DepthSplat~\cite{depthsplat} enable efficient novel-view synthesis by learning scene-agnostic priors, significantly reducing optimization time compared to per-scene training.
Recent approaches, such as STORM~\cite{yang2025storm} extend feed-forward NVS methods to driving scenarios but require LiDAR input.

Benchmarking all driving-specific NVS methods is challenging due to the high computational requirements - for example, DesireGS~\cite{desiregs}, PVG~\cite{pvg}, StreetGaussians~\cite{yan2024street} and similar optimization-based methods require up to 4 hours training time on a single scene. As such, we benchmark only a subset of these representative approaches~\cite{mildenhall2021nerf, kerbl20233d, desiregs, pvg, depthsplat, monosplat, mvsplat} based on their prominence and recency under a unified evaluation protocol for real-world, wide-baseline ground-to-ground and aerial-to-ground settings.

\section{Multi-View Multi-Vehicle Dataset}

In this section, we describe the collection and composition process of our proposed \MV dataset.
We adopt a multi-vehicle setup (\secref{subsec:sensor}) to collect a set of driving videos following specific protocols (\secref{subsec:collection}). Thereafter, the collected set of videos are divided into train and test sets(\secref{subsec:split}). Each image in \MV is registered to a single global coordinate system using a state-of-the-art camera pose estimation pipeline (\secref{subsec:pose_est}). The estimated camera poses are then verified using epipolar geometry (\secref{subsec:pose_verify}).

\MV follows the standard novel-view synthesis setup, where train images and train camera poses are used to obtain a 3D scene representation~\cite{mildenhall2021nerf,kerbl20233d,depthsplat} of the scene. The scene representation is then rendered from the viewpoint of the test images using the test camera poses. Finally, these rendered images are compared with ground-truth test images to obtain NVS performance metrics.

\subsection{Sensor Setup}
\label{subsec:sensor}
All the vehicles are equipped with GoPro 10 camera sensors. Two cameras, \{L(-eft), C(-entral)\} are fixed to the camera rig mounted on the car. We denote the set of images collected using these three cameras as $V{^L}, V{^C}$. Similarly, we denote the images collected using sensors, S and D placed on the scooty and the drone as $V{^S}$ and drone as $V{^D}$ respectively. All the sensors on the vehicles are forward-facing. Each camera captures images at a resolution of 1080 $\times$ 1980 at 60 FPS. The images collected from the above sensors are synchronized using the wall-clock.

\subsection{Dataset Collection}
\label{subsec:collection}
The proposed dataset is collected by driving the two vehicles through various urban and semi-urban roads for two hours over a period of 5 days. Lateral and aerial adjacency between the vehicles was maintained during data collection. The video data collected from each sensor is sampled at 2 FPS and divided into smaller video segments consisting of 100 frames. Video segments of roads in high traffic congestion, under-tunnel, misaligned lateral adjacency between vehicles and red-light traffic points are excluded. This results in a total of 200 scenes, with 5 camera sequences each.

\subsection{Evaluation Setup}
\label{subsec:split}

We consider two evaluation setups: \cartrain and \dronetrain, based on which vehicle camera stream is used for training the NVS models. For each setup, different test sets are formed by sampling images from the train sequence and image sequences from other vehicles. The test sets are referred to a $T_{X \rightarrow Y}$, where $X$ and $Y$ are the train and test cameras. For the \cartrain setup, we consider the following train and test splits.

\noindent \textit{\textbf{Train}}: Images collected from the center camera sensor, $V{^C}$.

\noindent \textit{\textbf{Test T}}$_{C \rightarrow C}$: Existing works~\cite{fischer2024dynamic,fischer2024multi,turki2023suds,yan2024street,yang2023emernerf} generate training and test sets by subsampling the train trajectory $C$, typically selecting every $k$-th frame for testing (e.g., $k{=}10$ for Argoverse~\cite{fischer2024multi} and Waymo (NOTR)~\cite{yang2023emernerf}, $k{=}4$ for KITTI and Virtual KITTI2~\cite{turki2023suds}). We adopt this protocol with $k{=}5$, forming the test set, \textit{\textbf{T}}$_{C \rightarrow C}$, with the highest viewpoint overlap with the training images. The train set introduced above excludes these frames.

\noindent \textit{\textbf{Test T}}$_{C \rightarrow L}$: Images from the left car-mounted camera sequence, $V{^L}$, sampled at indices $\{0,5,10,\ldots,95\}$, forms the test set, \textit{\textbf{Test T}}$_{C \rightarrow L}$. This set introduces a small lateral baseline relative to the train sequence, resulting in moderate viewpoint variation compared to the test set, \textit{\textbf{T}}$_{C \rightarrow C}$.

\noindent \textit{\textbf{Test T}}$_{C \rightarrow S}$: To introduce a more challenging setup, \MV adds images from scooty trajectory $V^S$ as the cross-vehicle test set, \textit{\textbf{Test T}}$_{C \rightarrow S}$. The camera sequence, $V_S$ is uniformly sampled at $\{0,5,10,\ldots,95\}$. Unlike the previous test sets, this cross-vehicle test set has a significant camera baseline with the training views, providing a rigorous assessment of cross-trajectory generalization.

 For the \dronetrain setup, the training set consists of images captured from the drone camera $V^D$. The test sets are composed using images from intra-vehicle and cross-vehicle sensors: \textit{\textbf{T}}$_{D \rightarrow D}$, \textit{\textbf{T}}$_{D \rightarrow C}$ and \textit{\textbf{T}}$_{D \rightarrow S}$, obtained by sampling $V{^D}$, $V{^C}$ and $V^S$, respectively, at the 5-frame interval ($0, 5, 10, \ldots, 95$). The following pairs of test sets, \{(\textit{\textbf{T}}$_{C \rightarrow C}$,\textit{\textbf{T}}$_{D \rightarrow C}$), (\textit{\textbf{T}}$_{C \rightarrow S}$,\textit{\textbf{T}}$_{D \rightarrow S}$), (\textit{\textbf{T}}$_{C \rightarrow D}$,\textit{\textbf{T}}$_{D \rightarrow C}$)\} seem equivalent but are not due to the pose verification step in \secref{subsec:pose_verify}.

\begin{figure*}[t]
\centering

\begin{subfigure}{0.49\textwidth}
    \centering
    \includegraphics[width=\linewidth]{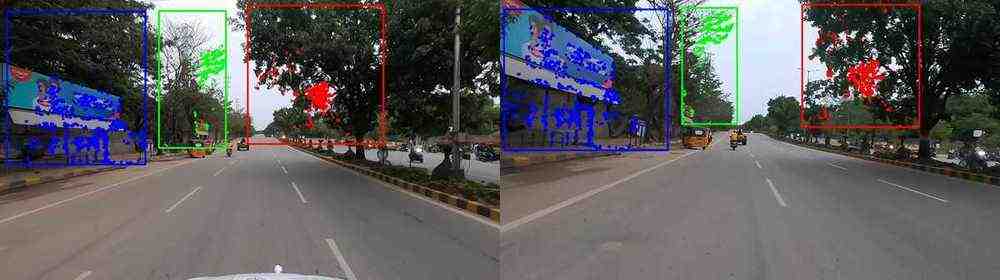}
    \caption{$V_C - V_S$ RoMA correspondences}
\end{subfigure}
\hfill
\begin{subfigure}{0.49\textwidth}
    \centering
    \includegraphics[width=\linewidth]{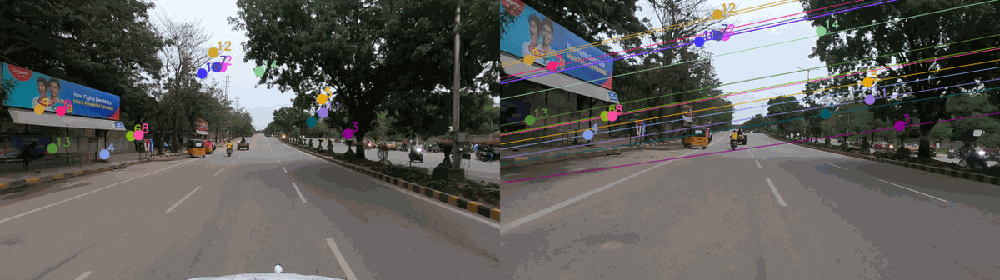}
    \caption{$V_C - V_S$ epipolar lines}
\end{subfigure}

\vspace{4pt}

\begin{subfigure}{0.49\textwidth}
    \centering
    \includegraphics[width=\linewidth]{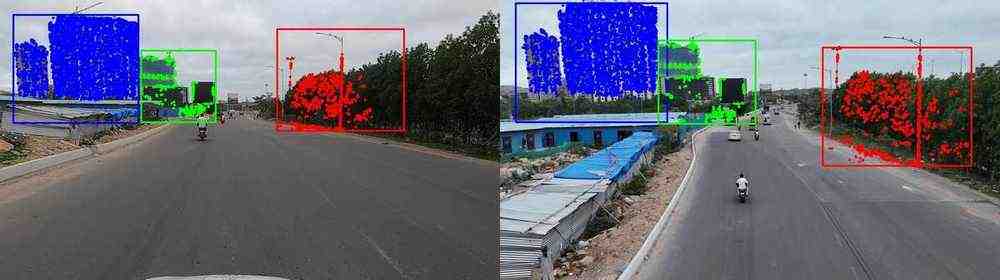}
    \caption{$V_D - V_C$ RoMA correspondences}
\end{subfigure}
\hfill
\begin{subfigure}{0.49\textwidth}
    \centering
    \includegraphics[width=\linewidth]{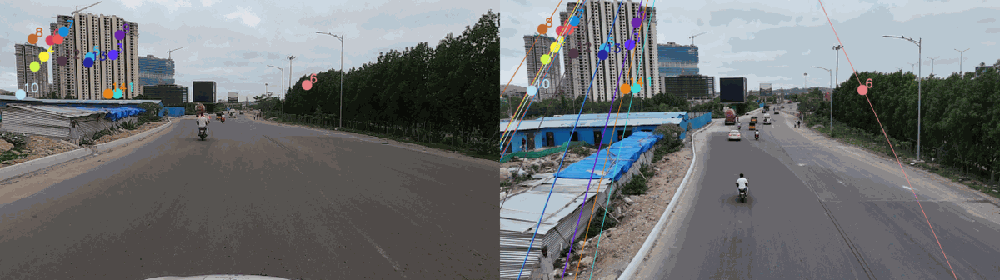}
    \caption{$V_D - V_C$ epipolar lines}
\end{subfigure}

\caption{\textbf{Annotation and Pose Evaluation.} Representative image pairs from $V_C\text{-}V_S$ (top row) $V_C\text{-}V_D$ (bottom row) with pixel correspondences obtained from annotated region correspondence and RoMA matches and epipolar lines (of randomly sampled RoMA correspondences for visualization purpose). The alignment of correspondences along epipolar lines demonstrates strong geometric consistency and reliable cross-view registration even under wide baselines.}
\label{fig:epi_anno}
\vspace{-10pt}

\end{figure*}


\begin{figure}[t]
    \centering
    \begin{tabular}{cc}
    \includegraphics[width=.49\textwidth] {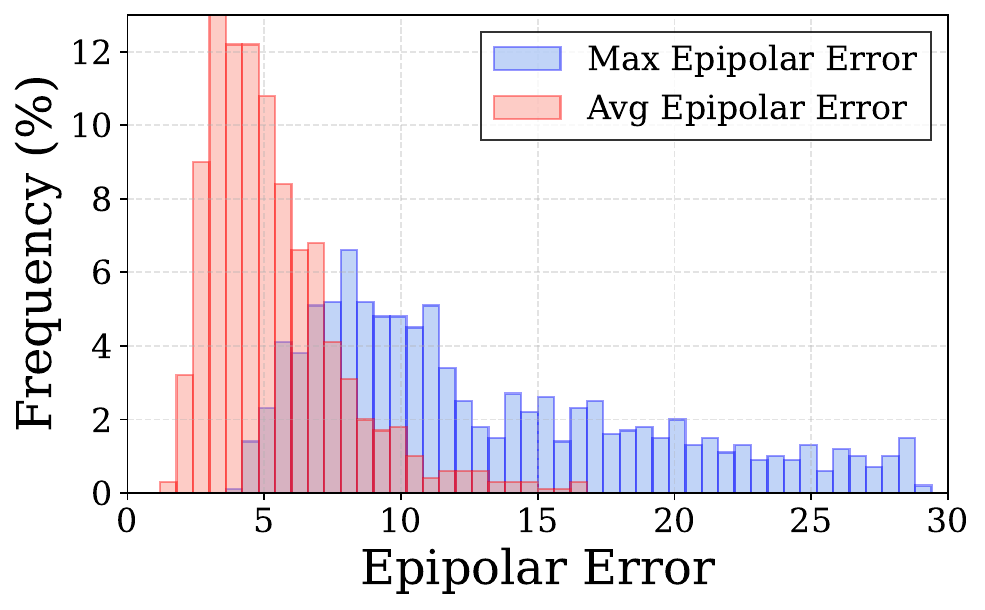} &
    \includegraphics[width=.49\textwidth] {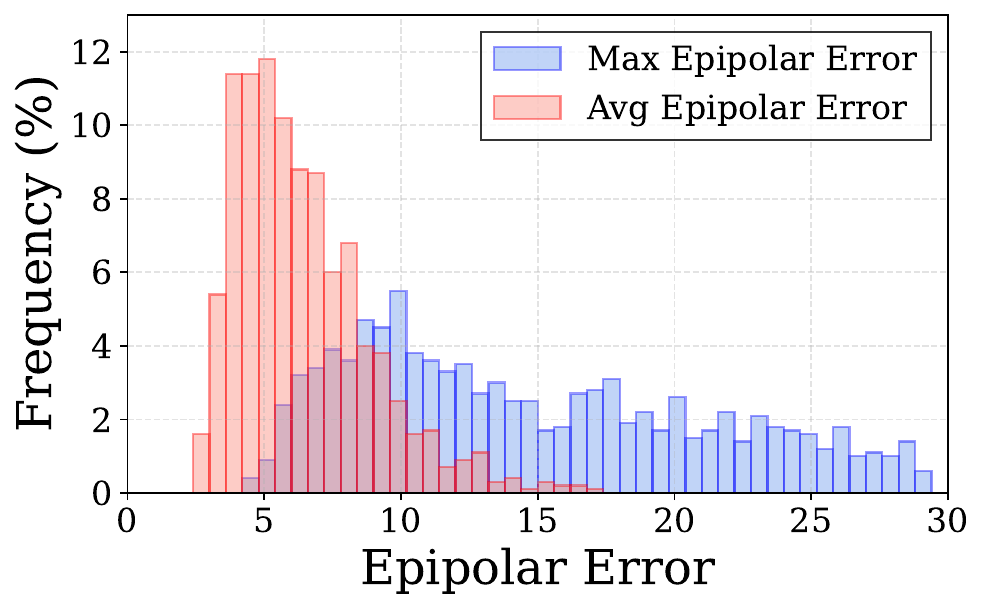} \\
    $T_{C \rightarrow S}$ &
    $T_{D \rightarrow C}$ \\
    \end{tabular}
    \vspace{-10pt}  
    \caption{\textbf{Epipolar error distribution} of mean and maximum epipolar errors (\eqref{eq:epi_error_ma}) over 50 scenes. Left: $T_{C \rightarrow S}$, Right: $T_{D \rightarrow C}$.}
    \label{fig:epierror_annot}
    \vspace{-15pt}  
\end{figure}


\subsection{Camera Parameters Estimation}
\label{subsec:pose_est}

We describe the process used to estimate both the intrinsic and camera poses (extrinsic) parameters of the images in the train and three test sets.

\noindent\textbf{Intrinsic Calibration.} We use a standard checkerboard calibration target to estimate the intrinsic parameters of the cameras, including focal lengths ($f_x$, $f_y$) and principal points ($c_x$, $c_y$). This calibration is performed prior to any pose estimation. The intrinsics are further refined in the following camera pose estimation steps.

\noindent \textbf{Camera Pose Estimation.} We estimate camera poses using a two-step pipeline. First, training images for each setup (\cartrain and \dronetrain) are registered with COLMAP using its standard configuration. Next, test images are localized with COLMAP’s localization module, using the 15 nearest training images as neighbors. While single-sequence (per-sensor) camera pose estimation produces accurate poses, jointly estimating train and test poses leads to degraded results many images from \textit{both the train and test set} fail to register or yield incorrect poses highlighting the large viewpoint diversity in \MV. The two-step strategy thus ensures (i) accurate, independently estimated training poses and (ii) the ability to filter out test images with failed or inconsistent localization without affecting the train poses. Details of this filtering process are provided next.

\subsection{Pose Verification}
\label{subsec:pose_verify}
Camera poses estimated by COLMAP are subject to errors arising from keypoint mismatches and inaccuracies in Structure-from-Motion (SfM) optimization. Since ground-truth poses are unavailable, absolute pose accuracy cannot be directly verified. Instead, we assess the accuracy of each train and test image’s absolute pose by evaluating its \textit{relative pose} with respect to its nearest training image. However, relative pose evaluation alone is insufficient to guarantee the correctness of an image’s absolute pose if the nearest training image has an inaccurate absolute pose, its relative pose with another image can still appear correct. To address this ambiguity, we first verify the accuracy of training image poses, as described below.

\noindent \textbf{Train Pose Verification}. We model the training image sets, $V_C$ and $V_D$, as sequential pose graphs, where each node corresponds to a training image camera pose and each edge represents the relative pose between consecutive images. Our hypothesis is that if every edge (relative pose) is accurate, then the corresponding nodes (absolute poses) can also be considered accurate. We evaluate each relative pose using manually annotated pixel correspondences as described next. To address the issue of camera pose drift, we also evaluate the relative poses between the first and last or the 50th image based on covisibility.

Given a correspondence $(\mathbf{x}_1, \mathbf{x}_2)$ between two cameras with intrinsics $\mathbf{K}_1, \mathbf{K}_2$ and extrinsics $(\mathbf{R}_1, \mathbf{t}_1)$ and $(\mathbf{R}_2, \mathbf{t}_2)$, the \textit{epipolar error} is defined as
\begin{equation}
e = \left| \mathbf{x}_2^\top \mathbf{F} \mathbf{x}_1 \right|,
\end{equation}, where
$\mathbf{F} = \mathbf{K}_2^{-\top} [\mathbf{t}]_\times \mathbf{R} \mathbf{K}_1^{-1},
\mathbf{R} = \mathbf{R}_2 \mathbf{R}_1^\top,
\mathbf{t} = \mathbf{t}_2 - \mathbf{R} \mathbf{t}_1$,
\label{eq:rel_pose}
\noindent $[t]_x$ is skew-symmetric matrix~\cite{hartley2003multiple}.

 For each candidate relative pose between an image pair, we compute the mean and maximum epipolar errors using $N$ manually annotated correspondences as
\begin{equation}
e_a = \tfrac{1}{N} \sum_i e_i, \quad e_m = \max_{i \in \{1,\dots,N\}} e_i.
\label{eq:epi_error_ma}
\end{equation}

A relative pose is deemed inaccurate if $e_m > 30$. For each of the 200 sequences, we verify all consecutive training image pairs in $V_C$ and $V_D$; a sequence is retained only if all relative poses satisfy $e_m \le 30$ in both training sets.

\noindent \textbf{Test Pose Verification}. For sequences with validated training camera poses, we then evaluate the relative pose between each test image and its nearest training image using the same epipolar error criterion, based on annotated correspondences. The nearest training image is selected from either $V_C$ or $V_D$, depending on the evaluation setup (\cartrain or \dronetrain). A test image is retained if its relative pose satisfies $e_m \le 30$.

\noindent \textbf{Correspondence Annotation} is performed in two stages. First, human annotators identify corresponding bounding boxes that observe the same 3D region in an image pair. These regions are cropped, resized, and processed using a dense neural matcher (e.g., RoMA~\cite{edstedt2025romav2harderbetter}) to obtain dense pixel correspondences within each bounding box. The resulting correspondence set is denoted as $\mathcal{X}_{\text{RoMA}}$ and shown in~\figref{fig:epi_anno} (a) and (c). All candidate image pairs in both the training and test sets are subsequently validated using epipolar consistency checks computed independently over correspondences in $\mathcal{X}_{\text{RoMA}}$. An example image showing epipolar lines for a subset of correspondences in $\mathcal{X}_{\text{RoMA}}$ is shown in~\figref{fig:epi_anno} (b) and (d).

\begin{table*}[ht]
\centering
\small
\begin{tabular}{l|ccc|ccc|ccc}
\toprule
\textbf{Model} 
& \multicolumn{3}{c|}{\textbf{PSNR} $\uparrow$}
& \multicolumn{3}{c|}{\textbf{SSIM} $\uparrow$}
& \multicolumn{3}{c}{\textbf{LPIPS} $\downarrow$} \\
\cmidrule(lr){2-4} \cmidrule(lr){5-7} \cmidrule(lr){8-10}
& $T_{C\rightarrow C}$ & $T_{C\rightarrow L}$ & $T_{C\rightarrow S}$
& $T_{C\rightarrow C}$ & $T_{C\rightarrow L}$ & $T_{C\rightarrow S}$
& $T_{C\rightarrow C}$ & $T_{C\rightarrow L}$ & $T_{C\rightarrow S}$ \\
\midrule
Depthsplat (12v) & 18.05 & 17.13 & 14.39 & 0.33 & 0.31 & 0.25 & 0.28 & 0.36 & 0.40 \\
Monosplat (12v) & 18.43 & 18.24 & 15.71 & 0.42 & 0.38 & 0.28 & 0.26 & 0.29 & 0.33 \\
Mvsplat (12v)   & 16.33 & 14.08 & 12.88 & 0.42 & 0.38 & 0.35 & 0.44 & 0.45 & 0.54 \\
\midrule
Dronesplat       & 21.25 & 19.50 & 17.19 & 0.74 & 0.70 & 0.67 & 0.45 & 0.45 & 0.48 \\
3DGS             & 24.22 & 22.52 & 16.78 & 0.71 & 0.69 & 0.64 & 0.38 & 0.39 & 0.47 \\
NeRF             & 20.93 & 19.38 & 15.73 & 0.64 & 0.58 & 0.52 & 0.42 & 0.47 & 0.59 \\
DesireGS         & 25.96 & 24.31 & 19.48 & 0.69 & 0.61 & 0.55 & 0.23 & 0.31 & 0.34 \\
PVG              & 27.01 & 25.44 & 20.23 & 0.72 & 0.64 & 0.58 & 0.20 & 0.28 & 0.31 \\
\midrule\midrule
\textbf{Model}
& \multicolumn{3}{c|}{\textbf{PSNR} $\uparrow$}
& \multicolumn{3}{c|}{\textbf{SSIM} $\uparrow$}
& \multicolumn{3}{c}{\textbf{LPIPS} $\downarrow$} \\
\cmidrule(lr){2-4} \cmidrule(lr){5-7} \cmidrule(lr){8-10}
& $T_{D\rightarrow D}$ & $T_{D\rightarrow C}$ & $T_{D\rightarrow S}$
& $T_{D\rightarrow D}$ & $T_{D\rightarrow C}$ & $T_{D\rightarrow S}$
& $T_{D\rightarrow D}$ & $T_{D\rightarrow C}$ & $T_{D\rightarrow S}$ \\
\midrule
Depthsplat (12v) & 16.78 & 9.09  & 7.87  & 0.49 & 0.16 & 0.14 & 0.45 & 0.52 & 0.56 \\
Monosplat (12v) & 18.54 & 10.21 & 8.32  & 0.38 & 0.25 & 0.23 & 0.47 & 0.49 & 0.52 \\
Mvsplat (12v)   & 16.50 & 9.20  & 7.06  & 0.37 & 0.20 & 0.18 & 0.54 & 0.54 & 0.56 \\
\midrule
Dronesplat       & 16.43 & 12.16 & 12.33 & 0.53 & 0.51 & 0.50 & 0.64 & 0.71 & 0.72 \\
3DGS             & 29.13 & 10.36 & 11.42 & 0.86 & 0.44 & 0.43 & 0.22 & 0.52 & 0.46 \\
NeRF             & 21.30 & 11.01 & 11.54 & 0.65 & 0.34 & 0.35 & 0.67 & 0.56 & 0.61 \\
DesireGS         & 26.84 & 12.01 & 13.05 & 0.76 & 0.44 & 0.43 & 0.40 & 0.60 & 0.55 \\
PVG              & 28.12 & 12.65 & 13.82 & 0.80 & 0.47 & 0.46 & 0.37 & 0.57 & 0.52 \\
\bottomrule
\end{tabular}
\caption{NVS methods are compared on different test sets in evaluation setups \cartrain (top) and \dronetrain (bottom). 12v indicates the number of input views for feed-forward 3DGS methods. Remaining views \{2,6\} are presented in \secref{suppsec:addn_results} in Supplementary.}
\label{tbl:main}
\vspace{-15pt}
\end{table*}

\noindent \textbf{Statistics.}
Out of 200 recorded sequences, 50 satisfy the training-set threshold (maximum epipolar error $< 30$). Test images meeting the same criterion relative to validated train poses are retained. Example annotated correspondences with their epipolar lines are shown in Figure~\ref{fig:epi_anno} and the distribution of mean and maximum epipolar errors for the selected sequences are shown in Figure~\ref{fig:epierror_annot}. Statistics of selected test images in each test set are presented in \secref{suppsec:dataset_stats} in the Supplementary.

\noindent \textbf{Time Synchronization} between sensors is validated using epipolar consistency checks on pixel correspondences annotated on dynamic objects that are simultaneously observed by multiple sensors. For each sequence, we randomly select two image pairs from the beginning and end segments and manually annotate four pixel correspondences on the dynamic object in each pair. Epipolar errors computed from these correspondences using the estimated camera parameters are used to verify synchronization. Example pixel annotations, epipolar lines and epipolar error distributions are shown in the Supplementary.

\section{Benchmark Methods}

\noindent \textbf{Static NVS Methods.} We evaluate classic novel-view synthesis methods, Nerf~\cite{mildenhall2021nerf} and 3DGS~\cite{kerbl20233d}. In particular, we evaluate the NerfStudio variants, \textit{nerfacto} and \textit{splatfacto} respectively. For 3DGS, we initialize the centers of 3D Gaussians using the sparse 3D point cloud of the scene computed using SfM~\cite{colmap}(\cf \secref{subsec:pose_est}.

\noindent \textbf{Dynamic NVS Methods.} Standard NVS methods are designed for static scenes, whereas driving scenes contain dynamic objects. We evaluate recent 3DGS-based methods for such settings: DroneSplat~\cite{dronesplat} and PVG~\cite{pvg}. The latter two require LiDAR supervision and dynamic object masks. For segmentation masks, we intersect outputs from DroneSplat (which may contain outliers) and SAM~\cite{kirillov2023segment} prompted for vehicles and pedestrians, ensuring only dynamic regions are retained (\cf \figref{fig:static_dynamic}).  


To replace LiDAR supervision, we use depth from the off-the-shelf estimator
DaV3~\cite{lin2025depth3recoveringvisual}. The predicted depth maps are resized to the training-image
resolution and aligned to COLMAP SfM depth using an affine scale-and-shift
transformation:
\begin{equation}
\mathcal{L}_{\mathrm{depth}} =
\sum_{p \in \mathcal{P}}
\left\| \alpha D_{\mathrm{DaV3}}(p) + \beta - D_{\mathrm{SfM}}(p) \right\|_1 ,
\end{equation}
where $\mathcal{P}$ denotes pixels with valid SfM depth, and $\alpha,\beta$
are estimated per scene. 

\noindent \textbf{Feed-Forward 3DGS Methods} directly predict pixel-aligned 3D Gaussian Splats from input context images using a neural network trained on thousands of image sequences~\cite{re10k,ling2024dl3dv}. We consider three methods in particular, Depthsplat~\cite{depthsplat}, Monosplat~\cite{monosplat} and Mvsplat~\cite{mvsplat}. We consider three setups of 2,6 and 12 context views for each of these feed-forward 3DGS methods. 


\section{Experiments}


\begin{figure*}[t]
\centering

\begin{subfigure}{0.49\textwidth}
    \centering
    \includegraphics[width=\linewidth]{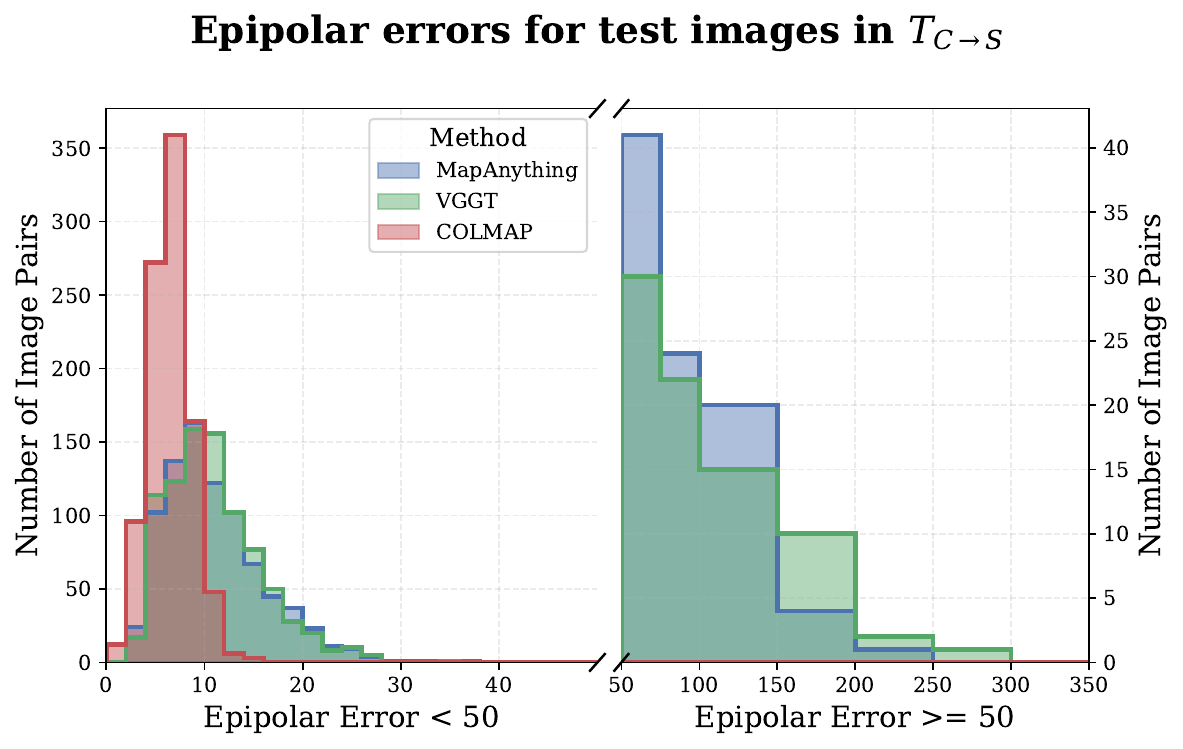}
    \caption{Epipolar errors for test images in $T_{C \rightarrow S}$}
\end{subfigure}
\hfill
\begin{subfigure}{0.49\textwidth}
    \centering
    \includegraphics[width=\linewidth]{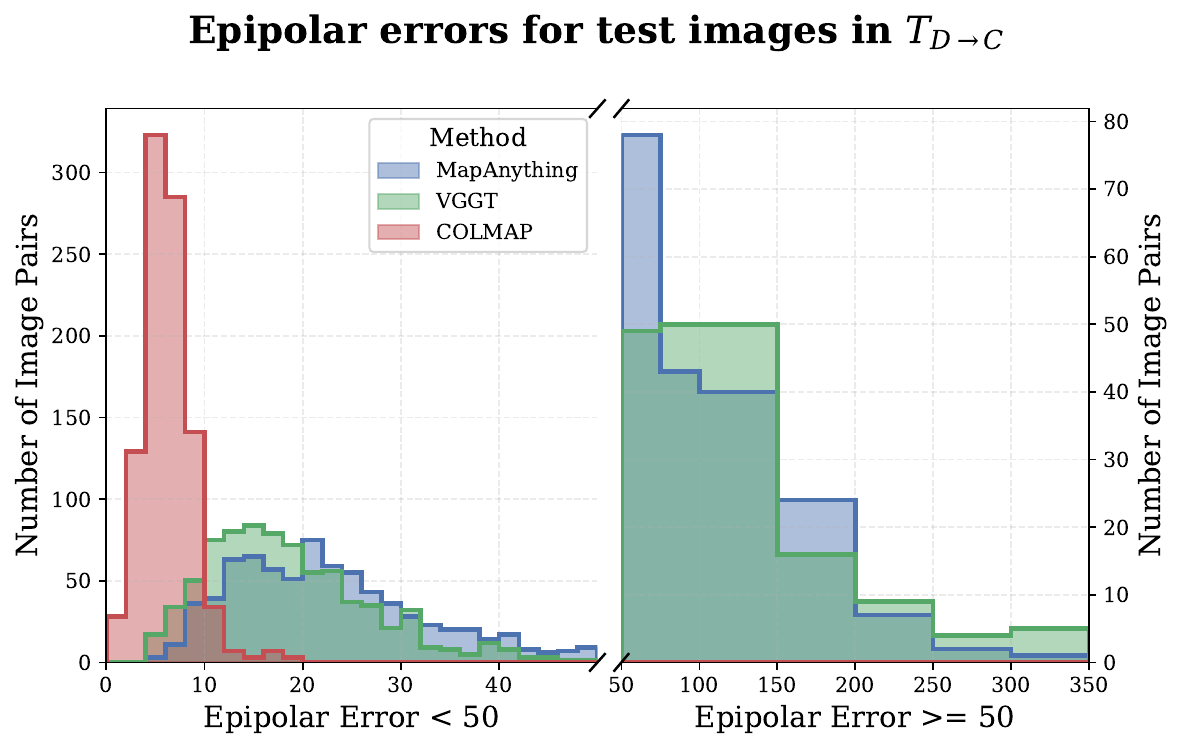}
    \caption{Epipolar errors for test images in $T_{D \rightarrow C}$}
\end{subfigure}

\vspace{4pt}


\caption{\textbf{Camera Pose Estimation.}COLMAP produces low average epipolar errors ($< 30$), whereas feed-forward pose estimators yield significantly larger errors ($> 50$). The number of high-error cases is substantially greater in $T_{D \rightarrow C}$ due to the wide-baseline aerial–ground viewpoint gap.}
\label{fig:epi_anno_cpe}
\vspace{-10pt}

\end{figure*}

\subsection{Evaluation Protocol}
We follow standard evaluation protocol measuring three metrics to assess the quality of rendered images - PSNR, SSIM, LPIPS~\cite{lpips} that evaluates the quality at pixel-level, patch-level and feature-level respectively. We compute these metrics for each image in the test sets in respective evaluation setups \cartrain and \dronetrain.

\subsection{Method Comparison}

We evaluate multiple NVS methods and summarize the results for both evaluation setups in \tabref{tbl:main}. On the standard test set, $T_{C \rightarrow C}$ in the \cartrain setup—where training and test views are sampled from the same car sequence $V_C$ - PVG~\cite{pvg} achieves the highest performance with PSNR $|$ SSIM $|$ LPIPS = $27.01 | 0.72 | 0.20$. In contrast, PVG exhibits a significant performance drop on the proposed cross-vehicle test sets, $T_{C \rightarrow S}$, where training and test images originate from car and scooty sensors, respectively. This degradation is consistent across all evaluated baselines. A similar trend is observed between $T_{C \rightarrow L}$ and $T_{C \rightarrow S}$, demonstrating that having an alternate camera sensor on the car is insufficient to assess the extrapolation capability of NVS models.

The same observation holds in the \dronetrain setup: 3DGS and PVG achieve the best results on the standard test set $T_{D \rightarrow D}$, but their performance declines on the cross-vehicle test sets $T_{D \rightarrow S}$ and $T_{D \rightarrow C}$. Interestingly, the aerial-specific method DroneSplat~\cite{dronesplat} is outperformed by the general-purpose NVS methods 3DGS and PVG. The results on dynamic objects NVS (\secref{suppsec:dynamic_objects} in Supplementary) follow the same trend
as the full-image evaluation: performance consistently degrades with increasing viewpoint variation. 

Accurate NVS is essential for driving simulation, which requires smooth and continuous camera trajectories (e.g., lane changes). Such simulations are only meaningful for downstream applications, including autonomous driving evaluation, if synthesized views are photorealistic and geometrically consistent. The poor NVS performance on $T_{C \rightarrow S}$ suggests that even simple scenarios, such as simulating a lane change from the car lane to the scooty lane, would result in unrealistic renderings due to poor synthesized scooty-view quality.


\vspace{-10pt}
\subsection{Dataset Comparison and Depth Filtering}

We compare PVG on \MV and Waymo Open Dataset (WOD)~\cite{waymo2020} to analyze whether
\MV remains comparable to standard driving NVS benchmarks under the
same-trajectory $T_{C\rightarrow C}$ setting. On WOD~\cite{waymo2020}, PVG~\cite{pvg} is commonly evaluated
with LiDAR supervision and train/test views sampled from the same vehicle
trajectory. Despite using single-camera input and monocular DaV3 depth instead
of LiDAR, PVG obtains comparable $T_{C\rightarrow C}$ performance on \MV
(Table~\ref{tab:depth_filter_ablation}), indicating that our benchmark remains
consistent with existing driving NVS settings in the standard interpolation
regime.

\begin{figure*}[h]
\centering

\begin{tabular}{cc|cc}
\includegraphics[width=0.24\textwidth]{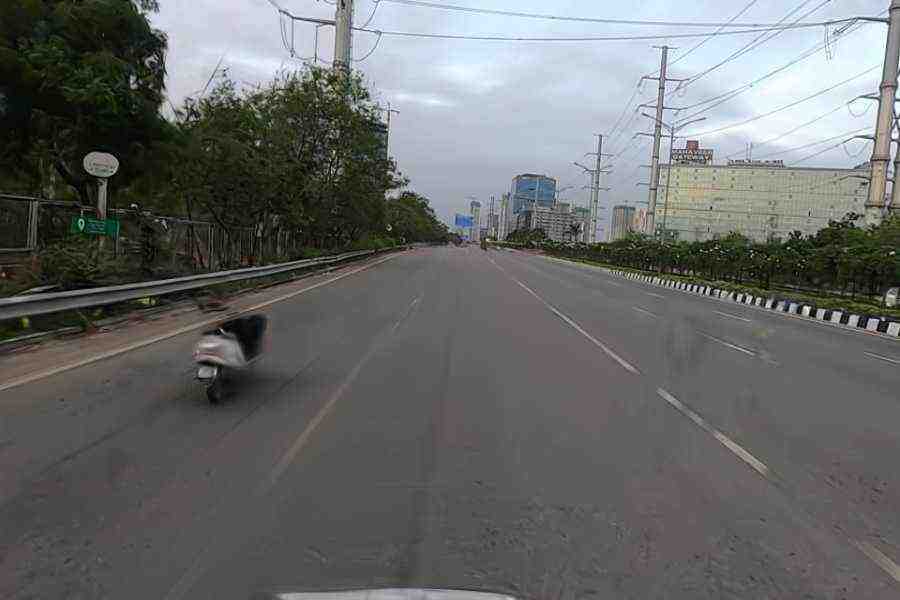} &
\includegraphics[width=0.24\textwidth]{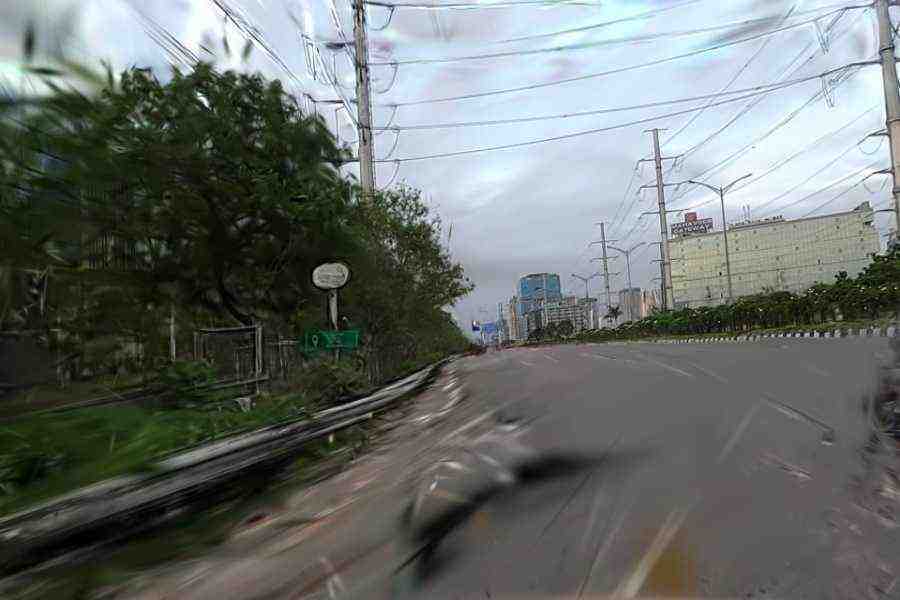} &
\includegraphics[width=0.24\textwidth]{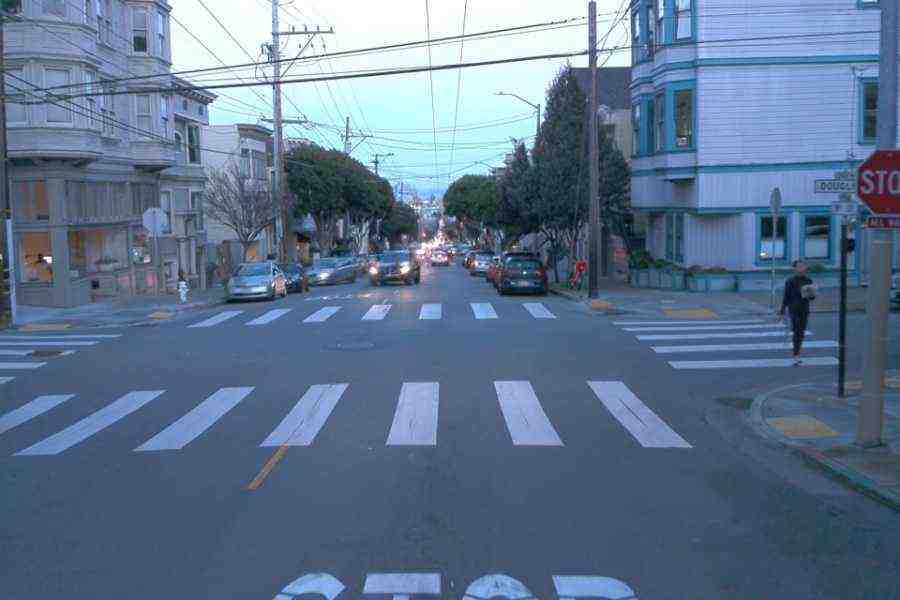} &
\includegraphics[width=0.24\textwidth]{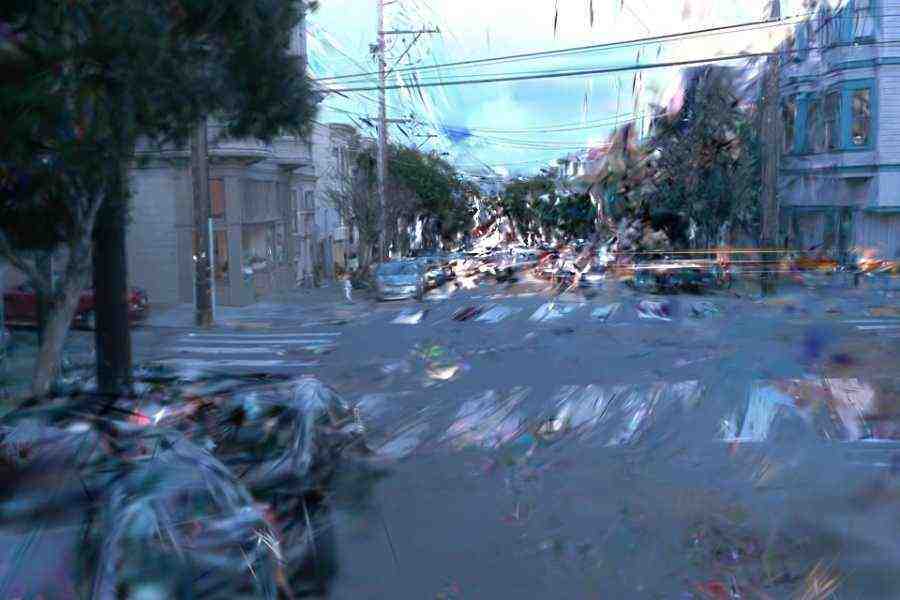}
\end{tabular}

\caption{\textbf{Lane-change scenario rendering comparison.}
Columns 1–2 show renderings from the \textbf{MV$^2$ dataset} ($T_{C \rightarrow C}$ and $T_{C \rightarrow S}$ views),
while Columns 3–4 show renderings from the \textbf{Waymo dataset} ($T_{C \rightarrow C}$ and cross-lane views).
BBoth examples illustrate a vehicle performing a lane-change maneuver.}
\label{fig:cross-lane-wod-mv2}
\end{figure*}

Since WOD does not provide a true cross-vehicle $T_{C\rightarrow S}$ setup, we
also perform a qualitative cross-lane study using PVG. Fig.~\ref{fig:cross-lane-wod-mv2}
shows that cross-lane renderings on WOD exhibit degradation similar to the
$T_{C\rightarrow S}$ renderings on \MV. This supports our main claim that
wide-baseline cross-lane synthesis remains challenging even when evaluated on
existing driving datasets.

We further study the role of depth supervision and filtering. For this analysis,
all WOD subset experiments use a matched single-camera PVG setup. We compare LiDAR depth, COLMAP depth (no DaV3), DaV3 with
IQR filtering, and DaV3 with $[10,30]$m filtering. As shown in
Table~\ref{tab:depth_filter_ablation}, restricting DaV3 supervision to the $[10,30]$m
range improves results on both WOD and \MV. On \MV, this improves both
$T_{C\rightarrow C}$ and $T_{C\rightarrow S}$, showing that depth filtering is
important for monocular-depth-based PVG. However, the gap between
$T_{C\rightarrow C}$ and $T_{C\rightarrow S}$ remains, confirming that improved
depth filtering strengthens the baseline but does not remove the cross-vehicle wide-baseline NVS challenge. Additional details on dataset splits for WOD and \MV datasets used in this ablation alongwith the motivation for depth filtering are provided in the Supplementary~\secref{suppsec:depth_filter}.

\vspace{-10pt}

\begin{table}[t]
\centering
\begin{tabular}{llcccccc}
\toprule
\multirow{2}{*}{\textbf{Setup}} &
\multirow{2}{*}{\textbf{Depth Supervision}} &
\multicolumn{3}{c}{\textbf{WOD}} &
\multicolumn{3}{c}{\textbf{\MV}} \\
\cmidrule(lr){3-5} \cmidrule(lr){6-8}
& & PSNR $\uparrow$ & SSIM $\uparrow$ & LPIPS $\downarrow$
& PSNR $\uparrow$ & SSIM $\uparrow$ & LPIPS $\downarrow$ \\
\midrule

\multirow{5}{*}{$T_{C\rightarrow C}$}
& LiDAR multi-view   & 28.45 & 0.912 & 0.145 & --    & --    & --    \\
& LiDAR single-view  & 28.66 & 0.792 & 0.121 & --    & --    & --    \\
& COLMAP             & 27.84 & 0.765 & 0.145 & 24.72 & 0.655 & 0.238 \\
& DaV3 + IQR               & 26.95 & 0.735 & 0.205 & 25.88 & 0.680 & 0.210 \\
& DaV3 $[10,30]$m     & 29.12 & 0.815 & 0.108 & 27.01 & 0.720 & 0.200 \\

\midrule
\multirow{2}{*}{$T_{C\rightarrow S}$}
& DaV3 + IQR              & --    & --    & --    & 19.43 & 0.550 & 0.330 \\
& DaV3 $[10,30]$m     & --    & --    & --    & 20.23 & 0.580 & 0.310 \\

\bottomrule
\end{tabular}
\caption{PVG comparison on WOD and \MV under different depth-supervision and filtering settings. LiDAR multi-view is the standard WOD PVG setup, while LiDAR single-view uses matched single-camera supervision. DaV3 + IQR uses monocular depth with IQR-based filtering, while DaV3 $[10,30]$m retains only valid depths in the 10--30m range. Since WOD lacks a true cross-vehicle setup, $T_{C\rightarrow S}$ is reported only on \MV.}
\label{tab:depth_filter_ablation}
\end{table}


\begin{figure*}[ht!]
\centering
\scriptsize
\setlength{\tabcolsep}{1pt}
\renewcommand{\arraystretch}{0.8}

\begin{adjustbox}{max width=\textwidth}
\begin{tabular}{ccccc}

\textbf{GT} & \textbf{3DGS} & \textbf{NeRF} & \textbf{DepthSplat} & \textbf{PVG} \\

\includegraphics[width=0.19\linewidth]{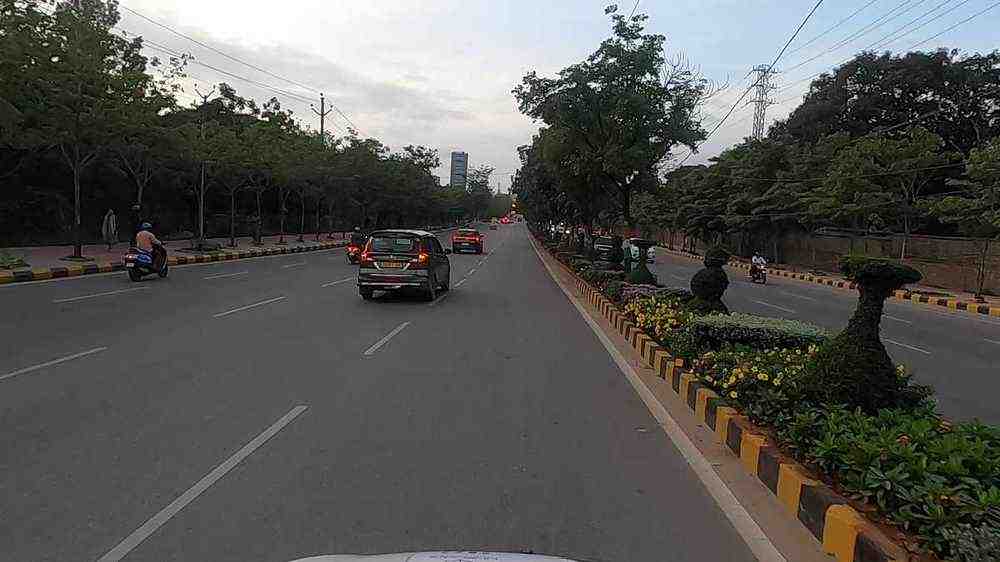} &
\includegraphics[width=0.19\linewidth]{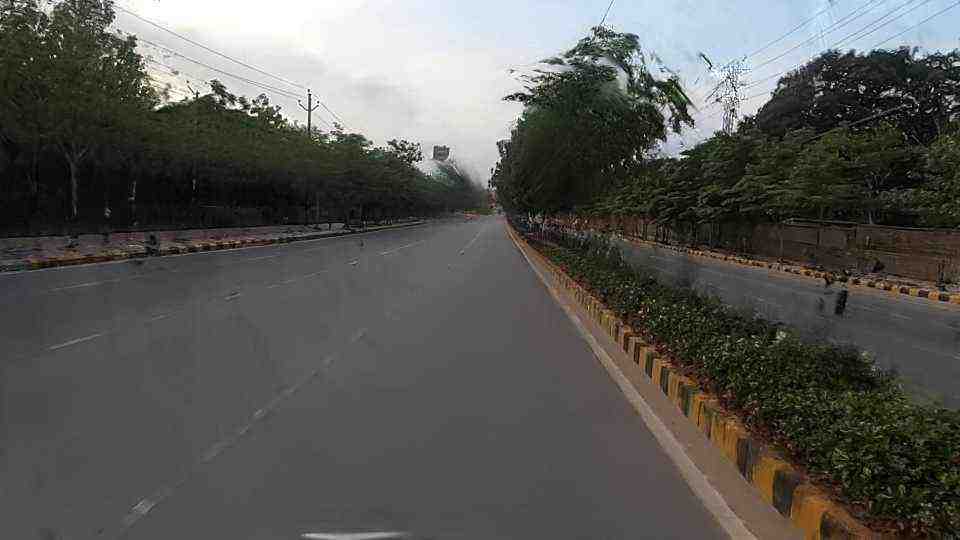} &
\includegraphics[width=0.19\linewidth]{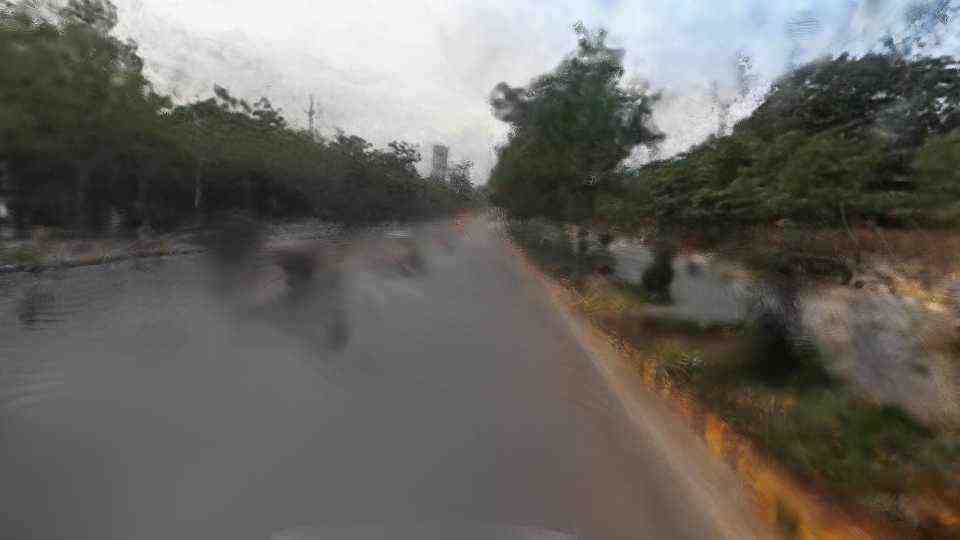} &
\includegraphics[width=0.19\linewidth]{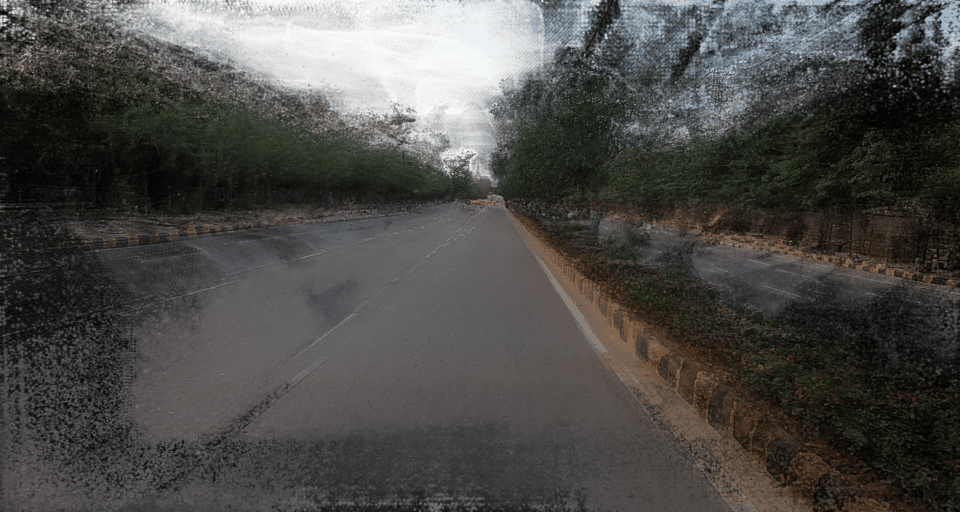} &
\includegraphics[width=0.19\linewidth]{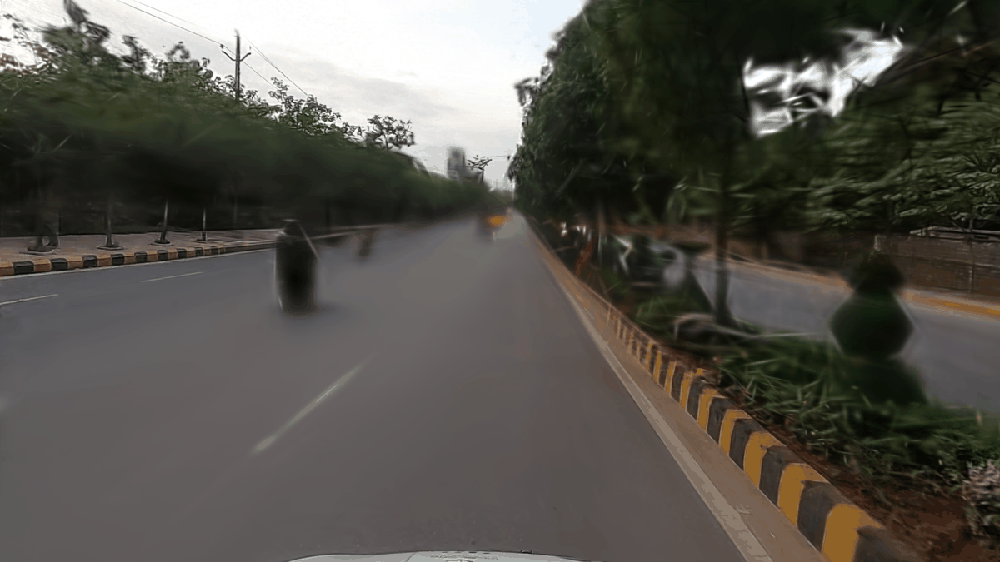} \\
& 21.49$|$0.73$|$0.33 & 17.89$|$0.51$|$0.61 & 14.33$|$0.39$|$0.53 & 25.91$|$0.72$|$0.19 \\

\includegraphics[width=0.19\linewidth]{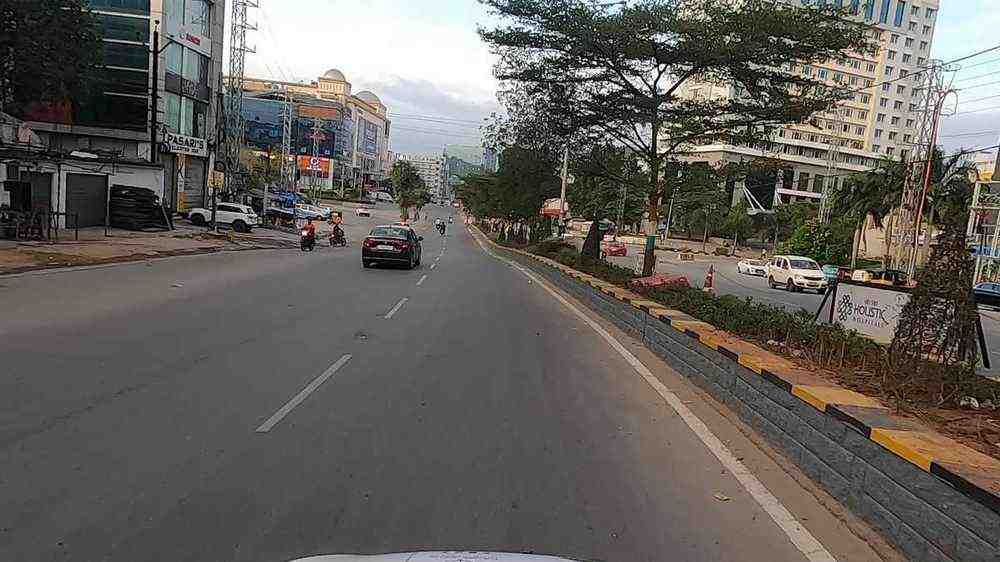} &
\includegraphics[width=0.19\linewidth]{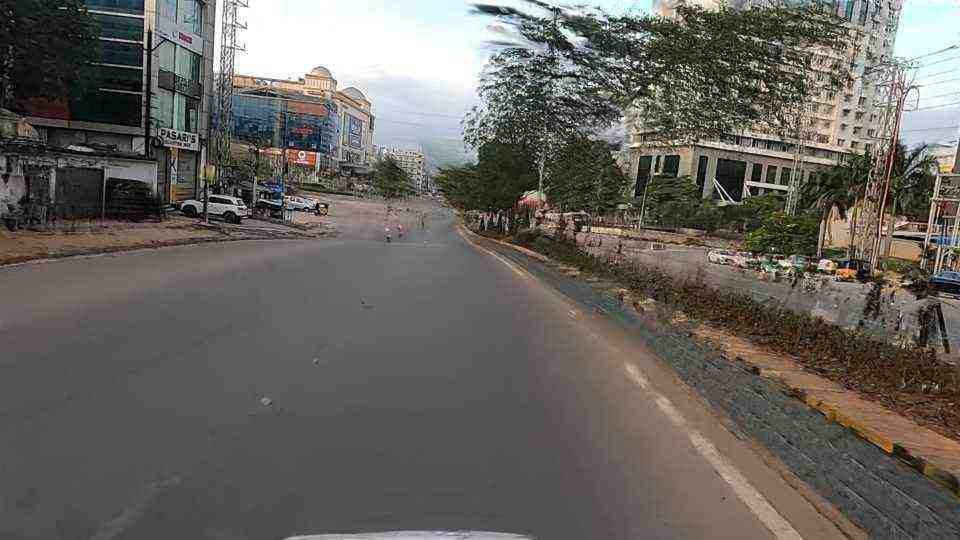} &
\includegraphics[width=0.19\linewidth]{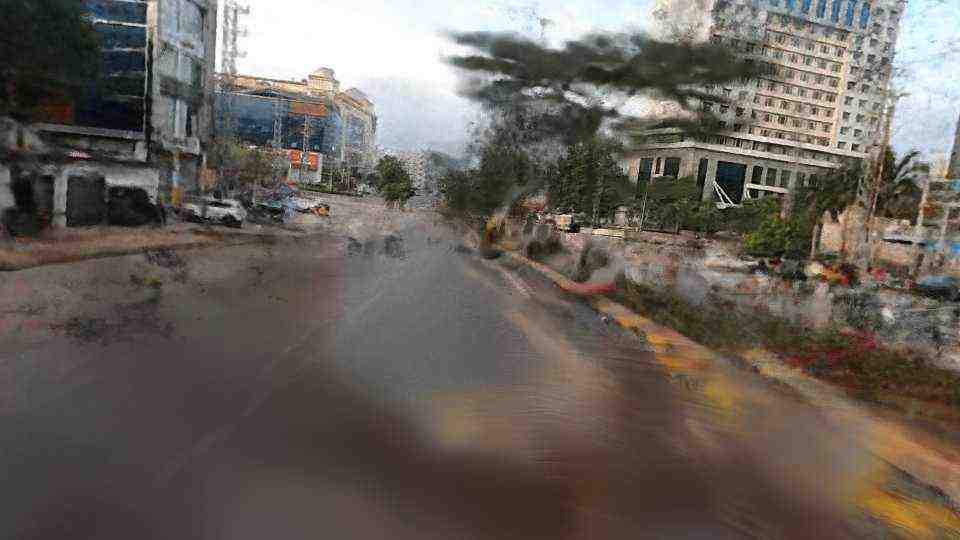} &
\includegraphics[width=0.19\linewidth]{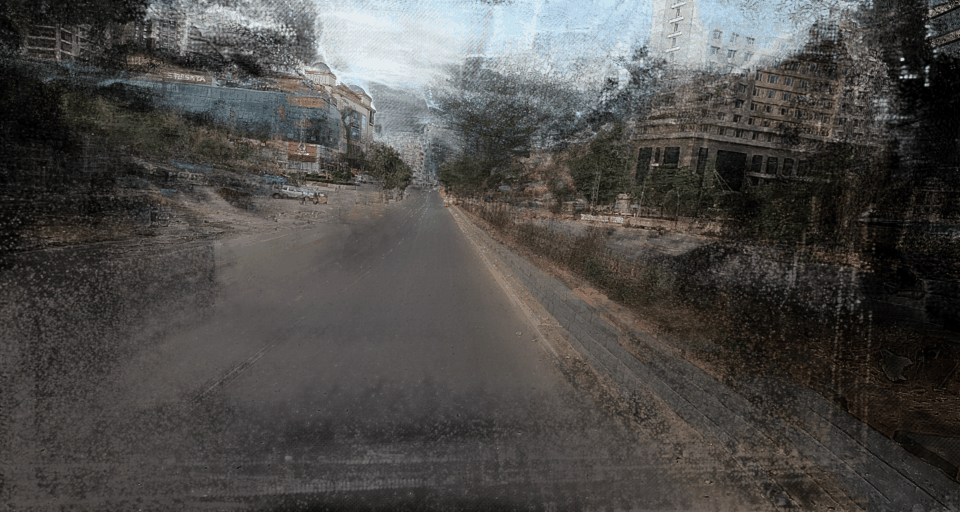} &
\includegraphics[width=0.19\linewidth]{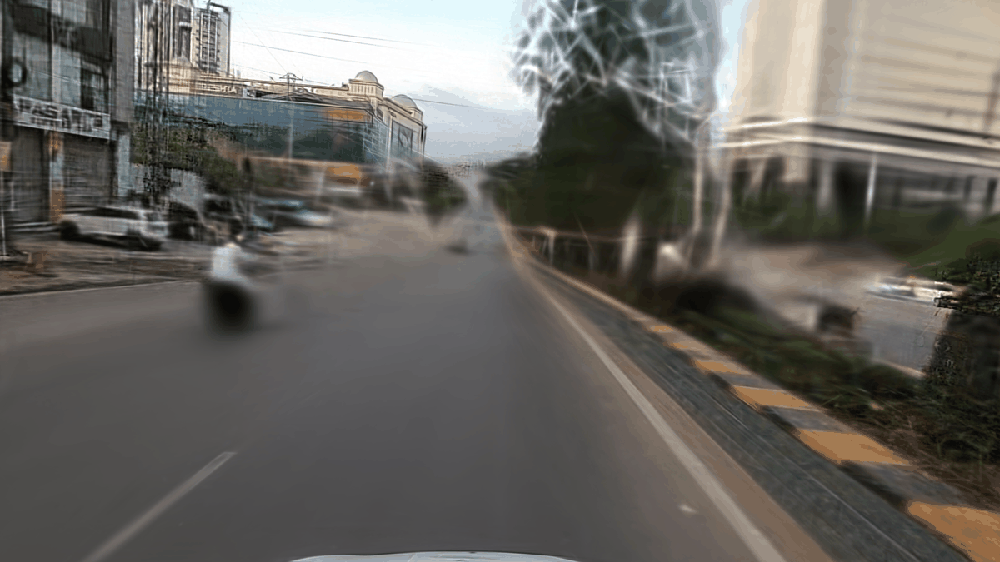} \\
& 19.00$|$0.74$|$0.34 & 14.39$|$0.42$|$0.61 & 12.99$|$0.33$|$0.59 & 24.56$|$0.64$|$0.29 \\

\includegraphics[width=0.19\linewidth]{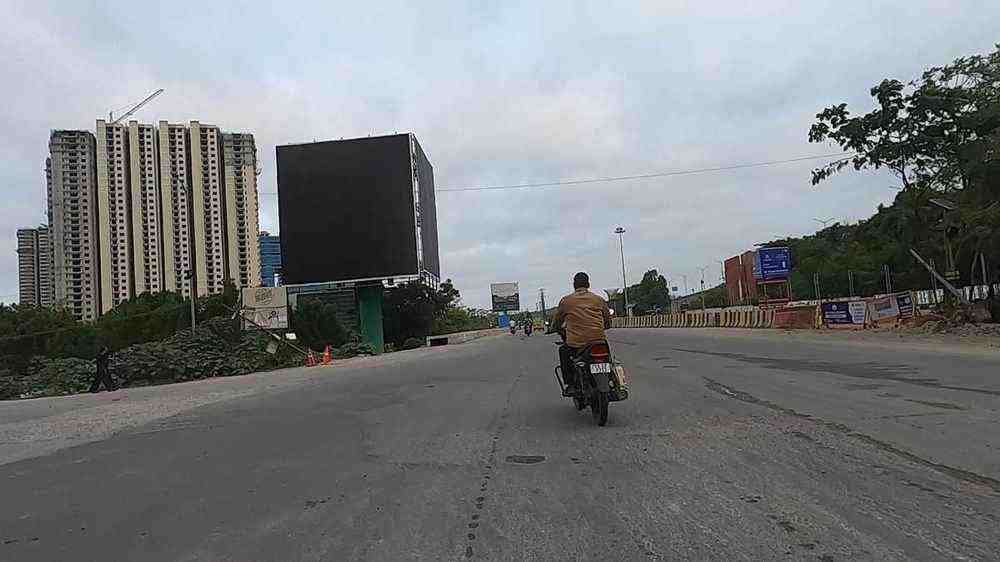} &
\includegraphics[width=0.19\linewidth]{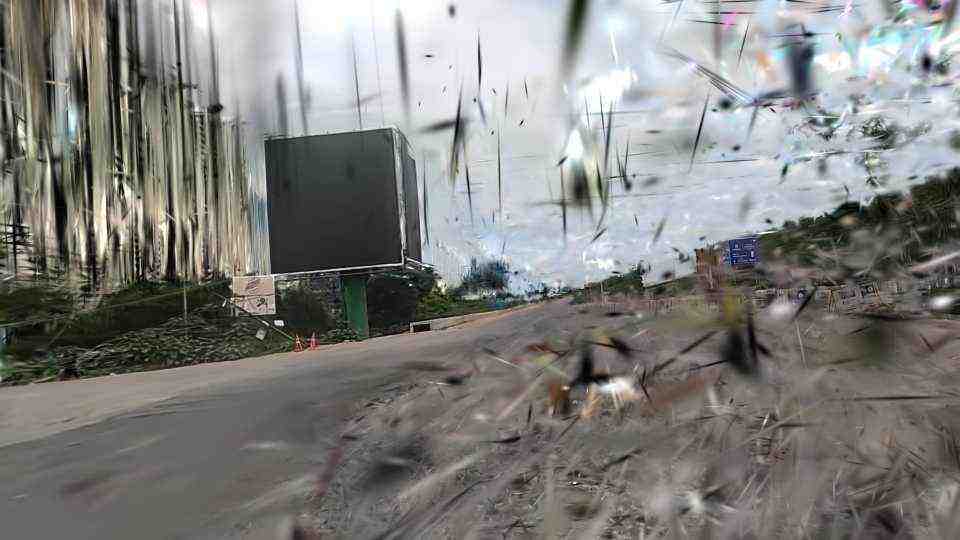} &
\includegraphics[width=0.19\linewidth]{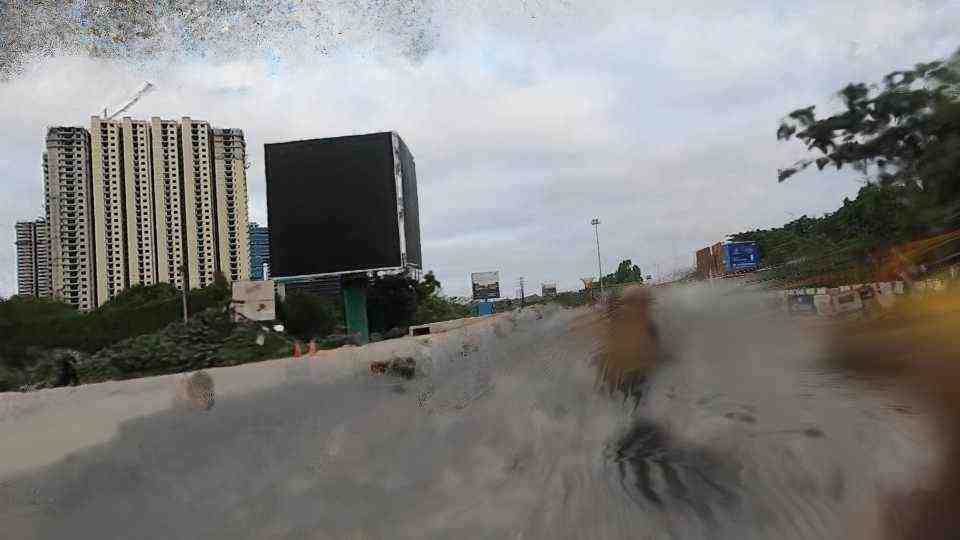} &
\includegraphics[width=0.19\linewidth]{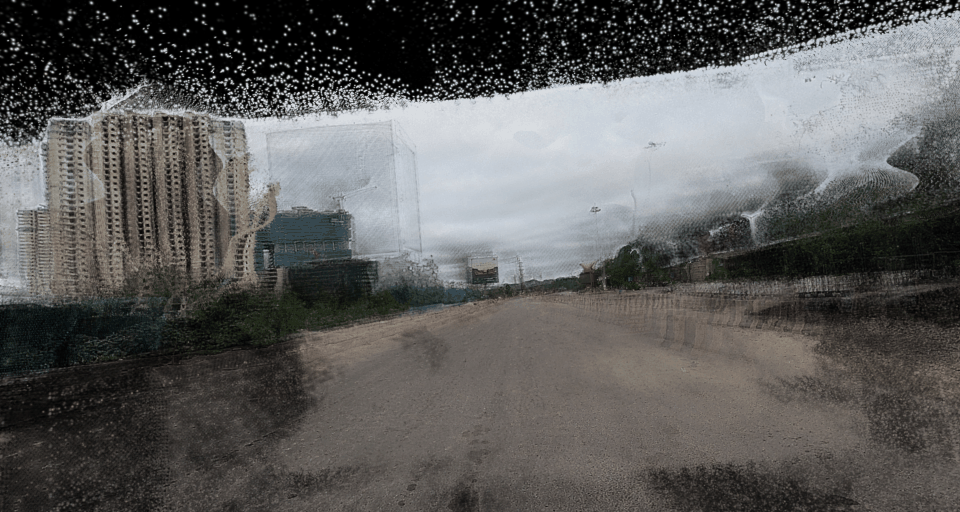} &
\includegraphics[width=0.19\linewidth]{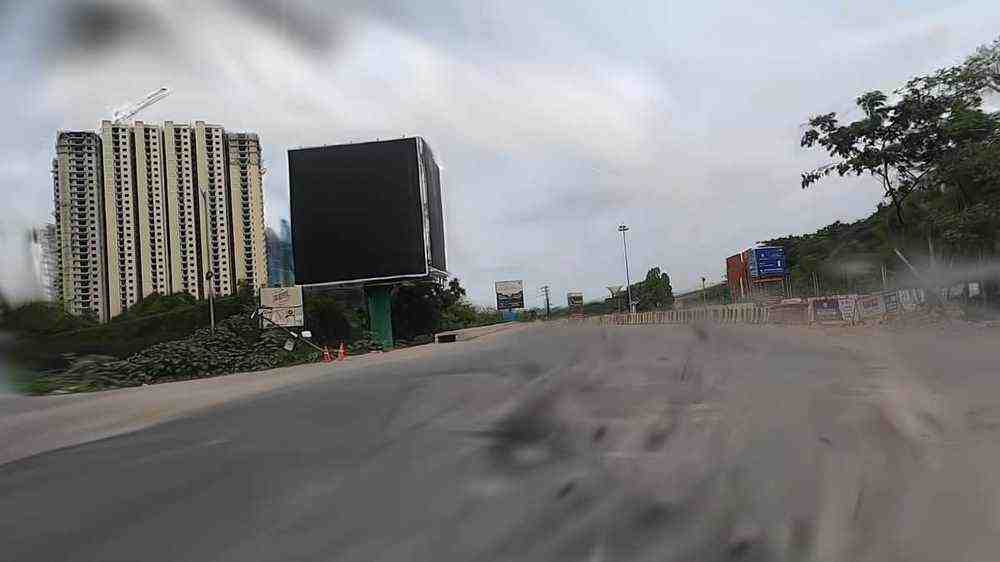} \\
& 13.96$|$0.42$|$0.63 & 18.05$|$0.58$|$0.46 & 9.03$|$0.28$|$0.71 & 20.02$|$0.75$|$0.41 \\

\includegraphics[width=0.19\linewidth]{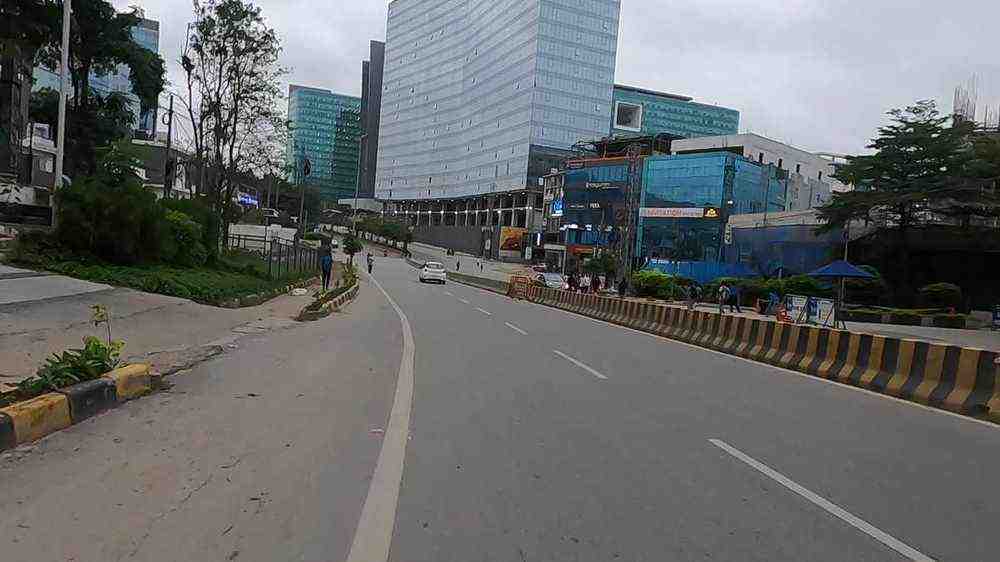} &
\includegraphics[width=0.19\linewidth]{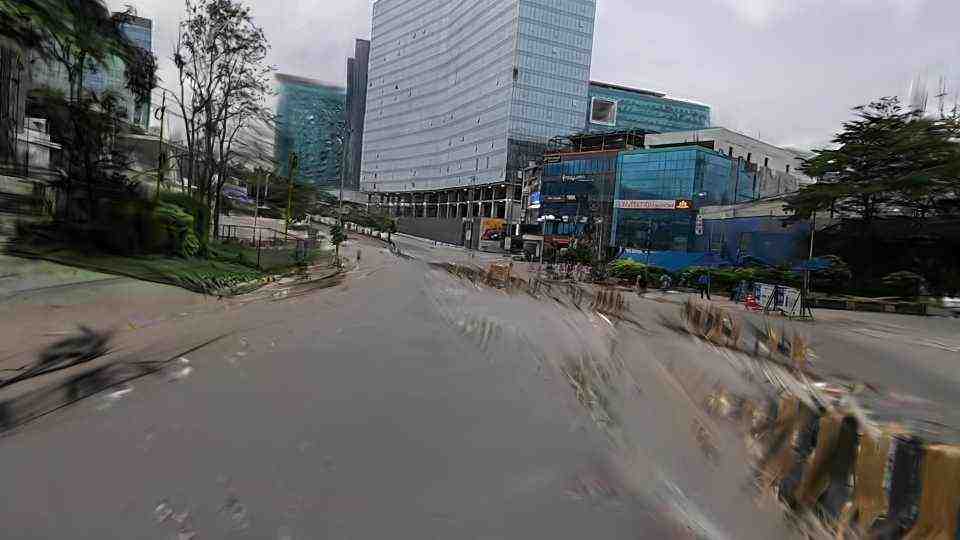} &
\includegraphics[width=0.19\linewidth]{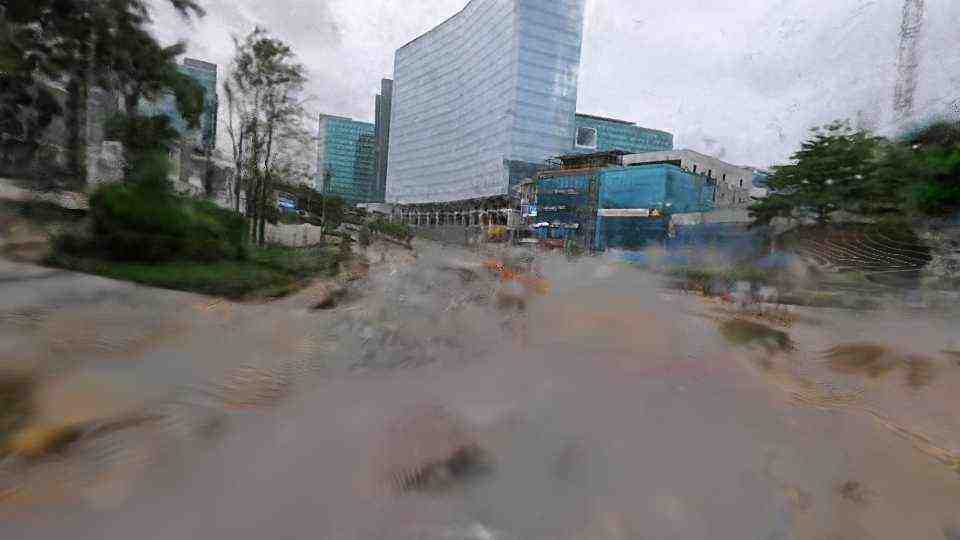} &
\includegraphics[width=0.19\linewidth]{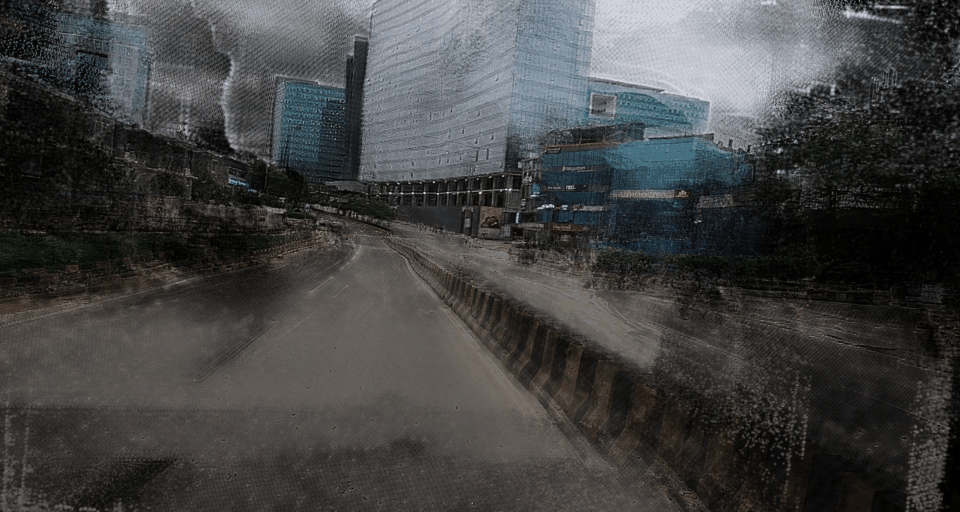} &
\includegraphics[width=0.19\linewidth]{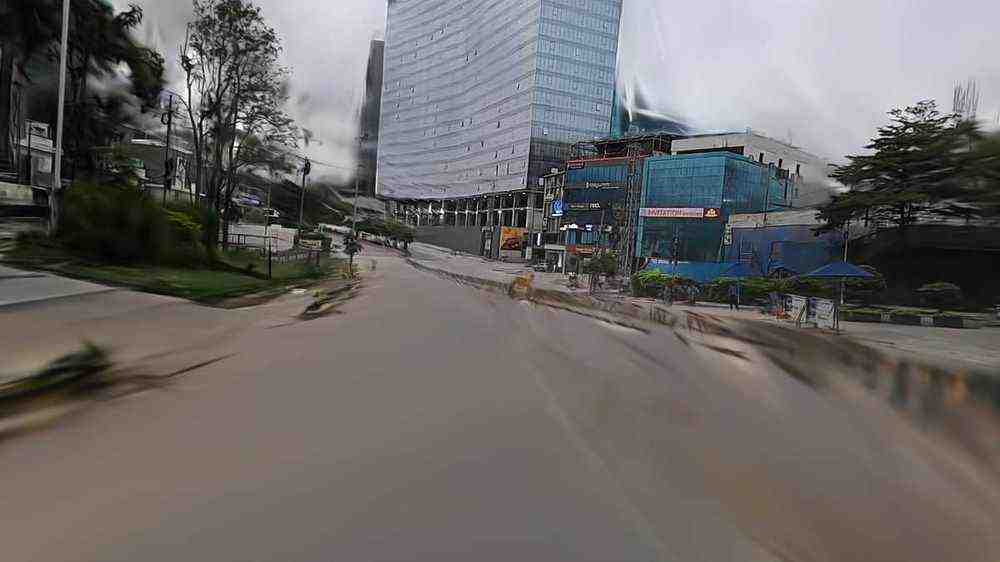} \\
& 20.08$|$0.66$|$0.39 & 14.88$|$0.46$|$0.62 & 10.61$|$0.27$|$0.69 & 19.86$|$0.71$|$0.42 \\

\includegraphics[width=0.19\linewidth]{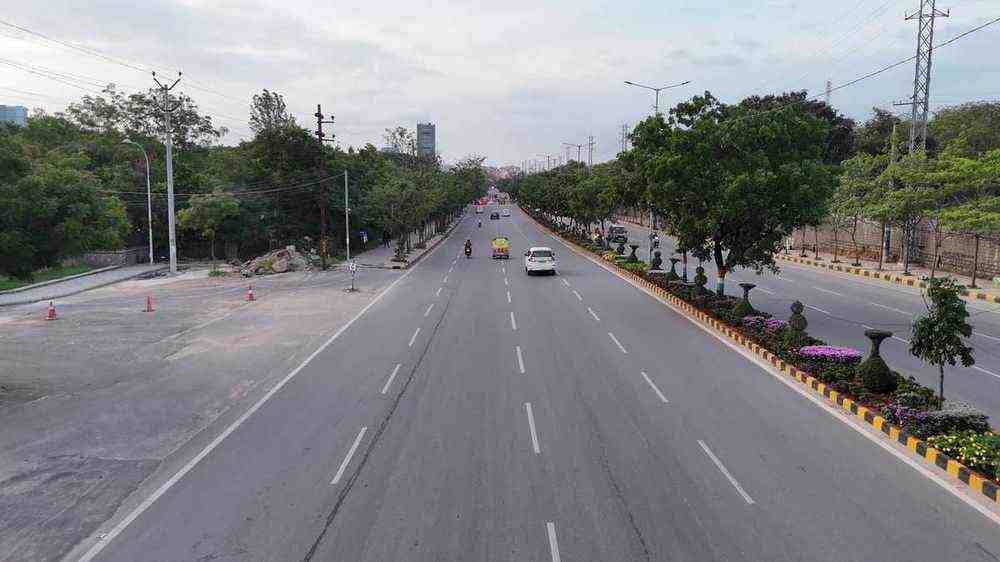} &
\includegraphics[width=0.19\linewidth]{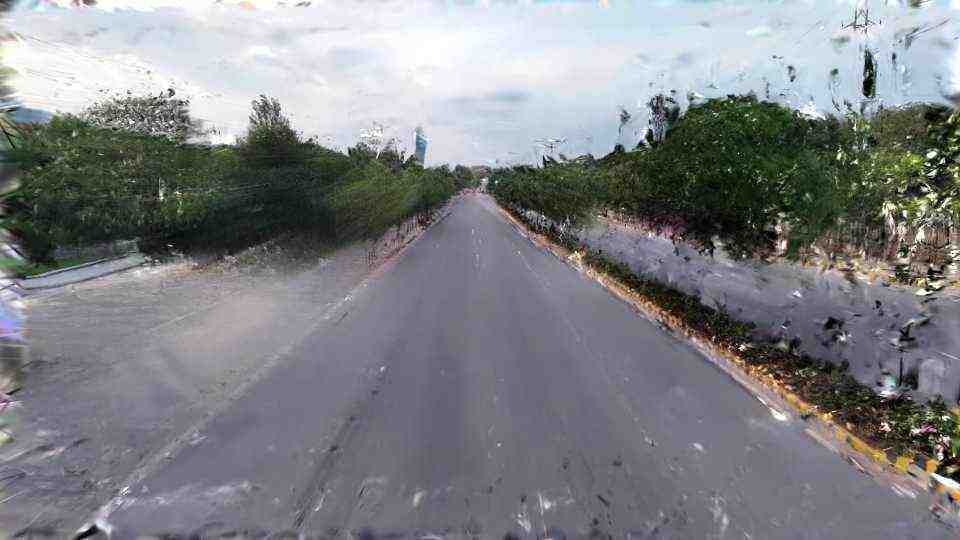} &
\includegraphics[width=0.19\linewidth]{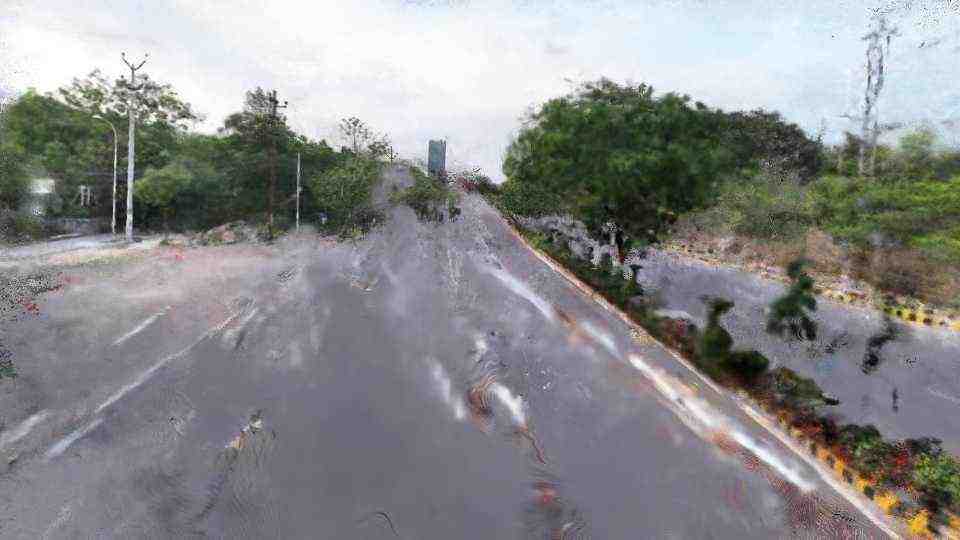} &
\includegraphics[width=0.19\linewidth]{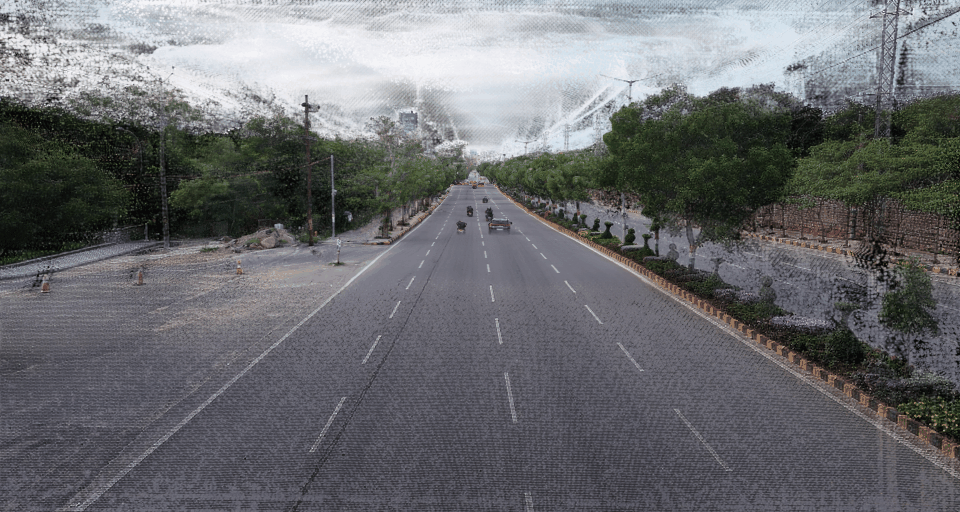} &
\includegraphics[width=0.19\linewidth]{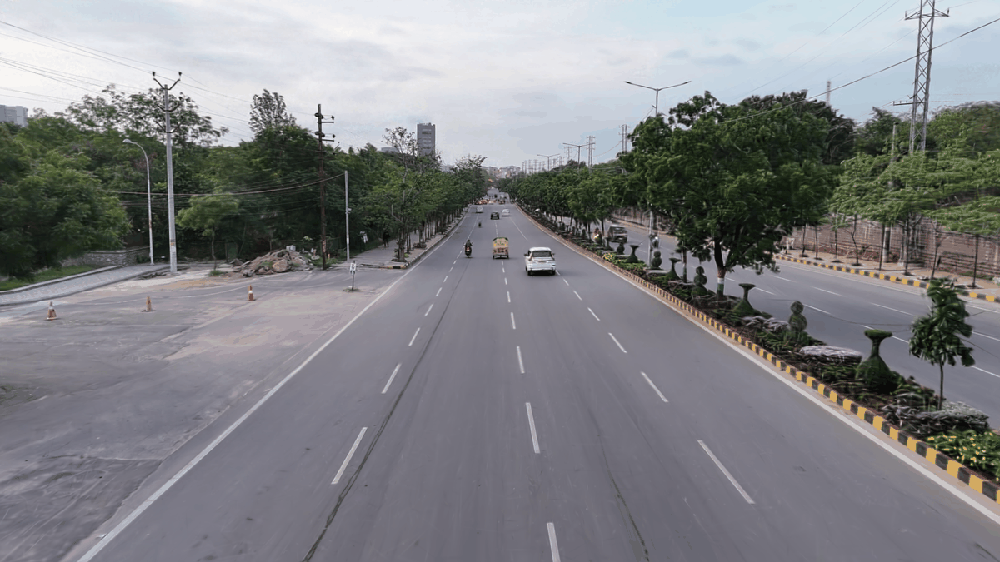} \\
& 19.89$|$0.58$|$0.31 & 17.08$|$0.54$|$0.51 & 19.65$|$0.59$|$0.26 & 28.34$|$0.86$|$0.10 \\

\includegraphics[width=0.19\linewidth]{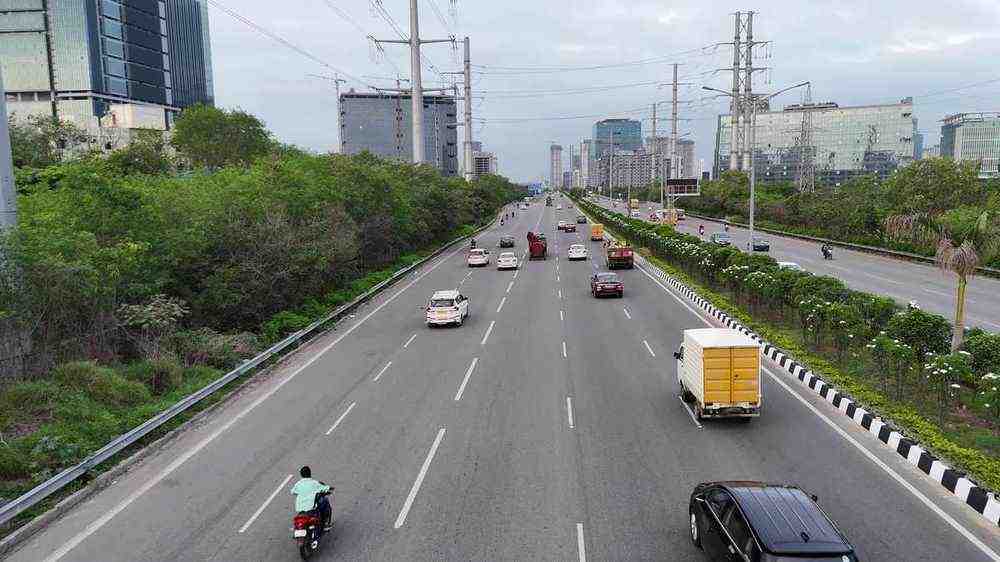} &
\includegraphics[width=0.19\linewidth]{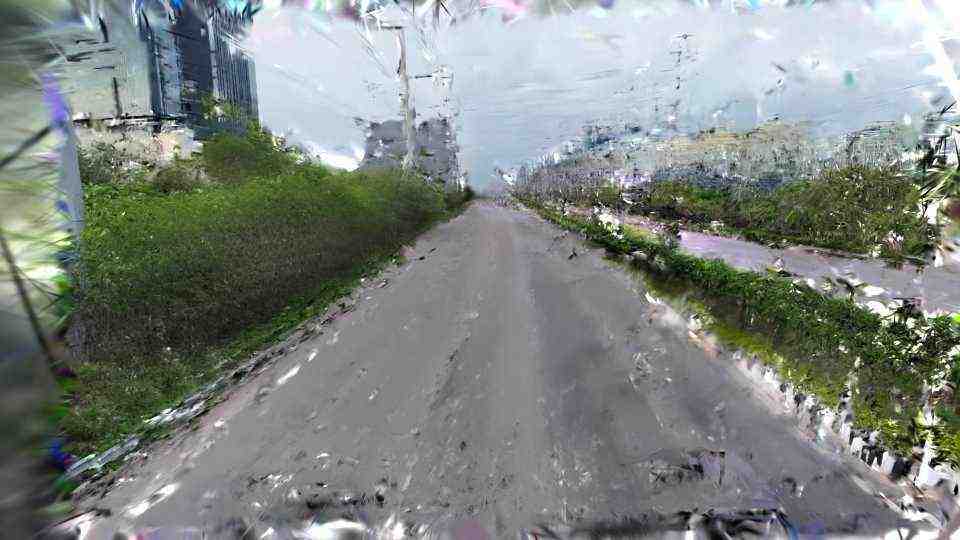} &
\includegraphics[width=0.19\linewidth]{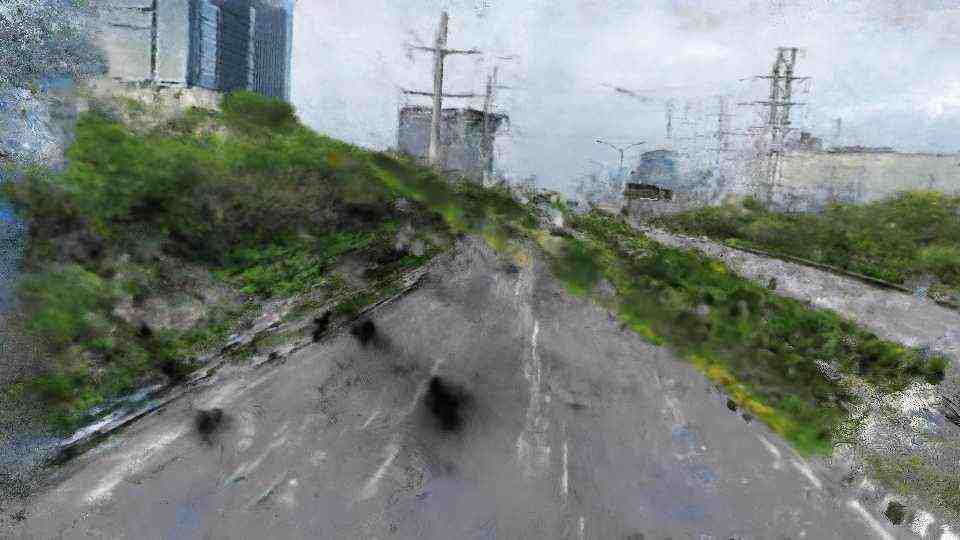} &
\includegraphics[width=0.19\linewidth]{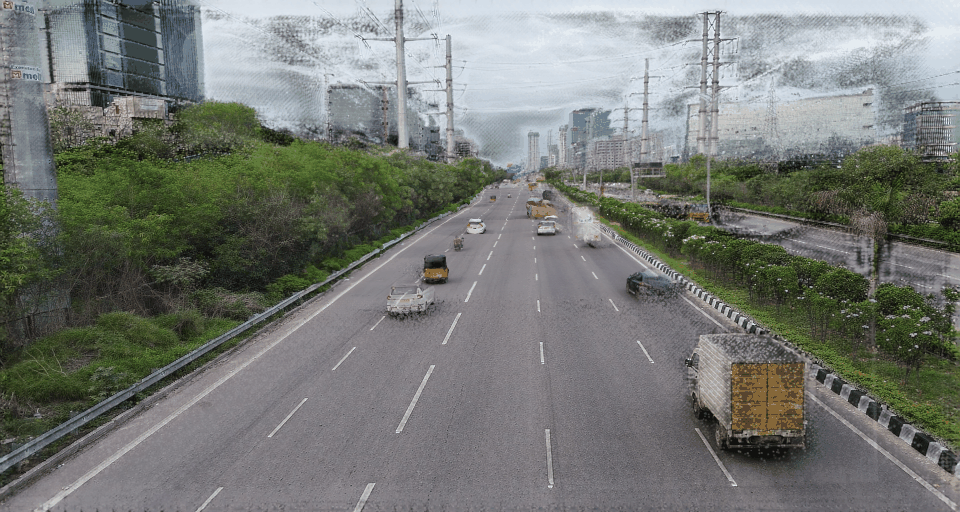} &
\includegraphics[width=0.19\linewidth]{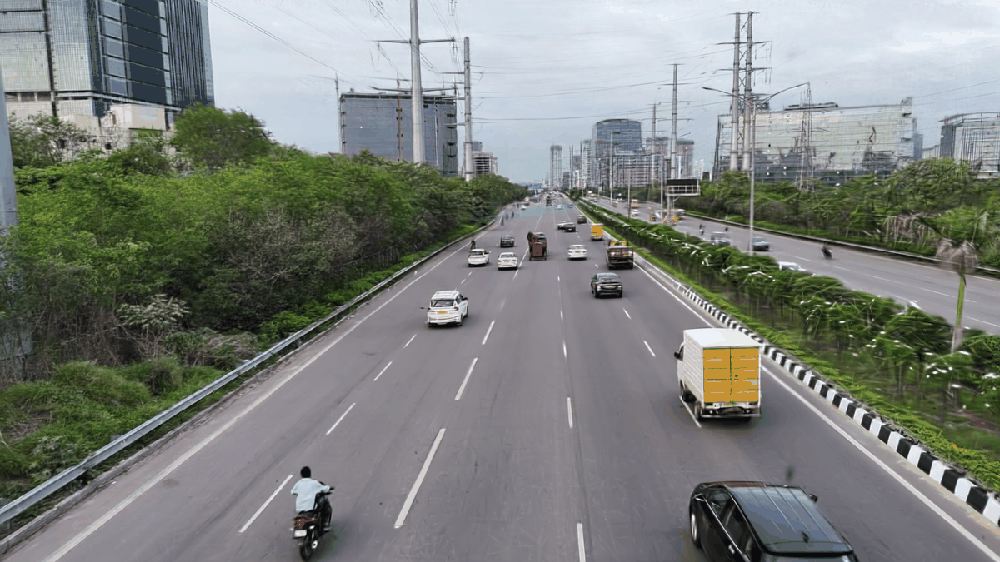} \\
& 18.12$|$0.50$|$0.48 & 16.43$|$0.50$|$0.63 & 19.01$|$0.53$|$0.29 & 28.41$|$0.88$|$0.13 \\

\includegraphics[width=0.19\linewidth]{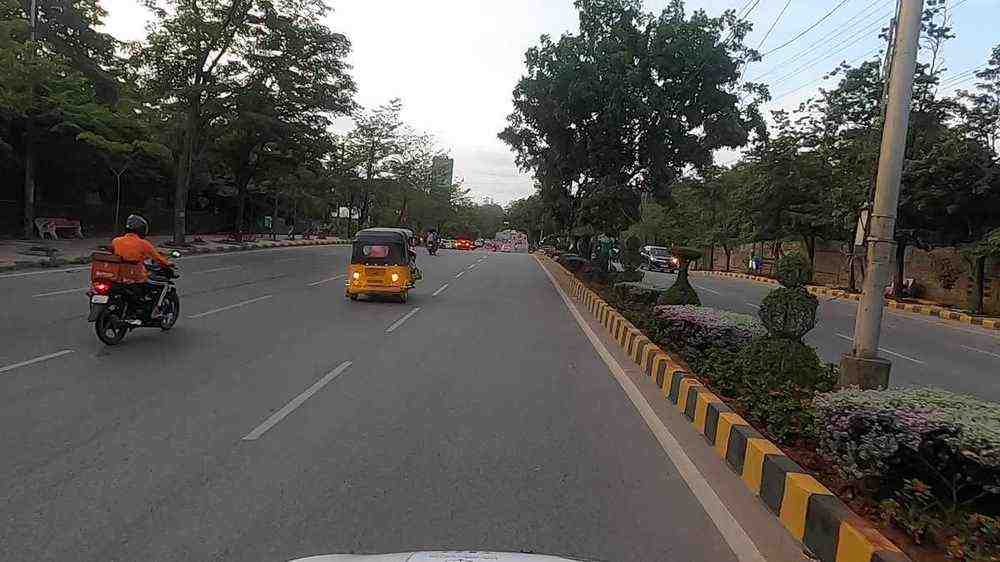} &
\includegraphics[width=0.19\linewidth]{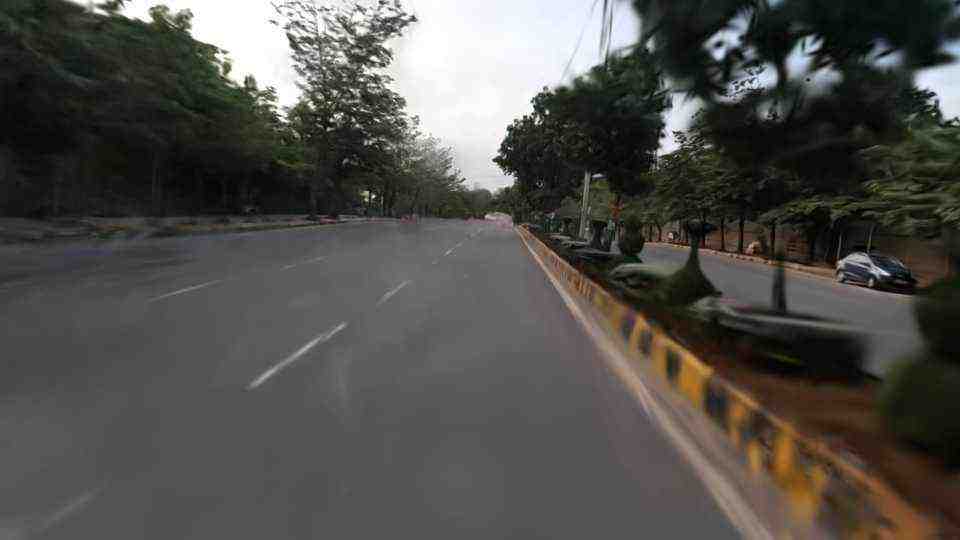} &
\includegraphics[width=0.19\linewidth]{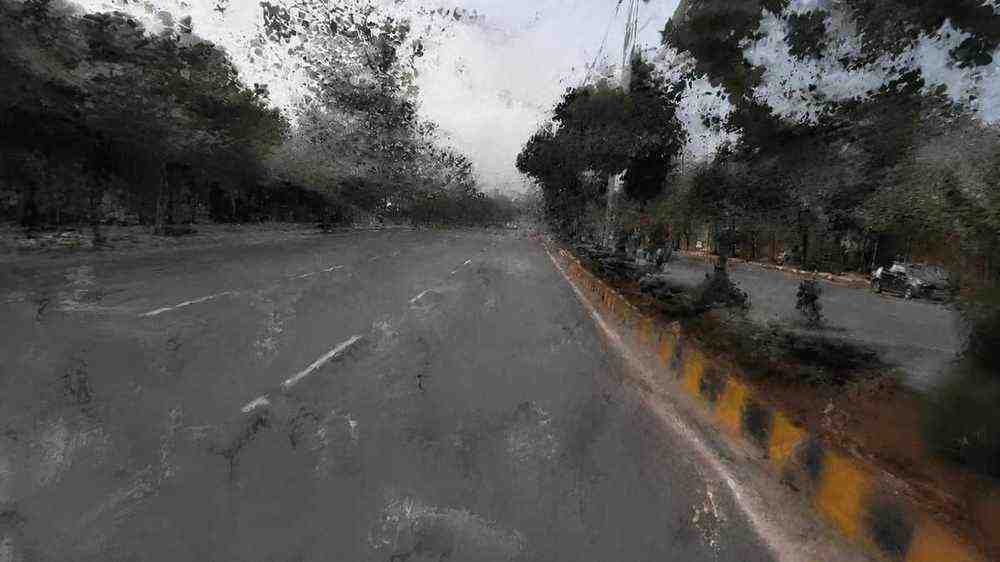} &
\includegraphics[width=0.19\linewidth]{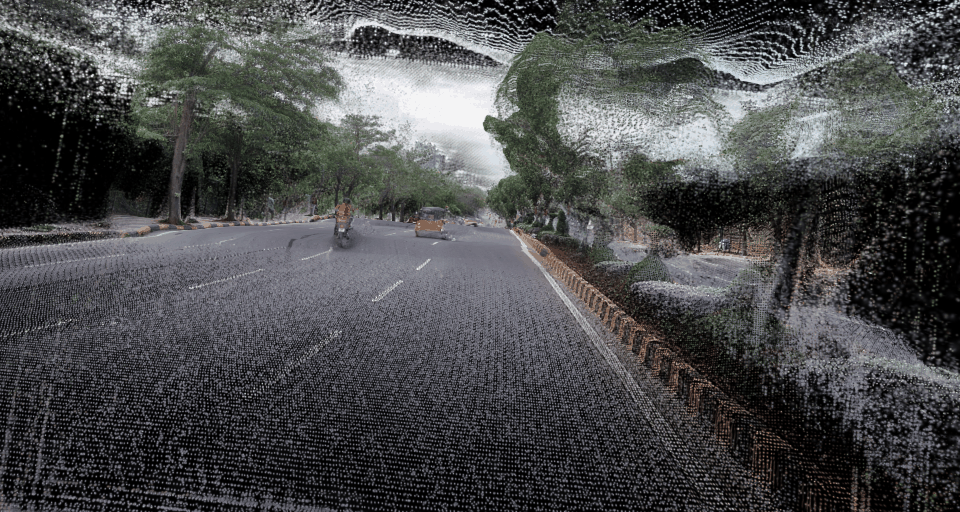} &
\includegraphics[width=0.19\linewidth]{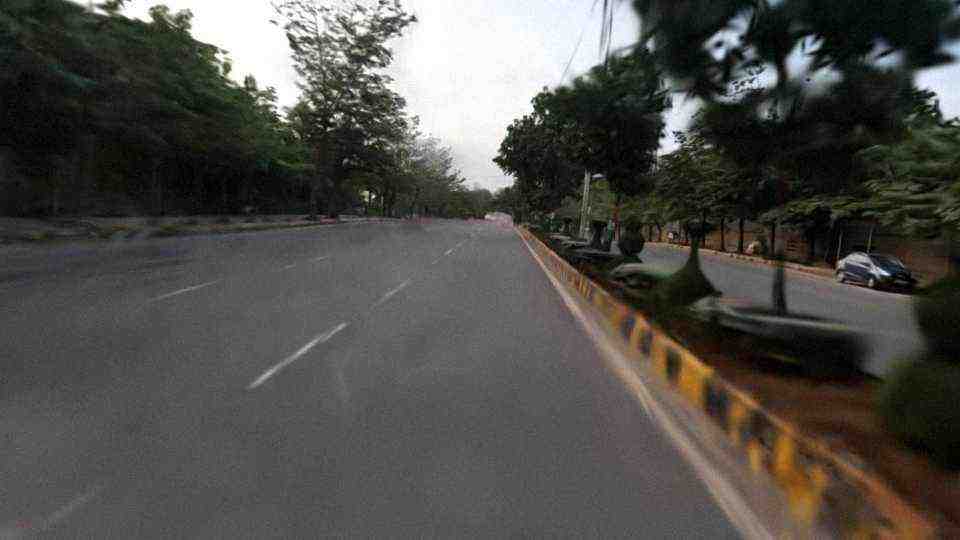} \\
& 15.37$|$0.61$|$0.47 & 15.51$|$0.55$|$0.45 & 15.21$|$0.59$|$0.43 & 19.64$|$0.66$|$0.26 \\

\includegraphics[width=0.19\linewidth]{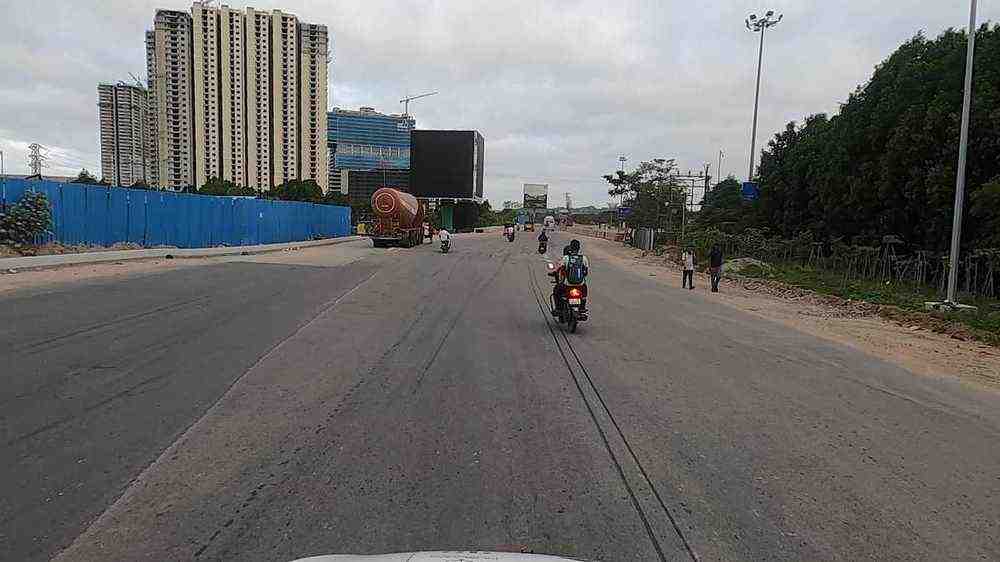} &
\includegraphics[width=0.19\linewidth]{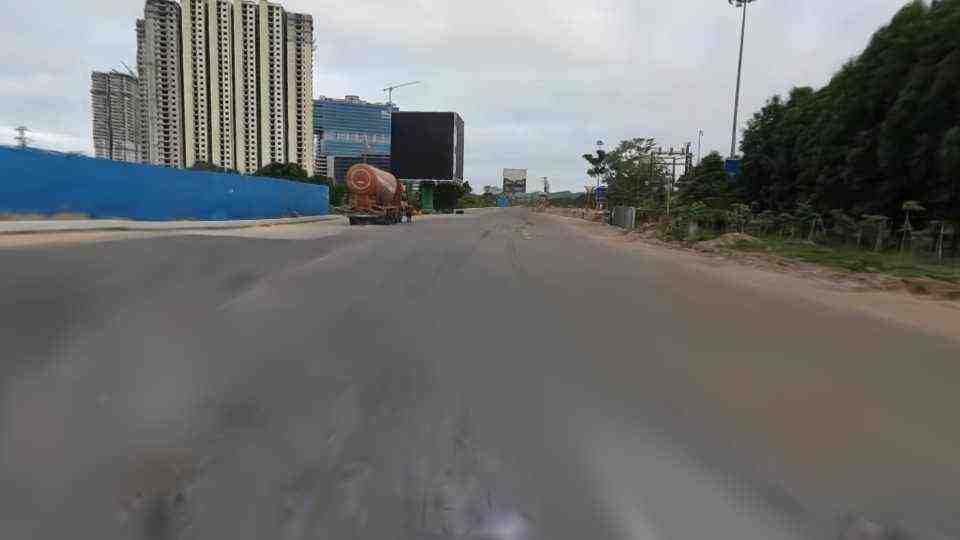} &
\includegraphics[width=0.19\linewidth]{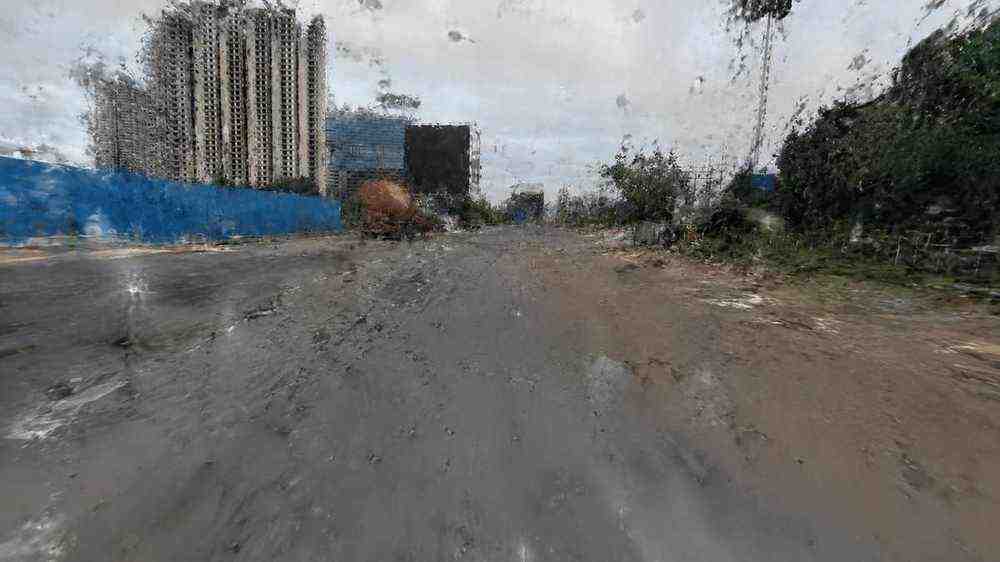} &
\includegraphics[width=0.19\linewidth]{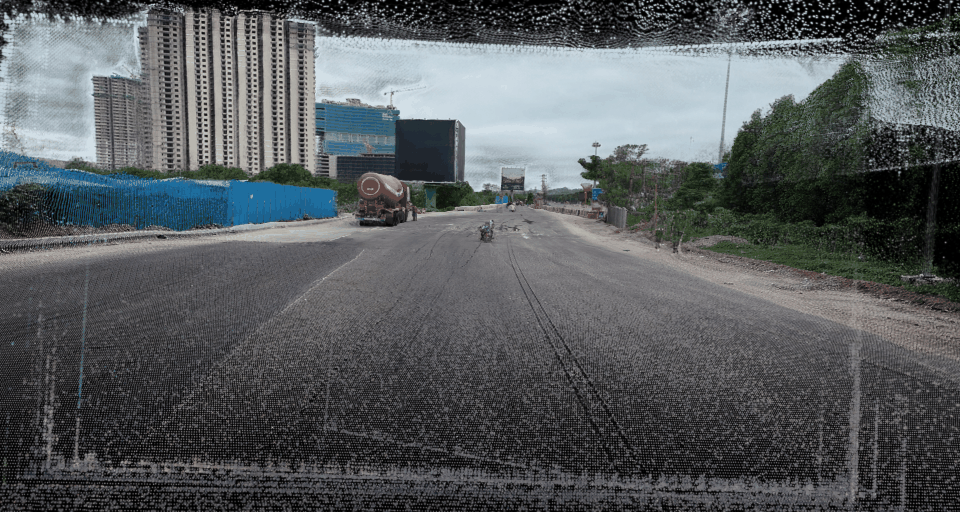} &
\includegraphics[width=0.19\linewidth]{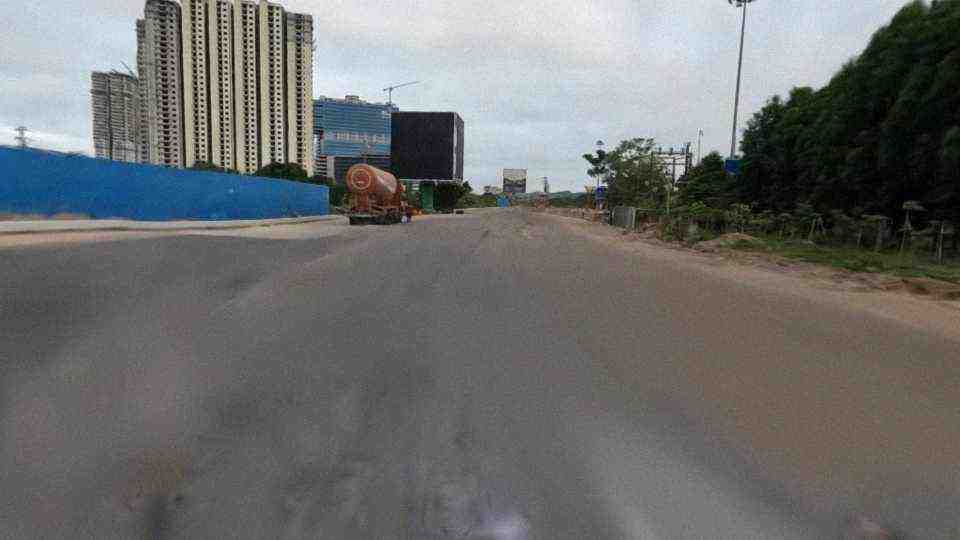} \\
& 15.24$|$0.55$|$0.53 & 15.39$|$0.55$|$0.44 & 15.67$|$0.56$|$0.39 & 18.53$|$0.58$|$0.31 \\

\includegraphics[width=0.19\linewidth]{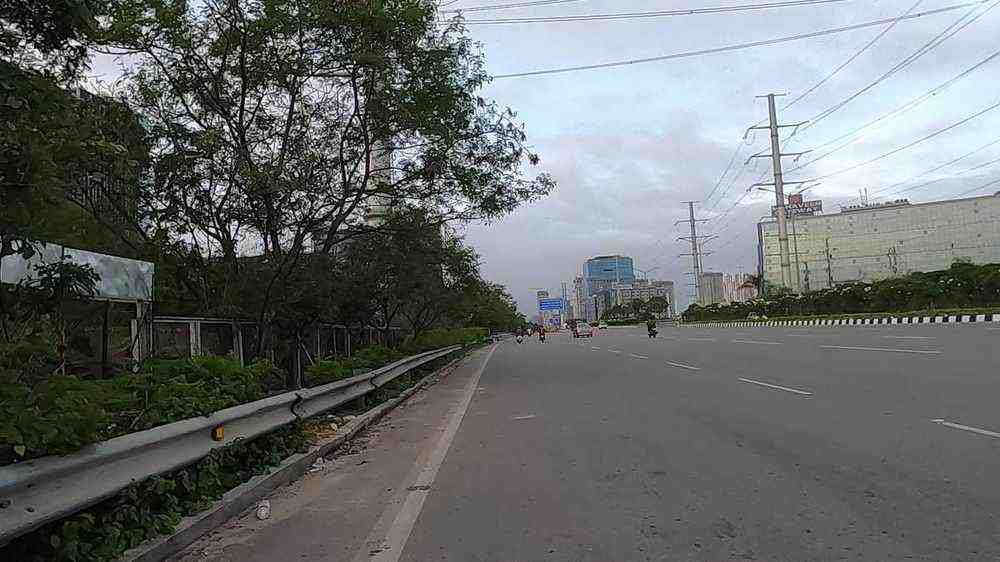} &
\includegraphics[width=0.19\linewidth]{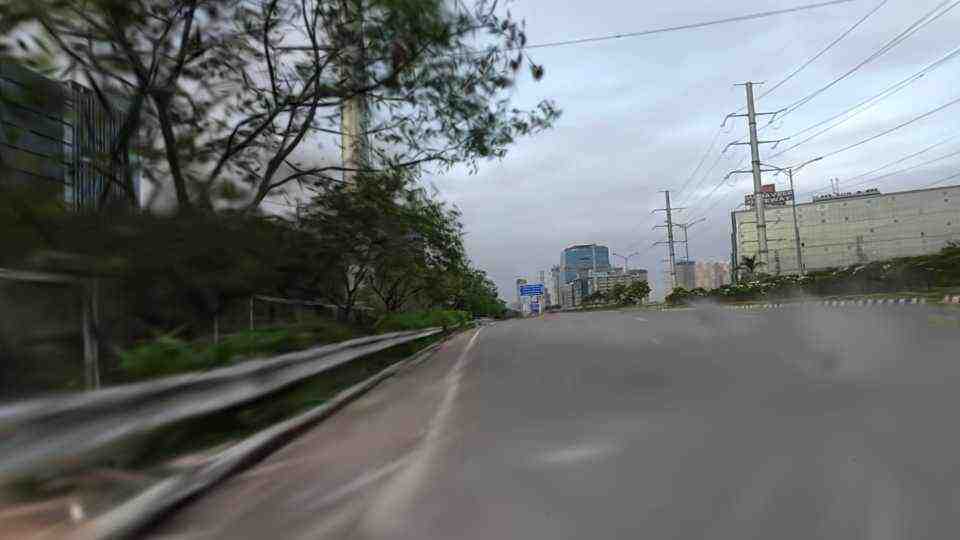} &
\includegraphics[width=0.19\linewidth]{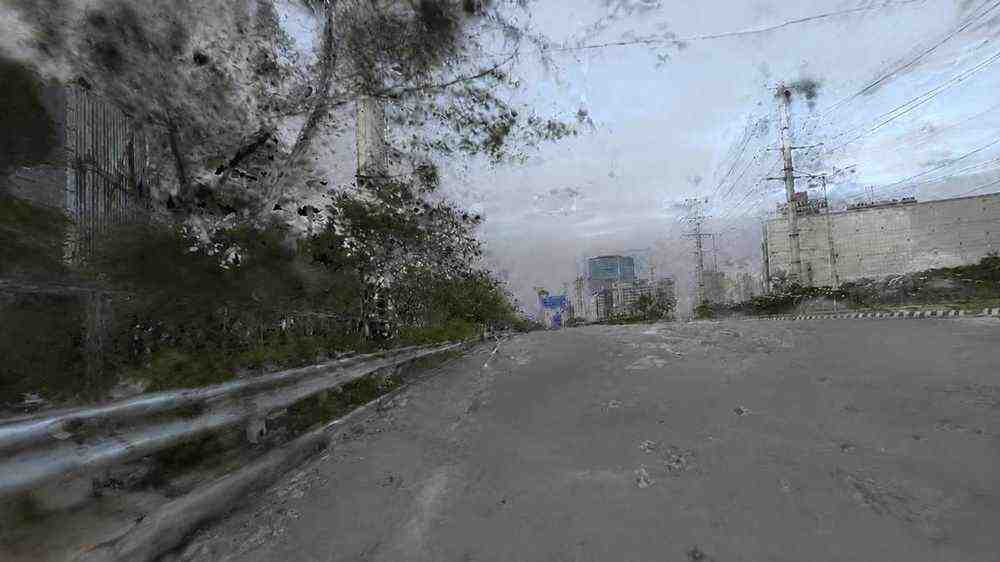} &
\includegraphics[width=0.19\linewidth]{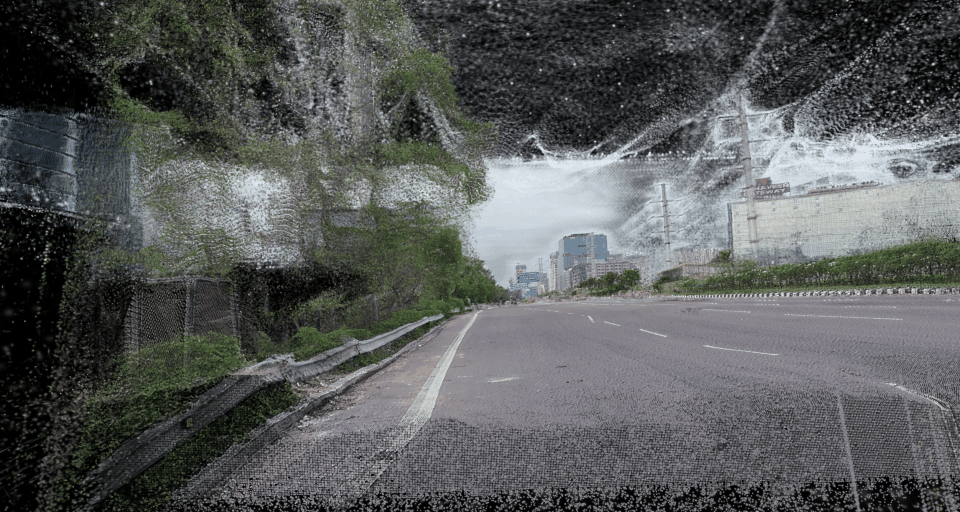} &
\includegraphics[width=0.19\linewidth]{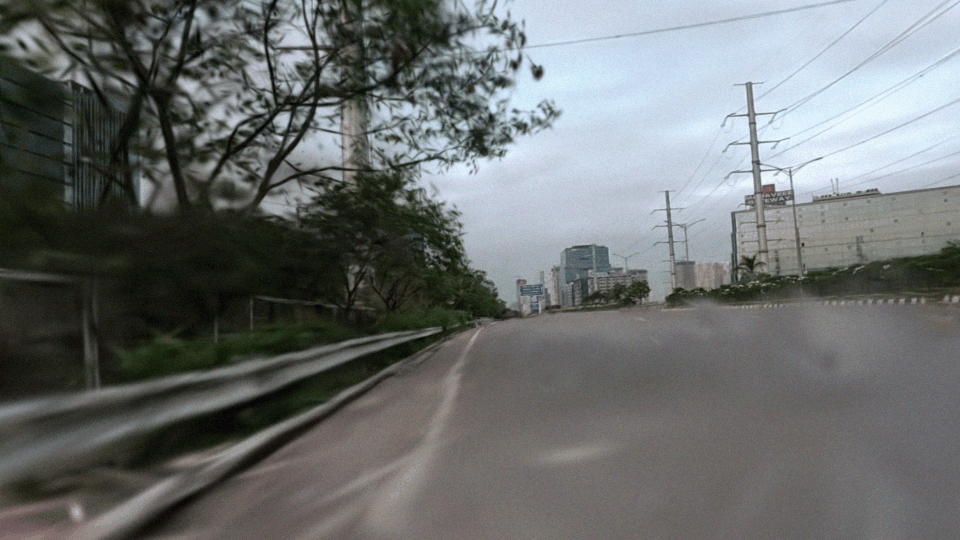} \\
& 13.21$|$0.44$|$0.52 & 13.08$|$0.49$|$0.47 & 13.87$|$0.52$|$0.44 & 16.61$|$0.53$|$0.47 \\

\includegraphics[width=0.19\linewidth]{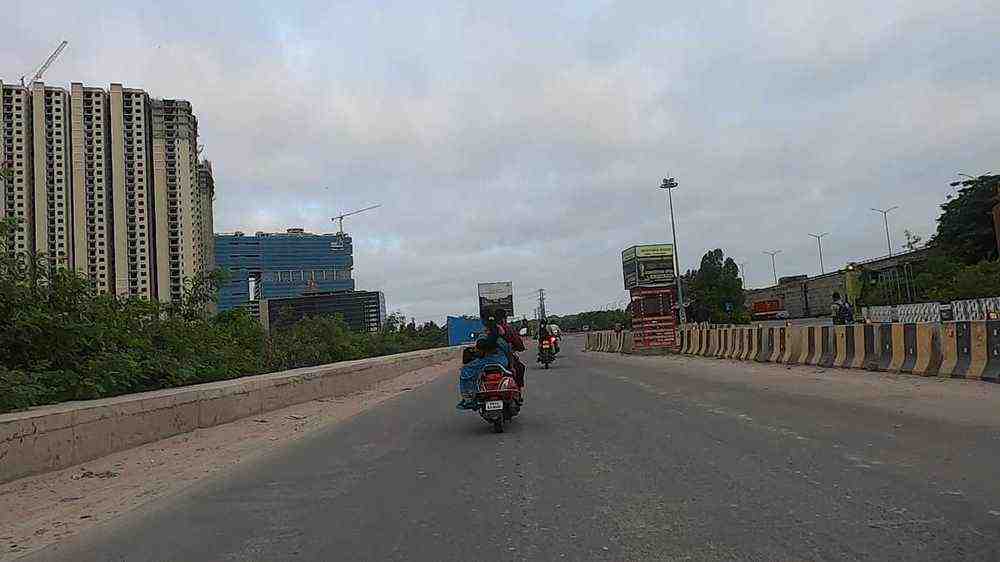} &
\includegraphics[width=0.19\linewidth]{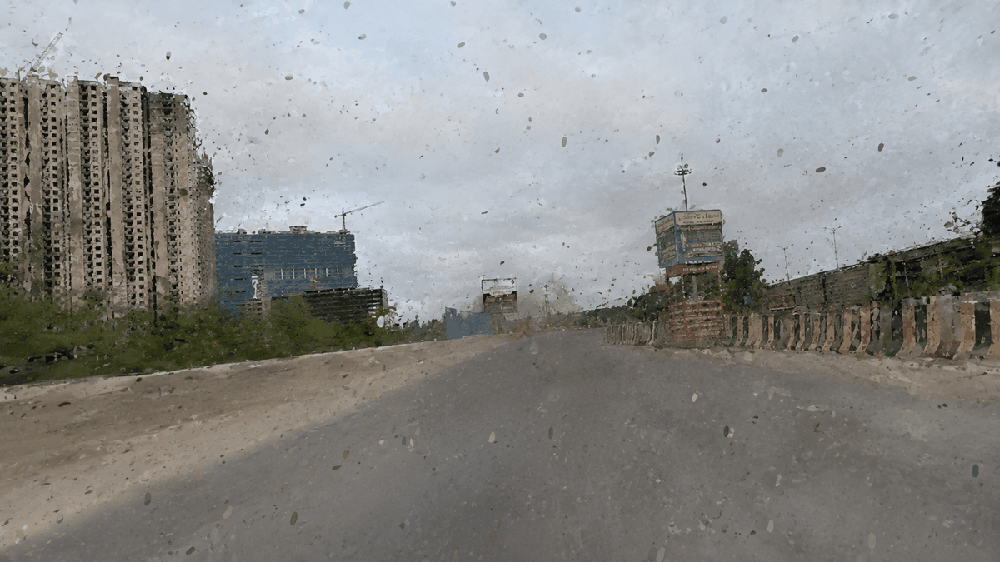} &
\includegraphics[width=0.19\linewidth]{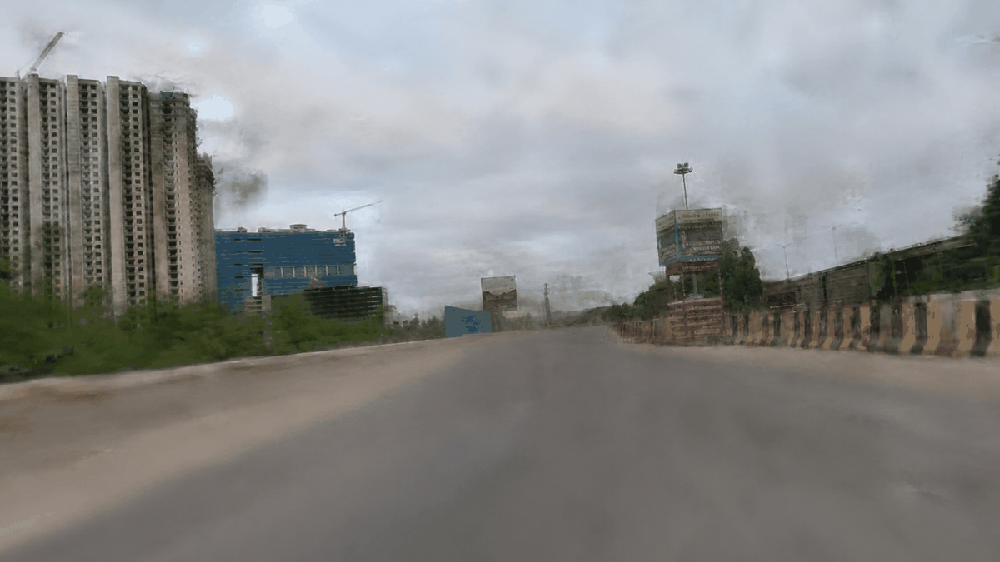} &
\includegraphics[width=0.19\linewidth]{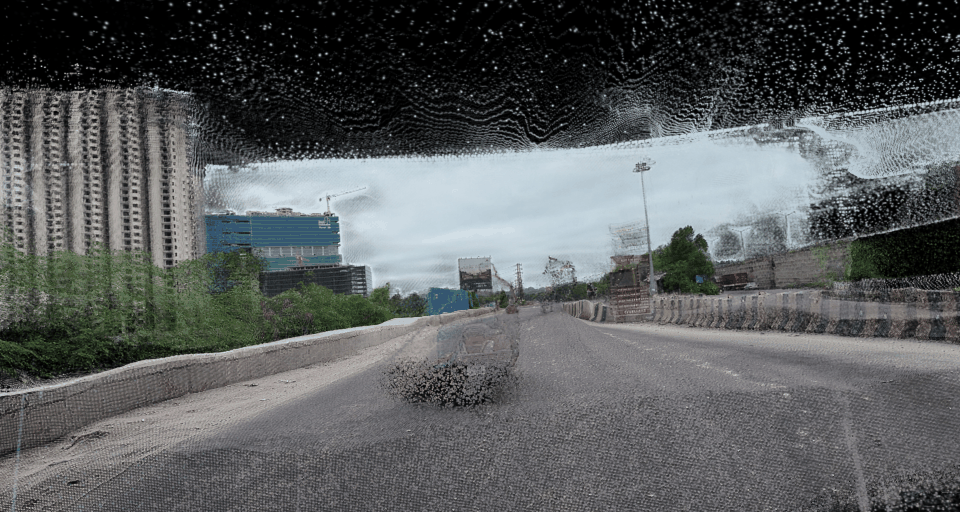} &
\includegraphics[width=0.19\linewidth]{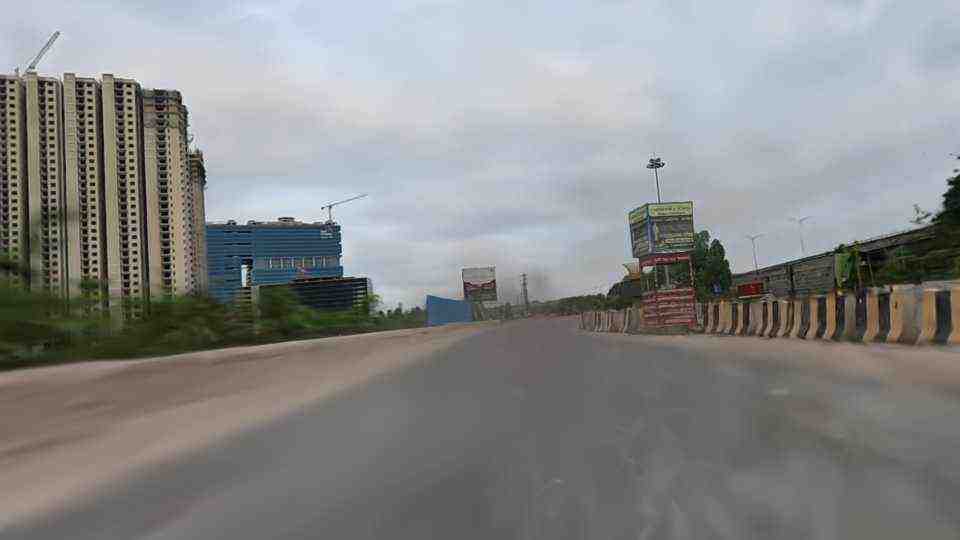} \\
& 13.42$|$0.51$|$0.39 & 15.23$|$0.54$|$0.30 & 14.3$|$0.46$|$0.41 & 16.98$|$0.50$|$0.35 \\

\end{tabular}
\end{adjustbox}

\caption{\textbf{Qualitative comparison of novel view synthesis across methods.}
Columns show GT, 3DGS, NeRF, DepthSplat, and PVG.
Rows correspond to rendering settings (top to bottom):
$T_{C\rightarrow C}$×2, $T_{C\rightarrow S}$×2, $T_{D\rightarrow D}$×2,
$T_{D\rightarrow C}$×2, and $T_{D\rightarrow S}$×2.
Metrics (PSNR$|$SSIM$|$LPIPS) are shown below each rendering.}
\label{fig:qualitative_results}

\end{figure*}
\vspace{-10pt}
\subsection{Dynamic Mask Ablation}

We evaluate the DS+SAM pseudo masks used for PVG dynamic-object supervision.
These masks are obtained by intersecting DroneSplat motion masks with
SAM-based vehicle and pedestrian masks; additional details of the dynamic-object
annotation and mask construction are provided in the Supplementary~\secref{suppsec:dynamic_objects}. Against GT
dynamic-object masks, DS+SAM pseudo masks achieve 63.69\% IoU, 70.93\%
Precision, and 86.19\% Recall.

We further evaluate PVG on the challenging $T_{C\rightarrow S}$ split under
three settings: no masks, DS+SAM pseudo masks, and GT dynamic masks.
Table~\ref{tab:mask_ablation} shows that DS+SAM pseudo masks substantially
improve over no masking and approach the GT-mask upper bound, validating
their use for dynamic-object supervision.


\begin{table}[t]
\centering
\small
\setlength{\tabcolsep}{4pt}
\renewcommand{\arraystretch}{1.1}
\begin{tabular}{lcccccc}
\hline
\multirow{2}{*}{Mask} &
\multicolumn{3}{c}{Dynamic Region} &
\multicolumn{3}{c}{Full Image} \\
\cline{2-7}
& PSNR $\uparrow$ & SSIM $\uparrow$ & LPIPS $\downarrow$
& PSNR $\uparrow$ & SSIM $\uparrow$ & LPIPS $\downarrow$ \\
\hline
None   & 13.98 & 0.32 & 0.62 & 14.72 & 0.34 & 0.63 \\
DS+SAM & 19.93 & 0.52 & 0.38 & 21.24 & 0.58 & 0.30 \\
GT     & 21.70 & 0.62 & 0.24 & 24.29 & 0.72 & 0.18 \\
\hline
\end{tabular}
\caption{PVG mask ablation on $T_{C\rightarrow S}$. DS+SAM denotes pseudo
dynamic masks obtained from the DroneSplat+SAM pipeline, while GT denotes
ground-truth dynamic-object masks.}
\label{tab:mask_ablation}
\end{table}


\vspace{-10pt}
\subsection{Camera Pose Estimation.}
\label{subsec:camera_pe}
We use the proposed benchmark to evaluate recent feed-forward camera pose estimation methods~\cite{wang2025vggt,mapanything,mast3r} alongside the classic optimization-based approach, COLMAP. Specifically, we assess how accurately these methods estimate test image poses. As described in \secref{subsec:pose_est}, COLMAP estimates test image poses through camera localization with respect to the 15 nearest training images. Similarly, each feed-forward method receives the test image and its 15 nearest training images as input to predict corresponding train and test camera poses.

For evaluation, we pair each test image with its nearest training image and compute the average epipolar error over RoMA~\cite{edstedt2025romav2harderbetter} pixel correspondences through the bounding box annotation (\cf \figref{fig:epi_anno}, \secref{subsec:pose_verify}). The distribution of average epipolar errors ($e_a$) for test sets $T_{C \rightarrow S}$ and $T_{D \rightarrow C}$ is shown in \figref{fig:epi_anno_cpe}. Feed-forward methods exhibit significantly higher epipolar errors, indicating less accurate pose estimation; larger epipolar errors correspond directly to larger pose errors. Detailed results are presented in \secref{suppsec:cpe} in the Supplementary.

Overall, the results reveal that (i) feed-forward pose estimation methods struggle under substantial viewpoint variation, and (ii) \MV also serves as a strong benchmark for evaluating camera pose estimation. While other optimization-based methods such as VIPE~\cite{huang2025vipe} exist, a comprehensive evaluation of such approaches is beyond the scope of this work and is left for future work.

\vspace{-10pt}




\subsection{Qualitative Analysis.}
\label{subsec:qualt}
We present qualitative examples of rendered images by different NVS methods
in Fig.~\ref{fig:qualitative_results} along with their respective NVS performance metrics.
For the standard $T_{C\rightarrow C}$ setting, PVG achieves the best visual quality
and metrics among the evaluated methods, with improved reconstruction of both
static and dynamic scene elements. However, rendering quality drops substantially
in cross-vehicle and aerial-to-ground setups, including $T_{C\rightarrow S}$,
$T_{D\rightarrow C}$, and $T_{D\rightarrow S}$. These qualitative results further
support our main observation that large viewpoint changes across vehicles remain
challenging for current NVS methods. More qualitative examples are in \secref{suppsec:qualitative_figs} in the Supplementary.



\section{Conclusions}
\vspace{-5pt}
We introduce \MV, a challenging NVS dataset and benchmark for driving scenes with viewpoints more diverse than existing ground and aerial driving datasets. Our results show that driving-scene NVS remains unsolved and needs major improvements for applications such as simulation. While state-of-the-art dynamic NVS methods perform well in standard settings, training-free baselines often match them in challenging cross-vehicle setups. We also demonstrate \MV's utility for camera pose estimation, showing that feed-forward state-of-the-art methods struggle under large viewpoint variations.

\section{Acknowledgements}
\vspace{-5pt}
This work was supported by the Ministry of Electronics and Information Technology (MeitY), Government of India, through the NLTM-Bhashini project, and by the Anusandhan National Research Foundation (ANRF) under Grant ANRF/ECRG/2025/005736/ENS. AWS resources were provided by the National Infrastructures for Research and Technology (GRNET) and funded by the EU Recovery and Resilience Facility.

We thank Azhar, Ram Sharma, Mahender Reddy, Pranav Varudkar, and the members of the CVIT Annotation Lab for their technical support and assistance with data collection and annotation.




%
%

\bibliographystyle{splncs04}
\bibliography{main}

\clearpage
\setcounter{page}{1}

\begin{center}
{\LARGE \textbf{MV$^2$: Multi-View Multi-Vehicle Driving Dataset for Novel View Synthesis}}\\
\vspace{0.2cm}
{\Large \textbf{Supplementary Material}}
\end{center}

\section{Dataset and Statistics}
\label{suppsec:dataset_stats}
This section provides a detailed overview of the proposed \MV dataset, highlighting its visual diversity, sensor configurations, and statistical characteristics. Our goal is to illustrate the scale, complexity, and multi-vehicle multi-view nature of the data, which together enable challenging cross-view reconstruction and novel view synthesis tasks.

\subsection{Statistical Analysis}







\begin{figure*}
    \centering
    \includegraphics[width=\linewidth]{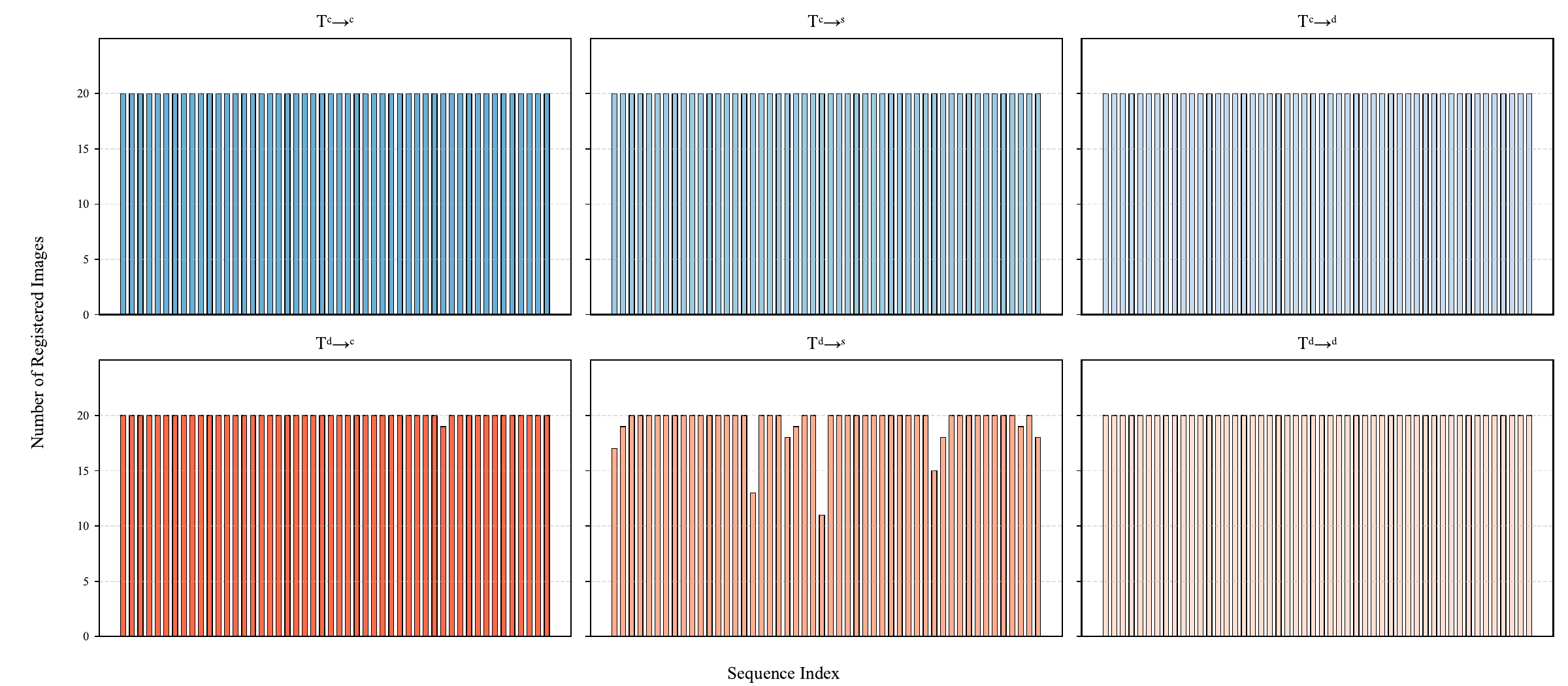}
    \caption{\textbf{Registered test images per sensor pair.} 
$V^C$ mappings ($T_{C\rightarrow x}$) register all frames, while $V^D$ cross-sensor mappings ($T_{D\rightarrow y}$, $y\neq D$) show slight drops due to larger viewpoint gaps.}
    \label{fig:test_images_registered}
\end{figure*}
As shown in ~\figref{fig:test_images_registered}, we visualize per-sequence registration statistics across all six sensor-pair transformations: $T_{C\rightarrow C}$, $T_{C\rightarrow S}$, $T_{C\rightarrow D}$, $T_{D\rightarrow C}$, $T_{D\rightarrow S}$, and $T_{D\rightarrow D}$. The top row corresponds to the $V^{c}$, where all mappings $T_{C\rightarrow x}$ register \emph{every} image in every sequence.

\begin{figure*}[ht!]
\centering
\begin{tabular}{ccc}
\includegraphics[width=0.30\textwidth]{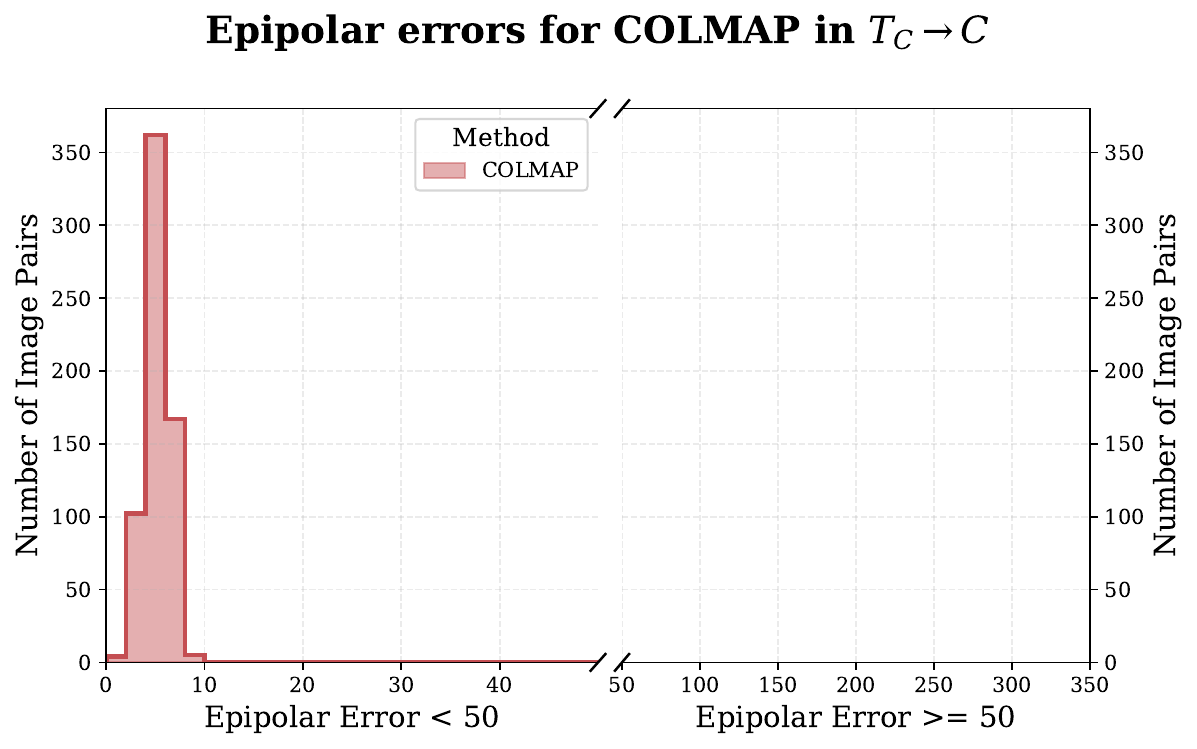} &
\includegraphics[width=0.30\textwidth]{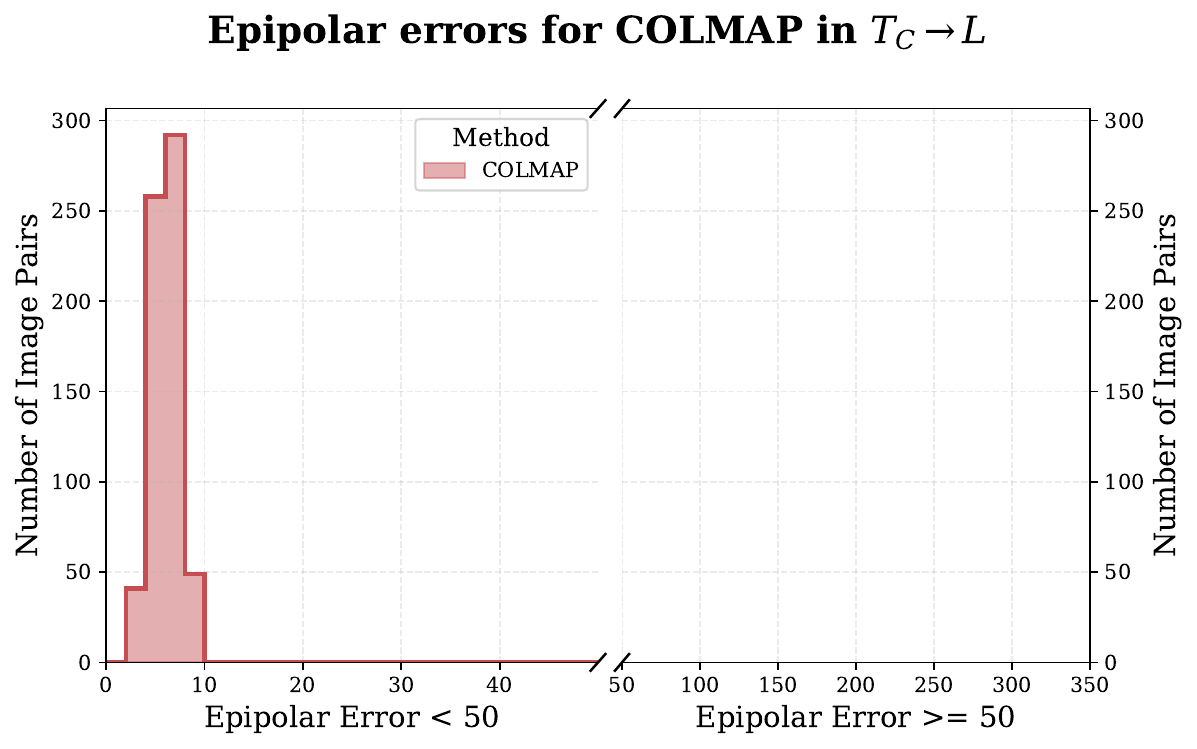} &
\includegraphics[width=0.30\textwidth]{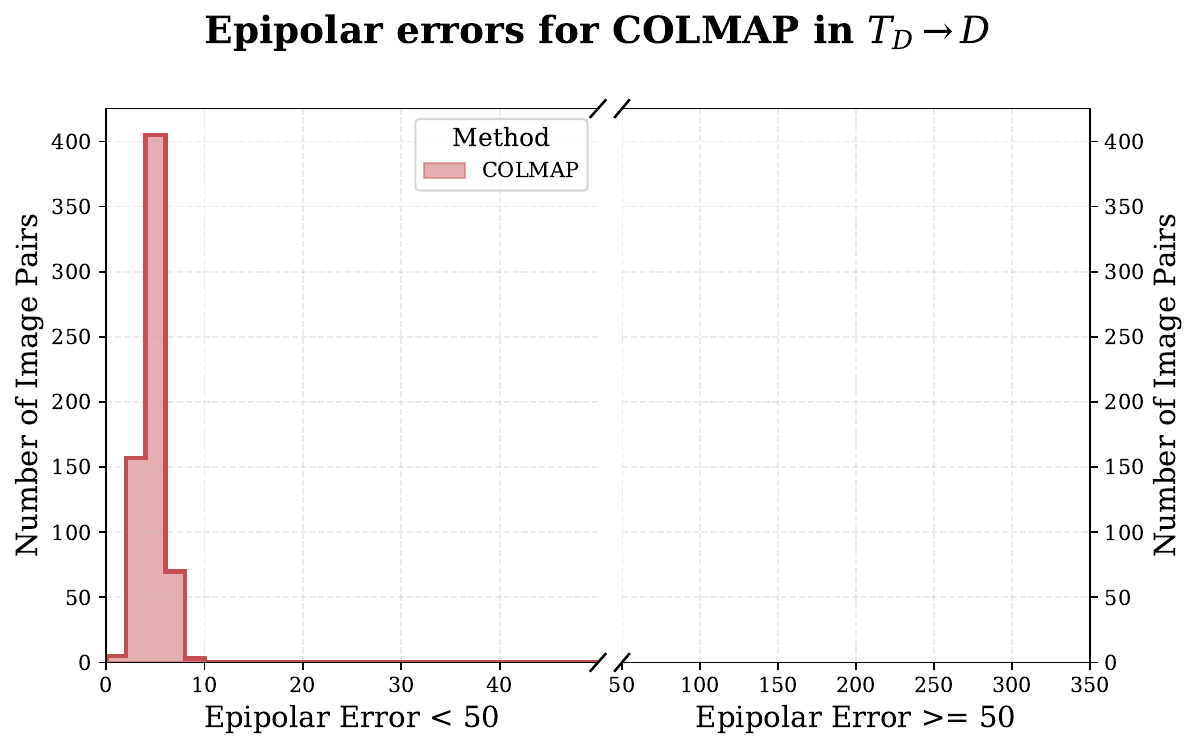} \\
(a) $T_{C \rightarrow C}$ & 
(b) $T_{C \rightarrow L}$ & 
(c) $T_{D \rightarrow D}$
\end{tabular}
\caption{\textbf{Epipolar error distributions across three key sensor-pair transformations using COLMAP.}
(a) For $T_{C \rightarrow C}$, epipolar errors are extremely low and tightly concentrated, reflecting stable same-sensor relative pose estimation.  
(b) For $T_{C \rightarrow L}$, the front→left-forward viewpoint change introduces mild geometric variation, yet COLMAP maintains low-error performance with only slight tail spread.  
(c) For $T_{D \rightarrow D}$, drone-only sequences show consistently low errors across aerial viewpoints, highlighting robust pose estimation in wide-baseline drone imagery.}
\label{fig:supply_cam_pose_three_panel}
\vspace{-5pt}
\end{figure*}

In contrast, the bottom row shows $V^{D}$. While $T_{D\rightarrow D}$ remains dense, the cross-sensor transformations $T_{D\rightarrow y}$ with $y \neq d$ exhibit modest reductions in frame availability across a few sequences. The mean epipolar error ($e_a$) for $T_{C \rightarrow C}, T_{C \rightarrow L}, T_{D \rightarrow D}$ are presented in \figref{fig:supply_cam_pose_three_panel}.

\subsection{Additional Viewpoints}

\setlength{\fboxrule}{1.5pt}
\begin{figure*}[ht!]
\centering
\begin{tabular}{ccccc}
  \textbf{Scene 1} & \textbf{Scene 2} & \textbf{Scene 3} & \textbf{Scene 4} \\

\fcolorbox{blue}{white}{\includegraphics[width=0.20\textwidth]{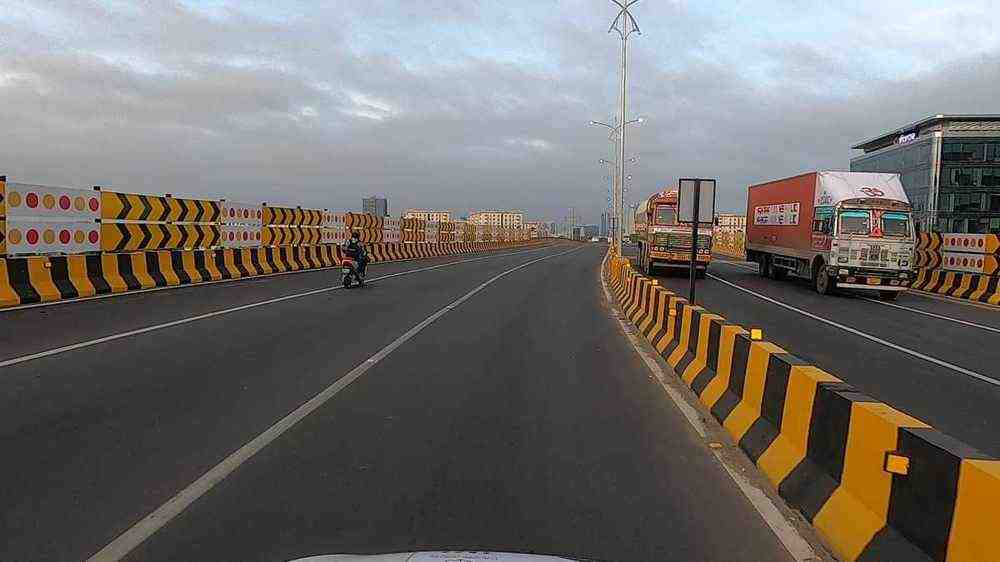}} &
\fcolorbox{blue}{white}{\includegraphics[width=0.20\textwidth]{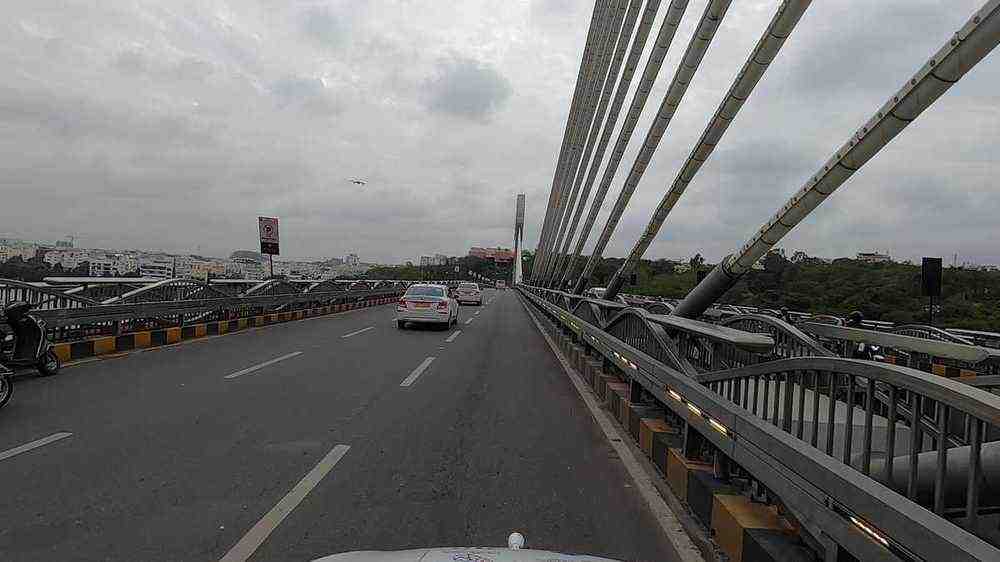}} &
\fcolorbox{blue}{white}{\includegraphics[width=0.20\textwidth]{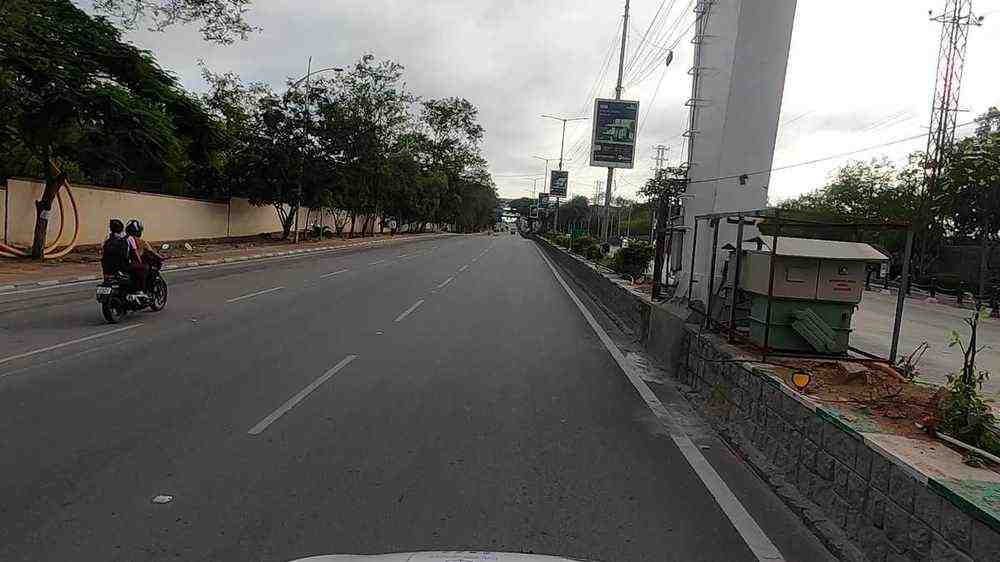}} &
\fcolorbox{blue}{white}{\includegraphics[width=0.20\textwidth]{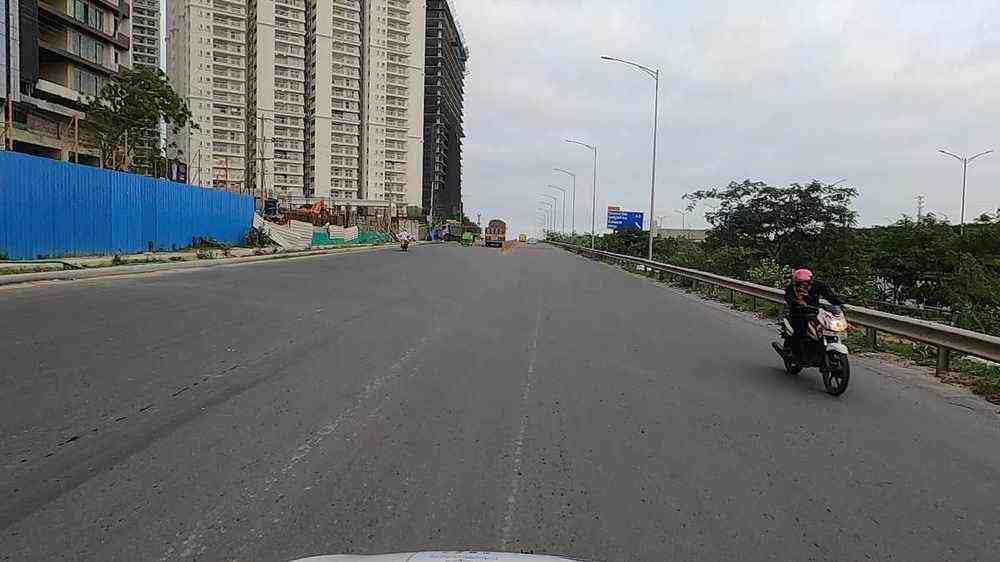}} \\[3pt]

\fcolorbox{green}{white}{\includegraphics[width=0.20\textwidth]{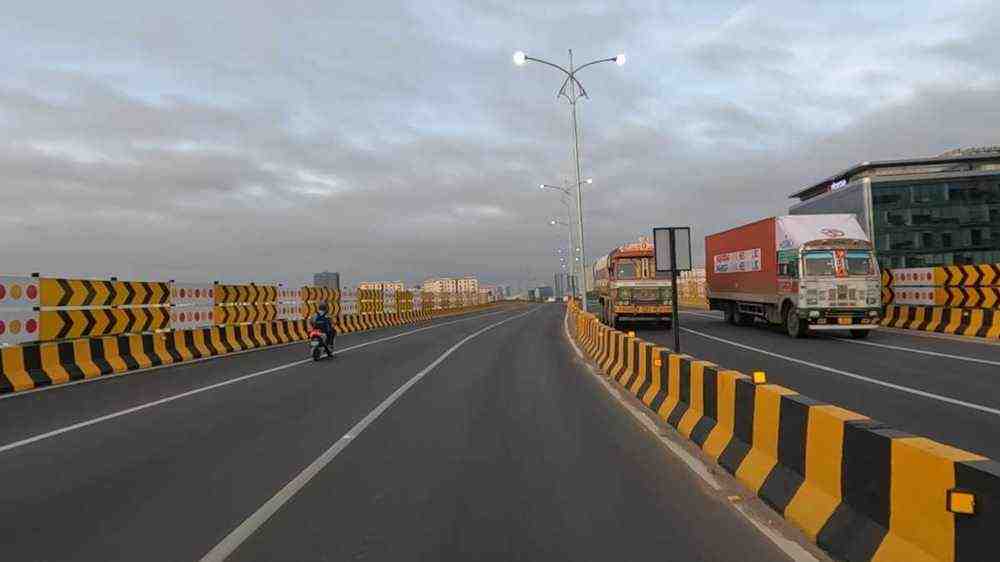}} &
\fcolorbox{green}{white}{\includegraphics[width=0.20\textwidth]{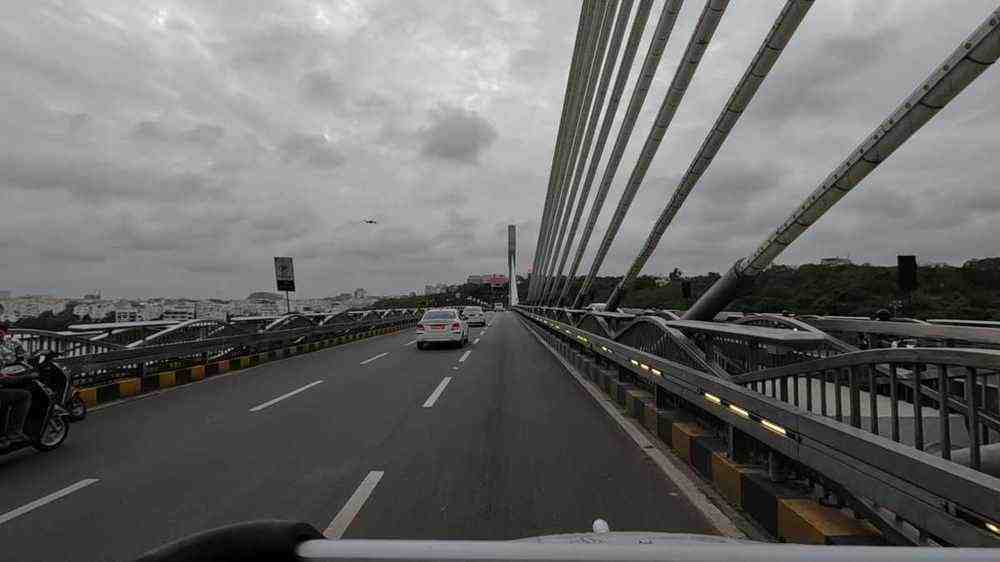}} &
\fcolorbox{green}{white}{\includegraphics[width=0.20\textwidth]{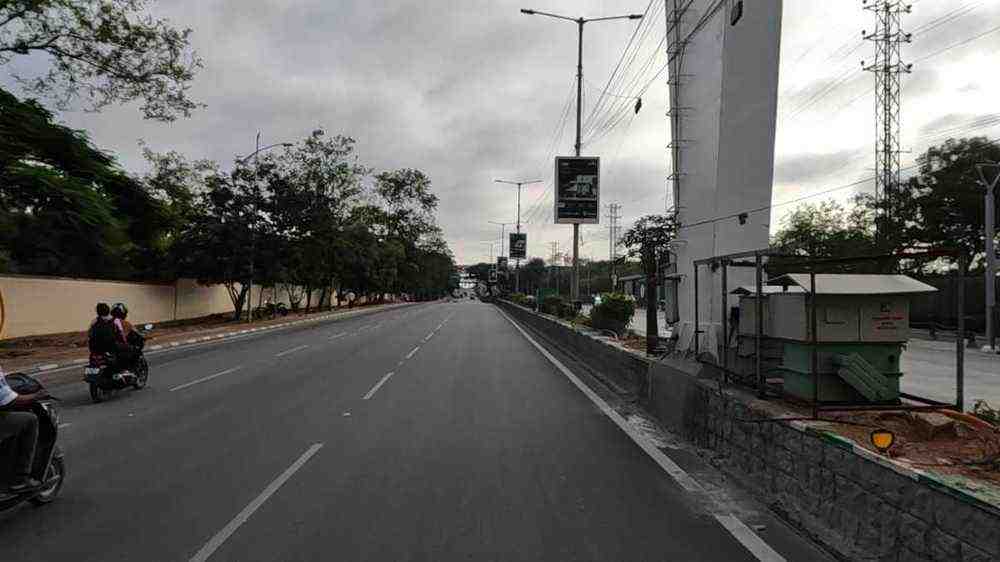}} &
\fcolorbox{green}{white}{\includegraphics[width=0.20\textwidth]{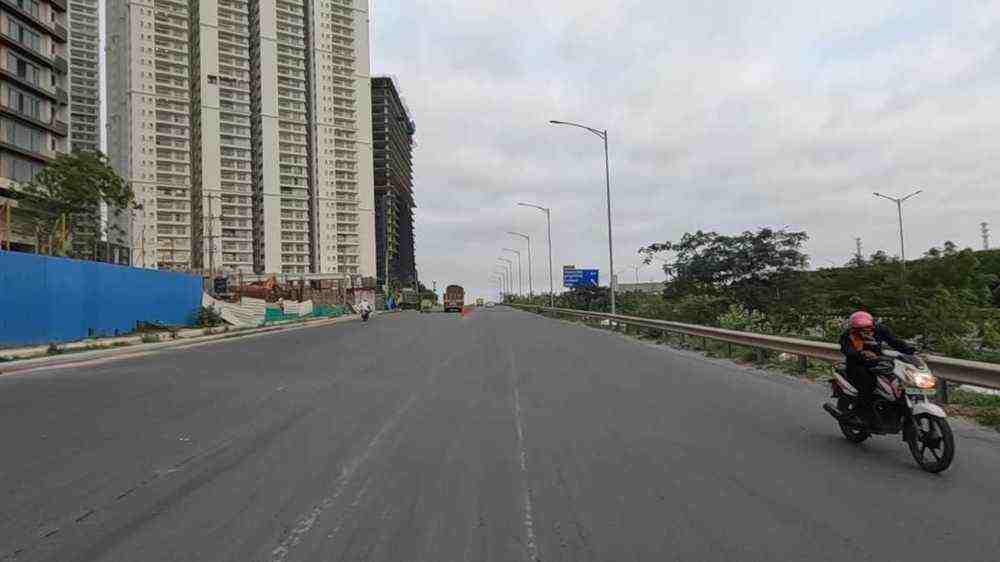}} \\[3pt]

\fcolorbox{brown}{white}{\includegraphics[width=0.20\textwidth]{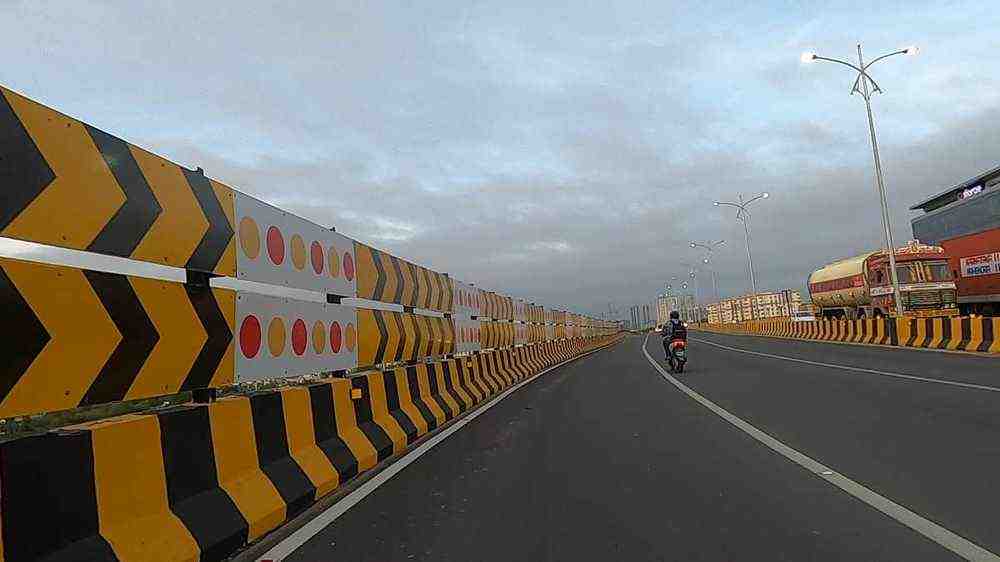}} &
\fcolorbox{brown}{white}{\includegraphics[width=0.20\textwidth]{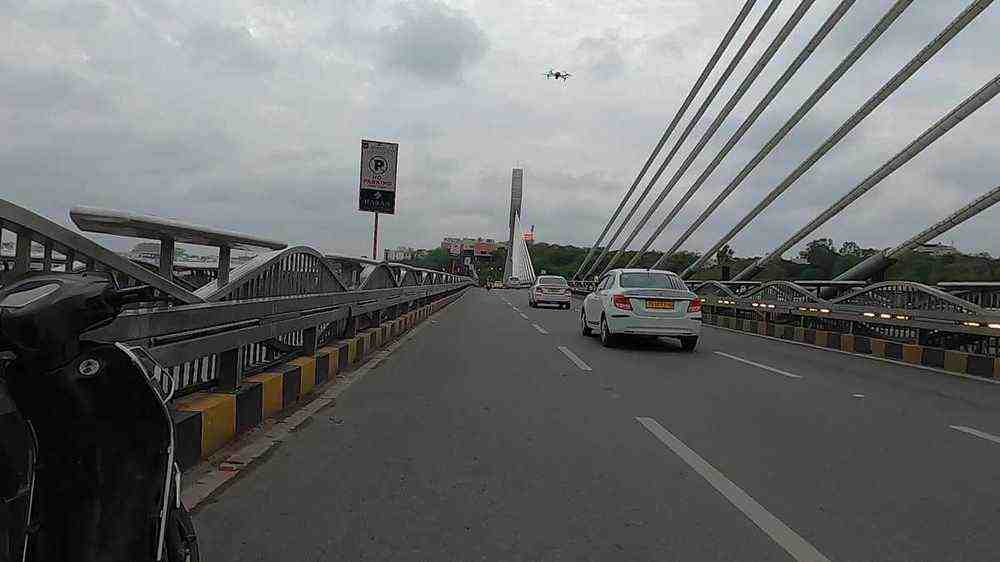}} &
\fcolorbox{brown}{white}{\includegraphics[width=0.20\textwidth]{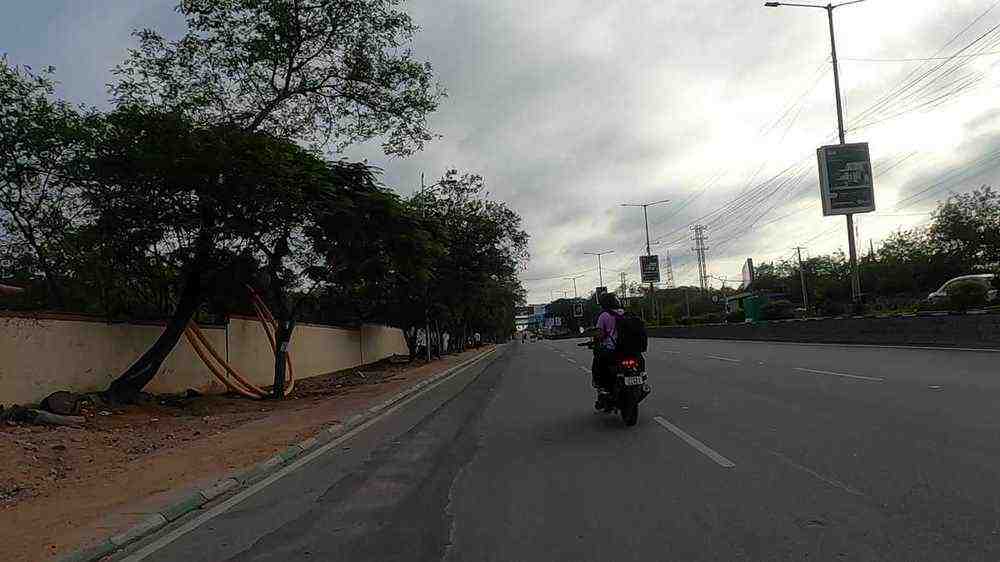}} &
\fcolorbox{brown}{white}{\includegraphics[width=0.20\textwidth]{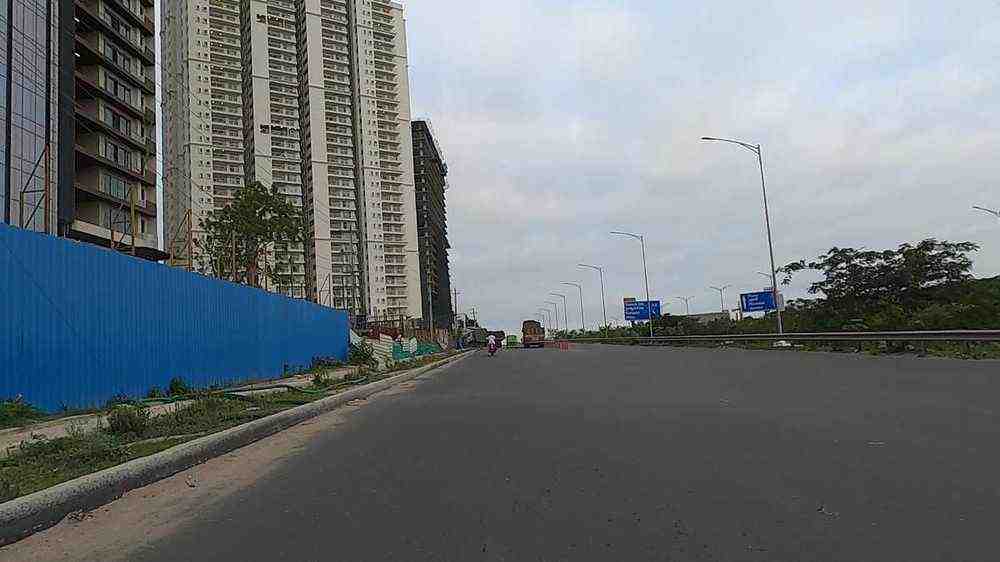}} \\[3pt]

\fcolorbox{violet}{white}{\includegraphics[width=0.20\textwidth]{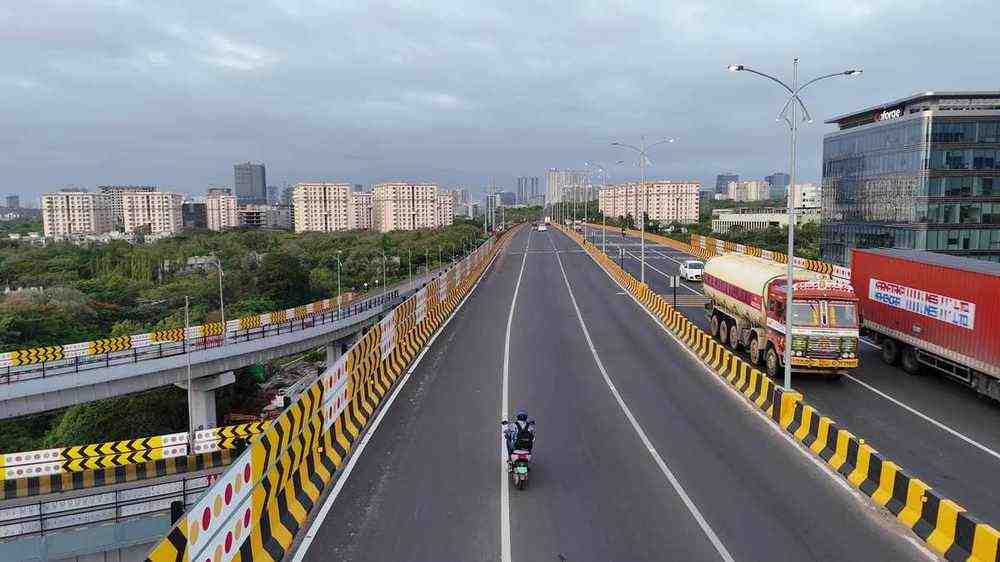}} &
\fcolorbox{violet}{white}{\includegraphics[width=0.20\textwidth]{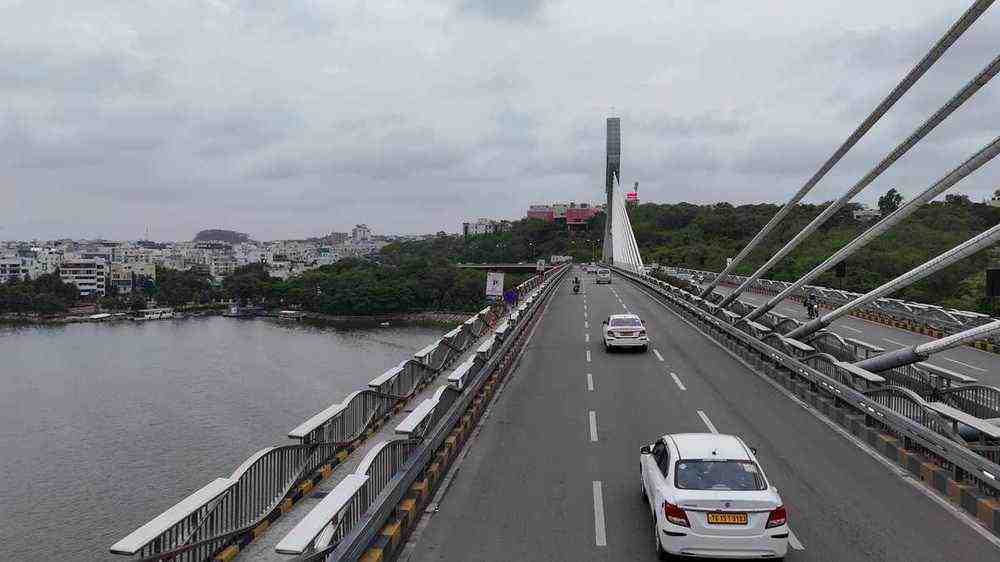}} &
\fcolorbox{violet}{white}{\includegraphics[width=0.20\textwidth]{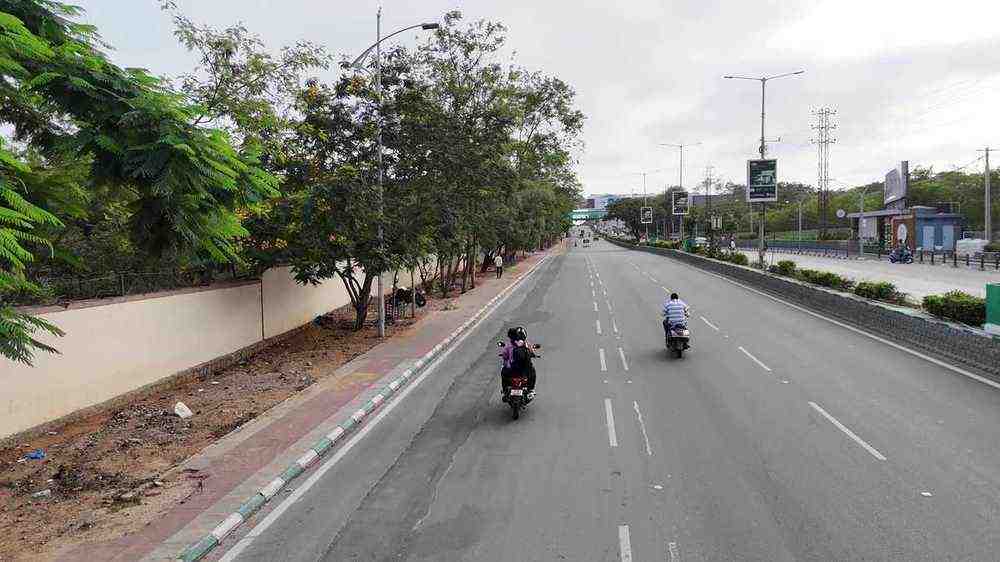}} &
\fcolorbox{violet}{white}{\includegraphics[width=0.20\textwidth]{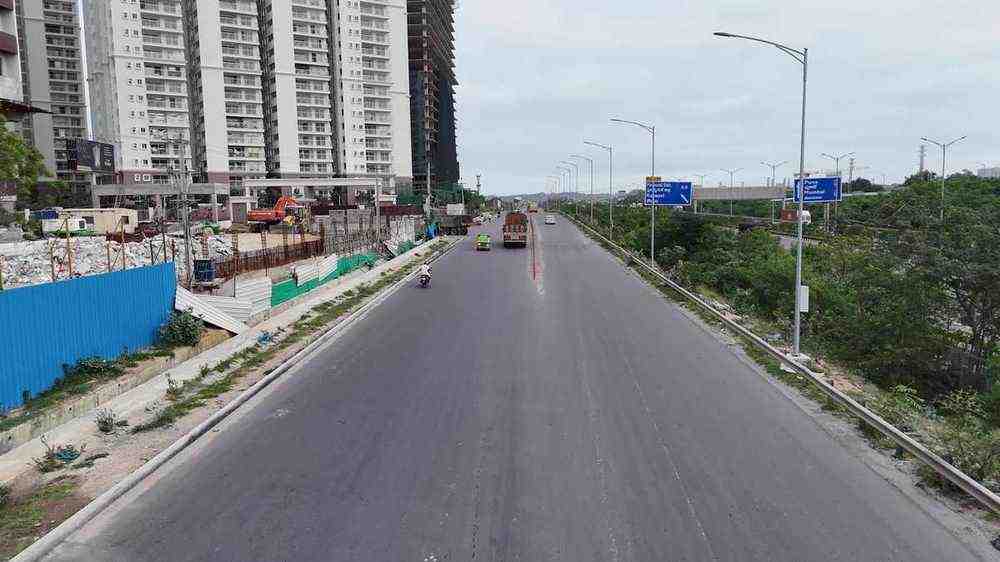}} \\[3pt]

\end{tabular}

\caption{Sample multi-view scenes captured across diverse environments which are used to calculate metrics. Each column represents a distinct scene, while each row corresponds to a different acquisition viewpoint \textcolor{blue}{$V^C$}, \textcolor{green}{$V^L$},\textcolor{violet}{$V^D$}, and \textcolor{brown}{$V^S$}. The colored borders highlight the viewpoint source of each image, helping visualize cross-view consistency across sensors.}
\label{fig:supply_sample_images}
\end{figure*}

\setlength{\fboxrule}{1.5pt}
\begin{figure*}[ht!]
\centering
\begin{tabular}{ccccc}
  \textbf{Scene 1} & \textbf{Scene 2} & \textbf{Scene 3} & \textbf{Scene 4} \\

\includegraphics[width=0.20\textwidth]{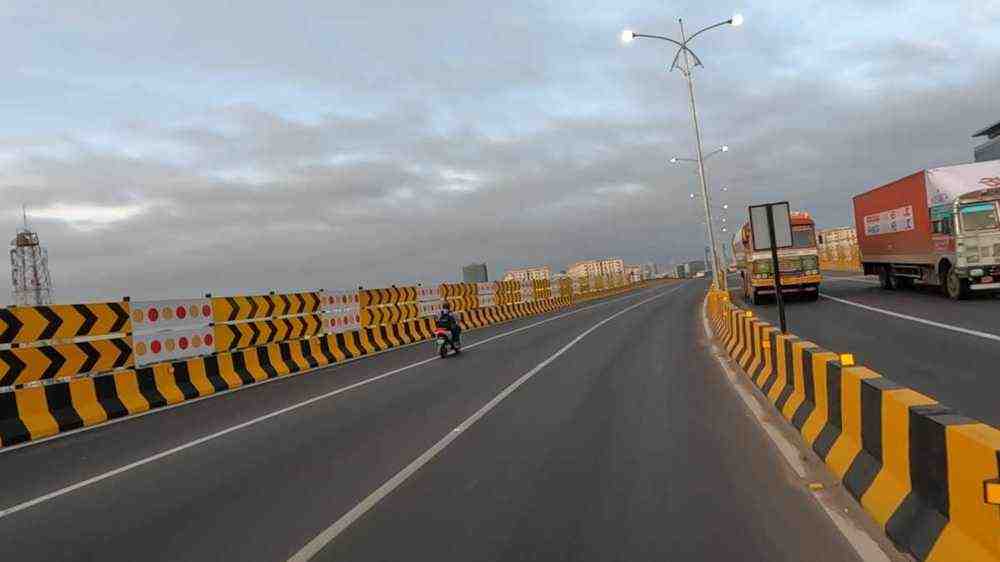} &
\includegraphics[width=0.20\textwidth]{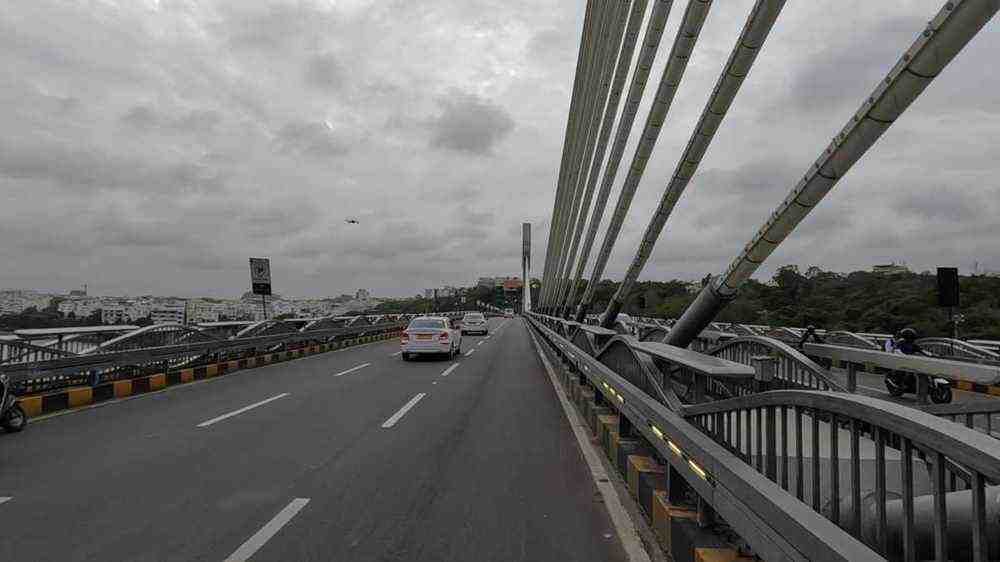} &
\includegraphics[width=0.20\textwidth]{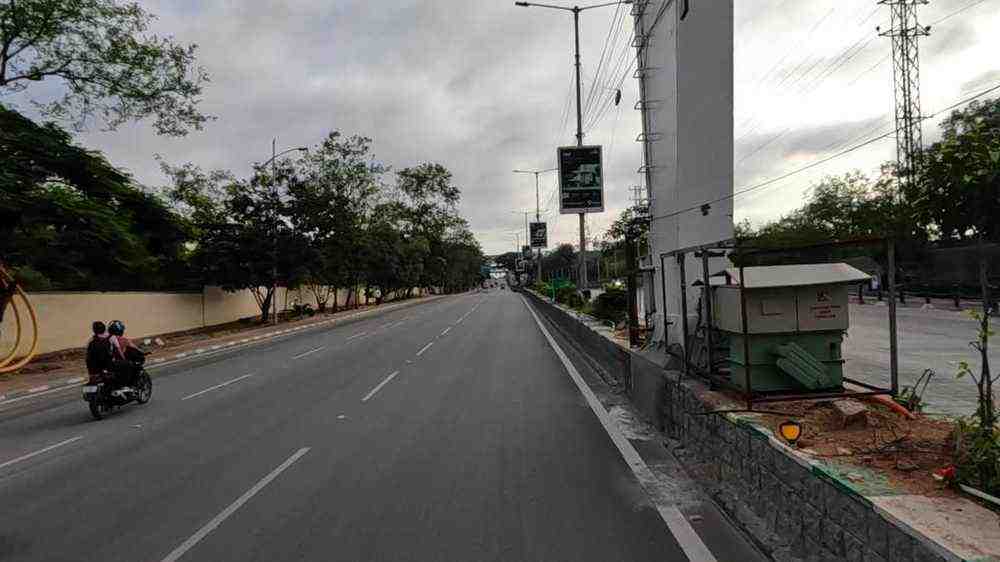} &
\includegraphics[width=0.20\textwidth]{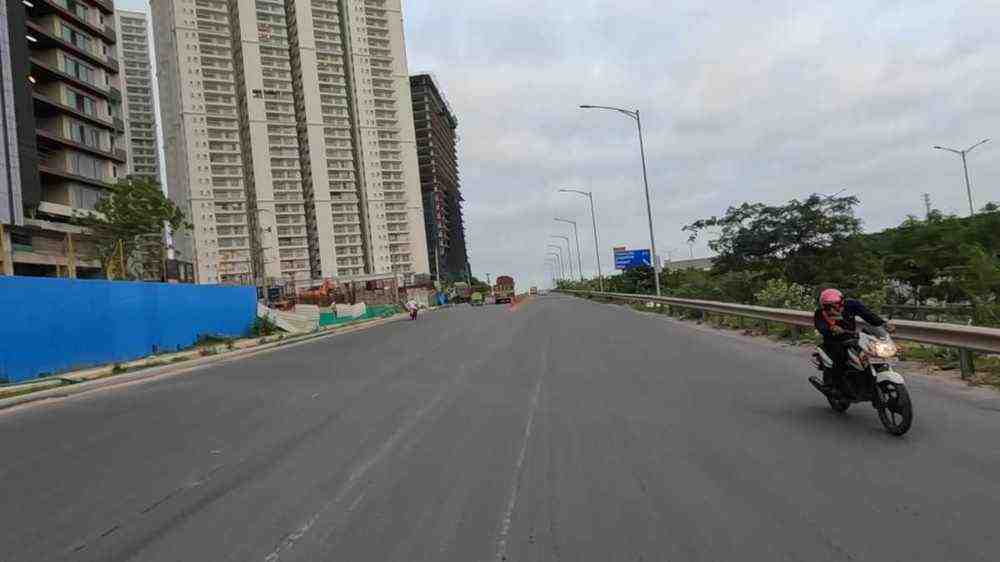} \\[3pt]

\includegraphics[width=0.20\textwidth]{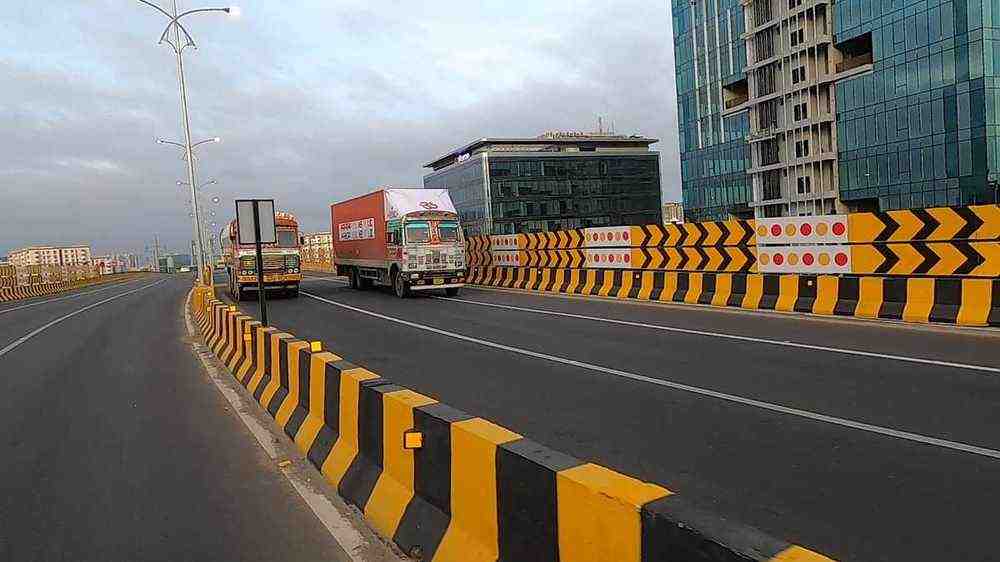} &
\includegraphics[width=0.20\textwidth]{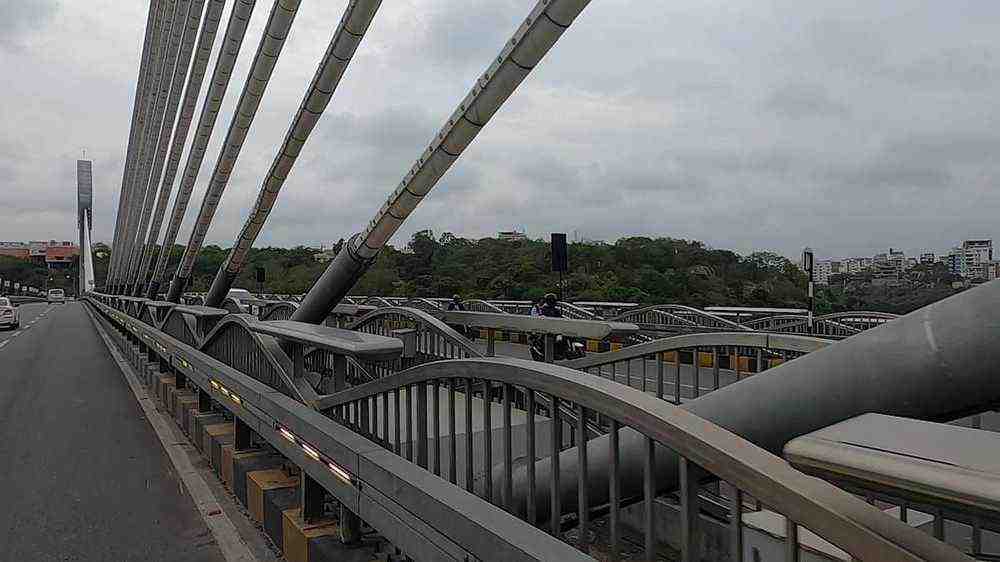} &
\includegraphics[width=0.20\textwidth]{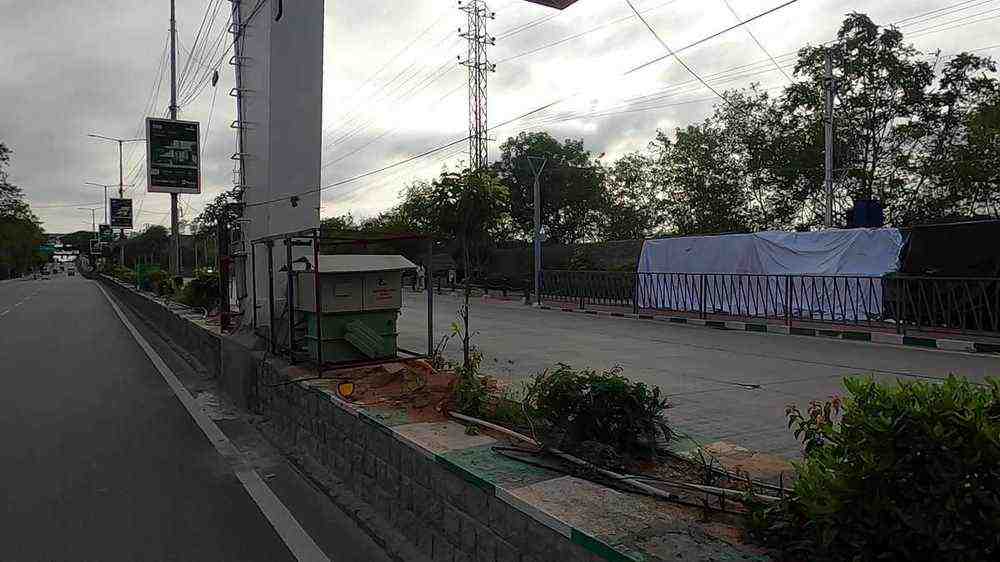} &
\includegraphics[width=0.20\textwidth]{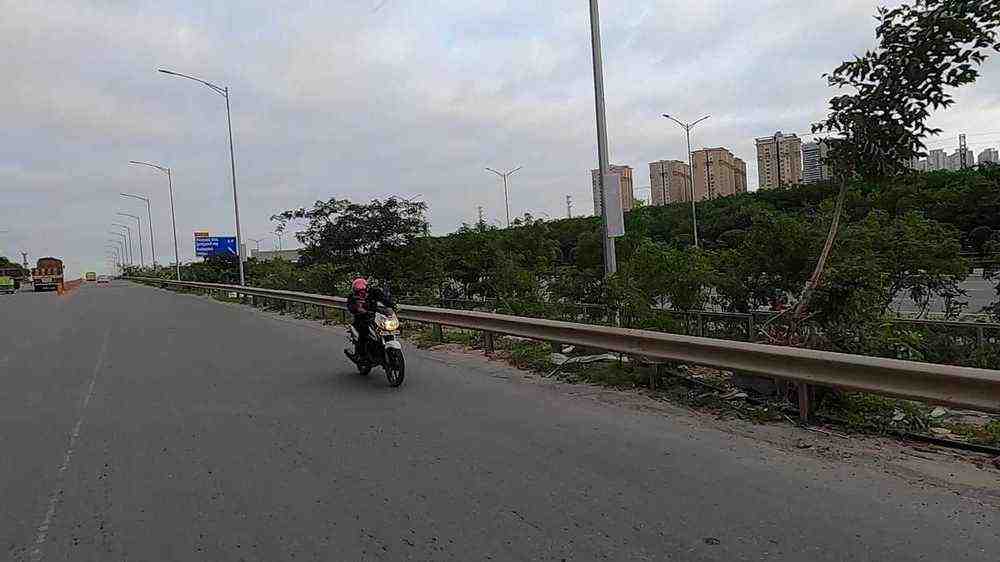} \\[3pt]

\includegraphics[width=0.20\textwidth]{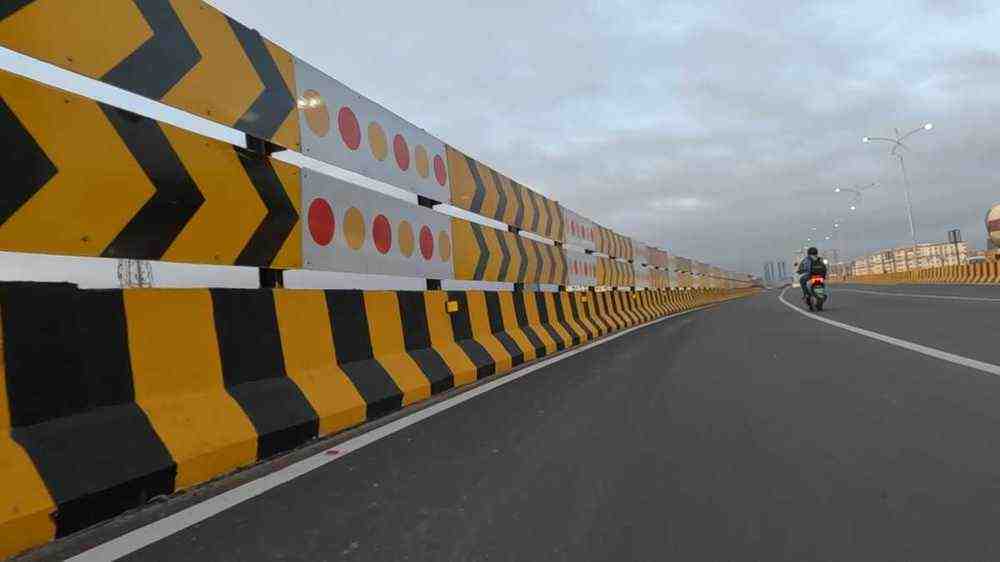} &
\includegraphics[width=0.20\textwidth]{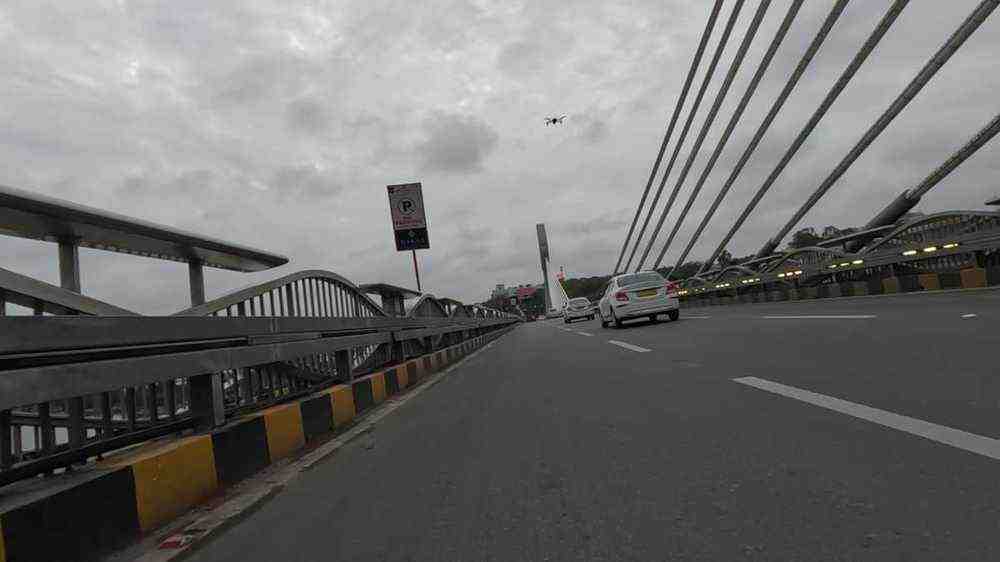} &
\includegraphics[width=0.20\textwidth]{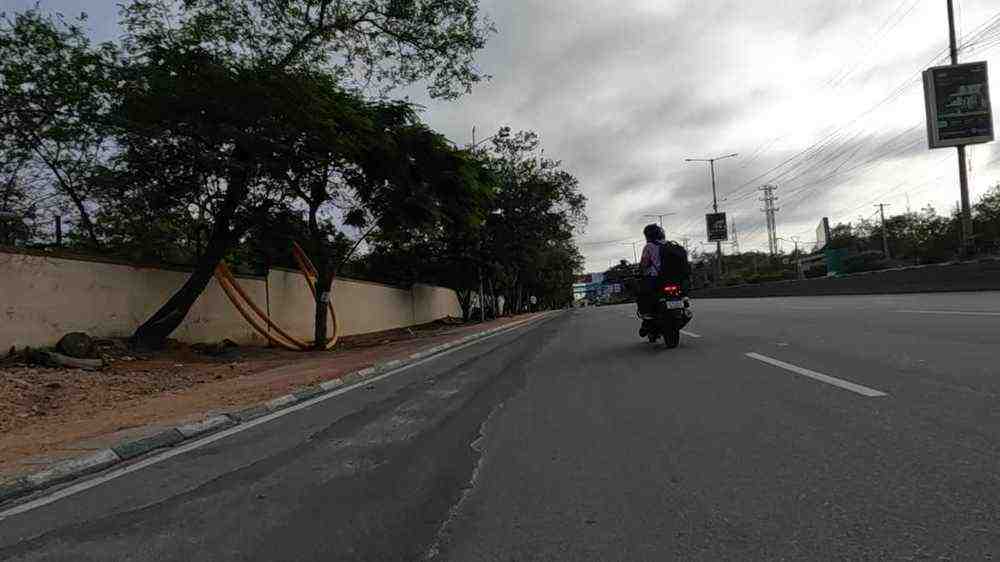} &
\includegraphics[width=0.20\textwidth]{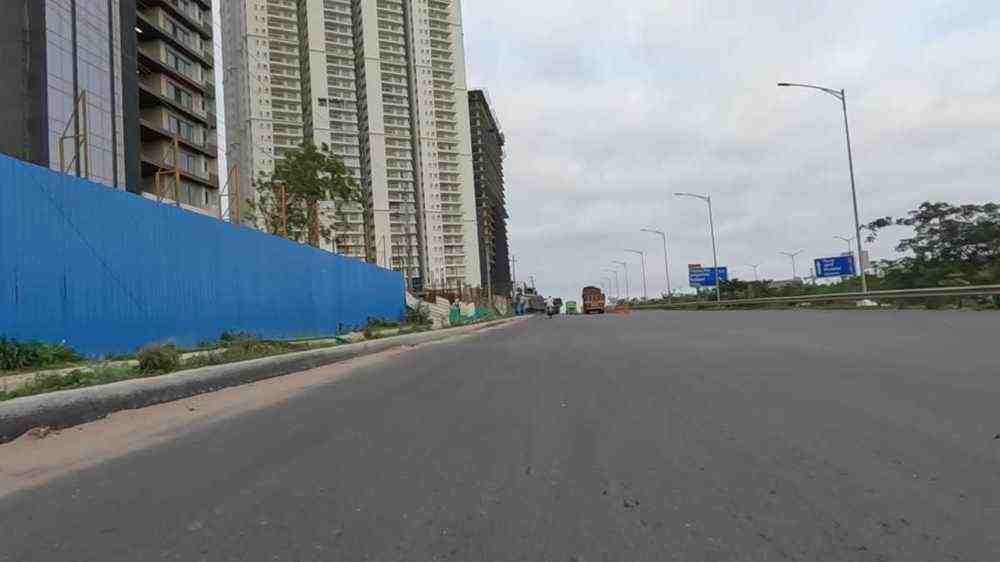} \\[3pt]

\includegraphics[width=0.20\textwidth]{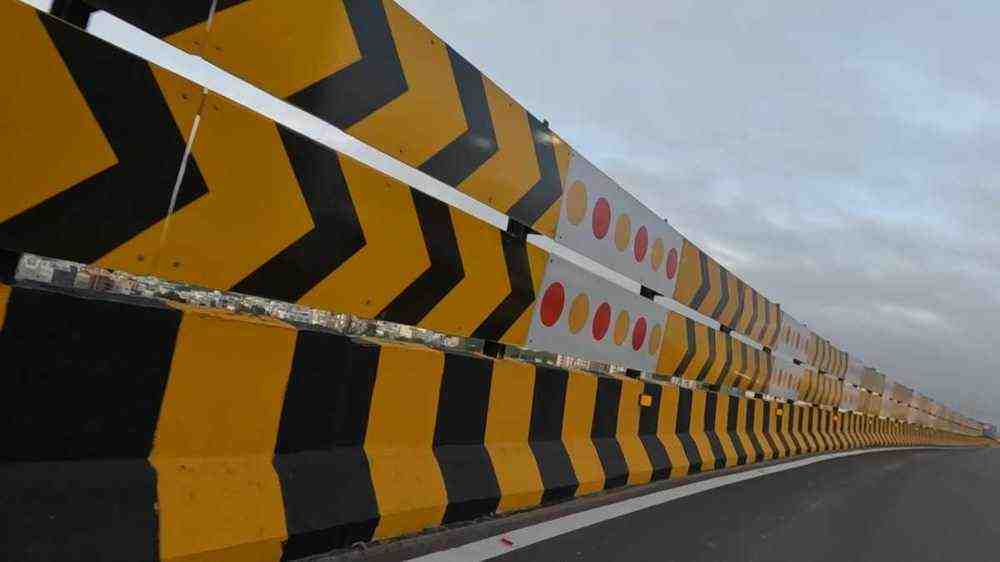} &
\includegraphics[width=0.20\textwidth]{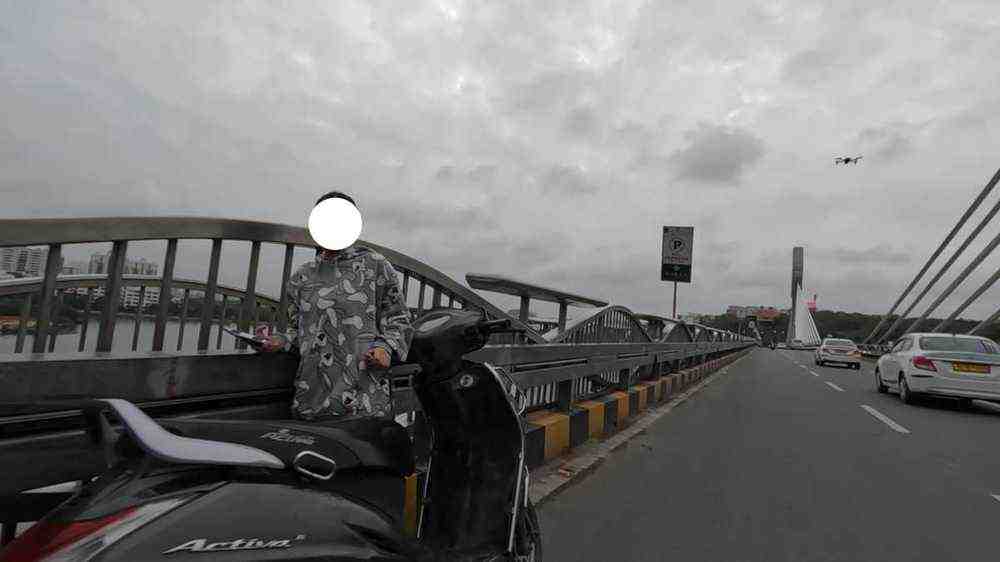} &
\includegraphics[width=0.20\textwidth]{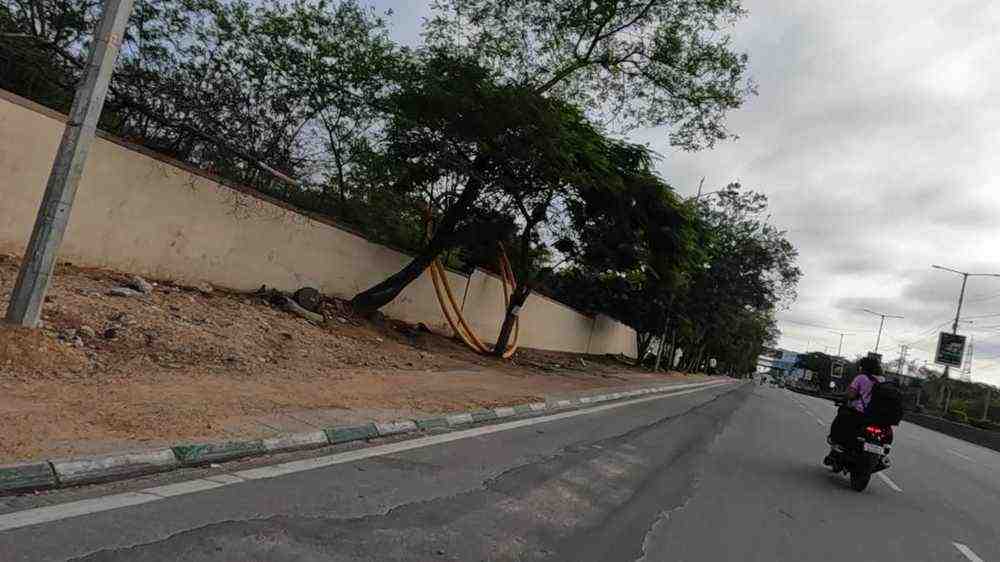} &
\includegraphics[width=0.20\textwidth]{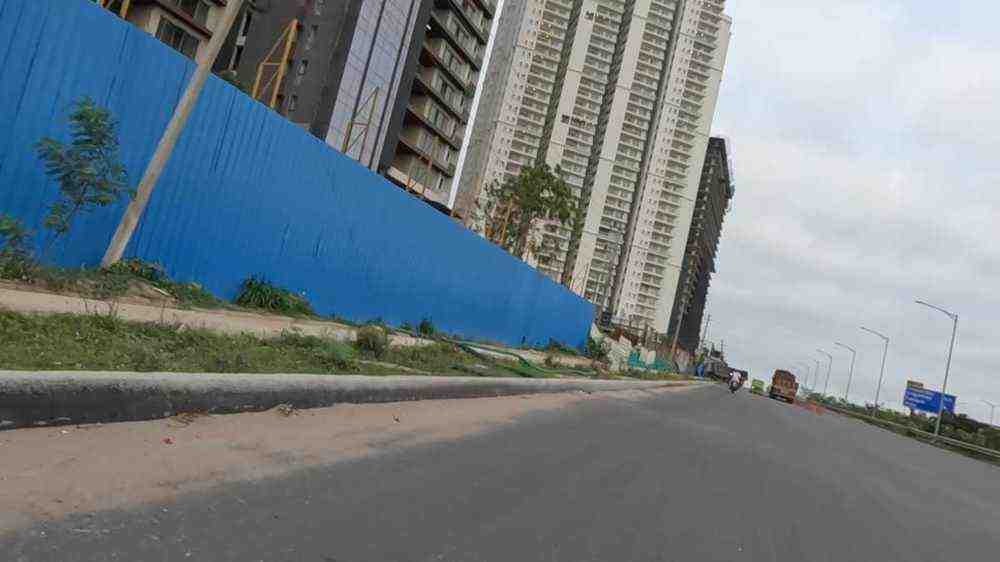} \\[3pt]

\end{tabular}

\caption{Sample images from the auxiliary sensors in our dataset. Each column corresponds to a distinct scene, while each row shows images captured from a specific auxiliary viewpoint: $\mathbf{V}^{CR}$, $\mathbf{V}^{CSR}$, $\mathbf{V}^{SSL}$, and $\mathbf{V}^{SB}$. These additional viewpoints provide lateral, stereo-offset, and near-ground perspectives, enriching the spatial and geometric diversity of the dataset and complementing the primary forward-facing sensors.}
\label{fig:supply_sample_images_addit}
\end{figure*}

~\figref{fig:supply_sample_images} shows examples from the primary sensors presented in the main paper, while ~\figref{fig:supply_sample_images_addit} illustrates images captured from the additional auxiliary sensors.

In addition to the primary forward-facing sensors $\mathbf{V}^{L}$, $\mathbf{V}^{C}$, $\mathbf{V}^{S}$, and $\mathbf{V}^{D}$ described in the main paper, our dataset also includes images captured from several auxiliary cameras mounted on both the car and the scooty. On the car, two additional GoPro~10 sensors are installed on the right side of the vehicle. The car-right camera, denoted as $\mathbf{V}^{CR}$, is mounted parallel to the vehicle’s forward direction, providing a lateral right-view perspective. The car stereo-right camera, denoted as $\mathbf{V}^{CSR}$, is positioned approximately $45^\circ$ offset from the primary car-left camera, forming a wider stereo baseline and capturing a slightly angled right-forward viewpoint.

Similarly, the scooty is equipped with two auxiliary sensors. The scooty stereo-left camera, denoted as $\mathbf{V}^{SSL}$, is mounted on the scooty’s handlebar, capturing a left-biased stereo viewpoint that complements the primary scooty sensor. The scooty-bottom camera, denoted as $\mathbf{V}^{SB}$, is mounted on the scooty’s front mudguard (above the wheel), providing a low-elevation, downward-facing perspective that captures ground-level details and immediate foreground structure.

All auxiliary sensors follow the same capture specifications as the primary ones, recording at a resolution of $1080 \times 1980$ at 60~FPS and precisely synchronized using the wall-clock. These additional viewpoints introduce complementary perspectives ranging from lateral offsets and stereo angular disparities to near-ground imagery thereby enriching the spatial and geometric diversity of the dataset and enabling more comprehensive multi-view analysis.

\textbf{Annotations} All auxiliary viewpoints are annotated following the same protocol as the primary sensors. Each auxiliary camera is geometrically aligned and indexed with respect to its corresponding primary forward-facing viewpoint, ensuring consistent calibration, pose association, and sequence synchronization.

\textbf{Note} These auxiliary sensors are included only for completeness; they are not used in any NVS or camera-pose experiment reported in the main paper.

\begin{figure}[t]
    \includegraphics[width=\columnwidth]{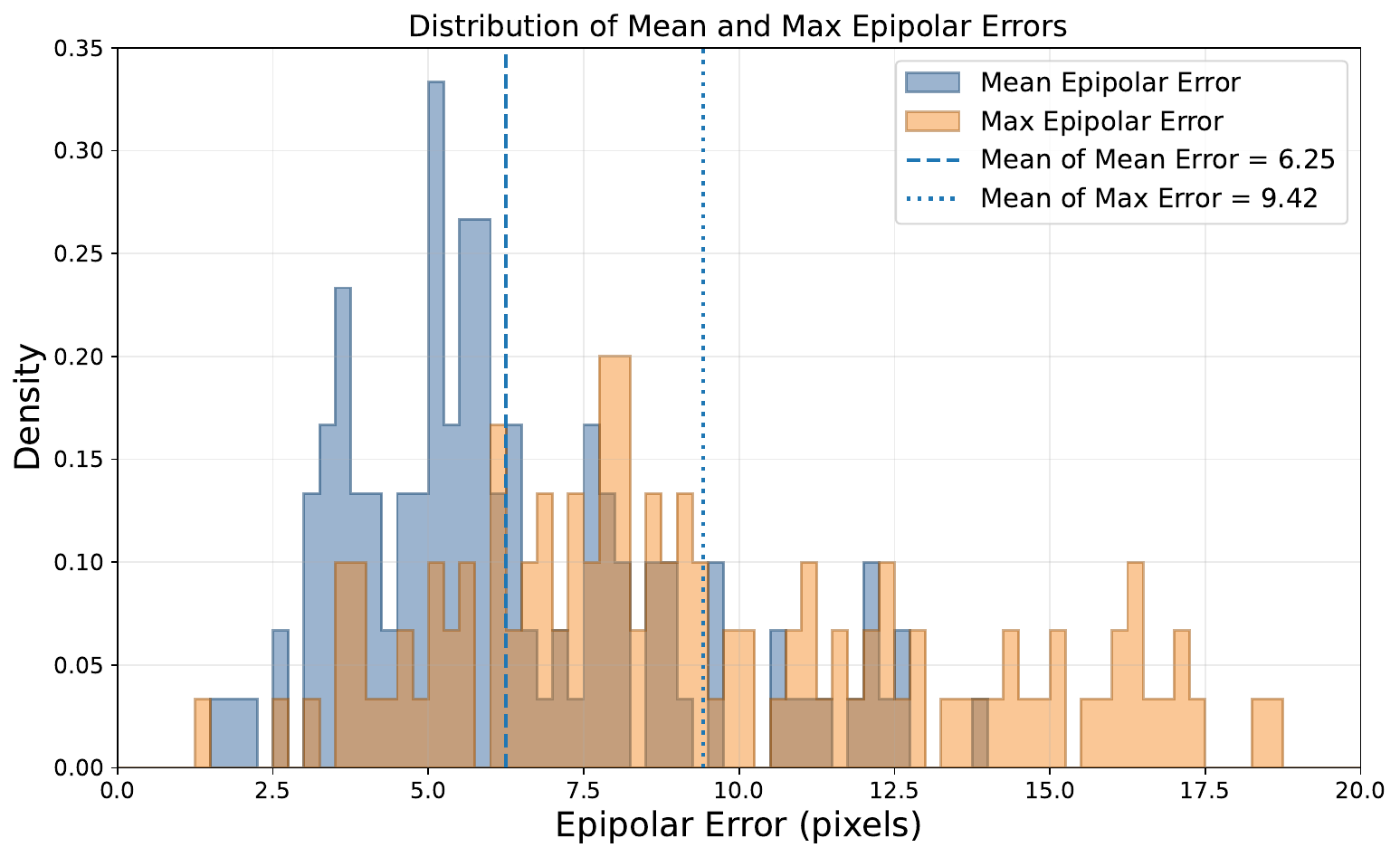}
    \caption{
\textbf{Epipolar error distribution for dynamic objects under $T_{C \rightarrow S}$.}
The histogram shows the density of mean and maximum epipolar reprojection errors computed from dynamic-object correspondences between the car view $V^{C}$ and the scooty view $V^{S}$. The strong concentration of errors at low pixel values indicates high epipolar consistency, confirming that the independently operating sensors remain temporally synchronized during capture.
    \label{fig:epipolar_sync_cs}
    \vspace{-15pt}
    }
\end{figure}

\subsection{Temporal Synchronization Validation via Epipolar Consistency}
Since $V^{C}$ and $V^{S}$ sensors operate without hardware-level triggering, verifying temporal alignment between the two streams is essential. We evaluate synchronization by analyzing epipolar consistency of corresponding pixels on dynamic objects observed simultaneously across both views under the transformation $T_{C \rightarrow S}$. If the sensors are temporally aligned, a moving object corresponds to the same physical state at a given timestamp, and therefore its projection in one view should lie close to the corresponding epipolar line in the other view. We compute the epipolar reprojection error for matched object points and analyze its distribution across sequences. As shown in \figref{fig:epipolar_sync_cs}, the majority of correspondences exhibit low errors, indicating strong geometric consistency between $V^{C}$ and $V^{S}$. In contrast, temporal misalignment would produce large deviations from the epipolar constraint due to motion-induced displacement. The observed low-error distribution therefore provides strong empirical evidence that the car and scooty streams remain effectively time synchronized during data capture.

\section{Additional Results}
\label{suppsec:addn_results}
\begin{table*}[ht!]
\centering
\resizebox{\textwidth}{!}{
\begin{tabular}{l|ccc|ccc|ccc}
\toprule
\textbf{Model (Views)} &
\multicolumn{3}{c|}{\textbf{PSNR} $\uparrow$} &
\multicolumn{3}{c|}{\textbf{SSIM} $\uparrow$} &
\multicolumn{3}{c}{\textbf{LPIPS} $\downarrow$} \\
\cmidrule(lr){2-4}\cmidrule(lr){5-7}\cmidrule(lr){8-10}
 & \textit{\textbf{T}}$_{C \rightarrow C}$ & \textit{\textbf{T}}$_{C \rightarrow L}$ & \textit{\textbf{T}}$_{C \rightarrow S}$ 
 & \textit{\textbf{T}}$_{C \rightarrow C}$ & \textit{\textbf{T}}$_{C \rightarrow L}$ & \textit{\textbf{T}}$_{C \rightarrow S}$ 
 & \textit{\textbf{T}}$_{C \rightarrow C}$ & \textit{\textbf{T}}$_{C \rightarrow L}$ & \textit{\textbf{T}}$_{C \rightarrow S}$ \\
\midrule

Depthsplat (2v) & 13.65 & 15.09 & 12.56 & 0.22 & 0.22 & 0.17 & 0.38 & 0.40 & 0.45 \\
Depthsplat (6v) & 15.71 & 15.78 & 13.08 & 0.28 & 0.27 & 0.24 & 0.34 & 0.37 & 0.41 \\

Monosplat (2v) & 16.11 & 15.26 & 13.14 & 0.35 & 0.31 & 0.22 & 0.32 & 0.35 & 0.38 \\
Monosplat (6v) & 18.31 & 17.72 & 15.10 & 0.41 & 0.37 & 0.27 & 0.27 & 0.30 & 0.35 \\

Mvsplat (2v) & 14.70 & 13.44 & 12.83 & 0.36 & 0.35 & 0.33 & 0.53 & 0.51 & 0.59 \\
Mvsplat (6v) & 16.31 & 14.06 & 12.87 & 0.41 & 0.38 & 0.35 & 0.45 & 0.46 & 0.55 \\

\midrule
\midrule

 & \textit{\textbf{T}}$_{D \rightarrow D}$ & \textit{\textbf{T}}$_{D \rightarrow C}$ & \textit{\textbf{T}}$_{D \rightarrow S}$ 
 & \textit{\textbf{T}}$_{D \rightarrow D}$ & \textit{\textbf{T}}$_{D \rightarrow C}$ & \textit{\textbf{T}}$_{D \rightarrow S}$ 
 & \textit{\textbf{T}}$_{D \rightarrow D}$ & \textit{\textbf{T}}$_{D \rightarrow C}$ & \textit{\textbf{T}}$_{D \rightarrow S}$ \\
\midrule

Depthsplat (2v) & 10.70 & 8.80 & 10.11 & 0.14 & 0.09 & 0.10 & 0.59 & 0.66 & 0.65 \\
Depthsplat (6v) & 13.05 & 9.07 & 11.03 & 0.28 & 0.12 & 0.13 & 0.48 & 0.63 & 0.62 \\

Monosplat (2v) & 14.28 & 9.53 & 11.67 & 0.28 & 0.21 & 0.23 & 0.52 & 0.63 & 0.62 \\
Monosplat (6v) & 16.54 & 10.19 & 12.68 & 0.37 & 0.27 & 0.29 & 0.41 & 0.58 & 0.56 \\

Mvsplat (2v) & 13.78 & 10.05 & 11.85 & 0.28 & 0.23 & 0.25 & 0.55 & 0.64 & 0.63 \\
Mvsplat (6v) & 16.04 & 9.84 & 12.10 & 0.35 & 0.26 & 0.29 & 0.45 & 0.62 & 0.58 \\

\bottomrule
\end{tabular}}
\caption{
NVS methods evaluated under the full cross-view setups in \textbf{Eval-Car-Train} and \textbf{Eval-Drone-Train}. Results are reported for all view-count settings (2v and 6v).
}
\label{tbl:main_supply}
\end{table*}

This section provides extended quantitative evaluations that complement the results presented in the main paper. We report detailed performance across all cross-view and cross-platform reconstruction settings using PSNR, SSIM, and LPIPS\cite{lpips}.

\subsection{Comparison with Existing Datasets}

\begin{figure*}[ht!]
\centering
\setlength{\arrayrulewidth}{1.2pt}

\begin{tabular}{|c|c c|c c|}
\hline

\textbf{GT} & \textbf{L-CC} & \textbf{L-Cross-Lane} & \textbf{D-CC} & \textbf{D-Cross-Lane} \\ 
\hline

\includegraphics[width=0.18\textwidth]{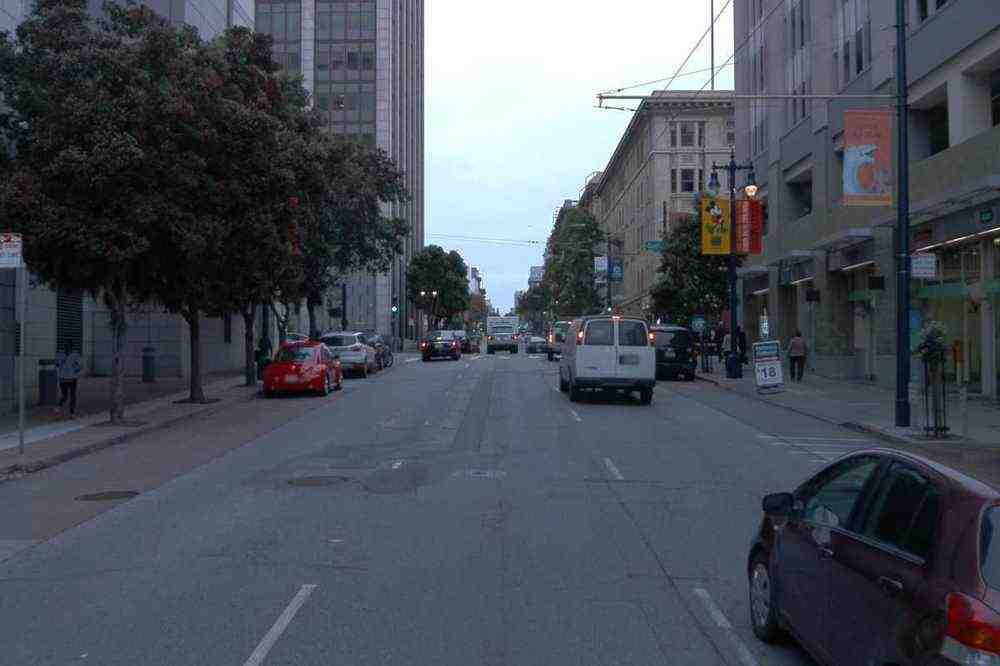} &
\includegraphics[width=0.18\textwidth]{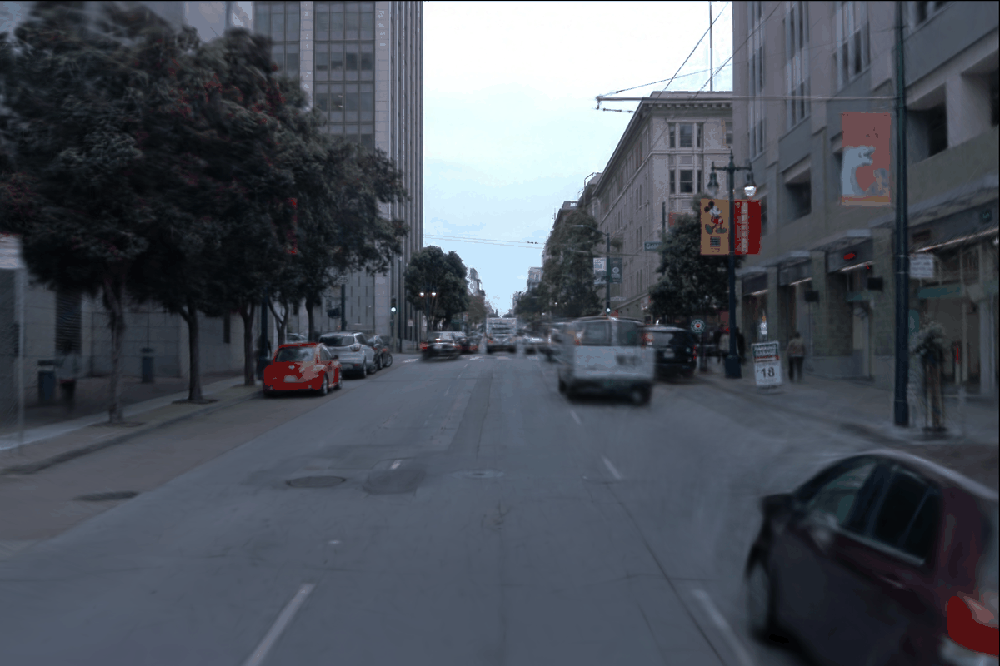} &
\includegraphics[width=0.18\textwidth]{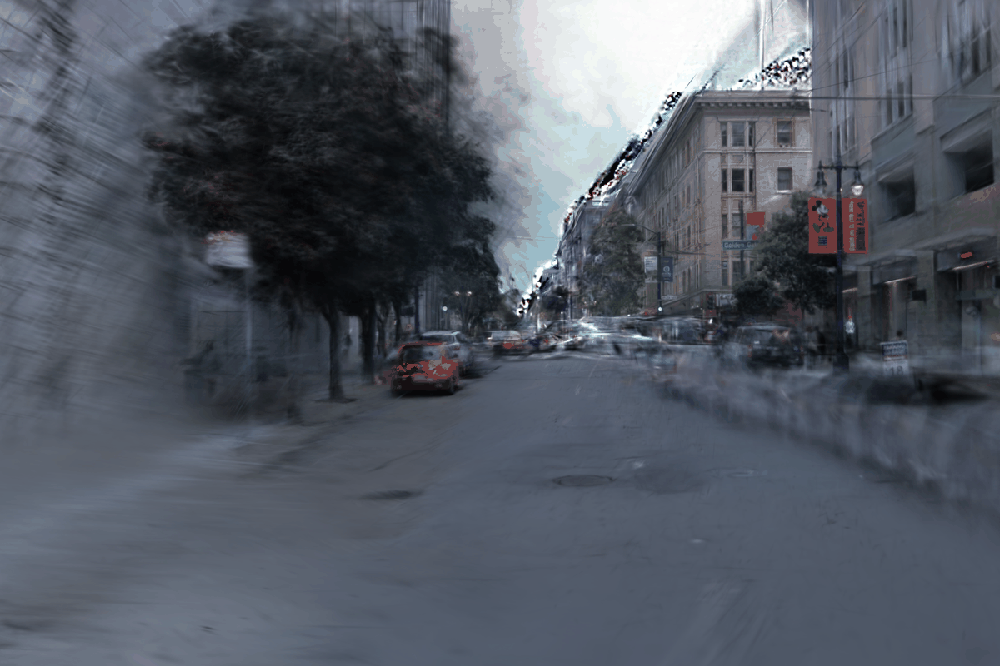} &
\includegraphics[width=0.18\textwidth]{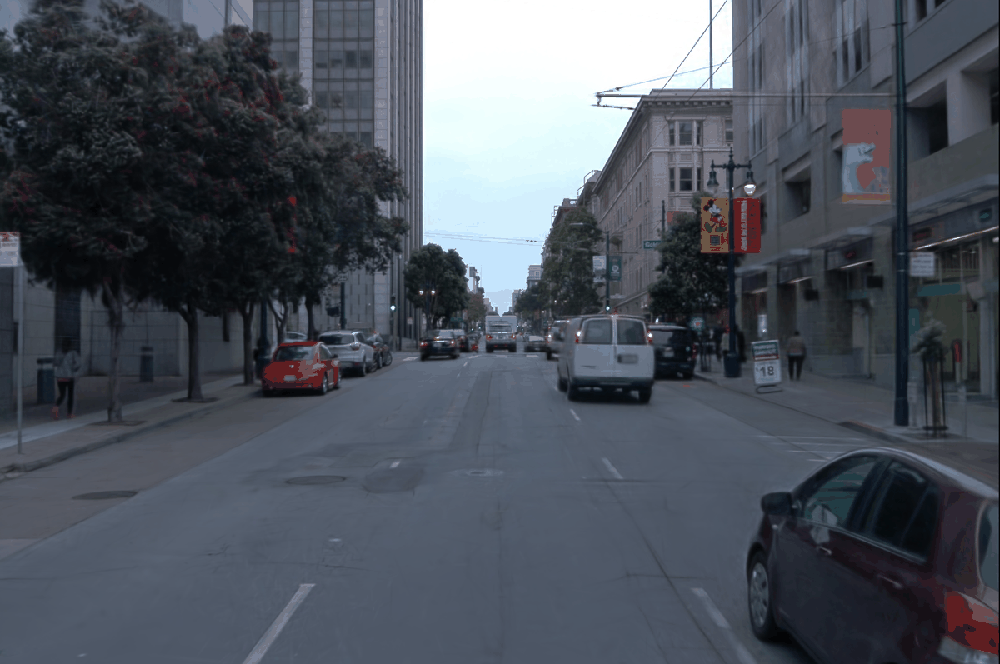} &
\includegraphics[width=0.18\textwidth]{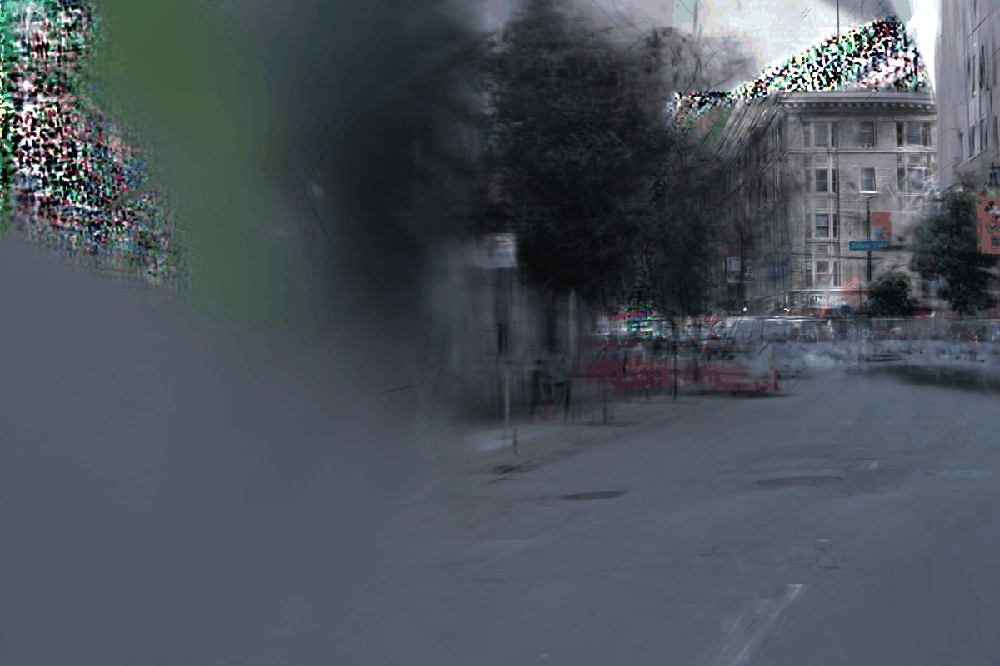} \\

& {\tiny 26.12 / 0.78 / 0.14} & &
{\tiny 27.01 / 0.82 / 0.12} & \\

\includegraphics[width=0.18\textwidth]{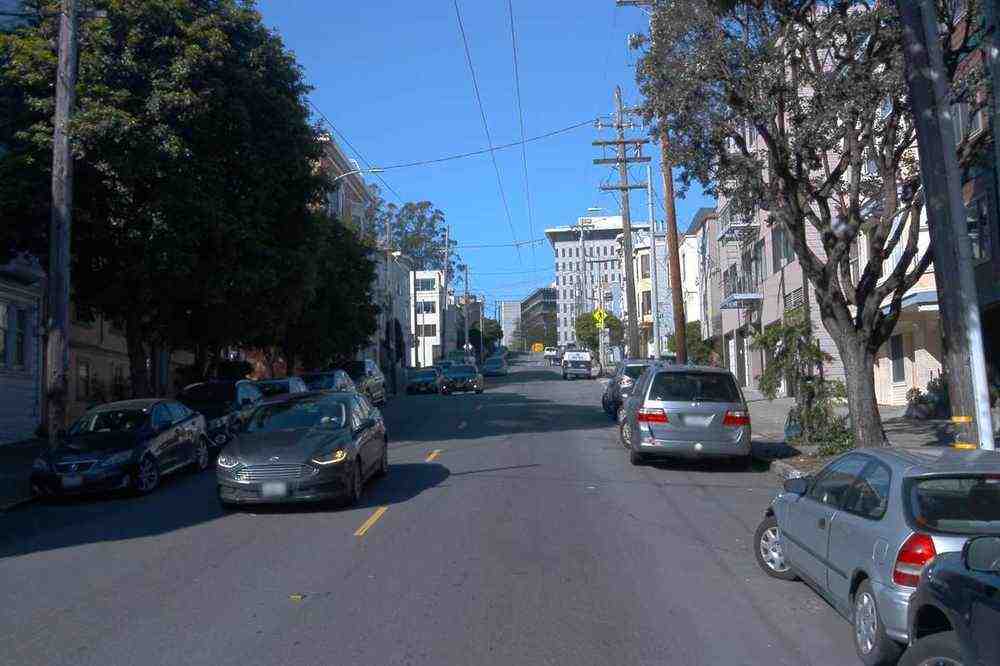} &
\includegraphics[width=0.18\textwidth]{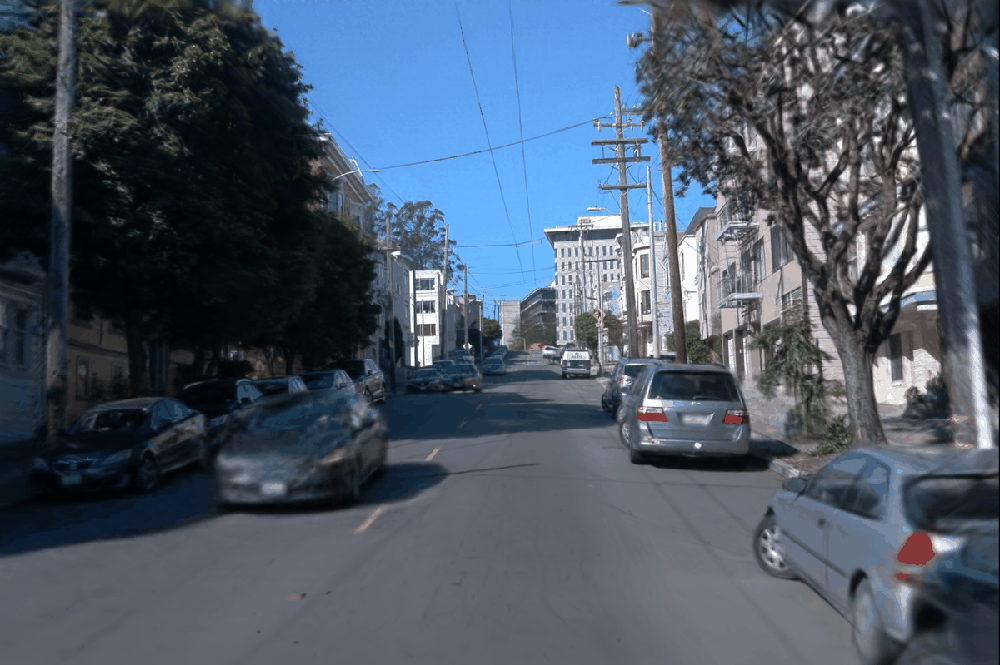} &
\includegraphics[width=0.18\textwidth]{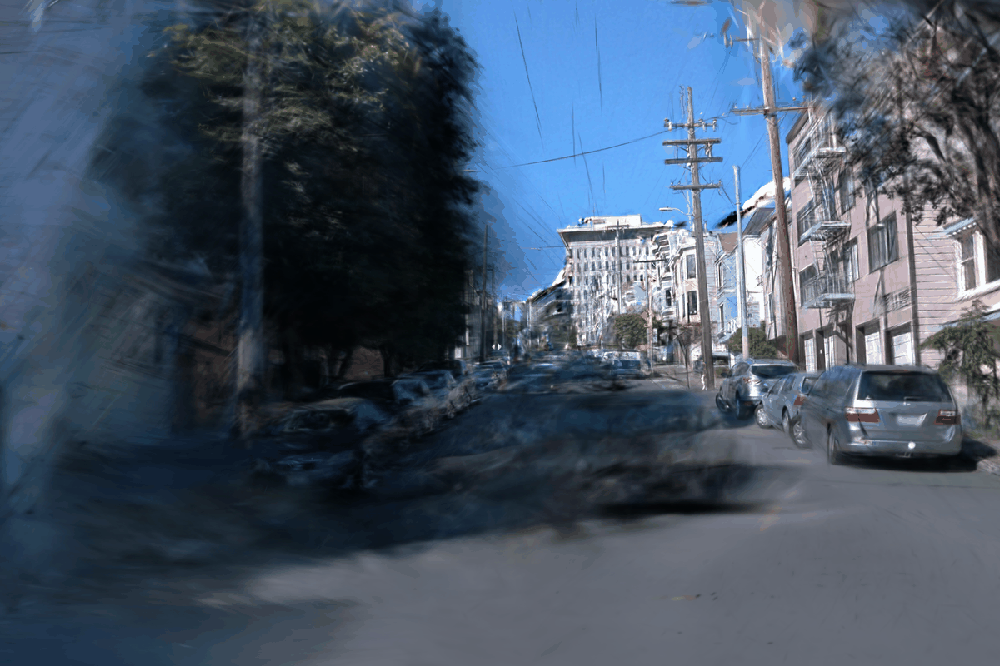} &
\includegraphics[width=0.18\textwidth]{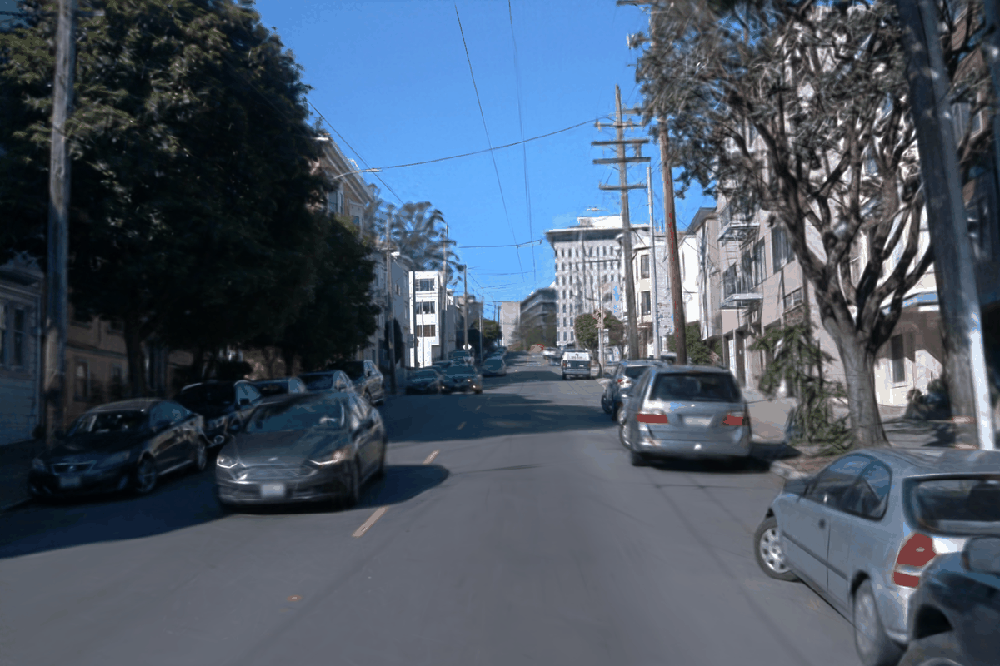} &
\includegraphics[width=0.18\textwidth]{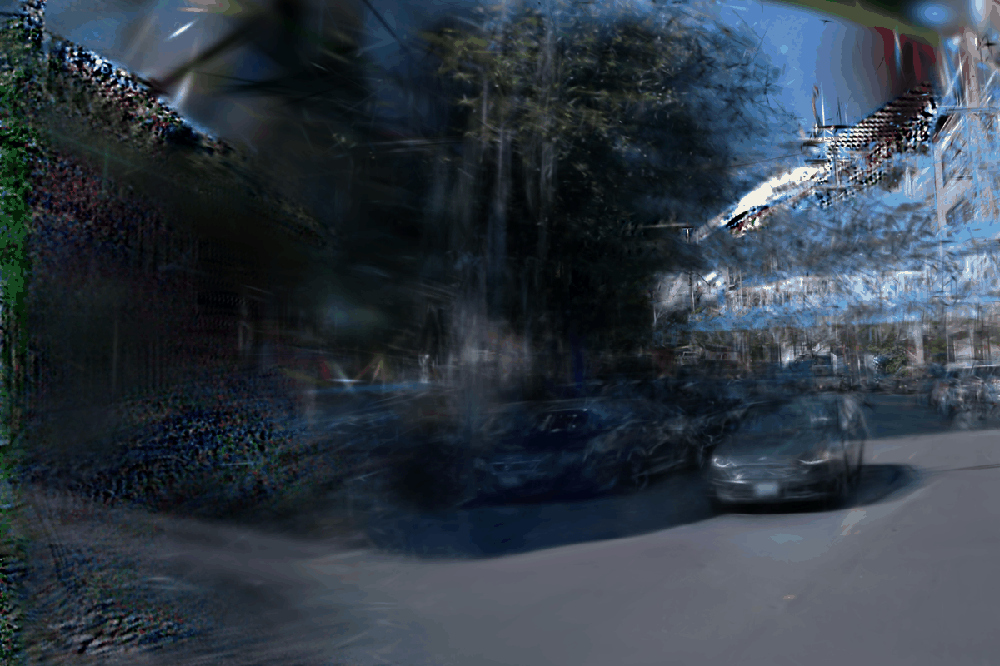} \\

& {\tiny 25.48 / 0.74 / 0.16} & &
{\tiny 26.33 / 0.80 / 0.13} & \\

\includegraphics[width=0.18\textwidth]{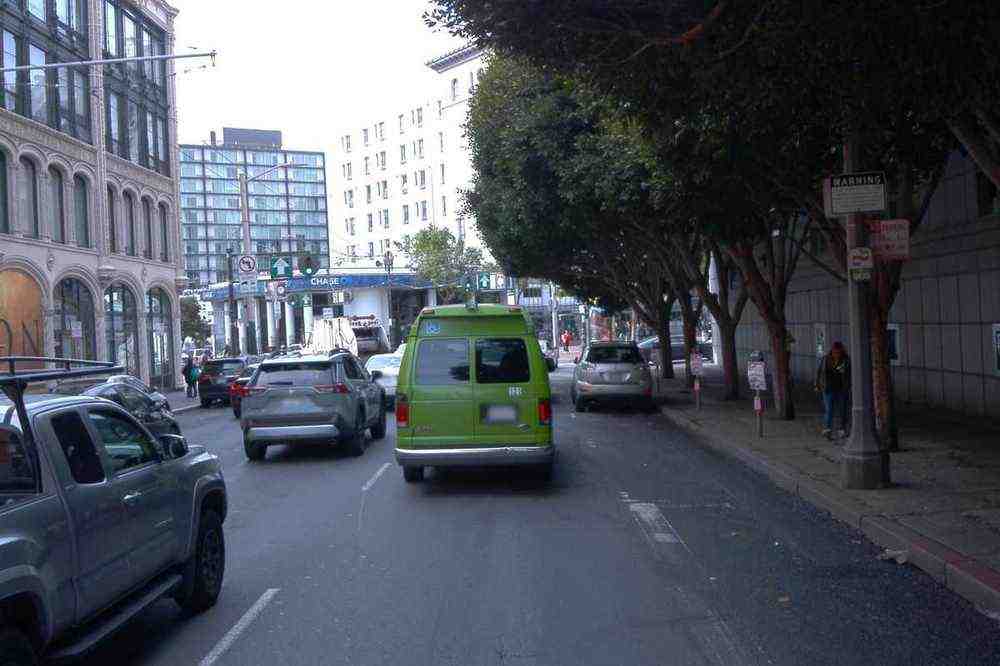} &
\includegraphics[width=0.18\textwidth]{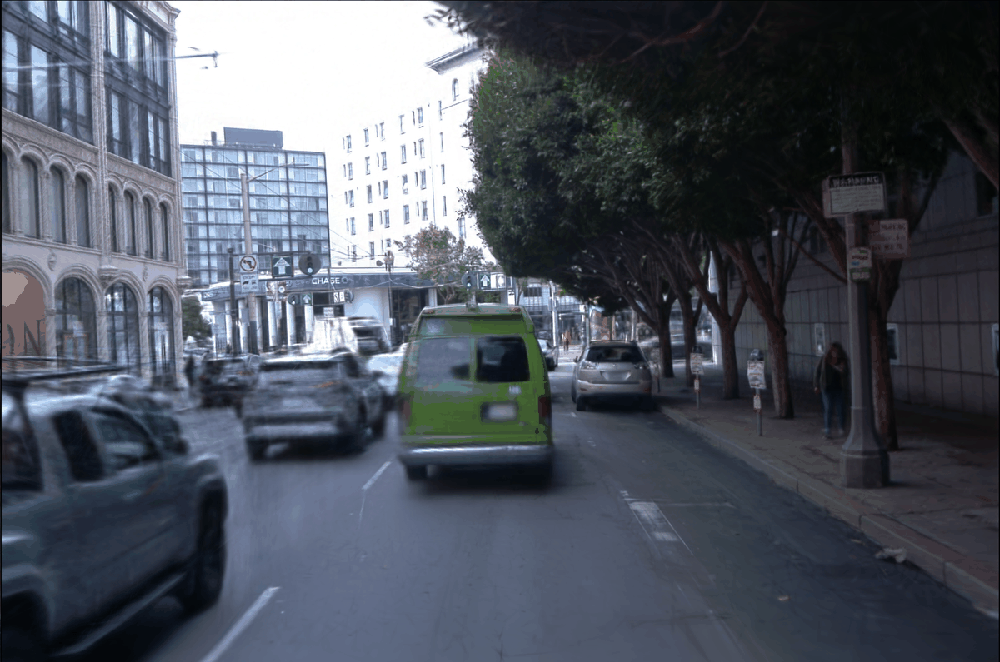} &
\includegraphics[width=0.18\textwidth]{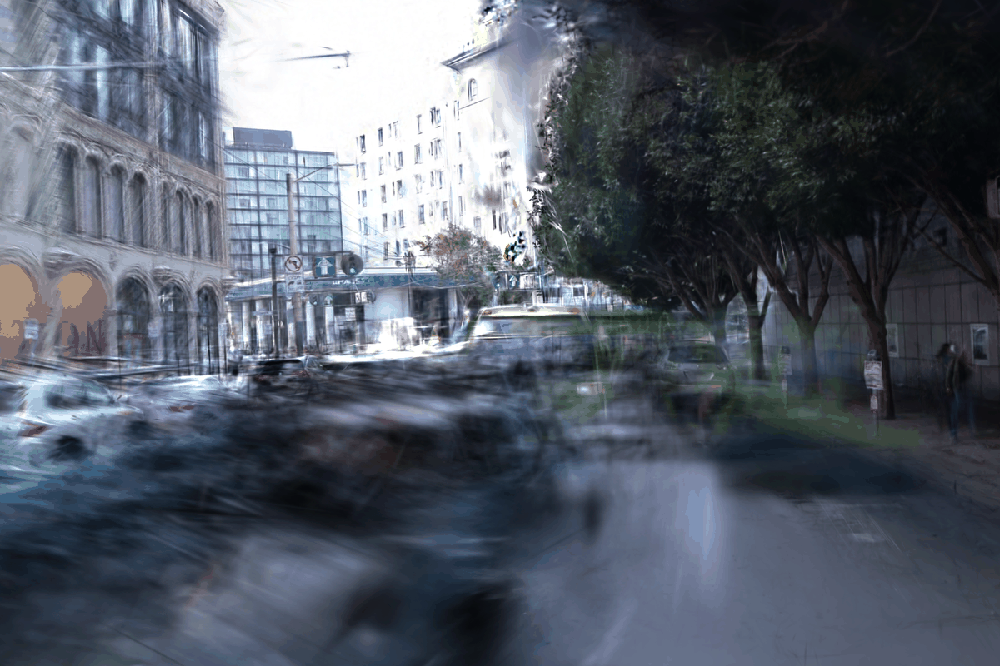} &
\includegraphics[width=0.18\textwidth]{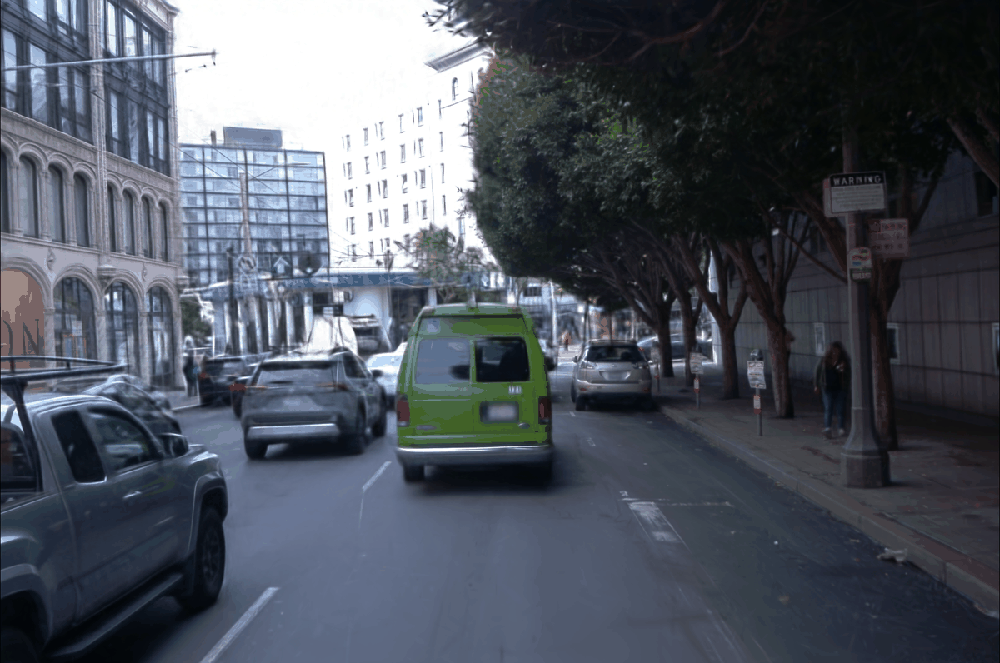} &
\includegraphics[width=0.18\textwidth]{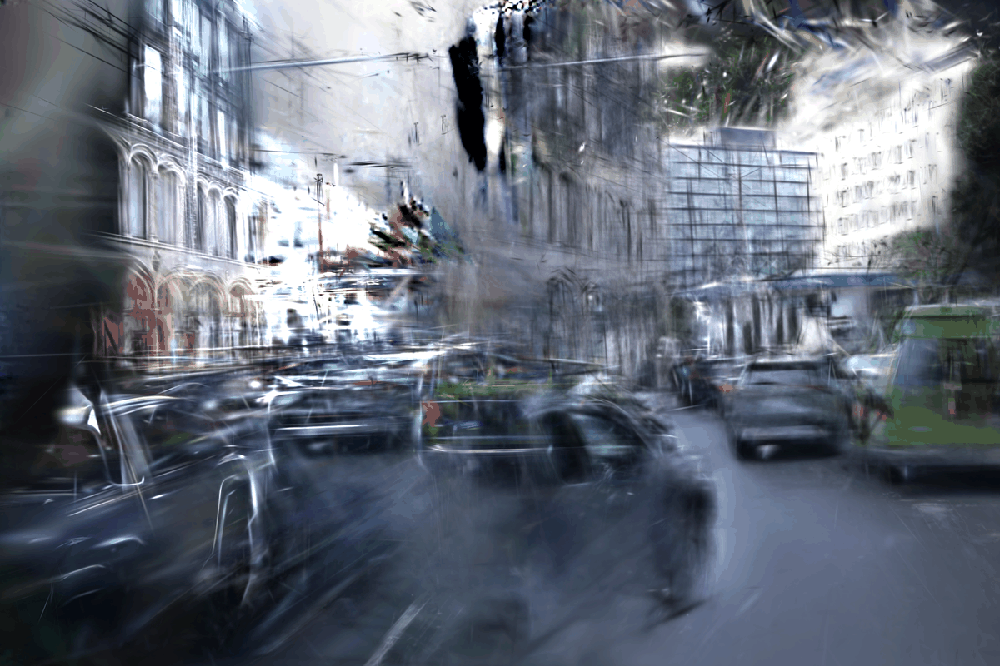} \\

& {\tiny 27.21 / 0.83 / 0.11} & &
{\tiny 26.02 / 0.76 / 0.15} & \\

\includegraphics[width=0.18\textwidth]{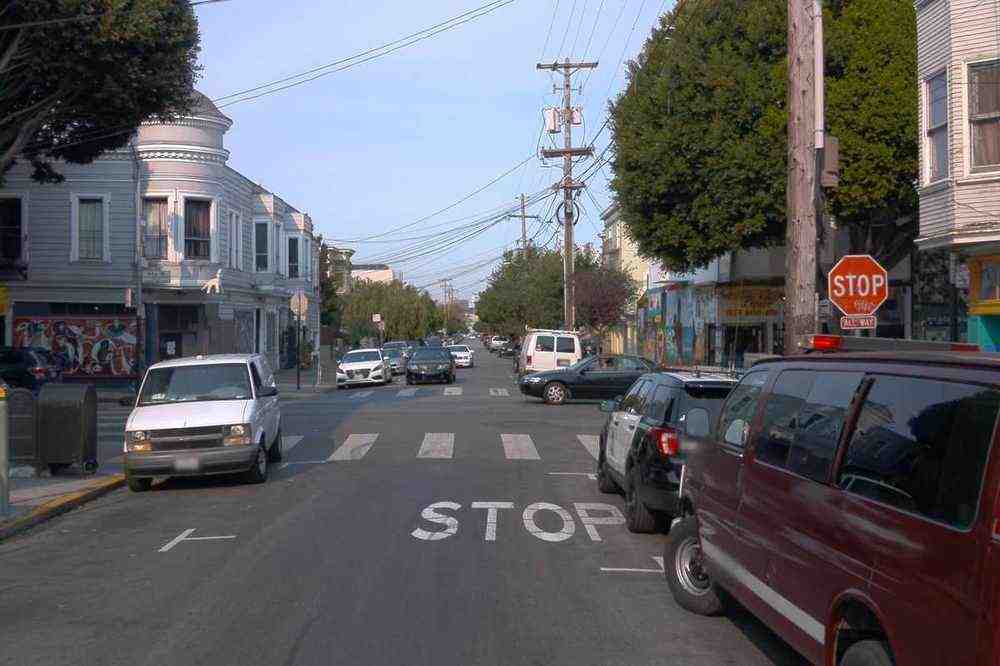} &
\includegraphics[width=0.18\textwidth]{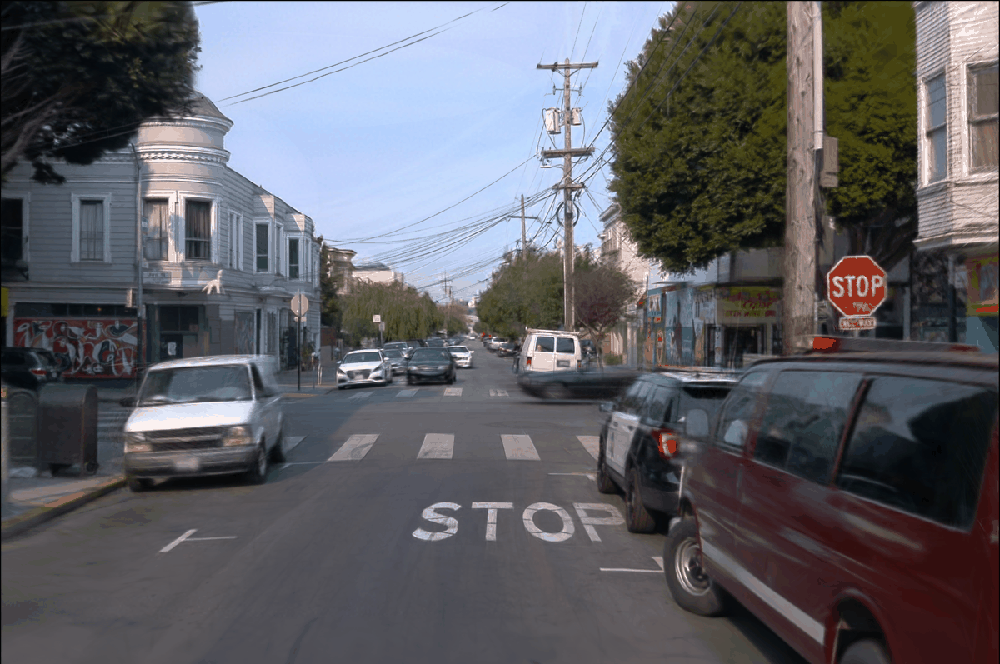} &
\includegraphics[width=0.18\textwidth]{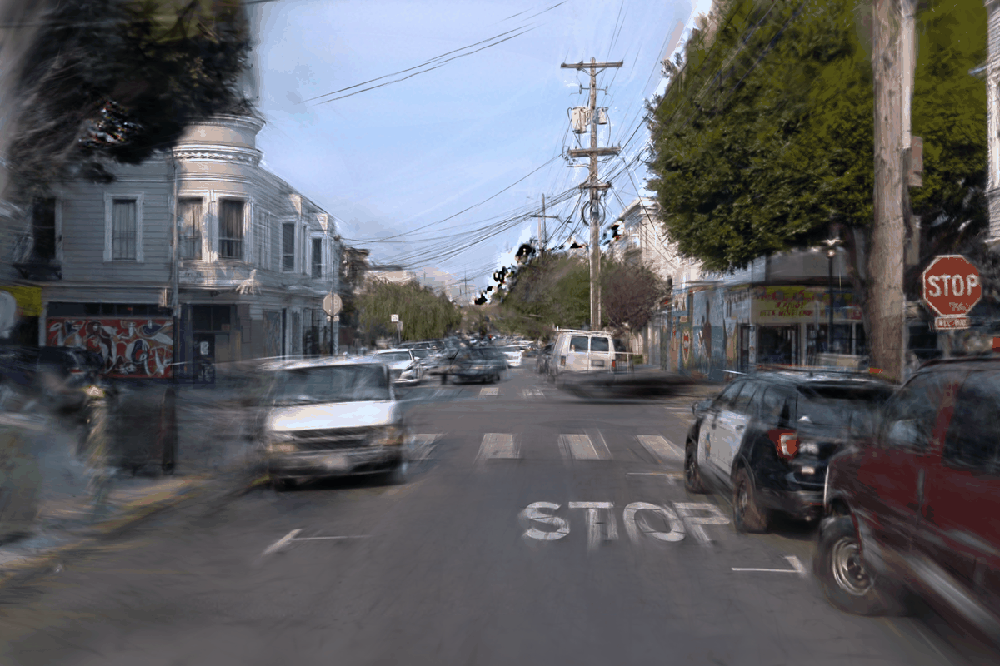} &
\includegraphics[width=0.18\textwidth]{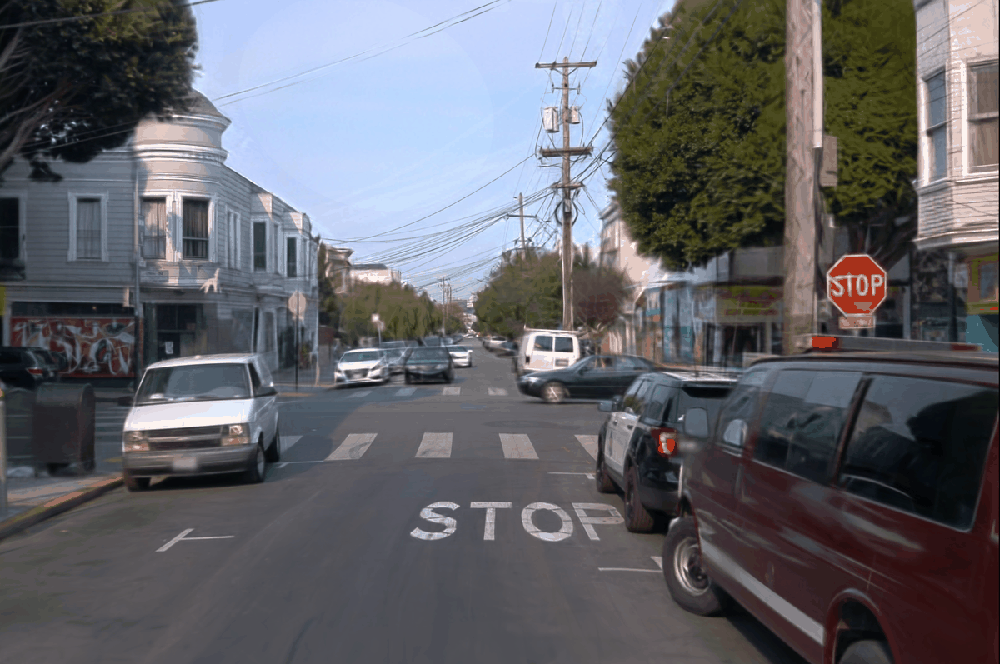} &
\includegraphics[width=0.18\textwidth]{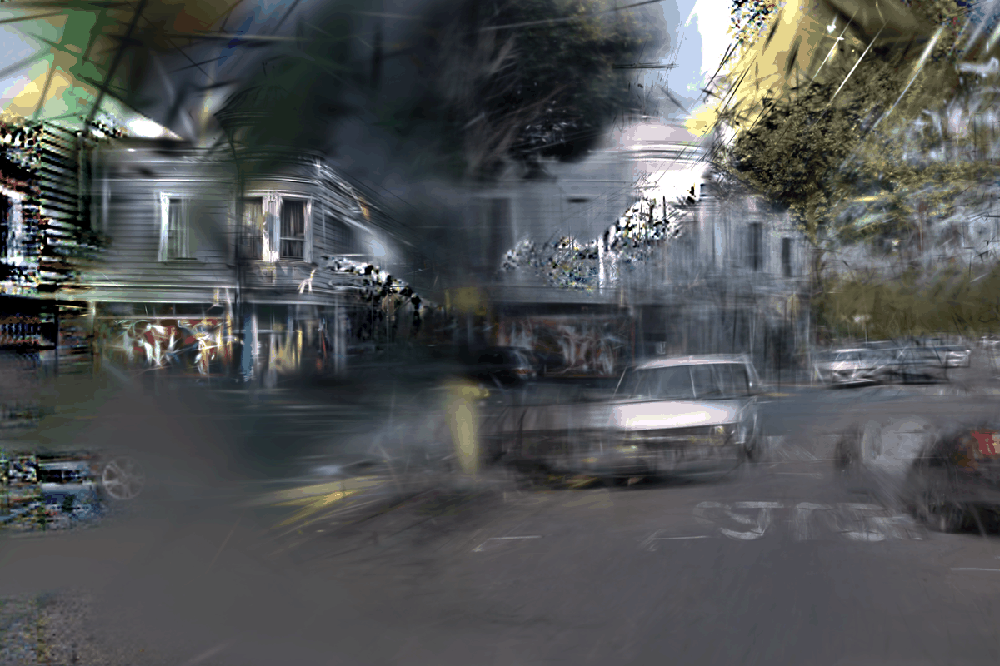} \\

& {\tiny 24.95 / 0.69 / 0.18} & &
{\tiny 25.88 / 0.75 / 0.16} & \\

\includegraphics[width=0.18\textwidth]{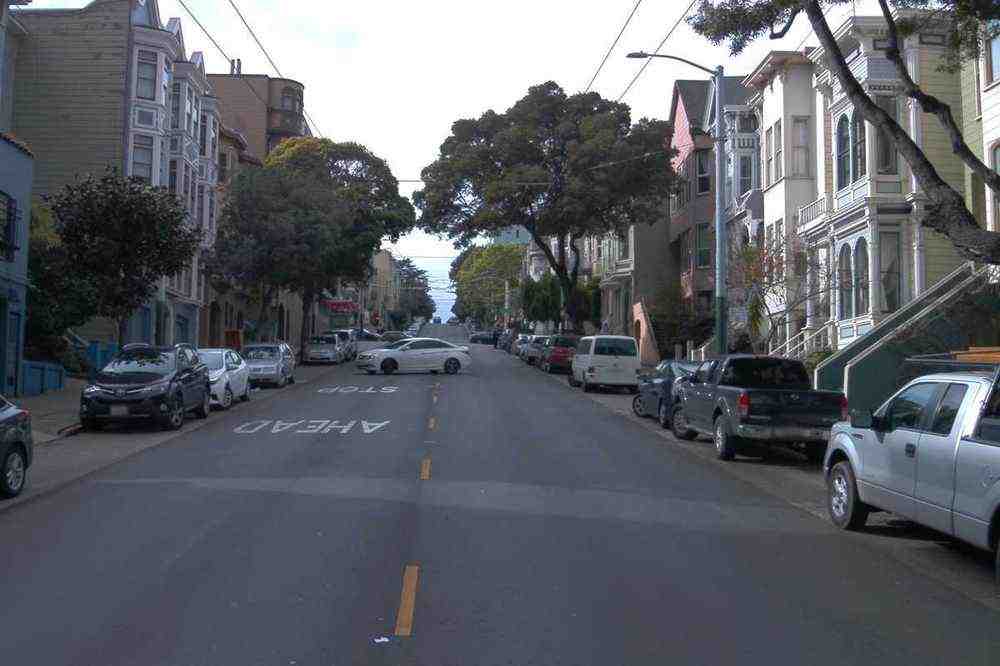} &
\includegraphics[width=0.18\textwidth]{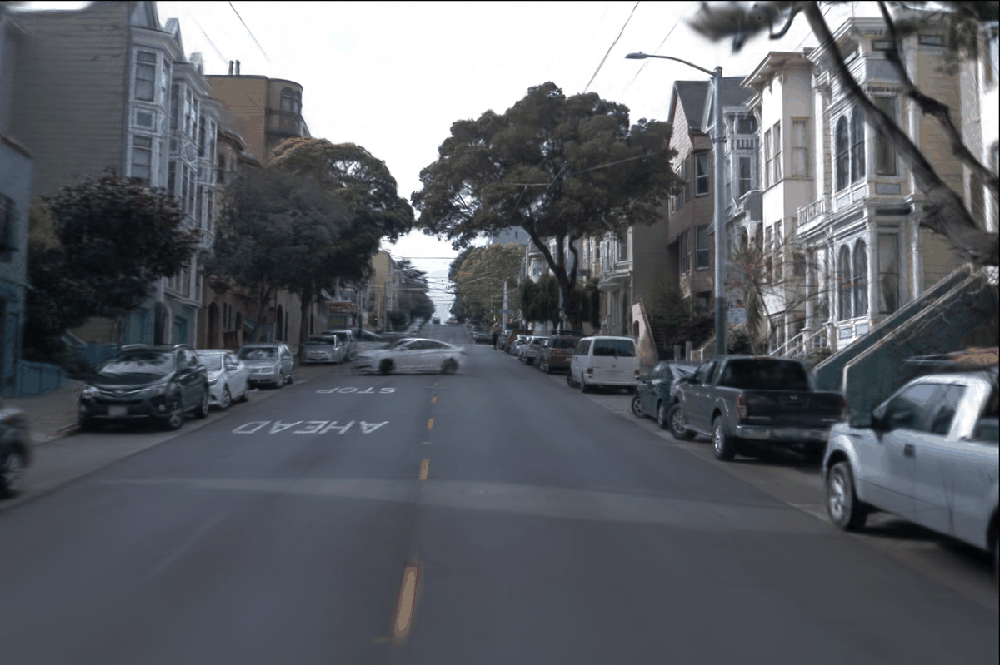} &
\includegraphics[width=0.18\textwidth]{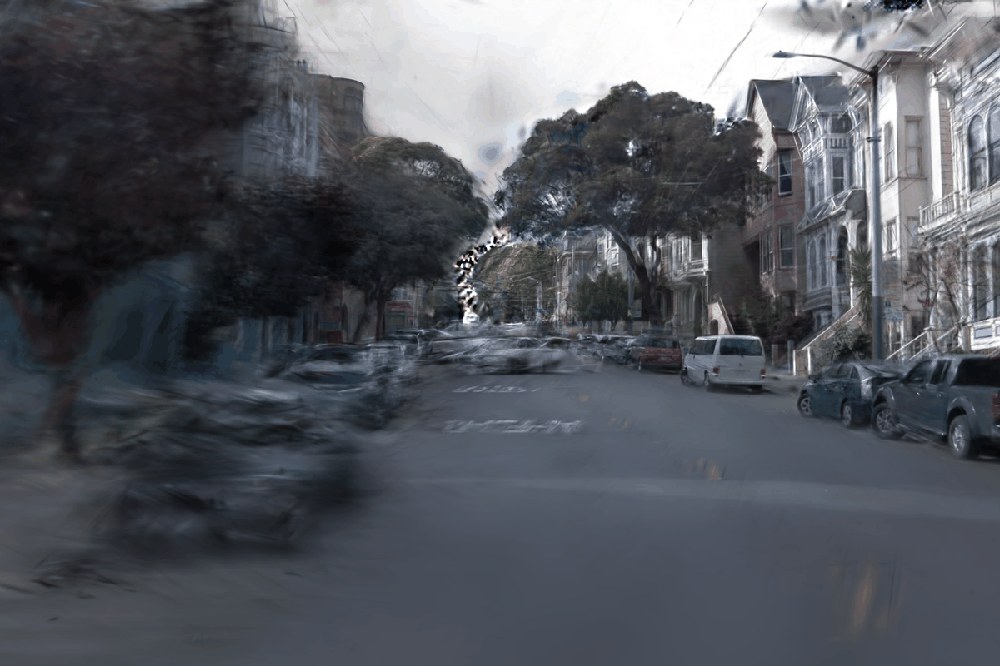} &
\includegraphics[width=0.18\textwidth]{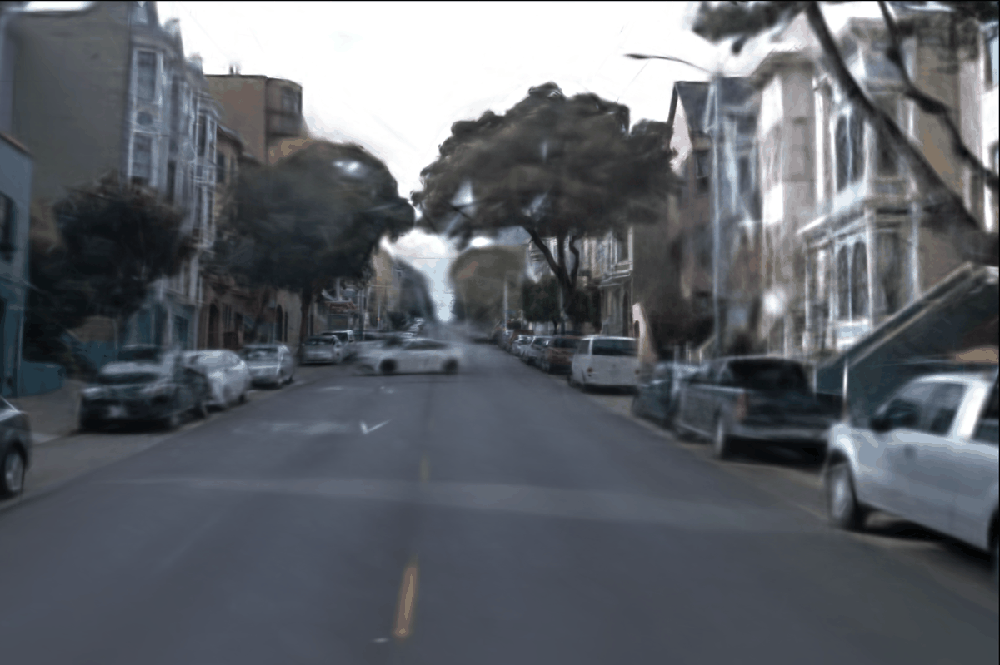} &
\includegraphics[width=0.18\textwidth]{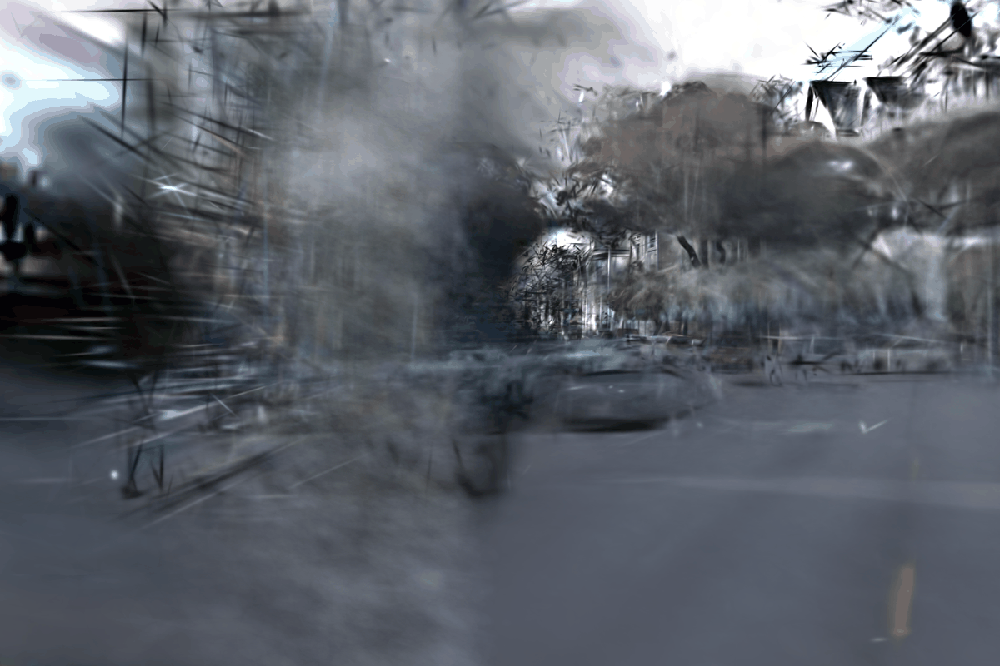} \\

& {\tiny 26.78 / 0.81 / 0.12} & &
{\tiny 27.43 / 0.84 / 0.10} & \\



\includegraphics[width=0.18\textwidth]{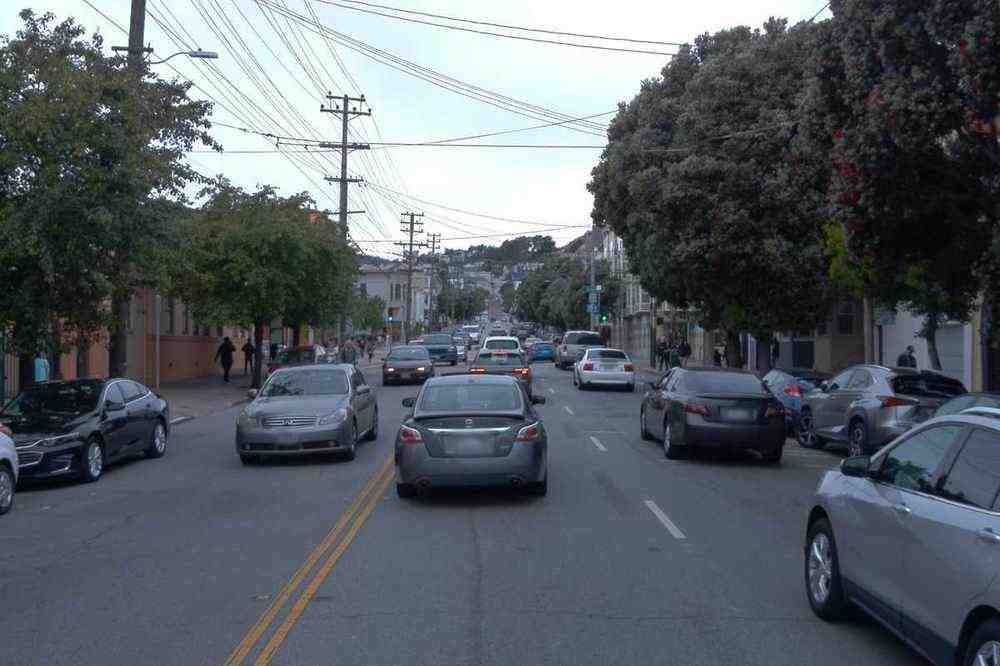} &
\includegraphics[width=0.18\textwidth]{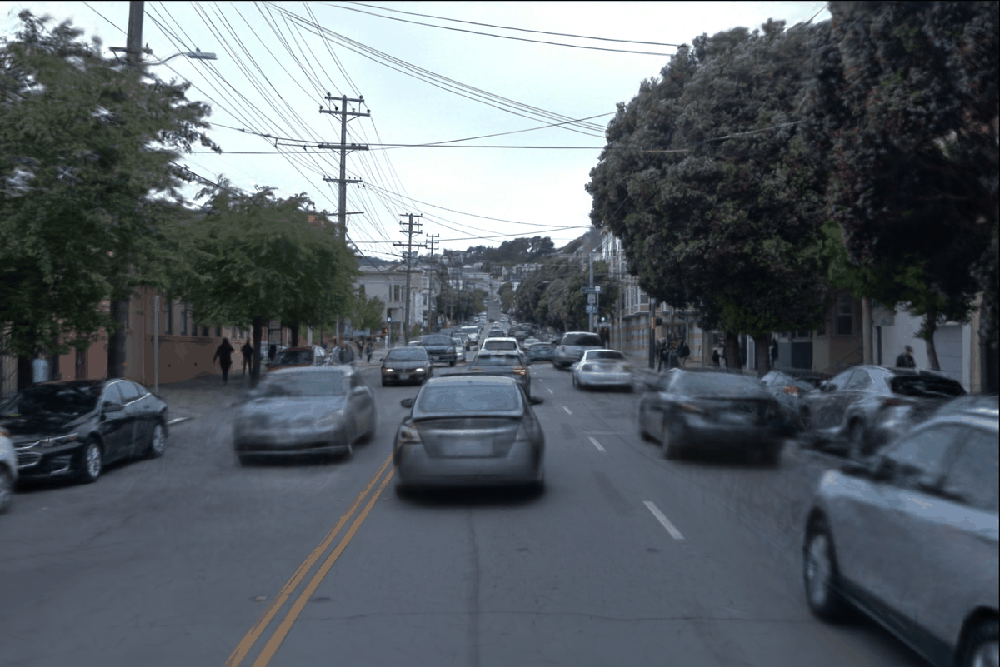} &
\includegraphics[width=0.18\textwidth]{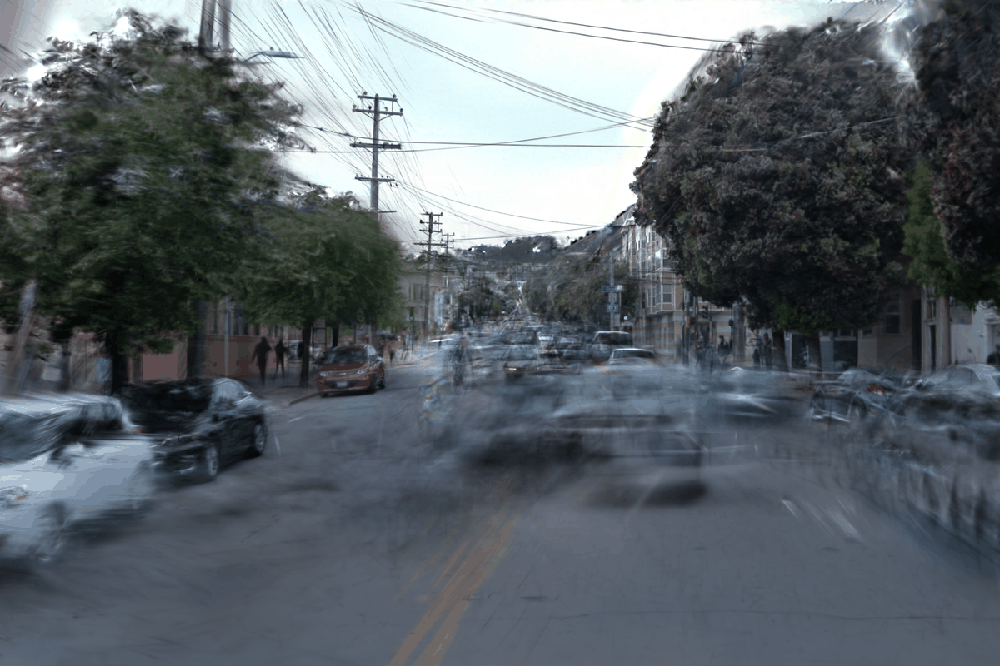} &
\includegraphics[width=0.18\textwidth]{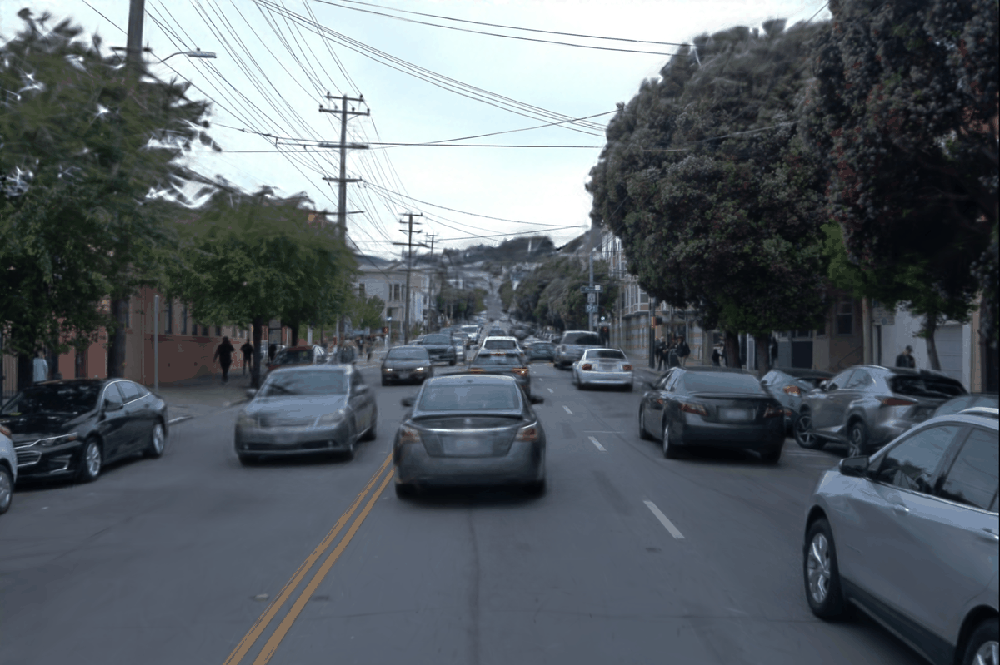} &
\includegraphics[width=0.18\textwidth]{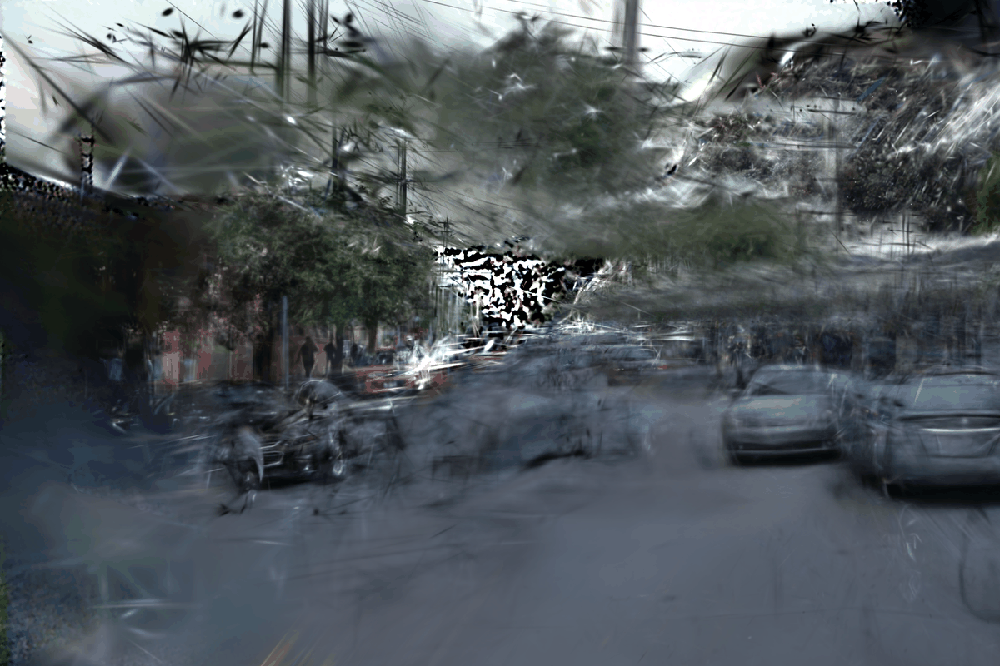} \\

& {\tiny 27.05 / 0.82 / 0.11} & &
{\tiny 26.66 / 0.77 / 0.13} & \\

\includegraphics[width=0.18\textwidth]{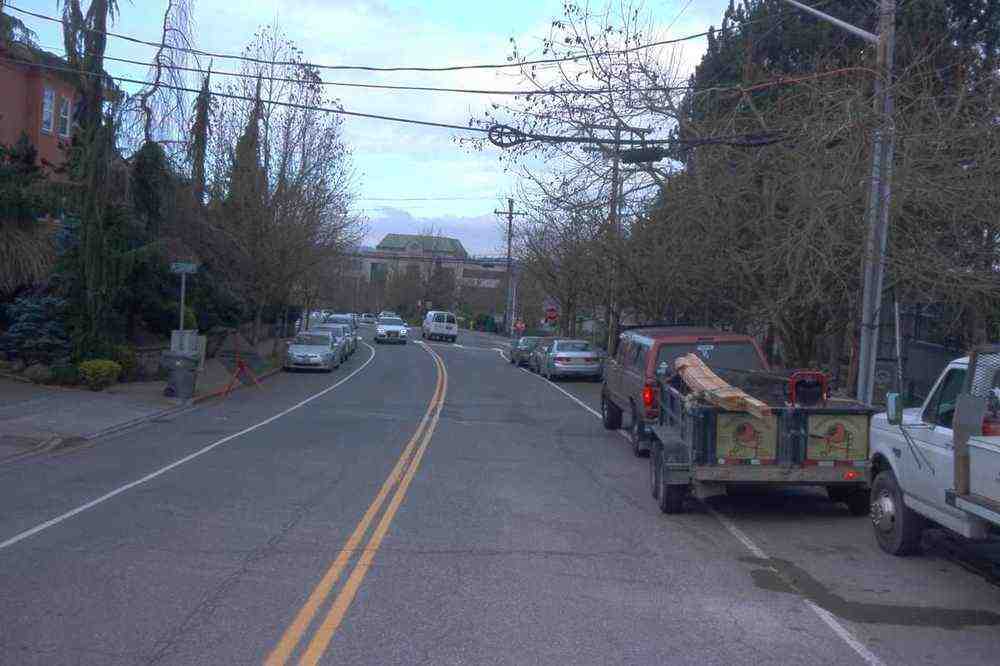} &
\includegraphics[width=0.18\textwidth]{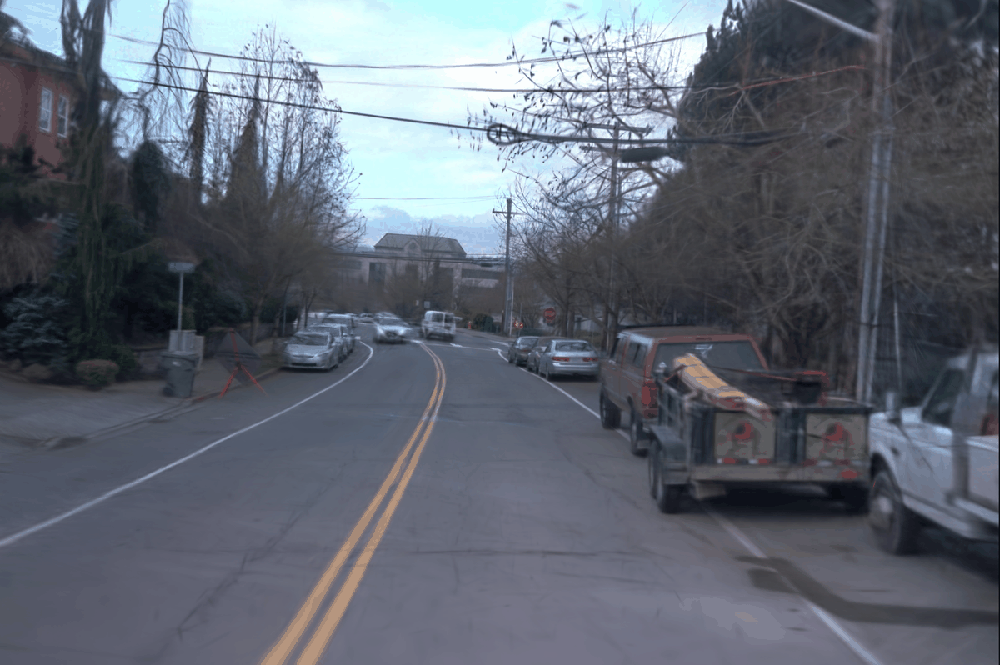} &
\includegraphics[width=0.18\textwidth]{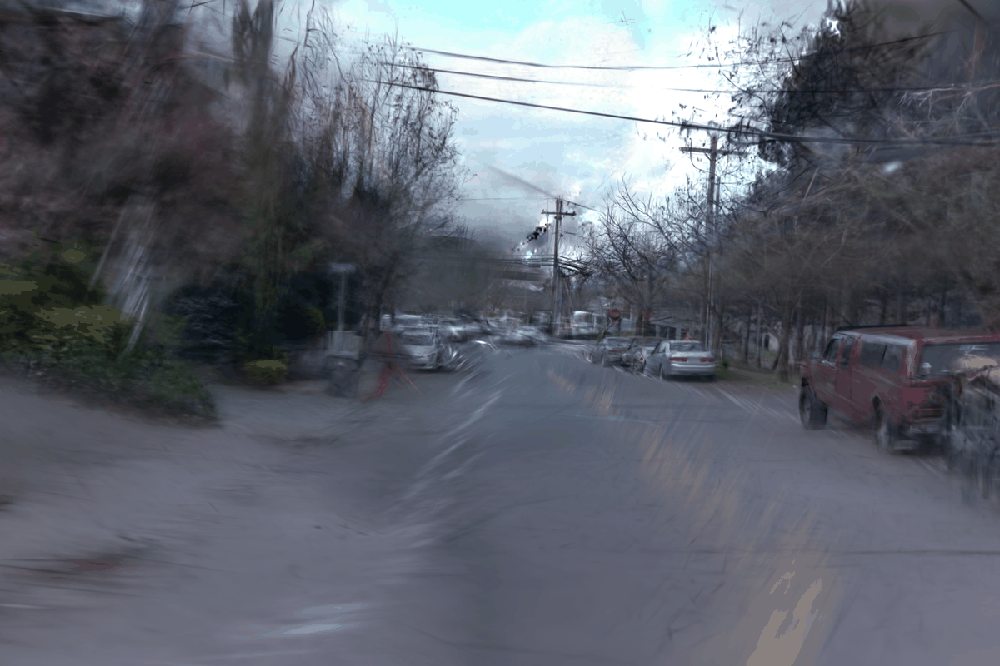} &
\includegraphics[width=0.18\textwidth]{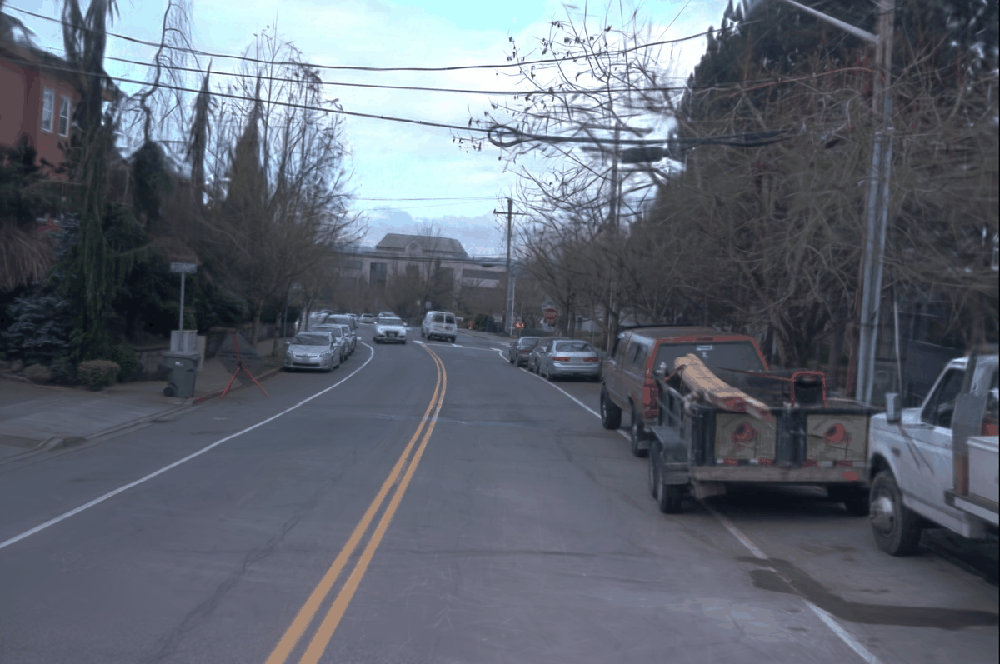} &
\includegraphics[width=0.18\textwidth]{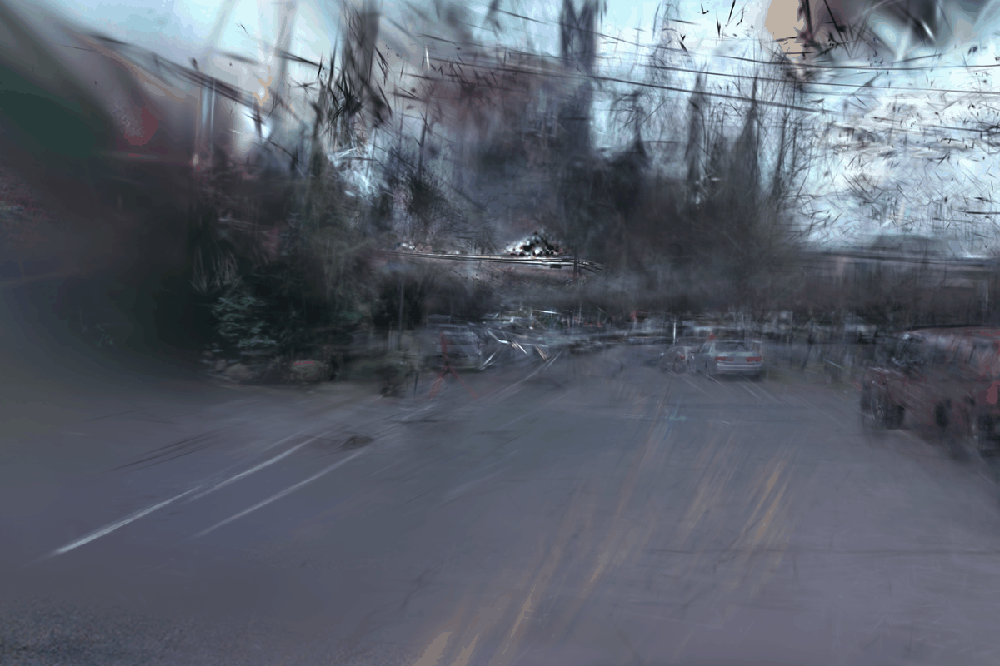} \\

& {\tiny 24.72 / 0.67 / 0.19} & &
{\tiny 25.39 / 0.71 / 0.17} & \\

\includegraphics[width=0.18\textwidth]{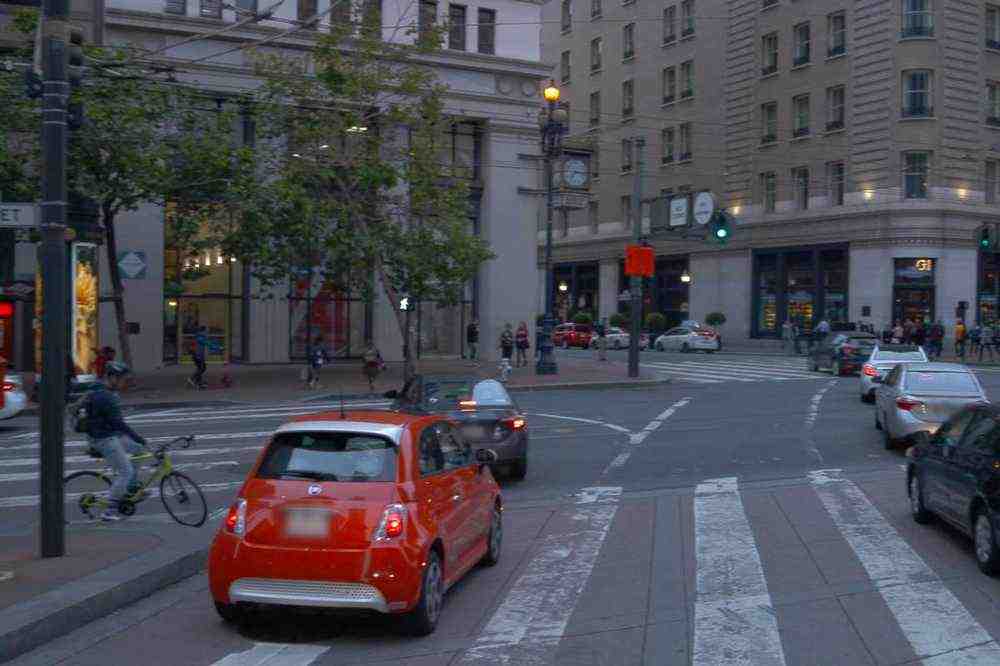} &
\includegraphics[width=0.18\textwidth]{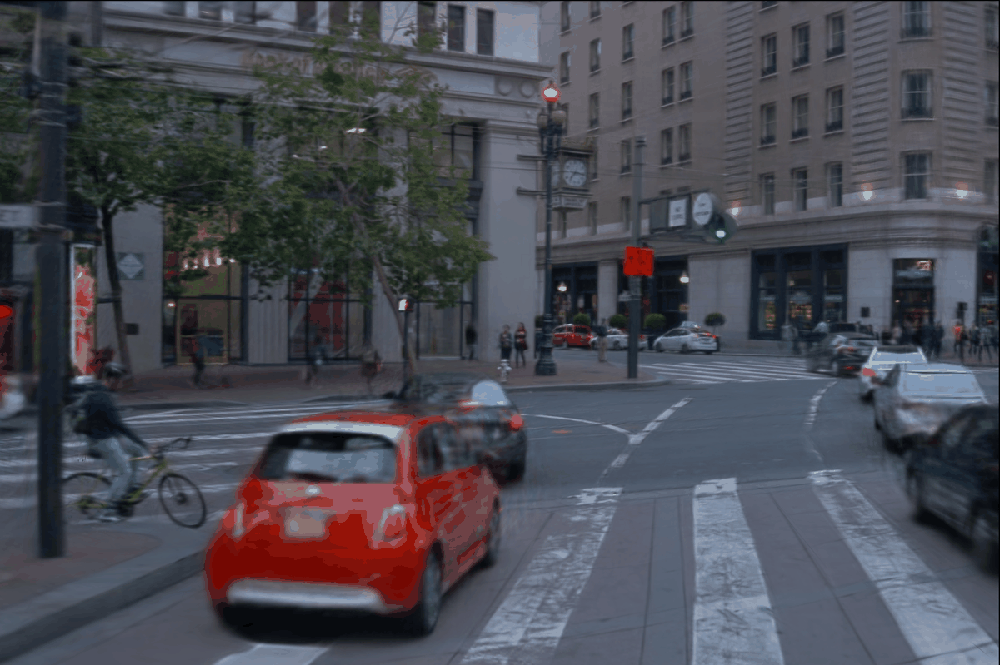} &
\includegraphics[width=0.18\textwidth]{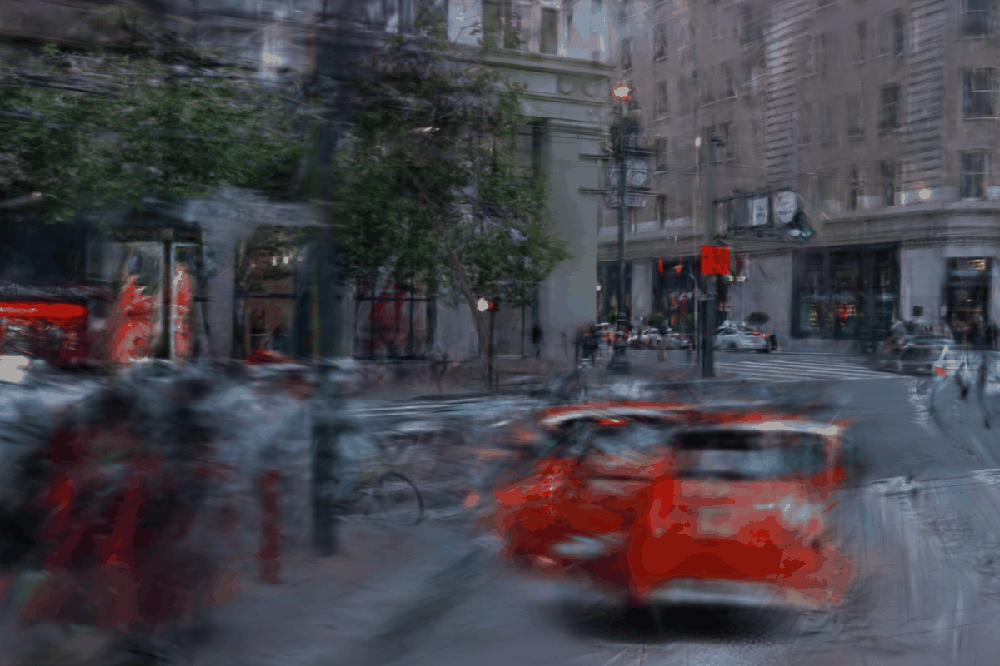} &
\includegraphics[width=0.18\textwidth]{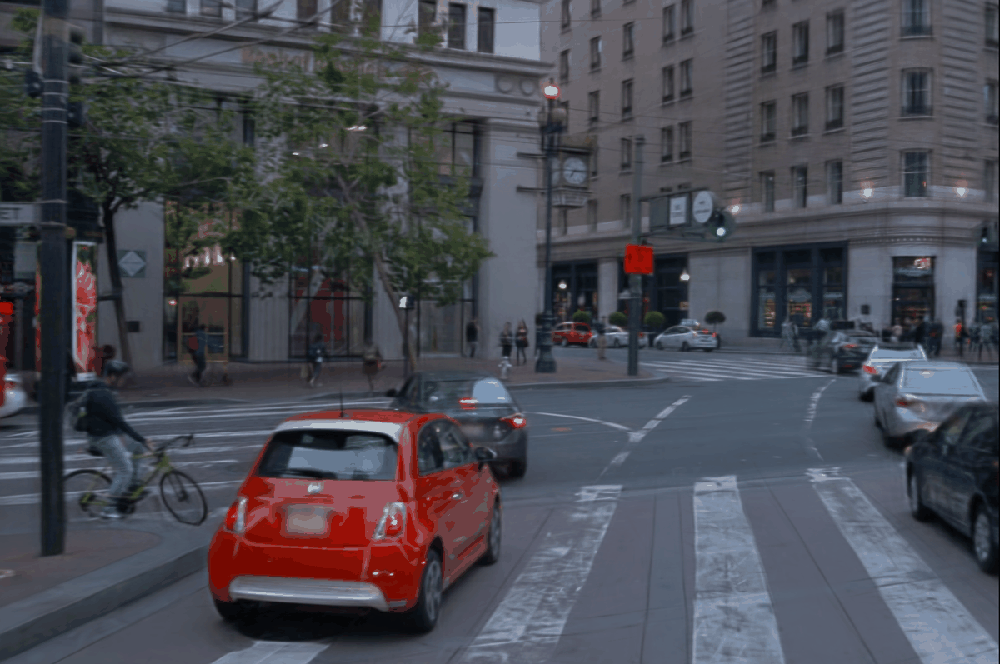} &
\includegraphics[width=0.18\textwidth]{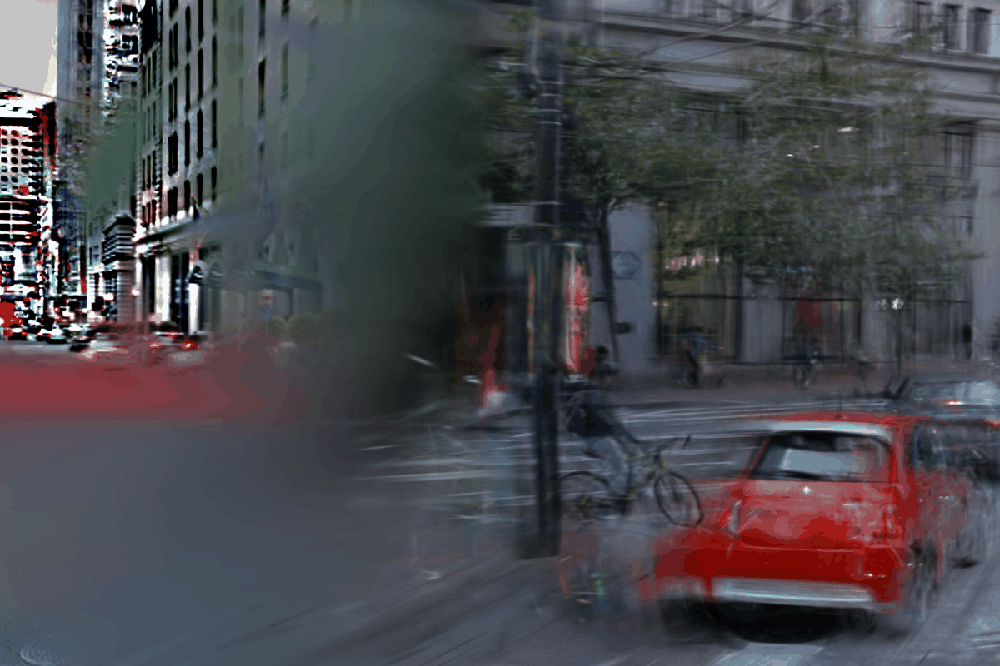} \\

& {\tiny 26.44 / 0.79 / 0.13} & &
{\tiny 27.12 / 0.83 / 0.11} & \\

\hline
\end{tabular}

\caption{\textbf{Qualitative comparison across different camera pair configurations.}
Each row corresponds to a frame. Columns show the ground truth (GT) and predictions for LiDAR-based (L) and DaV3 DaV3~\cite{lin2025depth3recoveringvisual}-based (D) depth supervision under same-camera (CC) and cross-lane transformations.}
\label{fig:waymo_qualitative_supply}
\vspace{-10pt}
\end{figure*}

We present additional results under the $T_{C\rightarrow C}$ setup on the dynamic split~\cite{yang2023emernerf} of the Waymo Open Dataset (WOD) in~\tabref{tab:depth_filter_ablation} and~\figref{fig:waymo_qualitative_supply}. First, our reproduced PVG results on WOD are better than those reported in the original benchmark (WOD* vs. WOD in~\tabref{tab:depth_filter_ablation}). Second, we observe a performance drop (row 2 vs. row 3 in~\tabref{tab:depth_filter_ablation}) when replacing the standard PVG inputs (multi-camera + LiDAR) with our minimal configuration (single camera + depth from an off-the-shelf model). Notably, PVG under this configuration achieves similar NVS metrics on WOD and on \MV for the same $T_{C\rightarrow C}$ setup (row 3 vs. row 4~\tabref{tab:depth_filter_ablation}).

To further analyze this behavior,~\figref{fig:waymo_qualitative_supply} provides qualitative comparisons under different input and NVS settings. Even with the standard inputs (multi-camera + LiDAR), cross-lane NVS renderings (column 3 in~\figref{fig:waymo_qualitative_supply} ) exhibit noticeable artifacts compared to in-lane ($T_{C\rightarrow C}$) renderings (column 2 in~\figref{fig:waymo_qualitative_supply}). Under our minimal-input configuration (single camera + off-the-shelf depth), the $T_{C\rightarrow C}$ renderings (column 4 in~\figref{fig:waymo_qualitative_supply}) remain visually comparable to the standard in-lane results (column 2 in~\figref{fig:waymo_qualitative_supply}). In contrast, cross-lane renderings in this setup (column 5) show the most severe degradation. The relative quality trends between in-lane and cross-lane renderings on WOD closely match those observed for PVG on \MV (\cf~\figref{fig:qual_car_car} and~\figref{fig:qual_car_scooty}), where $T_{C\rightarrow C}$ and $T_{C\rightarrow S}$ correspond to in-lane and cross-lane evaluation, respectively.

However, existing datasets such as WOD do not provide ground-truth cross-lane views, preventing quantitative evaluation of this degradation. In contrast, \MV explicitly supports such evaluation through the $T_{C\rightarrow S}$ protocol. The same limitation also applies to the \dronetrain setup.

\subsection{Additional Results for Feed-forward 3DGS models}

\tabref{tbl:main_supply} reports the complete results for the transformations
\begin{align*}
T_{C \rightarrow C},\; T_{C \rightarrow L},\; T_{C \rightarrow S},\; T_{D \rightarrow D},\; T_{D \rightarrow C},\; T_{D \rightarrow S}.
\end{align*}
We evaluate both 2-view and 6-view configurations for DepthSplat~\cite{depthsplat},
MonoSplat~\cite{monosplat}, and MVSplat~\cite{mvsplat}. These results complement
the 12-view models reported in the main paper.

Across all methods, we observe consistent trends that echo the findings in the main text. Increasing the number of conditioning views improves reconstruction fidelity, particularly for feed-forward splatting models. However, performance systematically degrades as the variation in the point of view increases for example, from $T_{C \rightarrow C}$ to $T_{C \rightarrow L}$ and further to $T_{C \rightarrow S}$. This reinforces our core conclusion that having an additional car-mounted sensor ($V_L$) is insufficient to emulate the challenging cross-vehicle viewpoint gap introduced in $T_{C \rightarrow S}$. Similar behavior is observed in the \textbf{Eval-Drone-Train} setting, where both 3DGS\cite{kerbl20233d} and PVG\cite{pvg} achieve strong performance on near-viewpoint tasks ($T_{D \rightarrow D}$) but experience notable degradation on cross-platform splits ($T_{D \rightarrow C}$ and $T_{D \rightarrow S}$).

\begin{figure*}
    \centering
    \includegraphics[width=\linewidth]{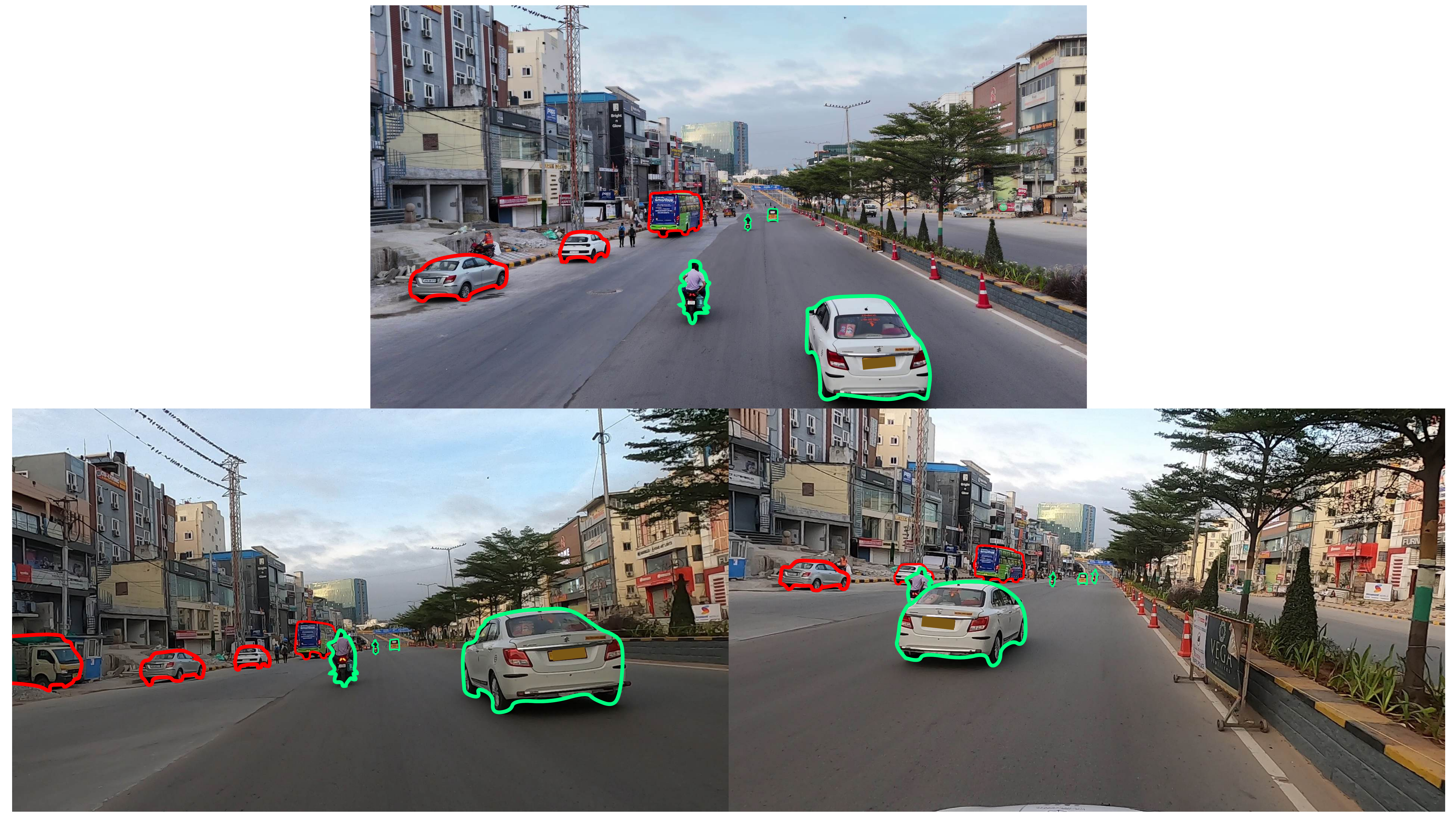}
    \caption{Illustration of dynamic-object masking using SAM and Dronesplat. The SAM masks correspond to the full red\,{+}\,green regions. Dynamic objects are obtained as the intersection of SAM and Dronesplat masks (green), whereas the remaining SAM-only regions (red) represent static objects. This visualization demonstrates how the combined masks separate dynamic and static scene elements.
}
    \label{fig:dynamic_object_example_image}
\end{figure*}

\begin{figure}[t]
    \includegraphics[width=\columnwidth]{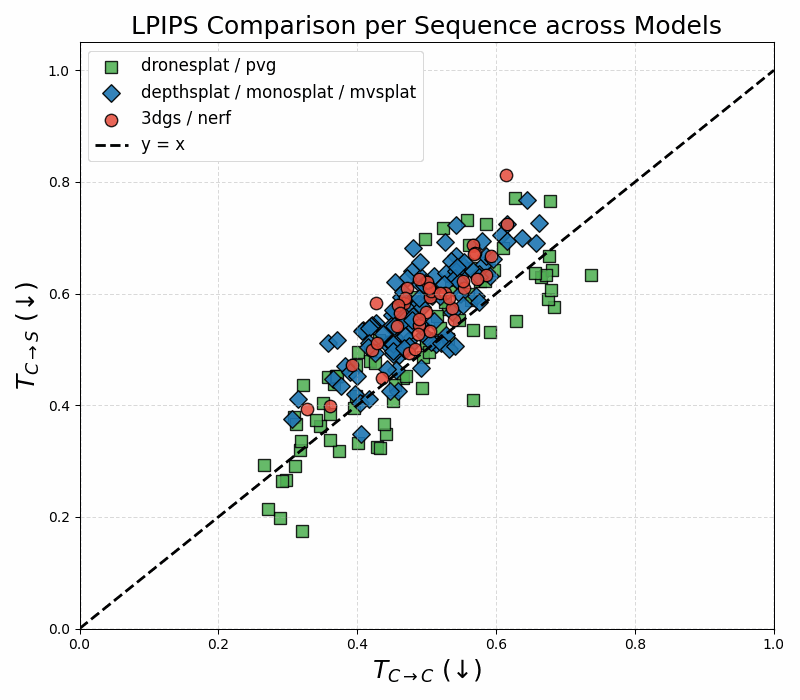}
    \caption{Sequence-level LPIPS comparison between the $\textbf{T}_{C \rightarrow C}$ (x-axis) and $\textbf{T}_{C \rightarrow S}$. Each point corresponds to LPIPS metric averaged over the corresponding test images in a single sequence. The methods are grouped as static (3DGS,NeRF), dynamic (DroneSplat, DesireGS, PVG), and feed-forward NVS methods.
    \label{fig:lpips_car_train}
    \vspace{-15pt}
    }
\end{figure}

\subsection{Sequence-Level LPIPS Consistency Across $V^C$ and $V^S$}

To analyze cross-vehicle generalization under the \textbf{Eval-Car-Train} setting, \figref{fig:lpips_car_train} shows sequence-level LPIPS comparison between $T_{C \rightarrow C}$ and $T_{C \rightarrow S}$. Each point represents the LPIPS averaged over test images in a single sequence (out of 50 sequences). Methods are grouped into static (3DGS, NeRF), dynamic (DroneSplat, DesireGS, PVG), and feed-forward NVS. For most sequences and methods, performance drops when the test viewpoint shifts from car to scooty, reflected by deviations from the $x{=}y$ line. The drop is largest for methods that perform well on the low-baseline set $T_{C \rightarrow C}$, and smaller—but still consistent—for methods that perform poorly on $T_{C \rightarrow C}$. Across methods, LPIPS on the challenging $T_{C \rightarrow S}$ set typically falls in the range 0.4–0.5. This scatter analysis complements the quantitative tables by highlighting sequence-dependent variations in cross-vehicle generalization.

\begin{table}[t]
\centering
\small
\setlength{\tabcolsep}{6pt}
\renewcommand{\arraystretch}{1.2}
\begin{tabular}{lcccc}
\hline
\textbf{Model} & \textbf{$T_{C\rightarrow C}$} & \textbf{$T_{C\rightarrow S}$} & \textbf{$T_{D\rightarrow D}$} & \textbf{$T_{D\rightarrow C}$} \\
\hline
GT         & 0.00   & 101.97 & 0.00   & 124.30 \\
DepthSplat & 121.66 & 152.30 & 110.00 & 208.94 \\
MonoSplat  & 140.77 & 156.91 & 127.97 & 216.68 \\
MVSplat    & 152.93 & 150.79 & 160.91 & 217.70 \\
DroneSplat & 144.25 & 127.07 & 142.72 & 195.34 \\
3DGS       & 90.90  & 195.73 & 58.55  & 215.59 \\
NeRF       & 132.92 & 190.44 & 105.96 & 236.46 \\
PVG        & 52.65  & 171.20 & 129.45 & 236.15 \\
\hline
\end{tabular}
\caption{Fréchet Inception Distance (FID $\downarrow$) comparison across models under different cross-domain rendering settings. Lower values indicate better image fidelity.}
\label{tab:fid_results}
\end{table}

\subsection{Additional Metrics}

In addition to standard NVS metrics, we also evaluate the Fréchet Inception Distance (FID) to measure distributional similarity between synthesized images and real images from the corresponding training domains. Specifically, we compute FID for the cross-view test sets $T_{C \rightarrow S}$ and $T_{D \rightarrow C}$ with respect to their corresponding training sequences $V^{C}$ and $V^{D}$. The results are summarized in \tabref{tab:fid_results}. 

Consistent with the trends observed in the ground-truth cross-vehicle evaluations, FID values increase as the viewpoint difference between training and test views becomes larger. This indicates that cross-platform rendering tasks introduce a larger domain gap compared to same-platform reconstructions.

However, the ranking induced by FID does not fully align with that obtained from standard NVS metrics. For example, PVG clearly outperforms MVSplat and DepthSplat on $T_{C \rightarrow S}$ in the \cartrain setting, yet it receives a lower ranking under FID. Similarly, FID ranks 3DGS below several feed-forward 3DGS variants despite its stronger performance according to PSNR, SSIM, and LPIPS. These discrepancies suggest that while FID can provide a coarse measure of distributional similarity under large viewpoint variations, it lacks the precision of standard NVS metrics that rely on ground-truth images. This observation further highlights the importance of the proposed dataset \MV, which enables reliable evaluation using paired ground-truth views.

\section{Dynamic Objects}
\label{suppsec:dynamic_objects}

Dynamic objects introduce significant challenges for view synthesis due to non-rigid motion, occlusion changes, and limited multi-view overlap. This section details how dynamic objects are identified in \MV, how reliable dynamic masks are obtained, and how NVS methods perform when evaluated specifically on moving agents.

\begin{figure}[t]
    \includegraphics[width=\columnwidth]{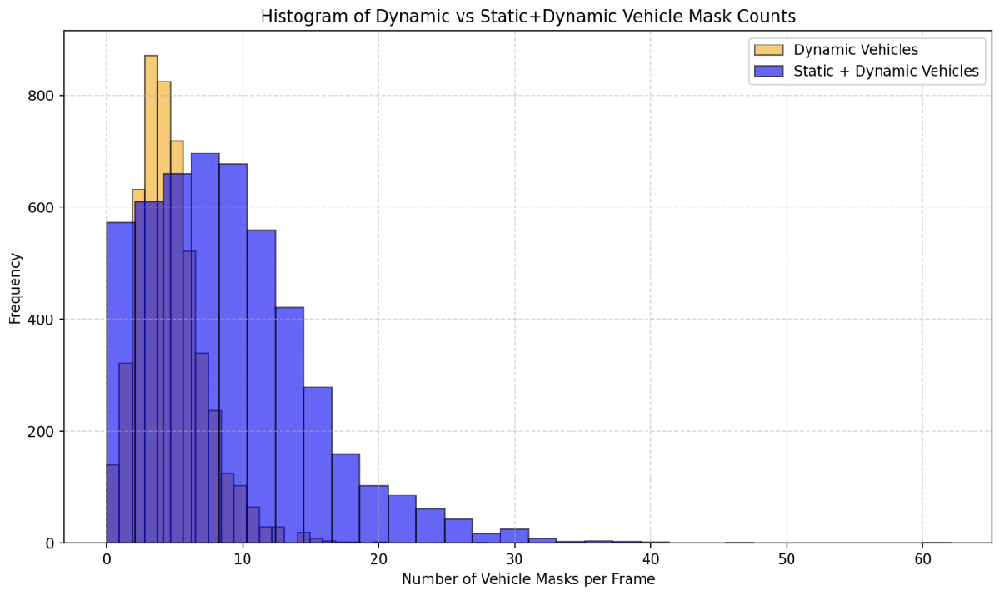}
    \caption{\textbf{Dynamic object distribution in \MV.}
    Histogram showing the relative frequency of dynamic vehicles compared to the total vehicle population. Dynamic agents represent a smaller portion of all vehicles, reflecting the natural imbalance in real driving scenarios and emphasizing the dataset’s realism for learning motion-aware reconstruction models.
    \label{fig:static_dynamic}}
\end{figure}
\setlength{\fboxrule}{1.5pt}
\begin{figure*}[ht!]
\centering
\begin{tabular}{ccc}

\includegraphics[width=0.31\textwidth]{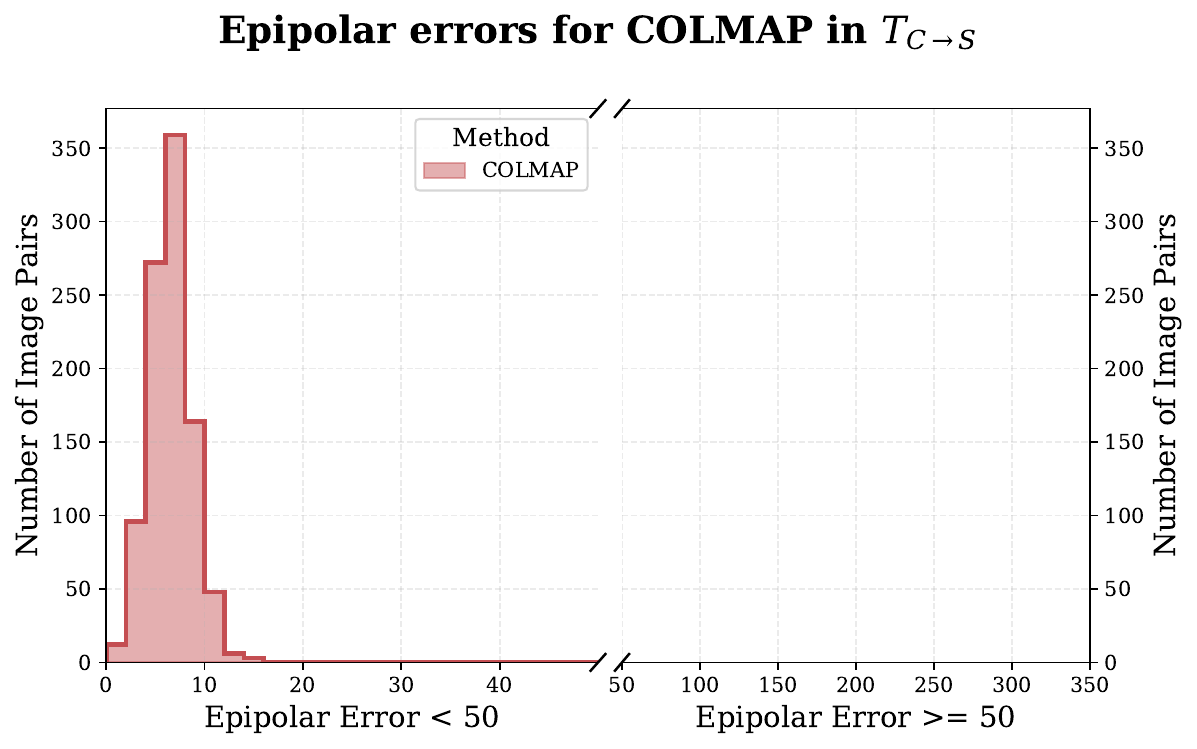} &
\includegraphics[width=0.31\textwidth]{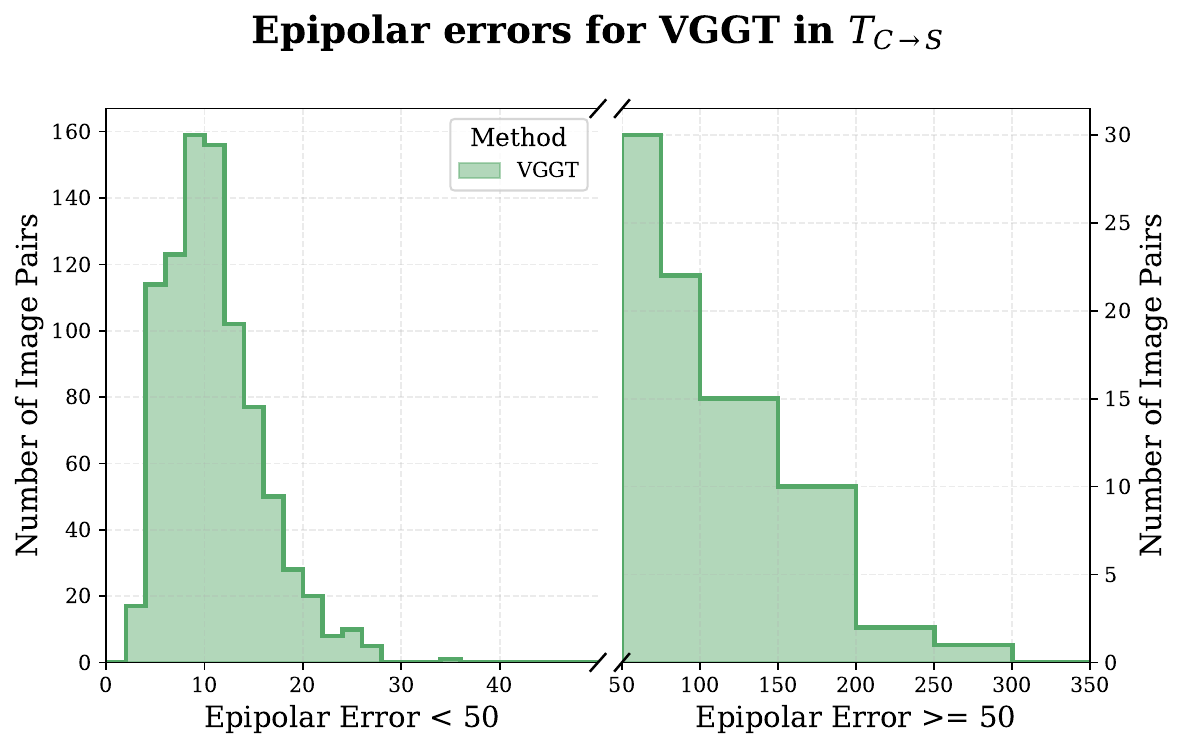} &
\includegraphics[width=0.31\textwidth]{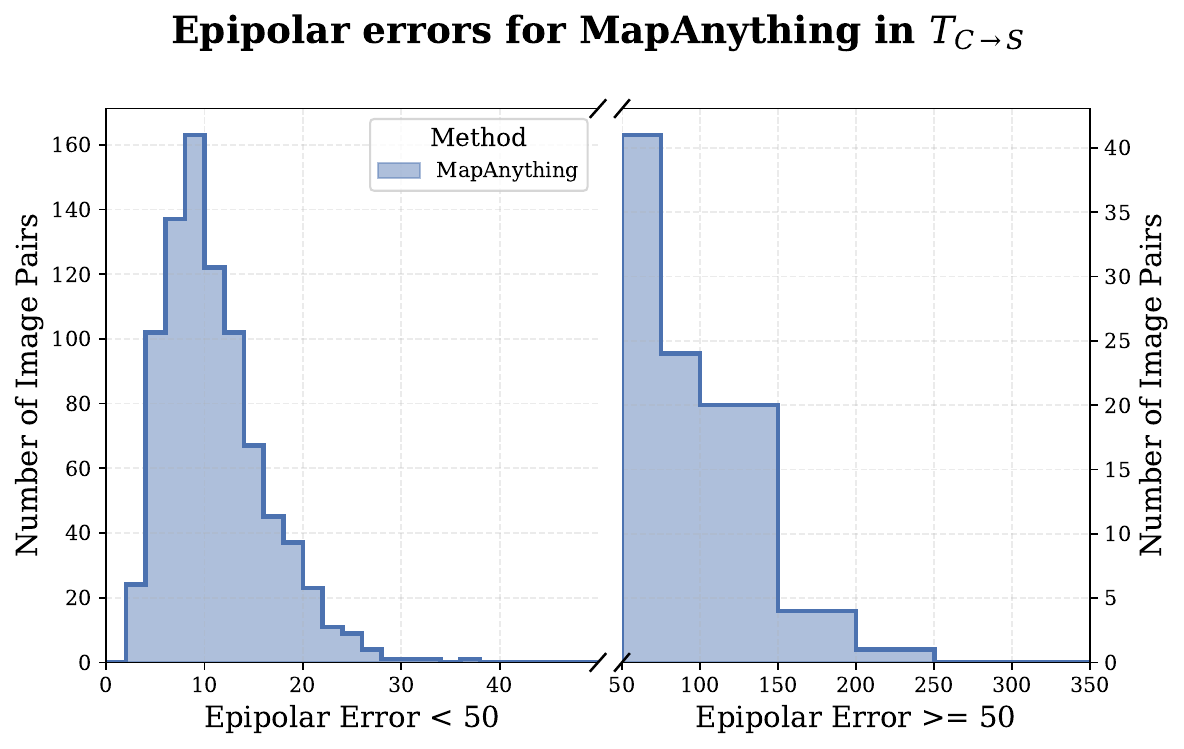} \\[3pt]

\end{tabular}
\caption{\textbf{Epipolar error distributions for $T_{C \rightarrow S}$.}
Comparison of COLMAP, MapAnything, and VGGT on the $T_{C \rightarrow S}$.
Errors are visualized across two ranges ($<50$ and $\geq 50$). 
COLMAP maintains consistently low errors, whereas MapAnything and VGGT exhibit heavier tails and numerous high-error outliers, indicating less reliable pose estimation.}

\label{fig:supply_cam_pose_car_train}
\vspace{-10pt}
\end{figure*}
\begin{figure*}[ht!]
\centering
\begin{tabular}{ccc}

\includegraphics[width=0.31\textwidth]{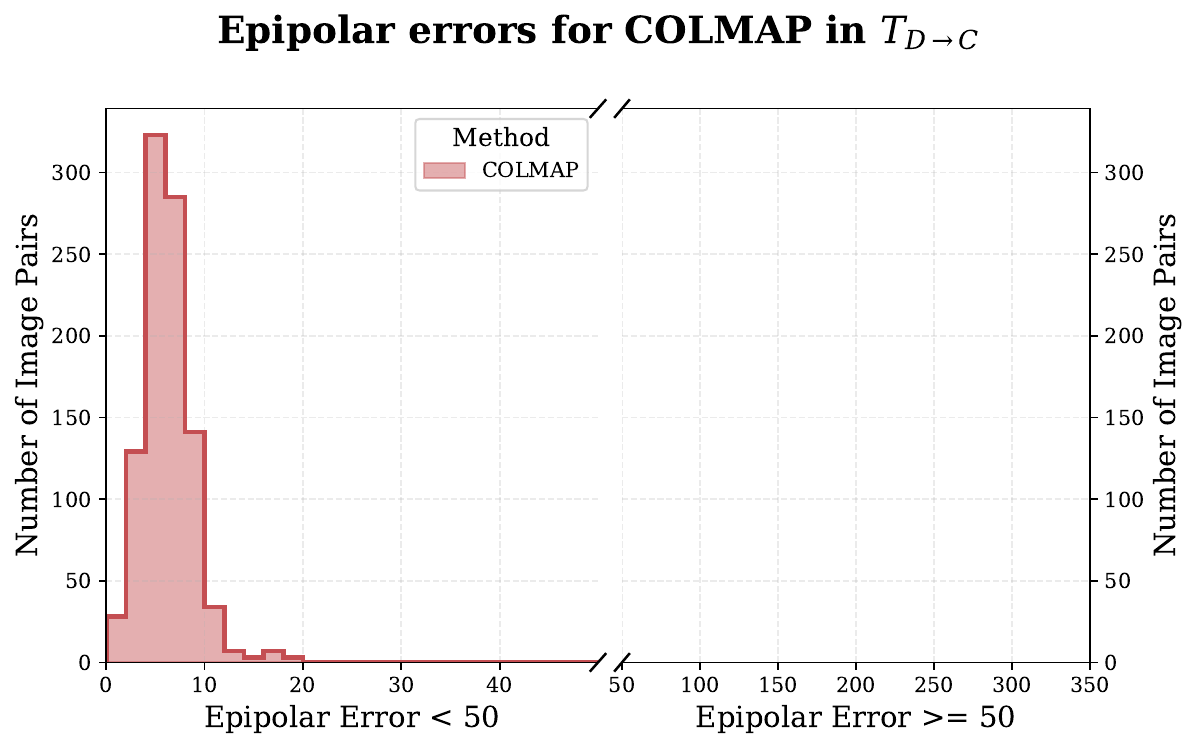} &
\includegraphics[width=0.31\textwidth]{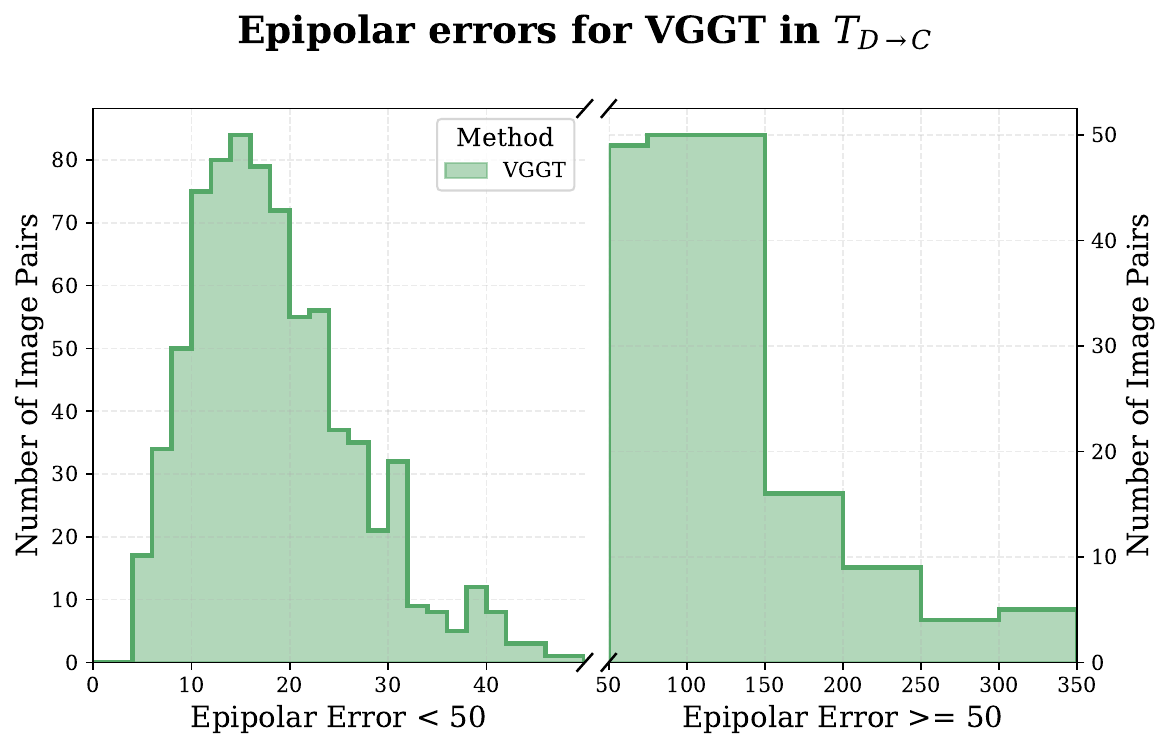} &
\includegraphics[width=0.31\textwidth]{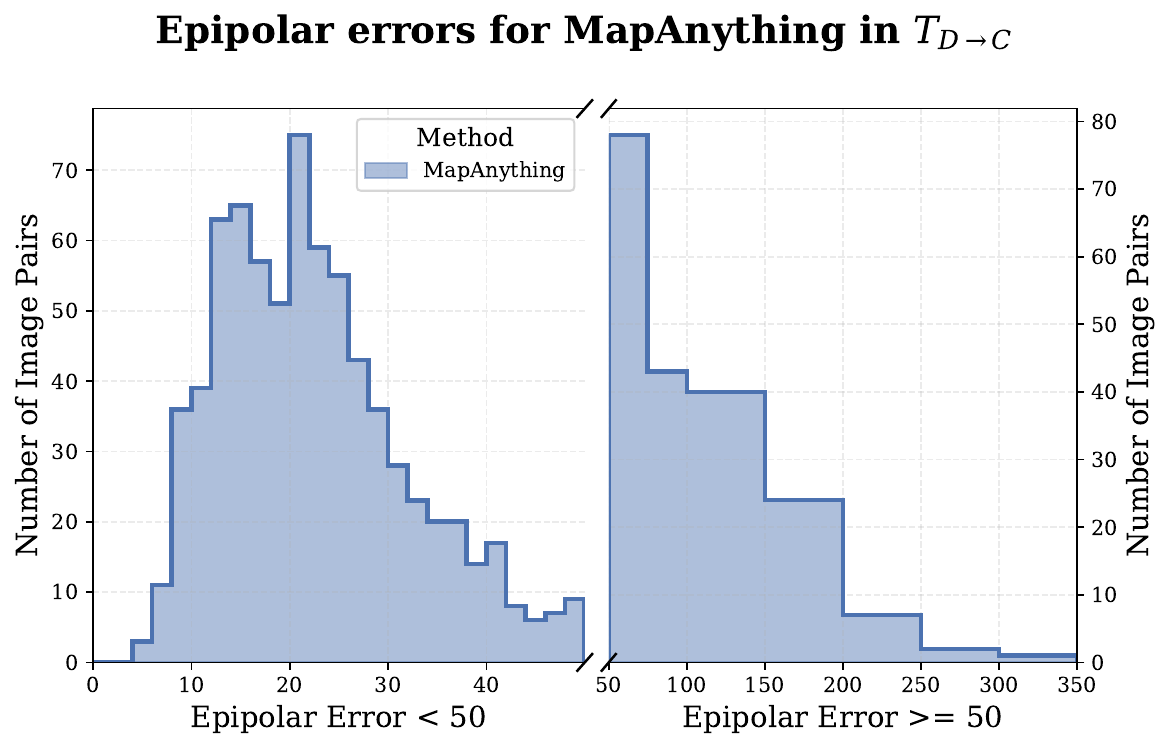} \\[3pt]

\end{tabular}
\caption{\textbf{Epipolar error distributions for $T_{D \rightarrow C}$.}
COLMAP, MapAnything, and VGGT evaluated on the $T_{D \rightarrow C}$ split.
The distributions are separated into the $<50$ and $\geq 50$ regions.
COLMAP performs robustly despite the large aerial–ground baseline, while MapAnything and VGGT produce significantly higher errors and long-tail failures.}

\label{fig:supply_cam_pose_drone_train}
\vspace{-10pt}
\end{figure*}

\subsection{Identifying Dynamic Objects}
Dynamic objects in \MV\ are defined as vehicles or pedestrians exhibiting measurable displacement between frames after compensating for camera motion. We use a combination of geometric cues and multi-view consistency to determine which instances are dynamic. As illustrated in \figref{fig:static_dynamic}, dynamic agents constitute only a subset of all visible vehicles, providing a realistic motion distribution for evaluating NVS methods under natural driving conditions.

\subsection{Dynamic Mask Generation}

Accurately segmenting dynamic objects is critical for isolating motion-sensitive regions during evaluation. We leverage both DroneSplat\cite{dronesplat} and SAM\cite{kirillov2023segment} to produce reliable dynamic masks.

\textbf{DroneSplat\cite{dronesplat} Motion Masks.}
DroneSplat\cite{dronesplat} provides pixel-level motion likelihoods using optical-flow cues and multi-view geometric constraints. However, its output can contain spurious activations and isolated outlier pixels, particularly near object boundaries or during rapid camera motion.

\textbf{SAM\cite{kirillov2023segment} Segmentation.}
Prompting SAM\cite{kirillov2023segment} with category queries such as “car,” “two-wheeler,” “truck,” “pedestrian,” and “auto” yields high-quality instance segmentation masks for all visible traffic agents. However, SAM\cite{kirillov2023segment} alone does not distinguish between static and dynamic instances, and therefore includes parked or stationary vehicles (\cf~\figref{fig:static_dynamic}).

\textbf{Intersection-Based Dynamic Masks.}
To obtain clean and temporally stable dynamic-object masks, we intersect SAM\cite{kirillov2023segment}'s category-based instance masks with DroneSplat\cite{dronesplat}’s motion predictions:
\[
\mathcal{M}_{\text{dyn}} = \mathcal{M}_{\text{SAM}} \cap \mathcal{M}_{\text{DroneSplat}},
\]
where $\mathcal{M}_{\text{SAM}}$ denotes pixels belonging to SAM-detected vehicles and pedestrians, and $\mathcal{M}_{\text{DroneSplat}}$ represents pixels flagged as dynamic.  
This intersection removes DroneSplat\cite{dronesplat} outliers while filtering out static SAM\cite{kirillov2023segment} segments, producing robust dynamic masks suitable for both qualitative and quantitative dynamic-object NVS evaluation.

We compare no masks, DS+SAM pseudo masks, and GT dynamic-object masks in Table~\ref{tab:mask_ablation} of the main paper is
performed on the $T_{C\rightarrow S}$ split using PVG. The \MV sequence IDs are:
\{1, 5, 10, 12, 17, 20, 22, 26, 30, 32, 37, 40, 42, 46, 49\}.

\subsection{Dynamic Mask Illustration}
\figref{fig:dynamic_object_example_image} provides a visual example of the dynamic-mask generation pipeline. This combined representation effectively disentangles static and dynamic agents and ensures that motion-specific errors are measured only on the relevant regions.

Overall, dynamic-object evaluation exposes fine-grained failure modes that are not apparent from whole-image metrics. These results highlight the need for motion-aware scene representations that can handle non-static content, which is essential for downstream tasks such as simulation, prediction, and autonomous driving evaluation.

\section{Qualitative Figures}
\label{suppsec:qualitative_figs}
This section provides extended qualitative comparisons across models and viewpoints under various sensor setups, complementing the quantitative results presented in the main paper. For each evaluation setup, we visualize predictions from multiple NVS methods alongside the ground-truth image and report their corresponding perceptual metric values. These examples highlight the strengths and weaknesses of each method under varying sensor viewpoints.

\begin{figure*}[ht!]
\centering
\scriptsize
\setlength{\tabcolsep}{1pt}
\renewcommand{\arraystretch}{0.8}

\begin{adjustbox}{max width=\textwidth}
\begin{tabular}{ccccc}

\textbf{GT} & \textbf{3DGS} & \textbf{NeRF} & \textbf{DepthSplat} & \textbf{PVG} \\

\includegraphics[width=0.19\linewidth]{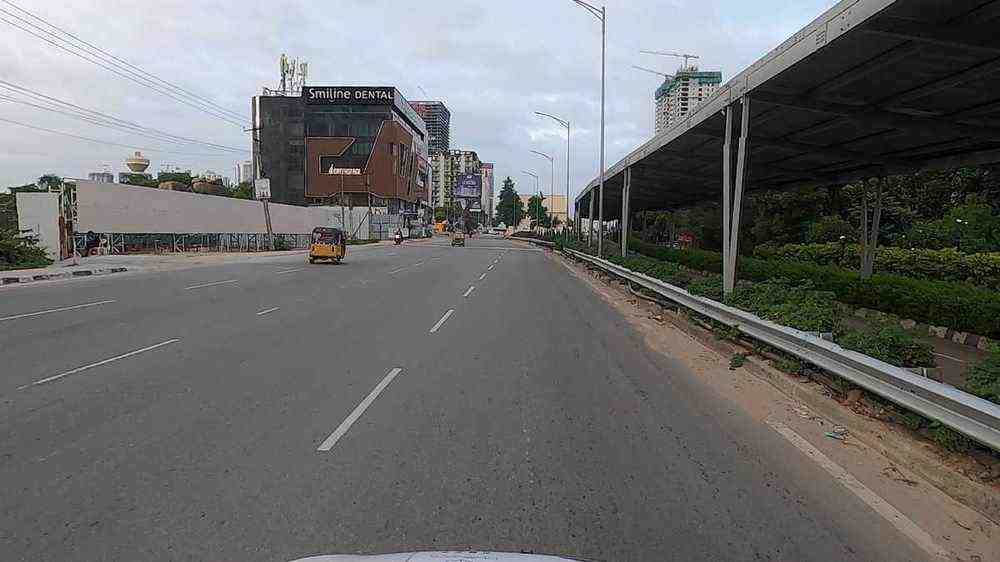} & \includegraphics[width=0.19\linewidth]{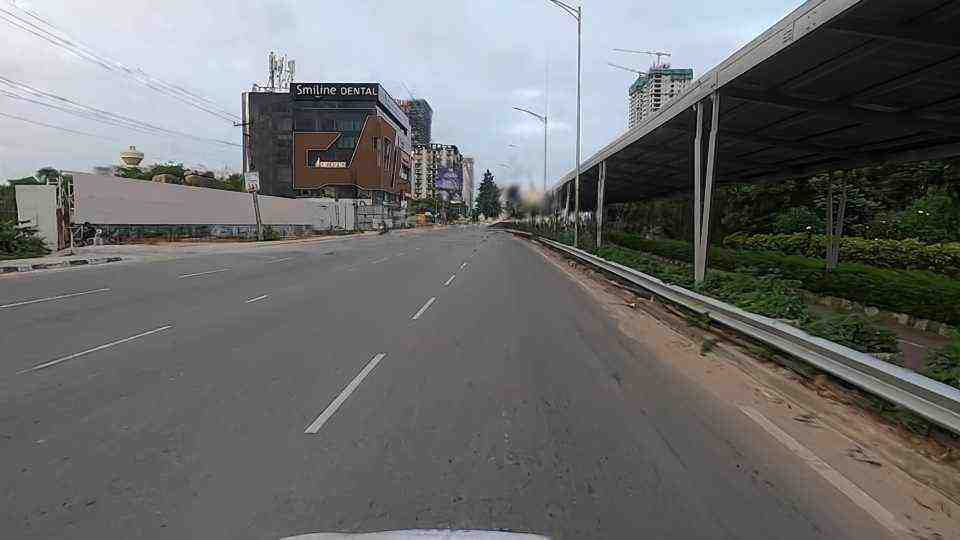} & \includegraphics[width=0.19\linewidth]{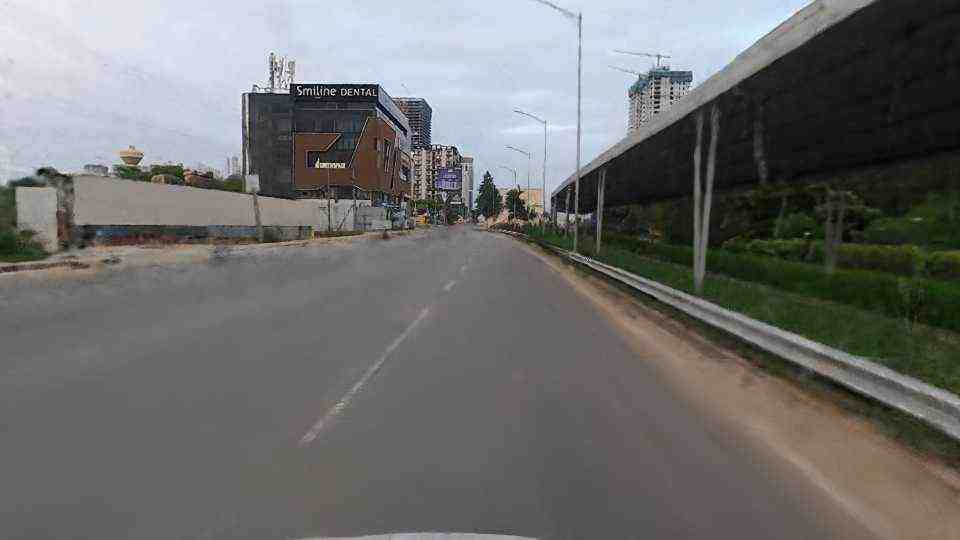} & \includegraphics[width=0.19\linewidth]{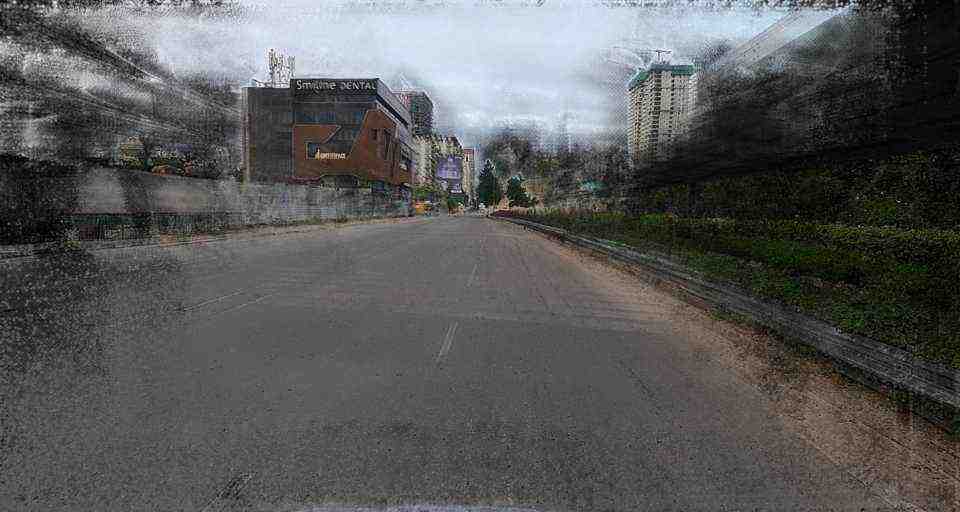} & \includegraphics[width=0.19\linewidth]{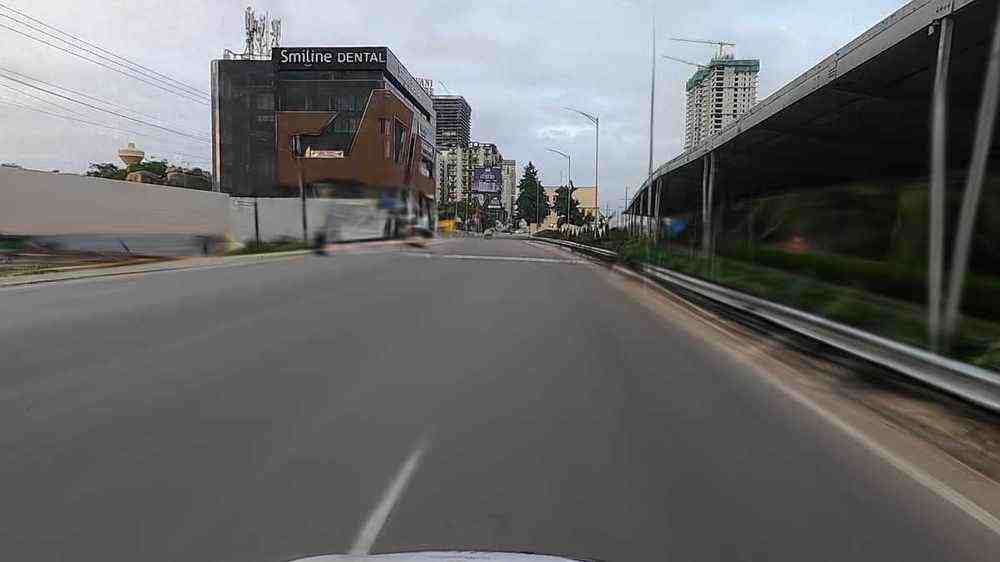} \\
& 21.45$|$0.68$|$0.55 & 15.15$|$0.59$|$0.64 & 15.02$|$0.54$|$0.56 & 26.65$|$0.82$|$0.35 \\
\includegraphics[width=0.19\linewidth]{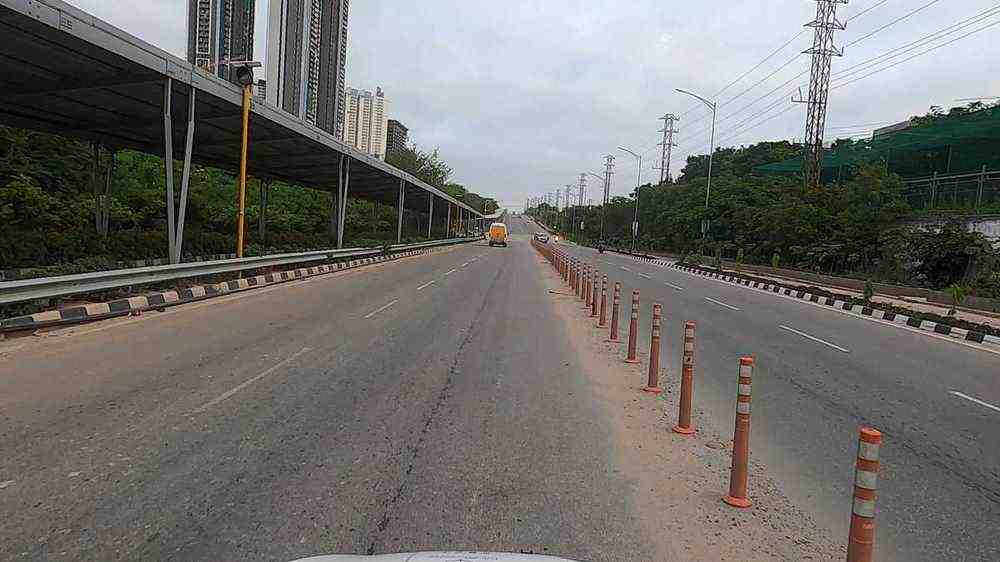} & \includegraphics[width=0.19\linewidth]{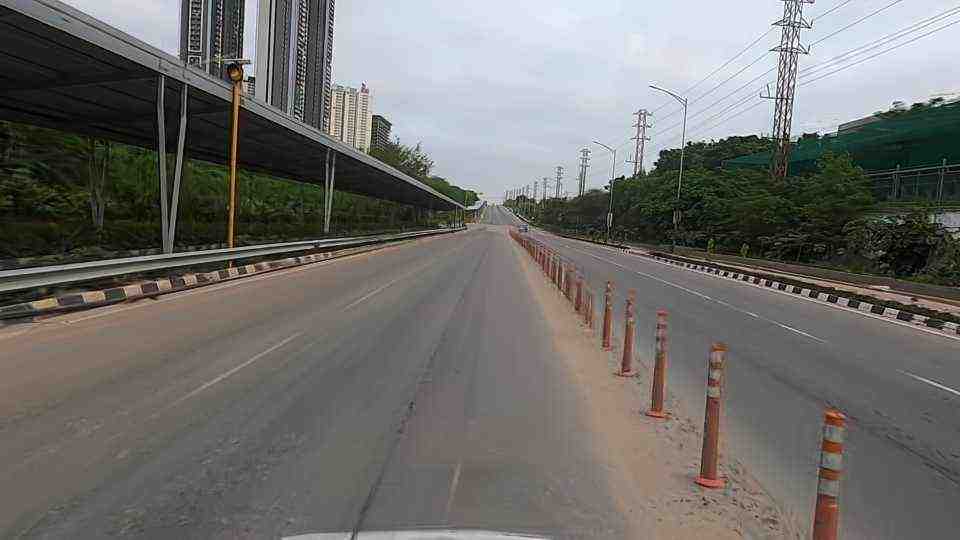} & \includegraphics[width=0.19\linewidth]{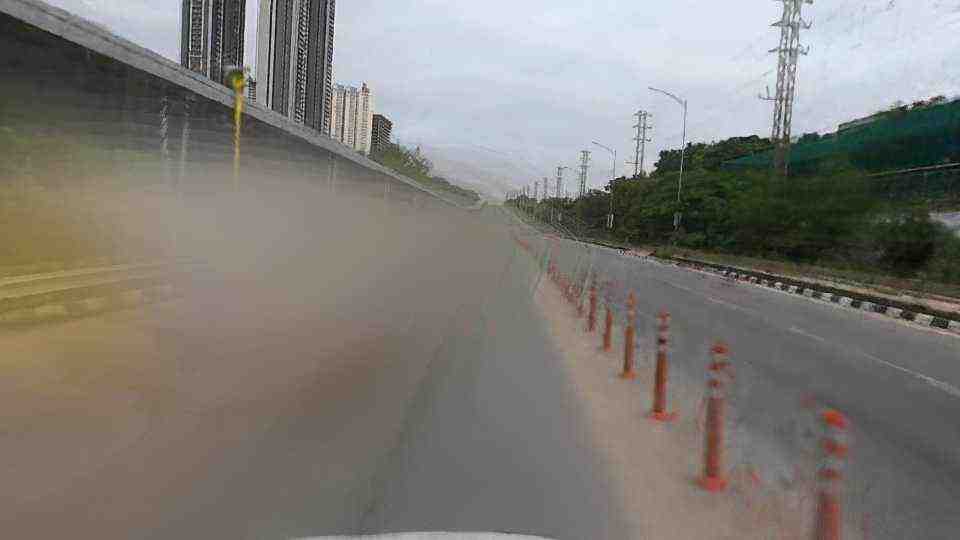} & \includegraphics[width=0.19\linewidth]{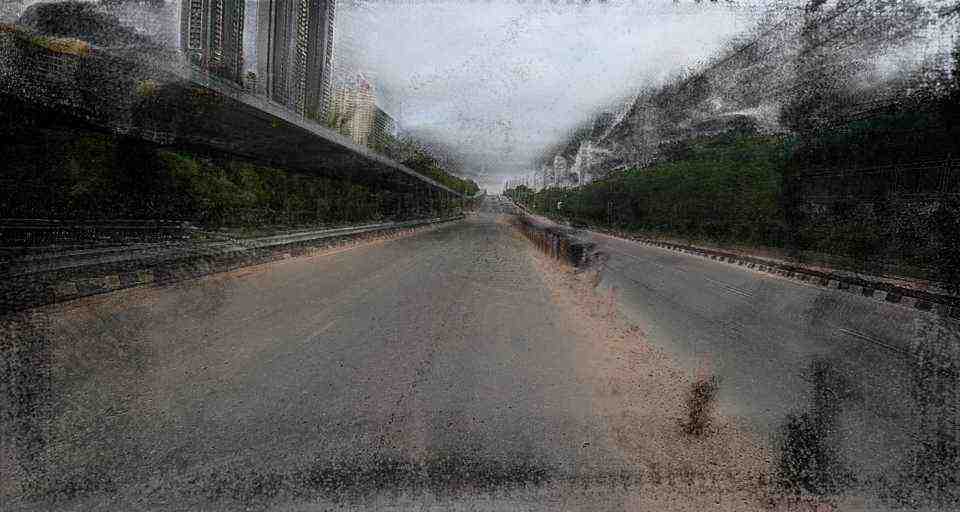} & \includegraphics[width=0.19\linewidth]{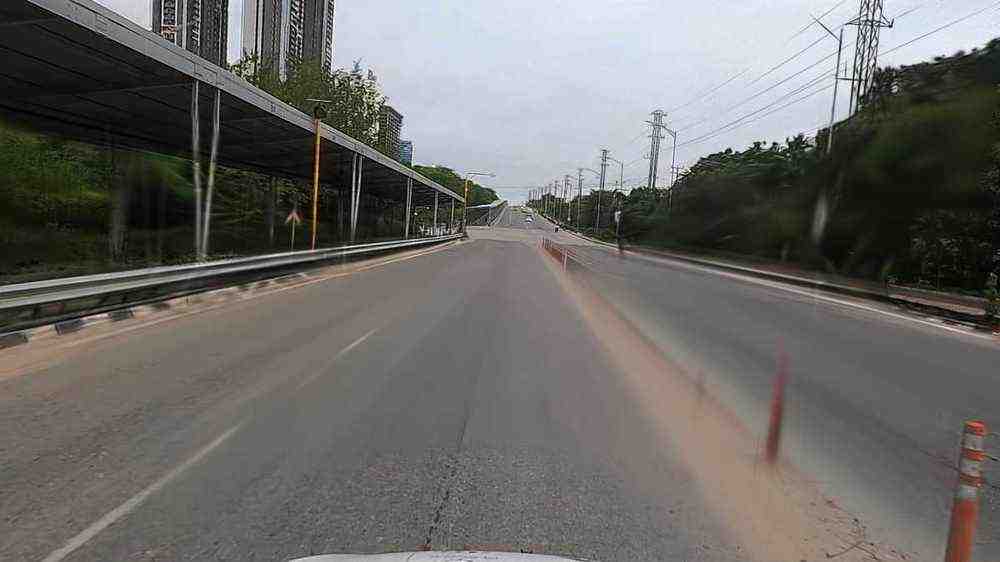} \\
& 16.93$|$0.58$|$0.62 & 15.94$|$0.52$|$0.58 & 15.10$|$0.44$|$0.58 & 26.44$|$0.75$|$0.42 \\
\includegraphics[width=0.19\linewidth]{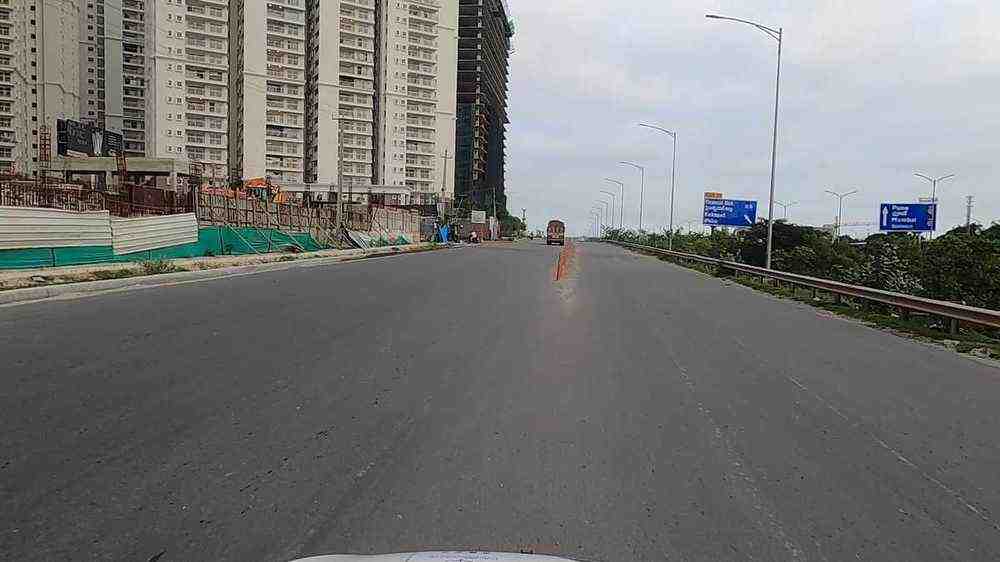} & \includegraphics[width=0.19\linewidth]{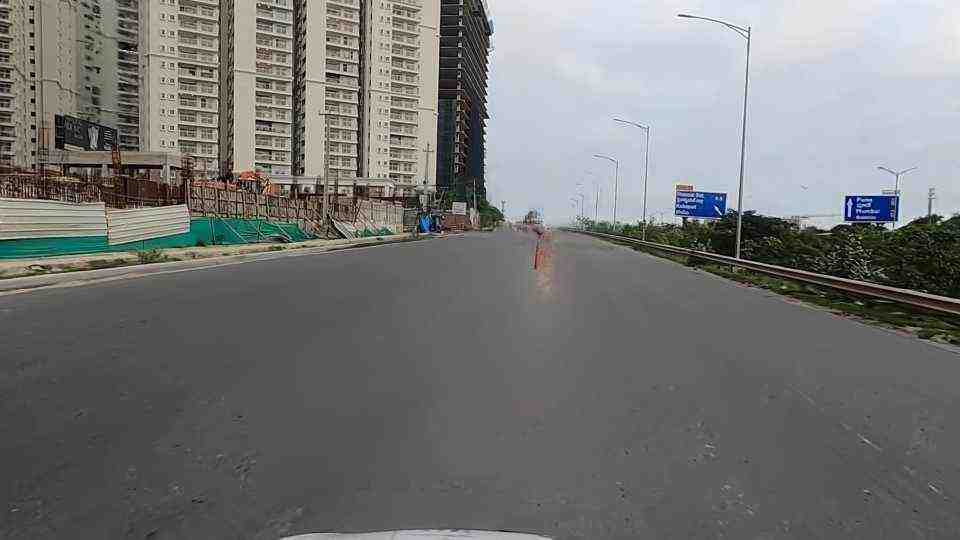} & \includegraphics[width=0.19\linewidth]{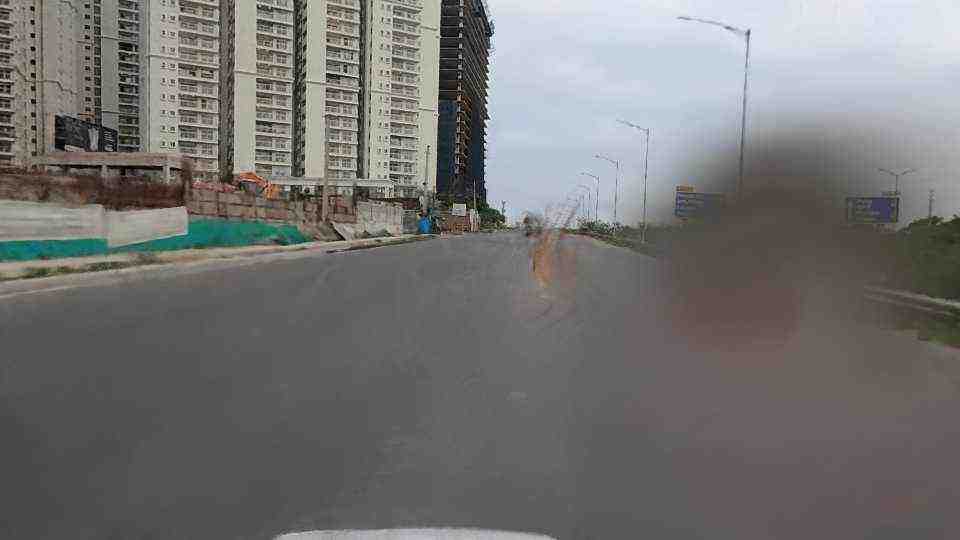} & \includegraphics[width=0.19\linewidth]{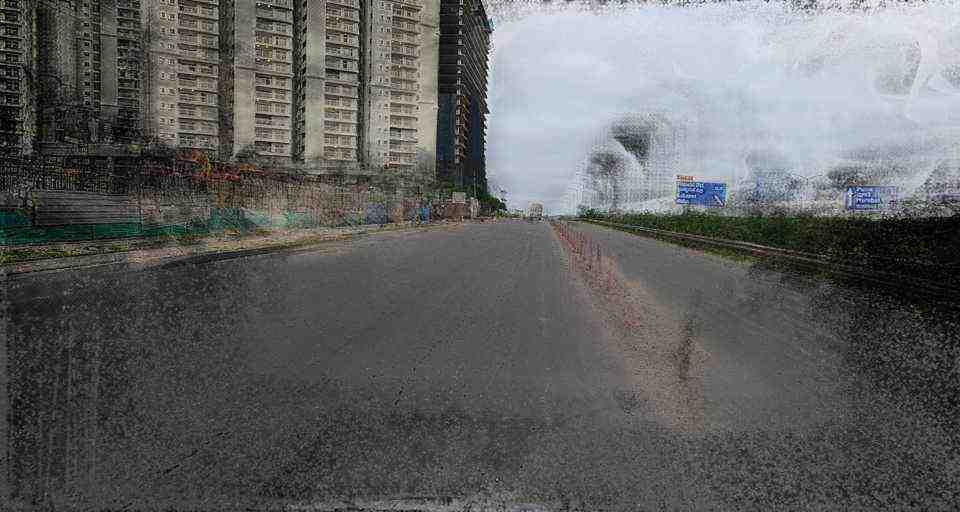} & \includegraphics[width=0.19\linewidth]{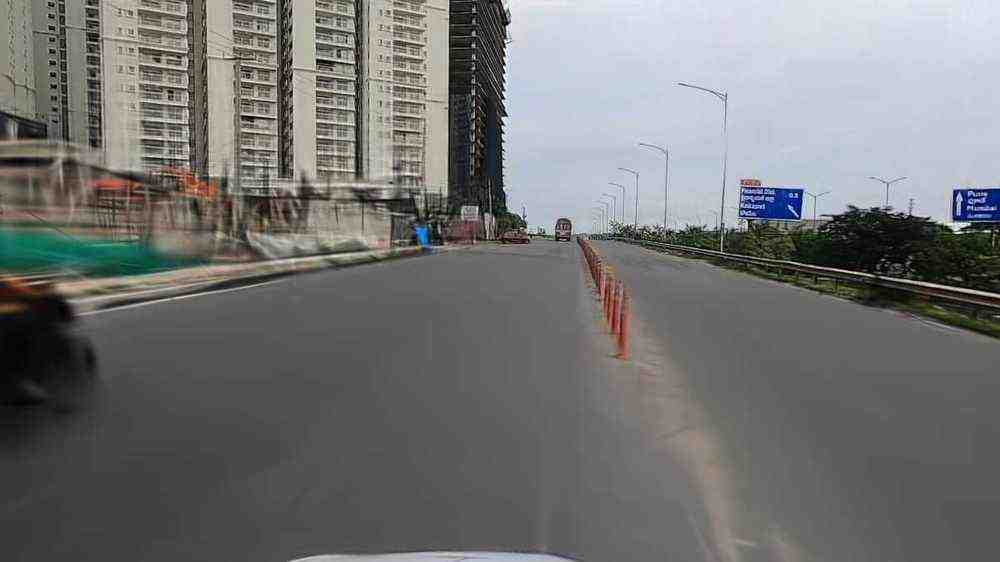} \\
& 19.22$|$0.72$|$0.46 & 16.11$|$0.62$|$0.56 & 17.21$|$0.58$|$0.52 & 27.08$|$0.86$|$0.33 \\
\includegraphics[width=0.19\linewidth]{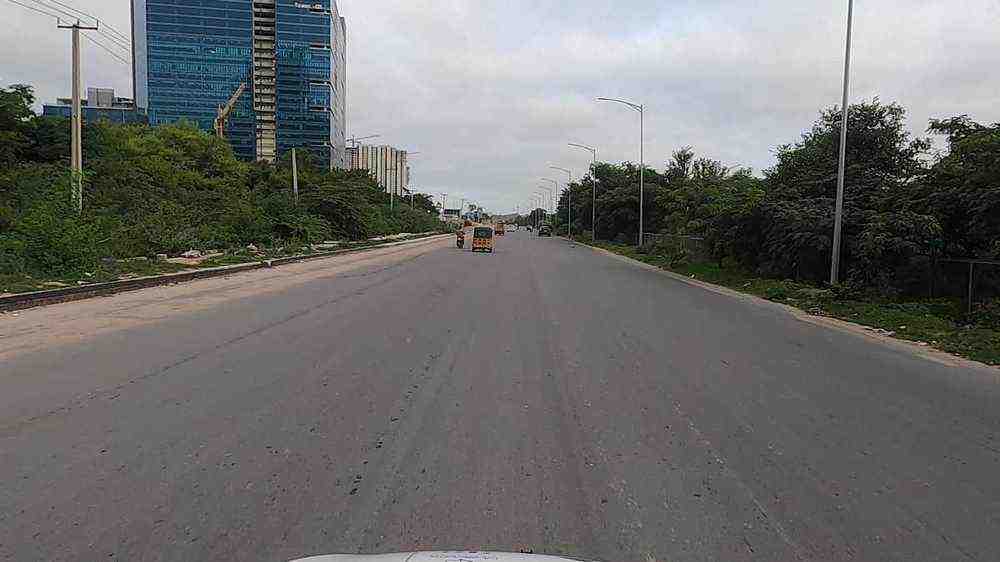} & \includegraphics[width=0.19\linewidth]{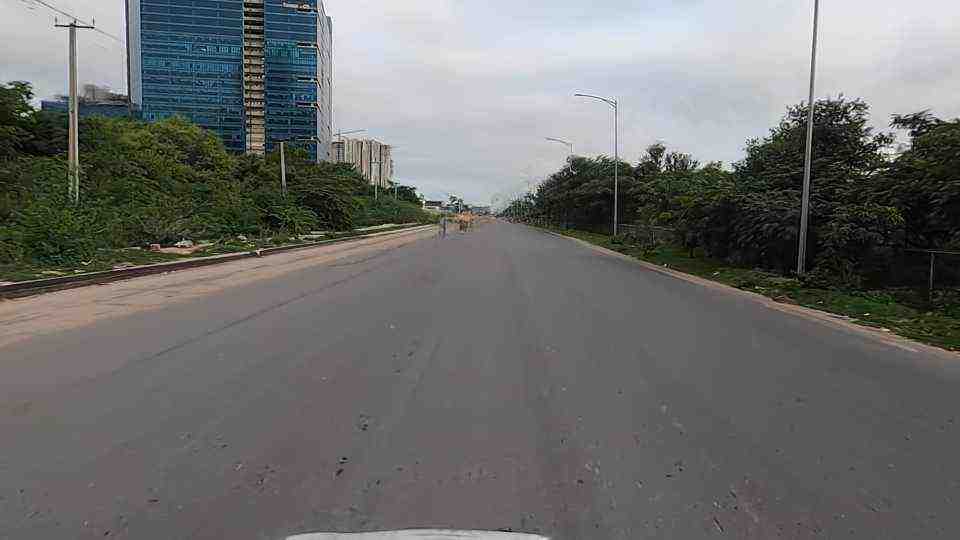} & \includegraphics[width=0.19\linewidth]{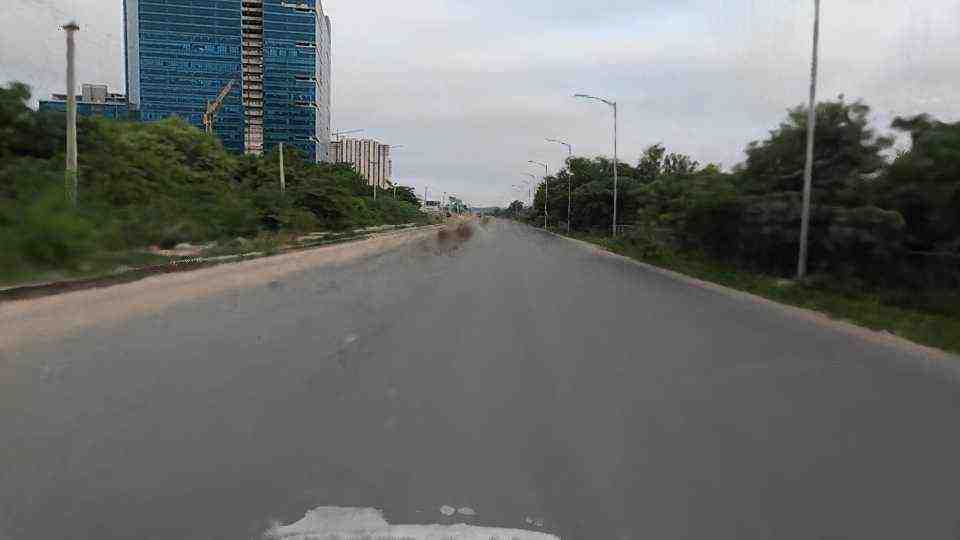} & \includegraphics[width=0.19\linewidth]{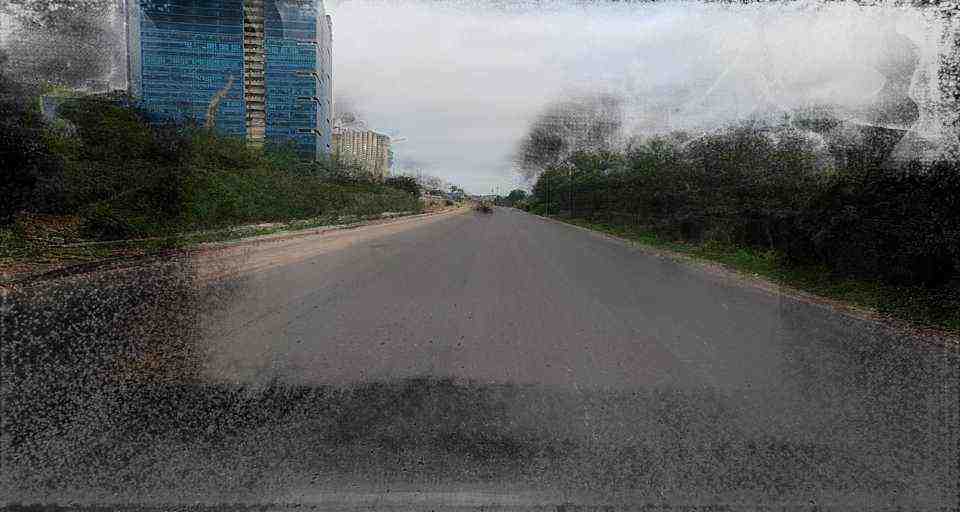} & \includegraphics[width=0.19\linewidth]{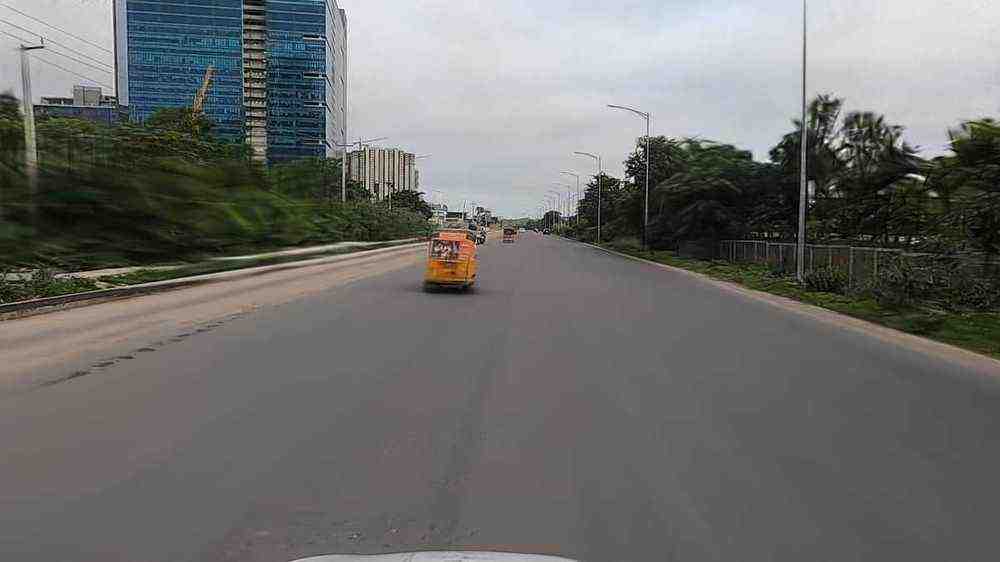} \\
& 23.40$|$0.71$|$0.50 & 18.60$|$0.63$|$0.55 & 16.71$|$0.54$|$0.56 & 27.75$|$0.84$|$0.38 \\
\includegraphics[width=0.19\linewidth]{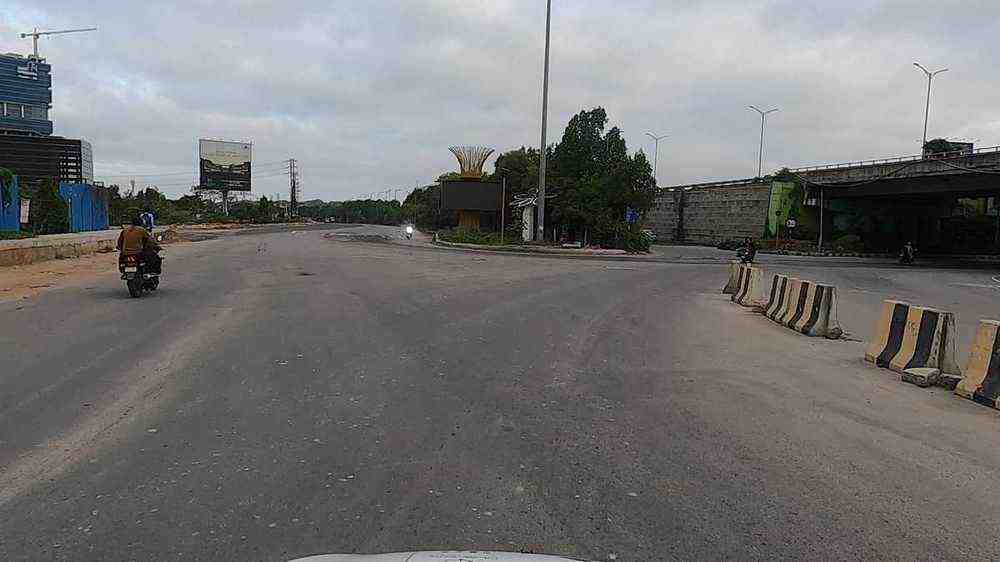} & \includegraphics[width=0.19\linewidth]{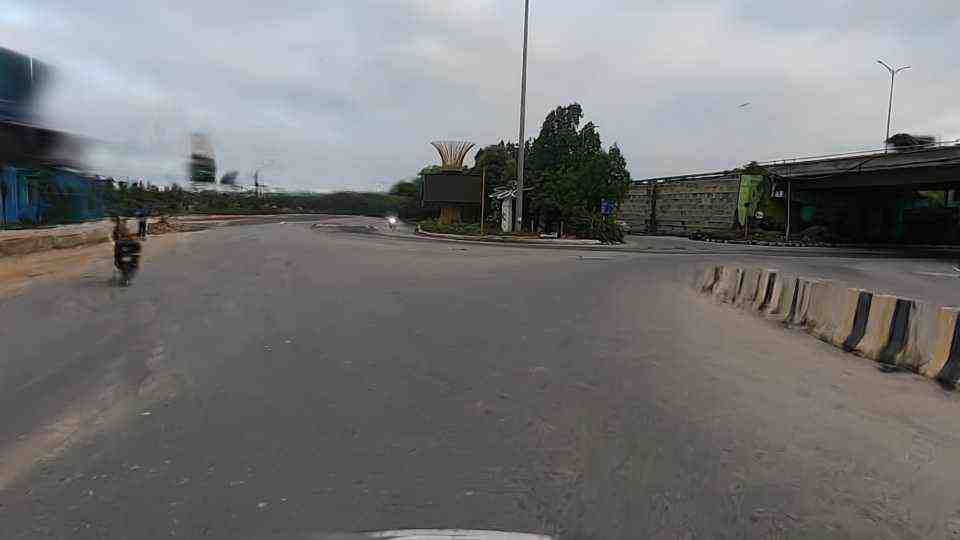} & \includegraphics[width=0.19\linewidth]{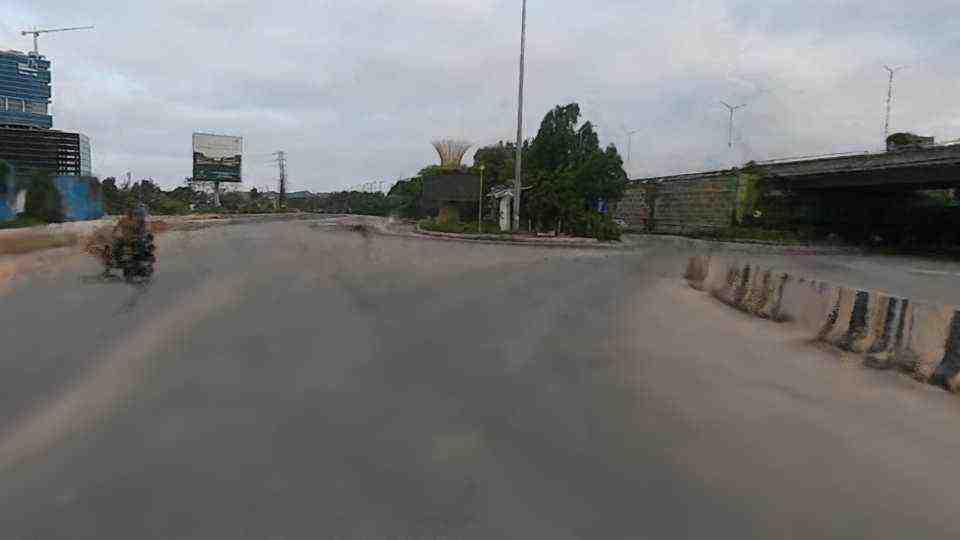} & \includegraphics[width=0.19\linewidth]{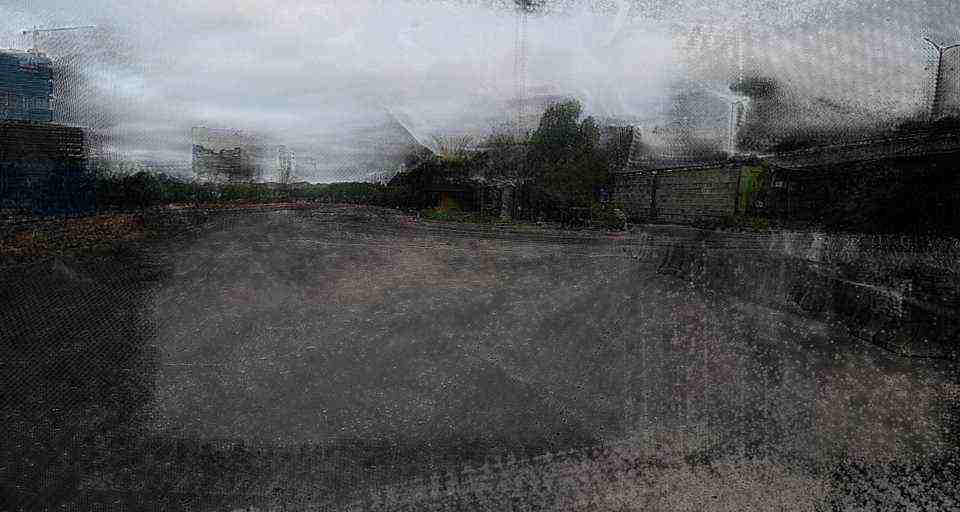} & \includegraphics[width=0.19\linewidth]{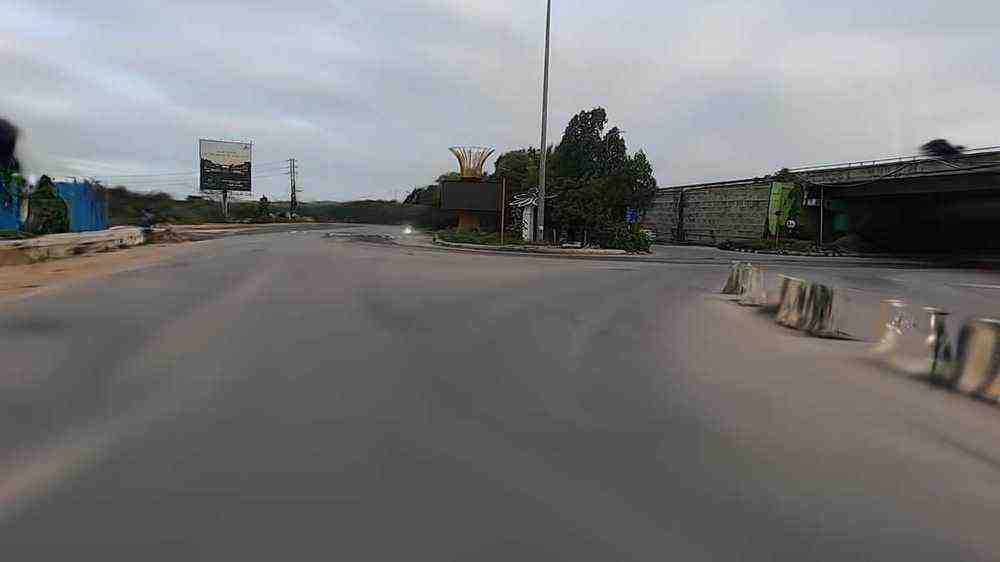} \\
& 23.12$|$0.71$|$0.55 & 21.63$|$0.77$|$0.52 & 15.04$|$0.50$|$0.59 & 24.71$|$0.77$|$0.47 \\
\includegraphics[width=0.19\linewidth]{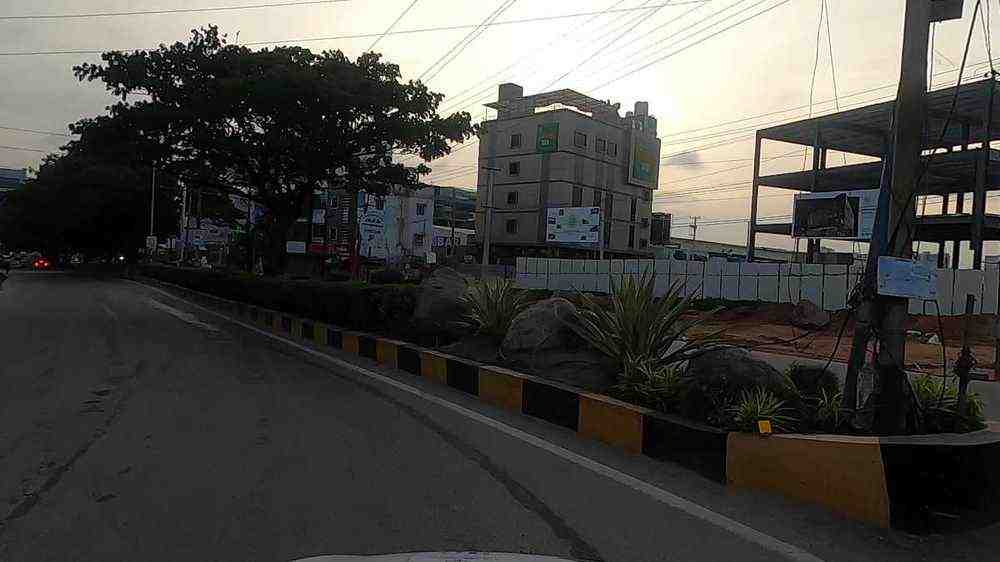} & \includegraphics[width=0.19\linewidth]{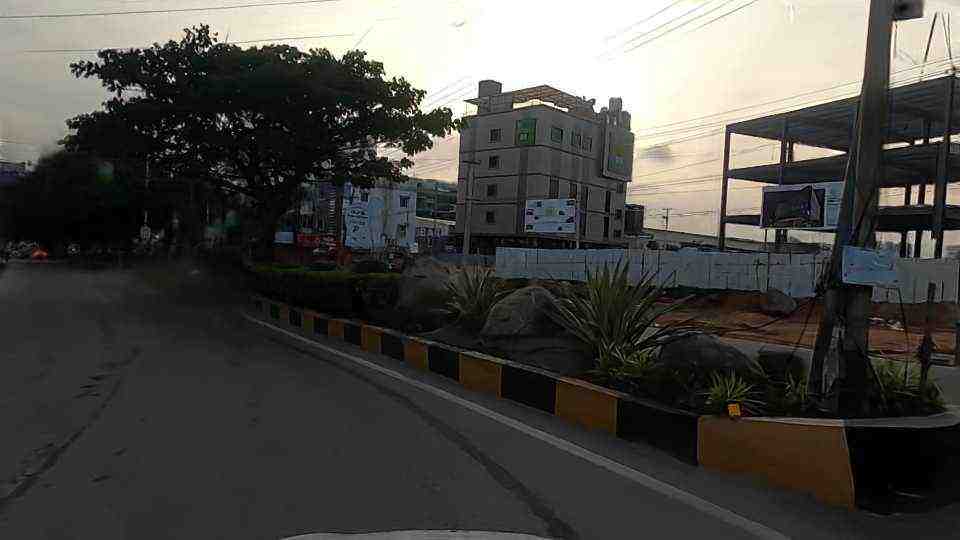} & \includegraphics[width=0.19\linewidth]{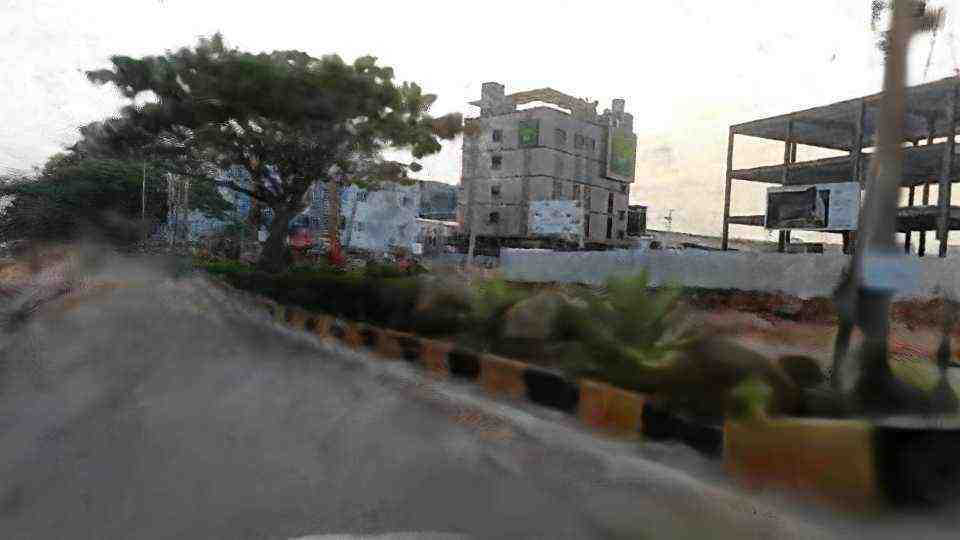} & \includegraphics[width=0.19\linewidth]{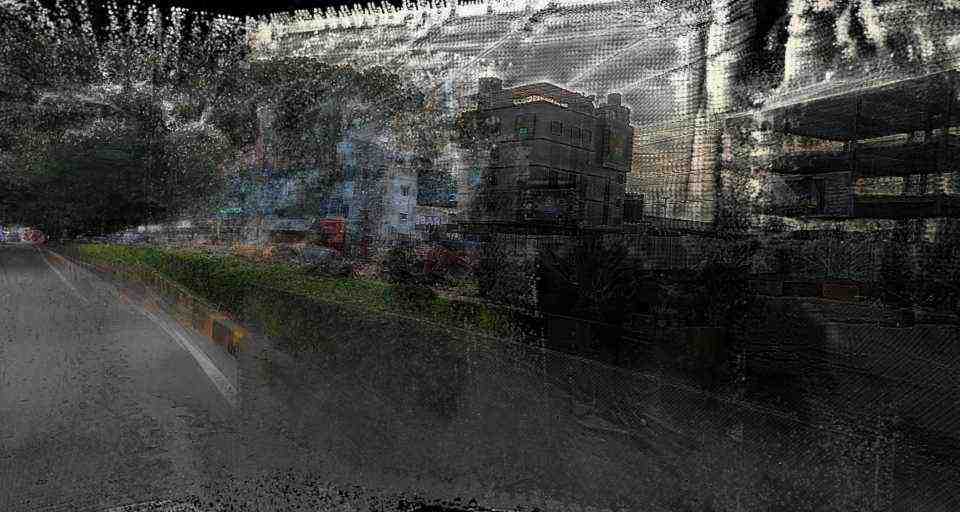} & \includegraphics[width=0.19\linewidth]{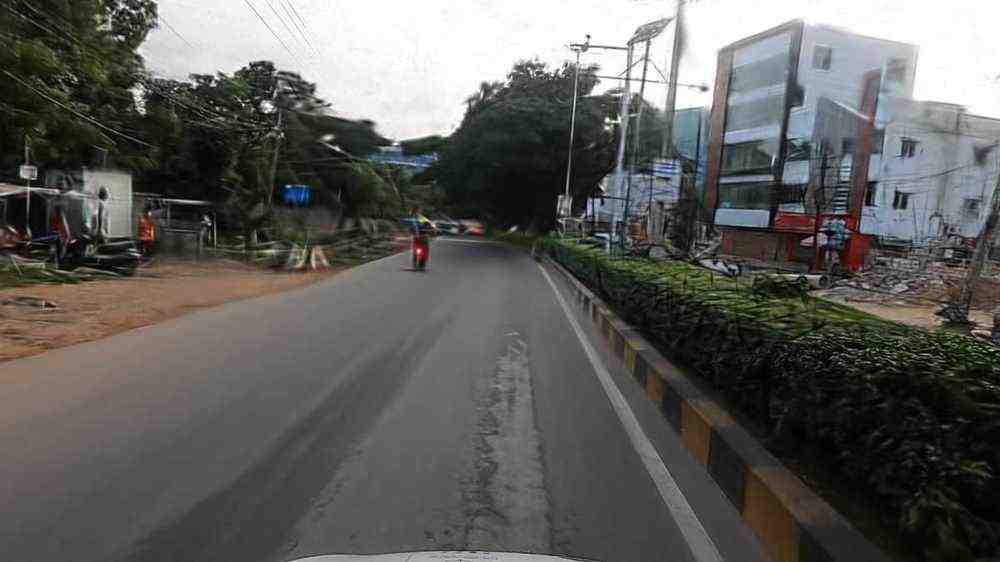} \\
& 13.92$|$0.51$|$0.64 & 10.89$|$0.39$|$0.75 & 12.18$|$0.36$|$0.74 & 26.69$|$0.82$|$0.36 \\
\includegraphics[width=0.19\linewidth]{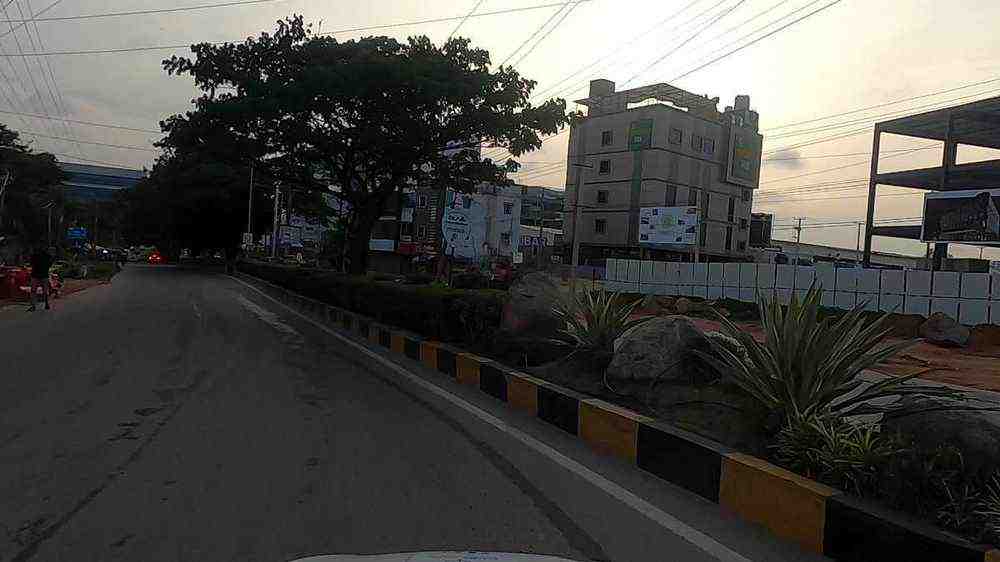} & \includegraphics[width=0.19\linewidth]{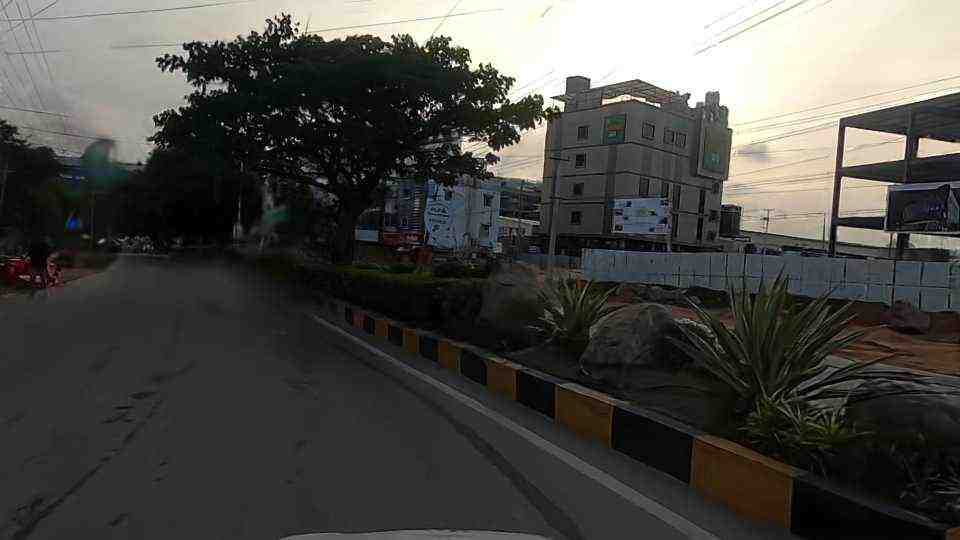} & \includegraphics[width=0.19\linewidth]{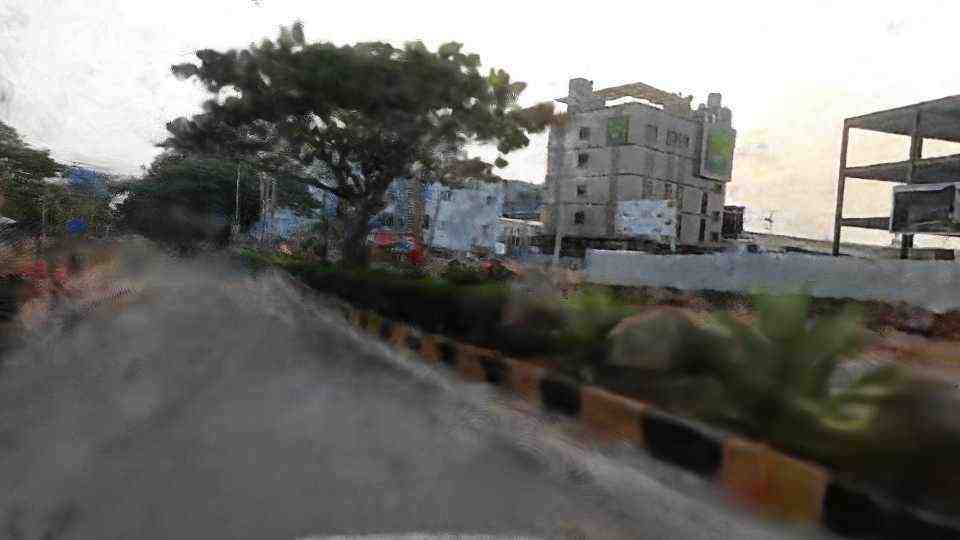} & \includegraphics[width=0.19\linewidth]{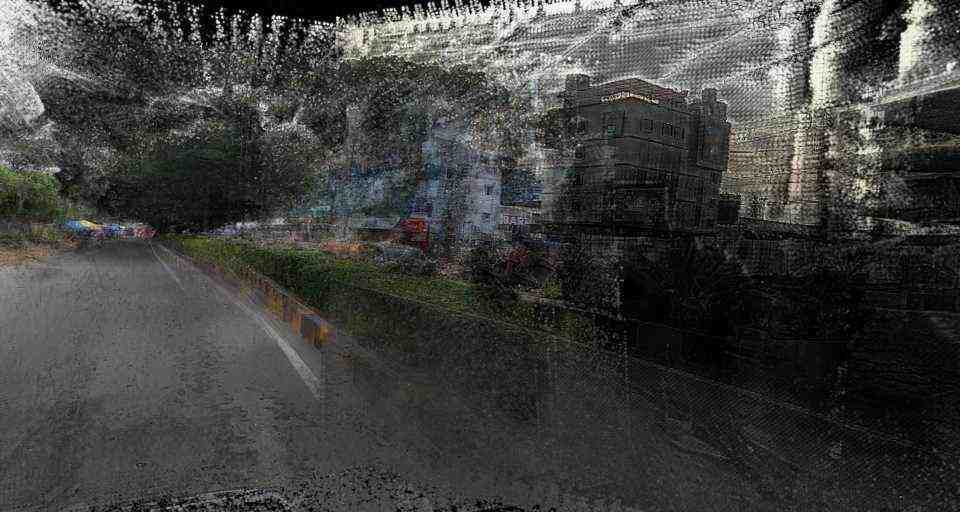} & \includegraphics[width=0.19\linewidth]{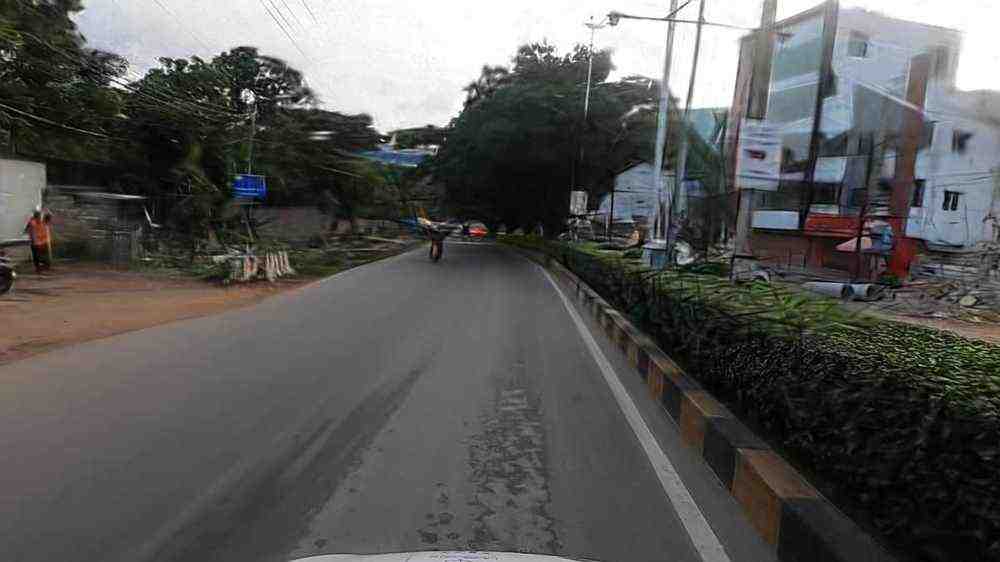} \\
& 14.08$|$0.52$|$0.64 & 10.99$|$0.41$|$0.73 & 11.91$|$0.34$|$0.75 & 26.85$|$0.82$|$0.35 \\
\includegraphics[width=0.19\linewidth]{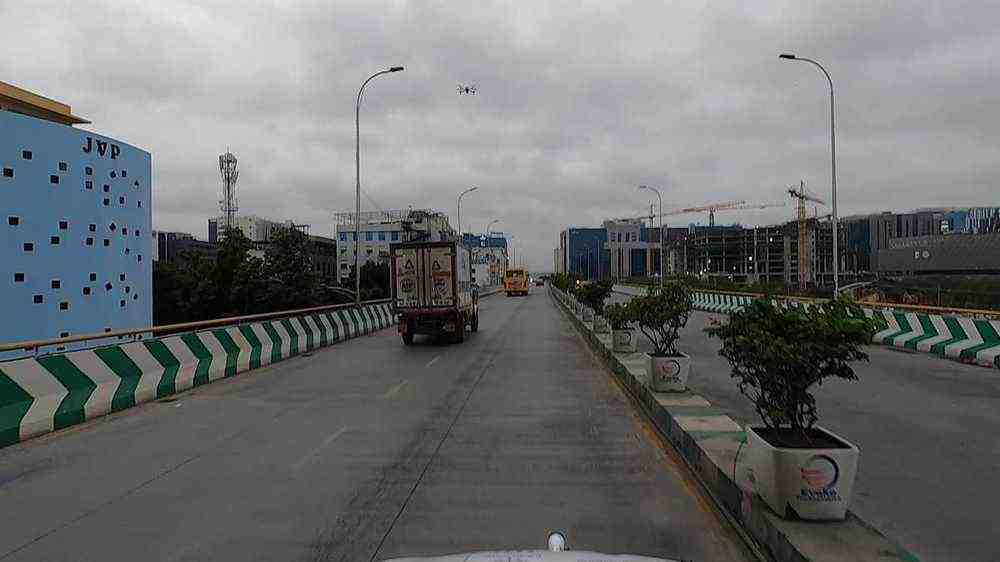} & \includegraphics[width=0.19\linewidth]{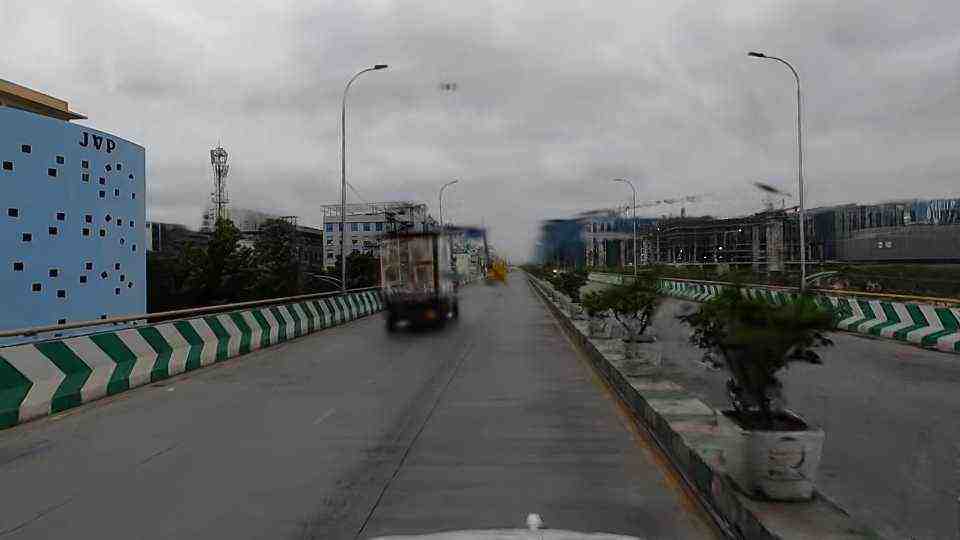} & \includegraphics[width=0.19\linewidth]{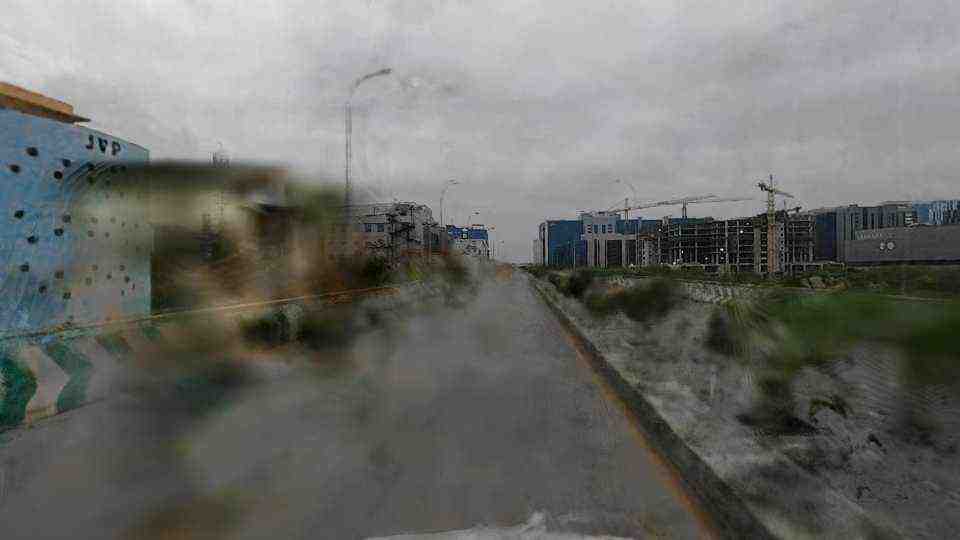} & \includegraphics[width=0.19\linewidth]{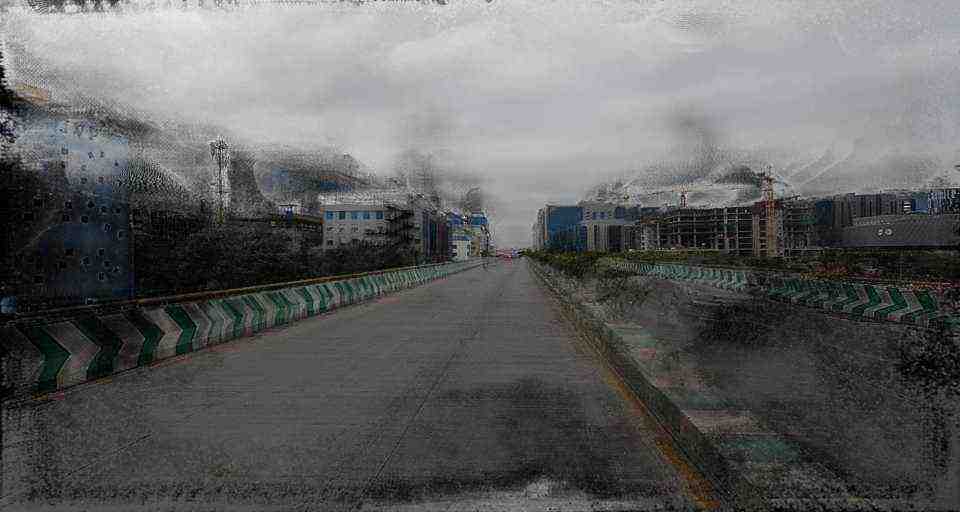} & \includegraphics[width=0.19\linewidth]{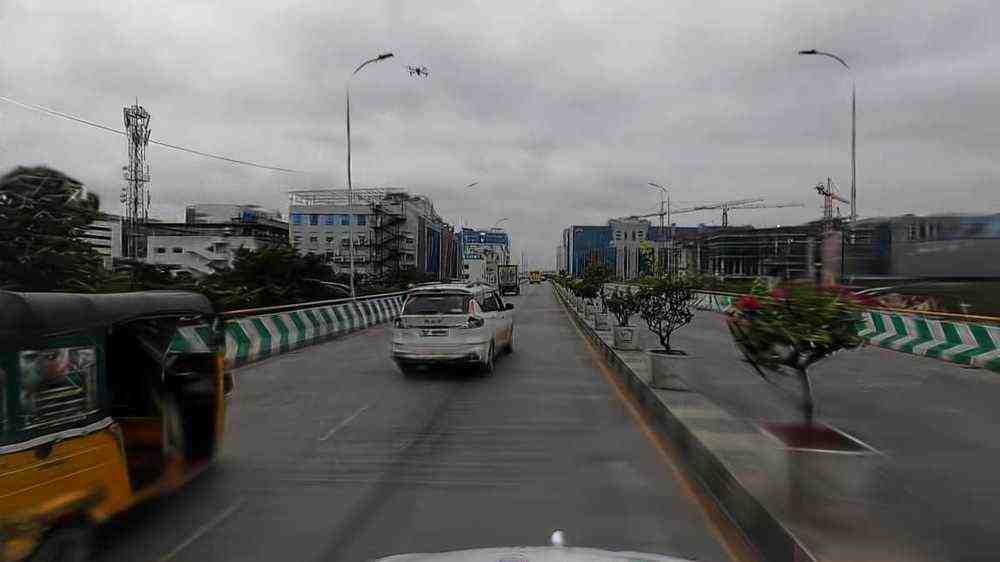} \\
& 19.38$|$0.70$|$0.52 & 16.77$|$0.62$|$0.53 & 17.61$|$0.62$|$0.57 & 28.10$|$0.85$|$0.34 \\
\includegraphics[width=0.19\linewidth]{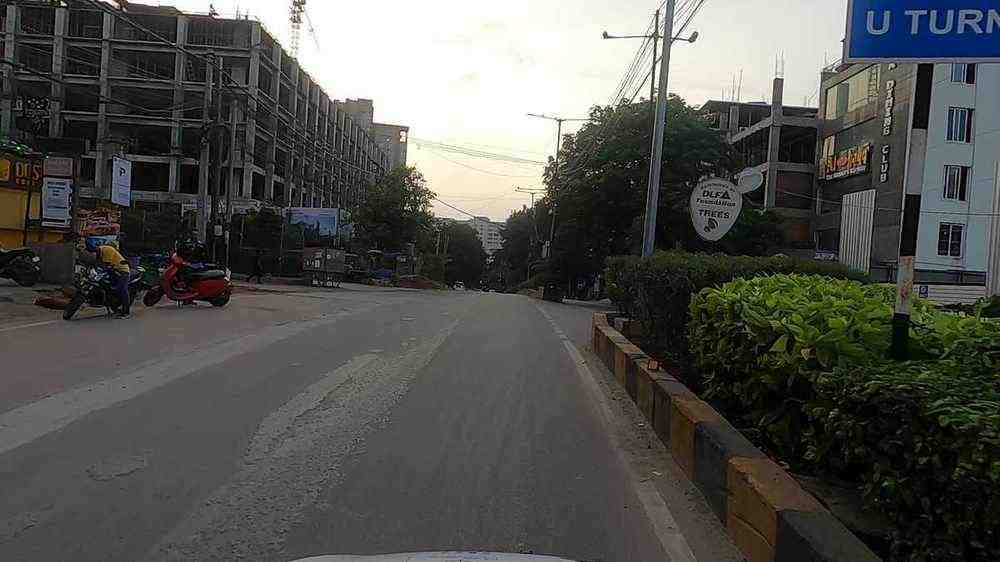} & \includegraphics[width=0.19\linewidth]{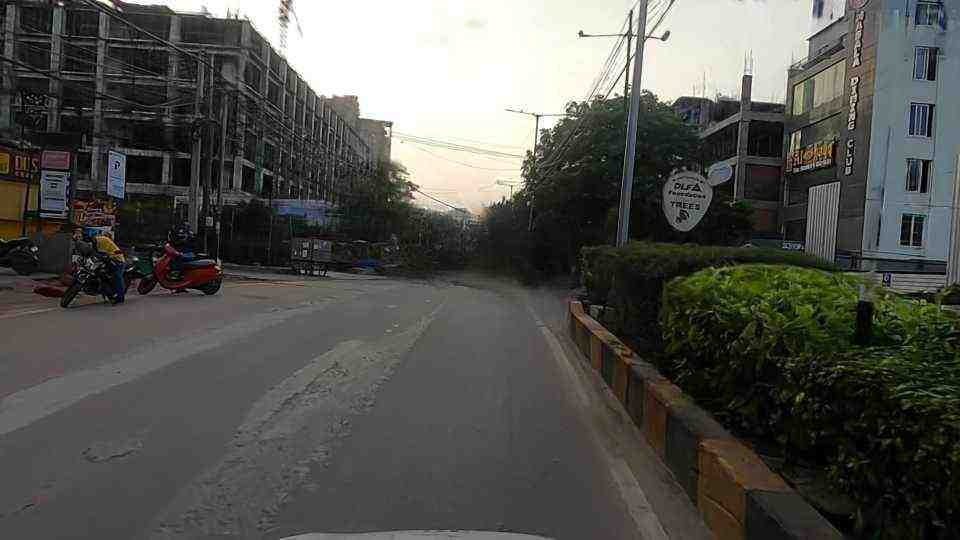} & \includegraphics[width=0.19\linewidth]{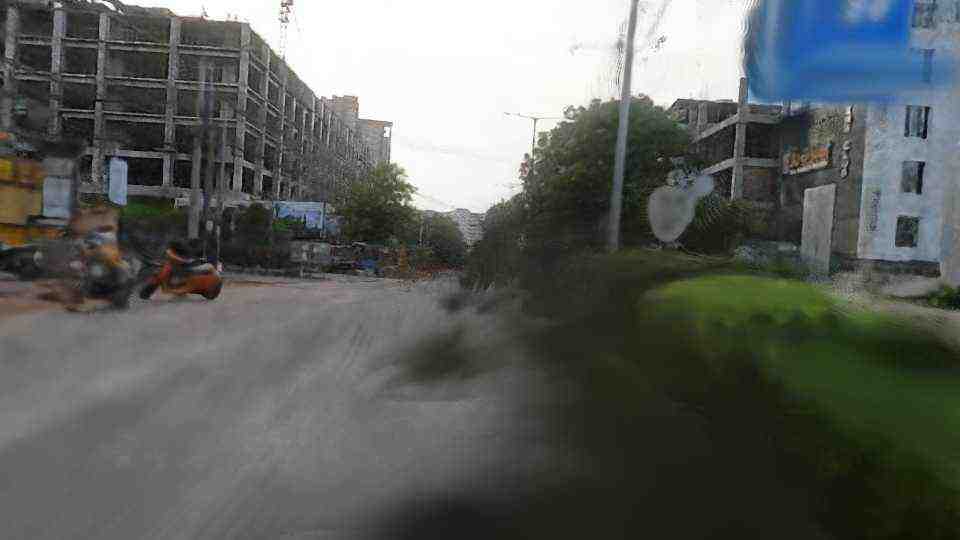} & \includegraphics[width=0.19\linewidth]{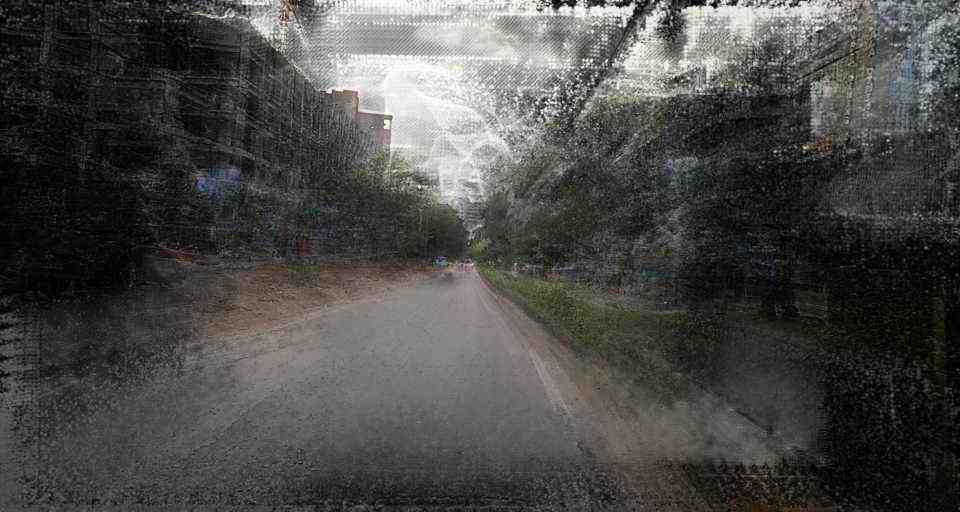} & \includegraphics[width=0.19\linewidth]{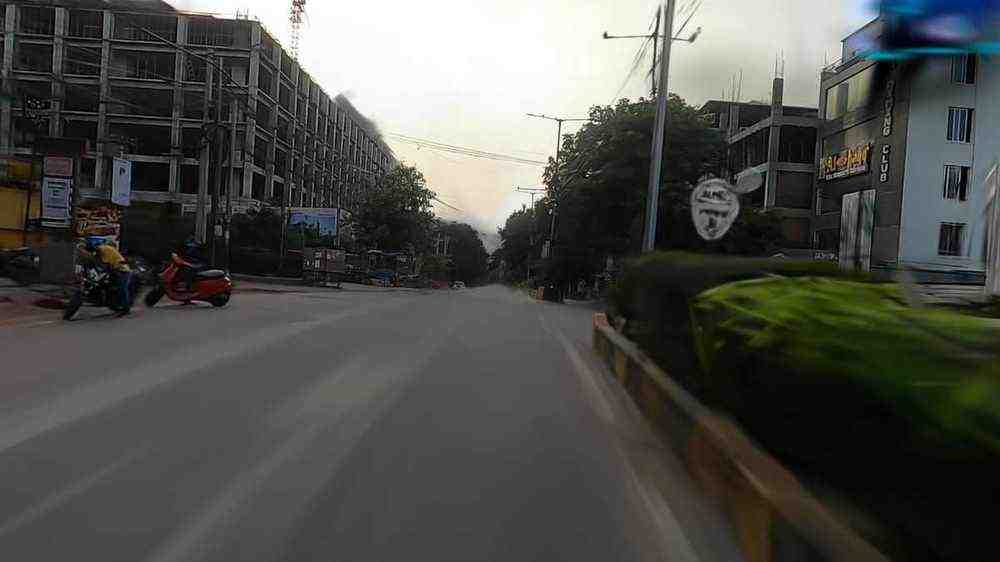} \\
& 16.89$|$0.54$|$0.63 & 22.11$|$0.73$|$0.44 & 14.76$|$0.37$|$0.72 & 24.35$|$0.78$|$0.36 \\
\includegraphics[width=0.19\linewidth]{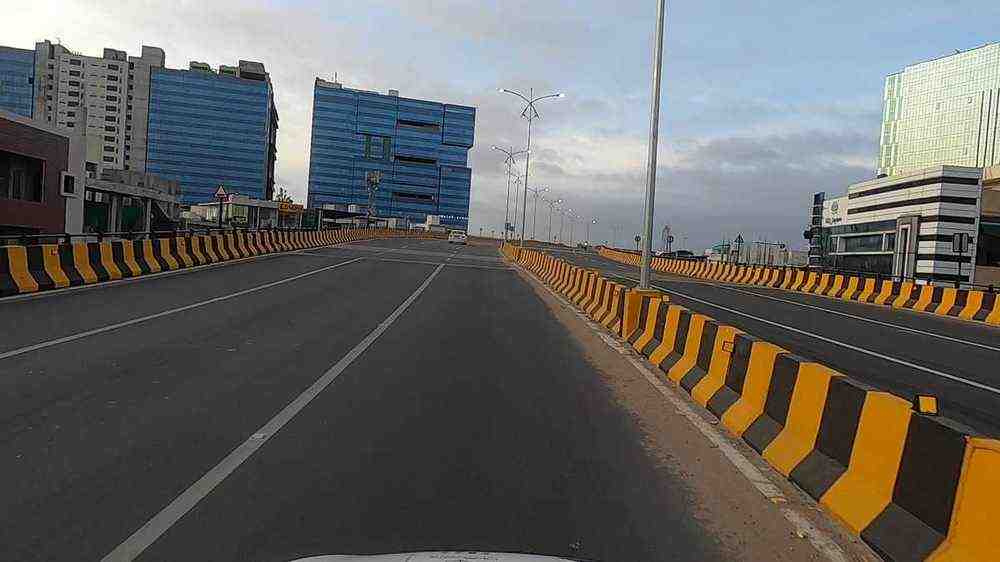} & \includegraphics[width=0.19\linewidth]{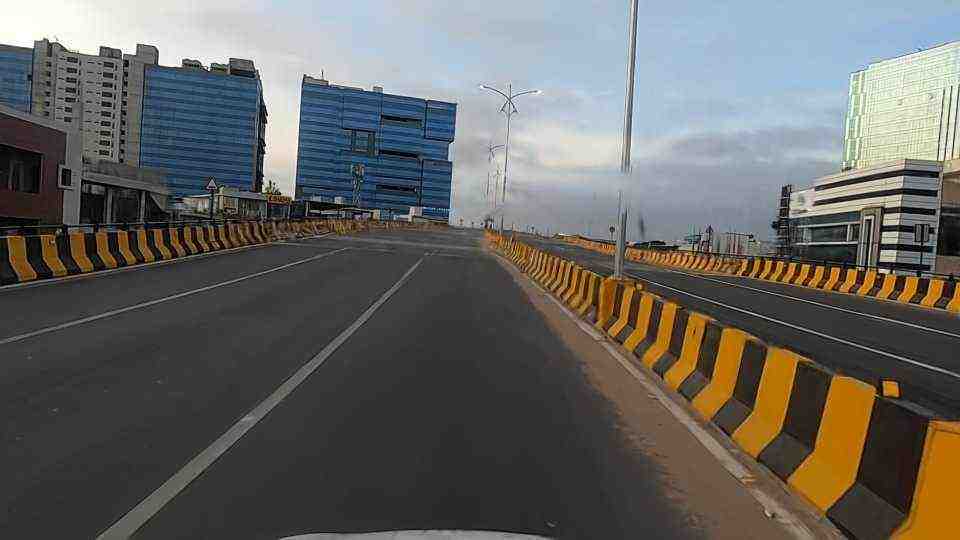} & \includegraphics[width=0.19\linewidth]{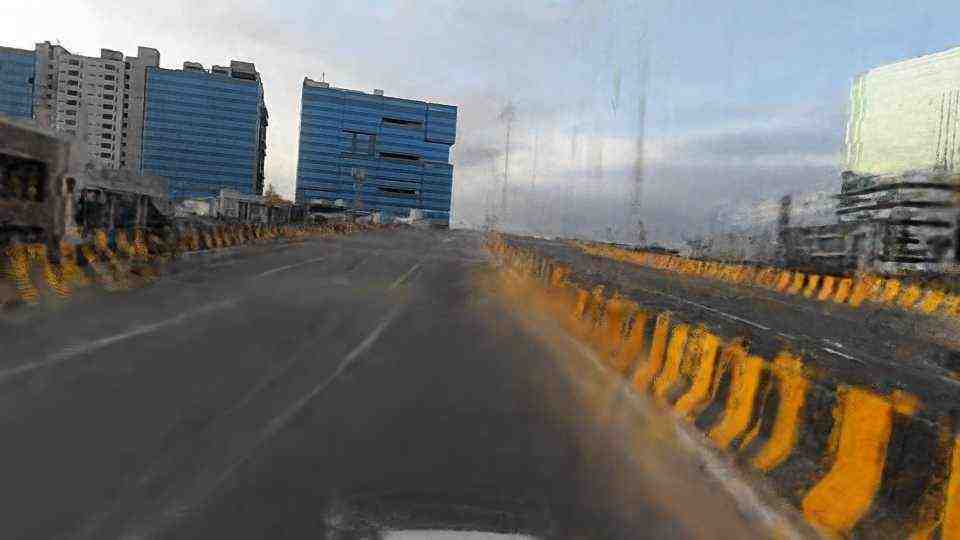} & \includegraphics[width=0.19\linewidth]{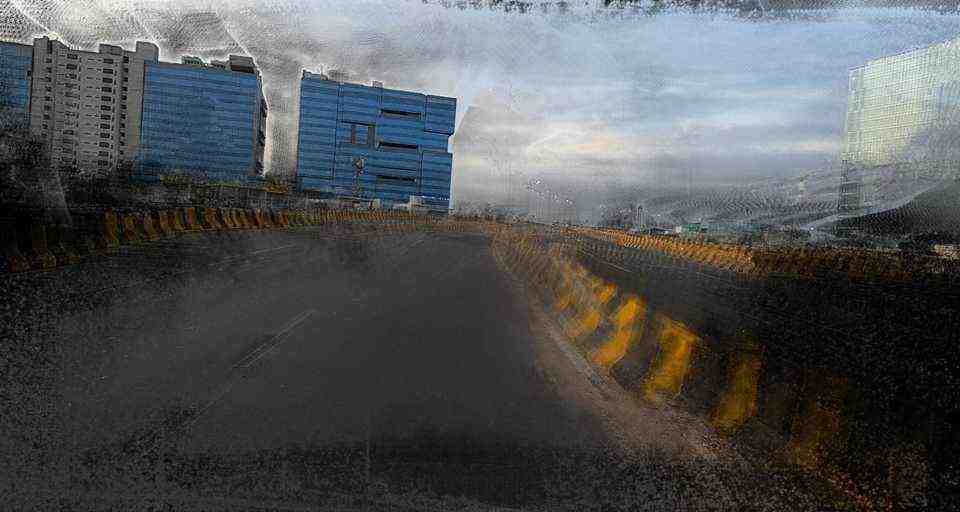} & \includegraphics[width=0.19\linewidth]{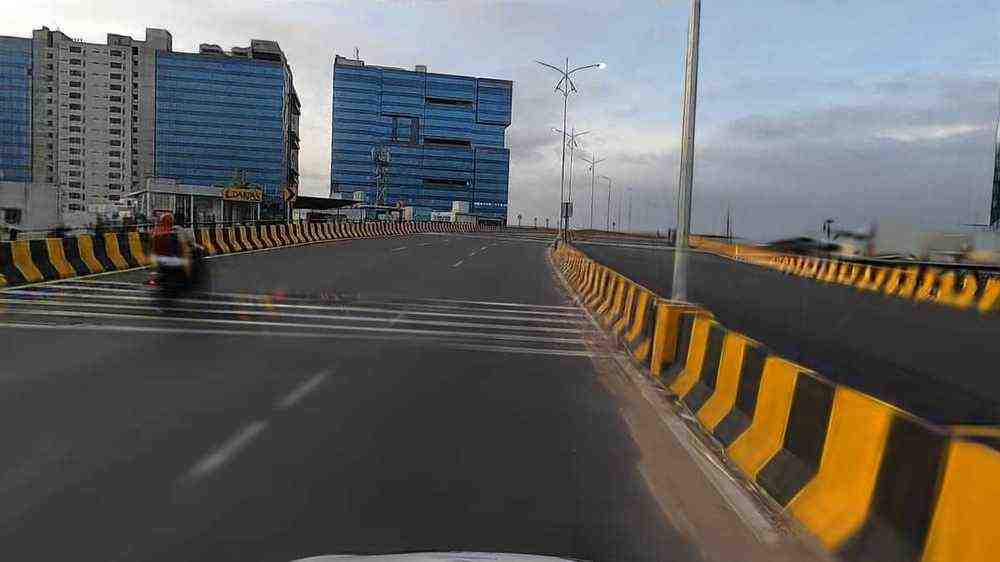} \\
& 18.08$|$0.62$|$0.53 & 15.26$|$0.56$|$0.58 & 16.33$|$0.56$|$0.60 & 27.48$|$0.85$|$0.27 \\

\end{tabular}
\end{adjustbox}

\caption{\textbf{Qualitative comparison across methods} Results from NVS methods trained on the $V^{C}$ sequences and rendered under the $T_{C\!\rightarrow C}$ viewpoint. 
Each column shows predictions from different NVS methods alongside the ground truth. 
The values beneath each rendered image correspond to the NVS metrics PSNR$|$SSIM$|$LPIPS.
}
\label{fig:qual_car_car}
\end{figure*}

\figref{fig:qual_car_car} presents results for models trained on the $V^{C}$ sequences and evaluated under the $T_{C\!\rightarrow C}$ configuration. Since the training and rendering viewpoints belong to the same domain, most methods produce stable reconstructions, though large-baseline view changes still expose limitations—particularly for geometry-sensitive regions such as vehicles and building facades.

\begin{figure*}[ht!]
\centering
\scriptsize
\setlength{\tabcolsep}{1pt}
\renewcommand{\arraystretch}{0.8}

\begin{adjustbox}{max width=\textwidth}
\begin{tabular}{ccccc}

\textbf{GT} & \textbf{3DGS} & \textbf{NeRF} & \textbf{DepthSplat} & \textbf{PVG} \\

\includegraphics[width=0.19\linewidth]{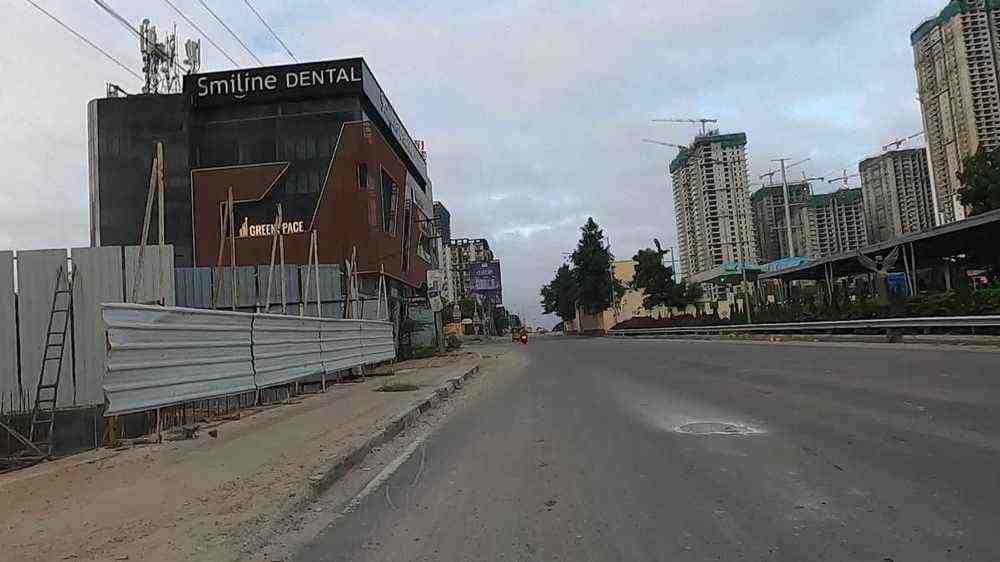} & \includegraphics[width=0.19\linewidth]{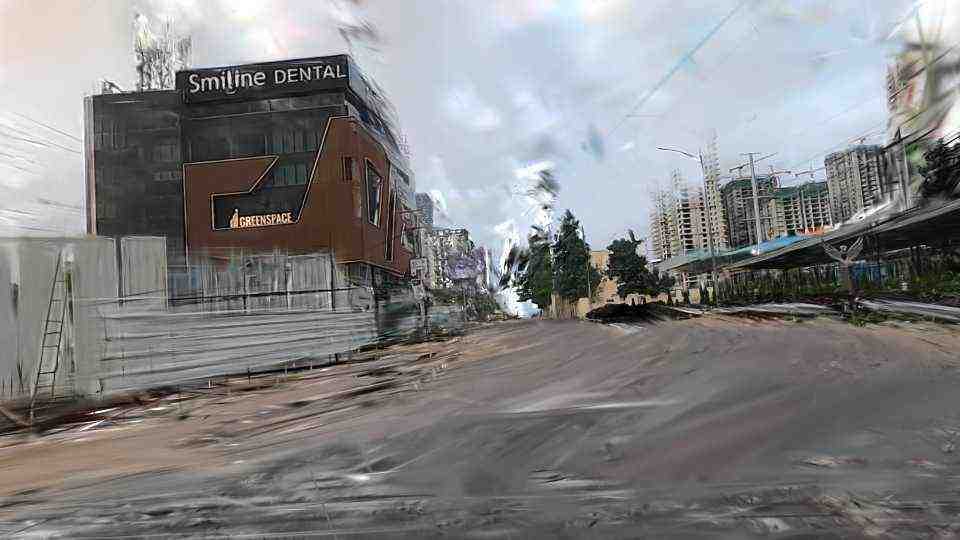} & \includegraphics[width=0.19\linewidth]{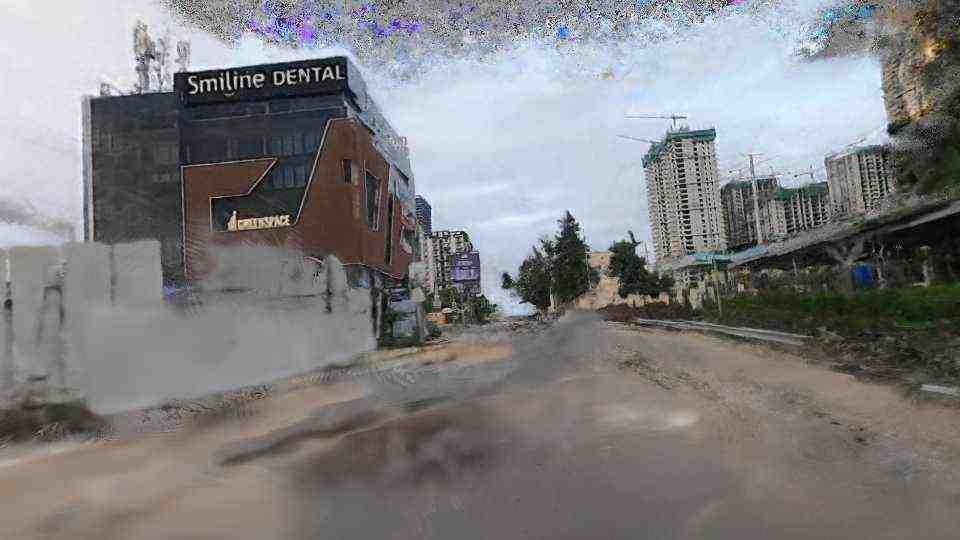} & \includegraphics[width=0.19\linewidth]{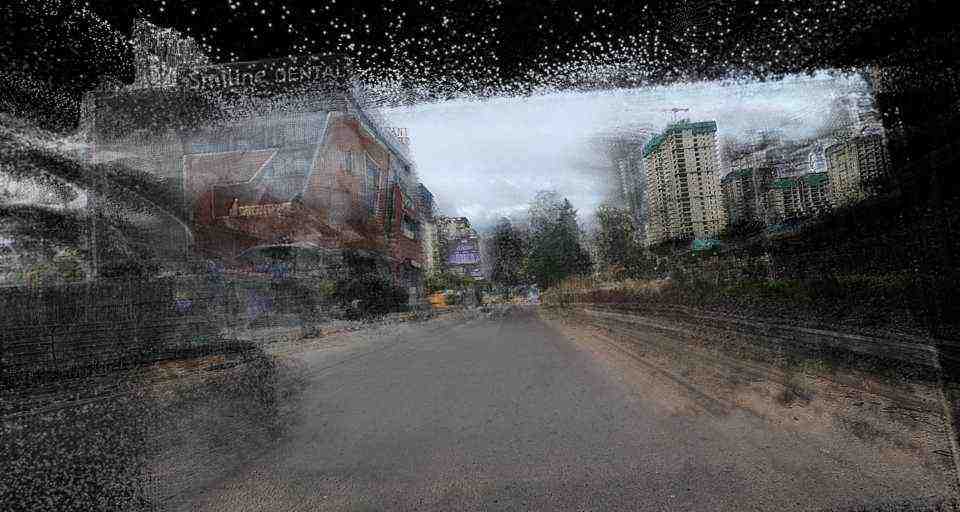} & \includegraphics[width=0.19\linewidth]{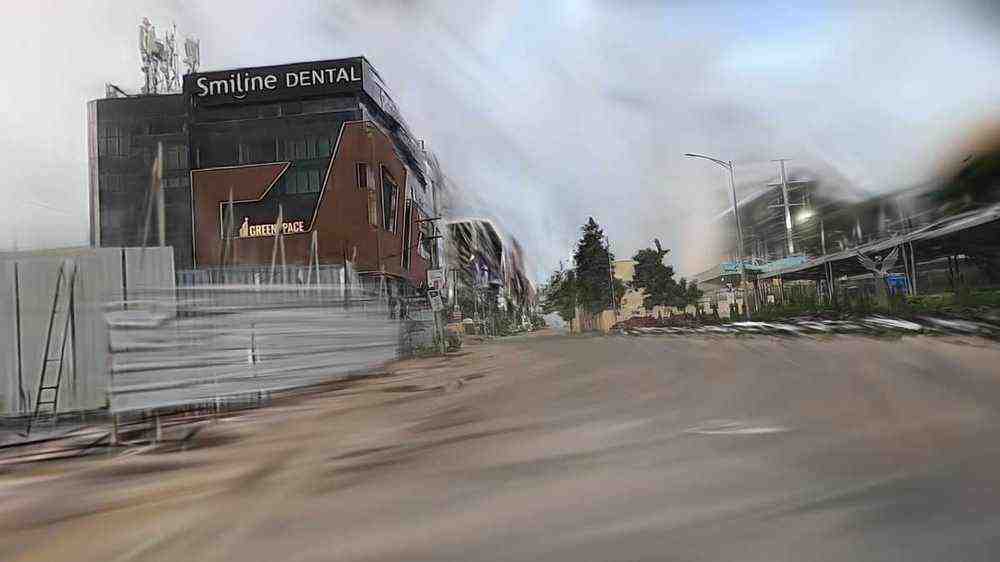} \\
& 16.84$|$0.58$|$0.57 & 16.30$|$0.65$|$0.51 & 9.53$|$0.36$|$0.71 & 17.06$|$0.61$|$0.53 \\
\includegraphics[width=0.19\linewidth]{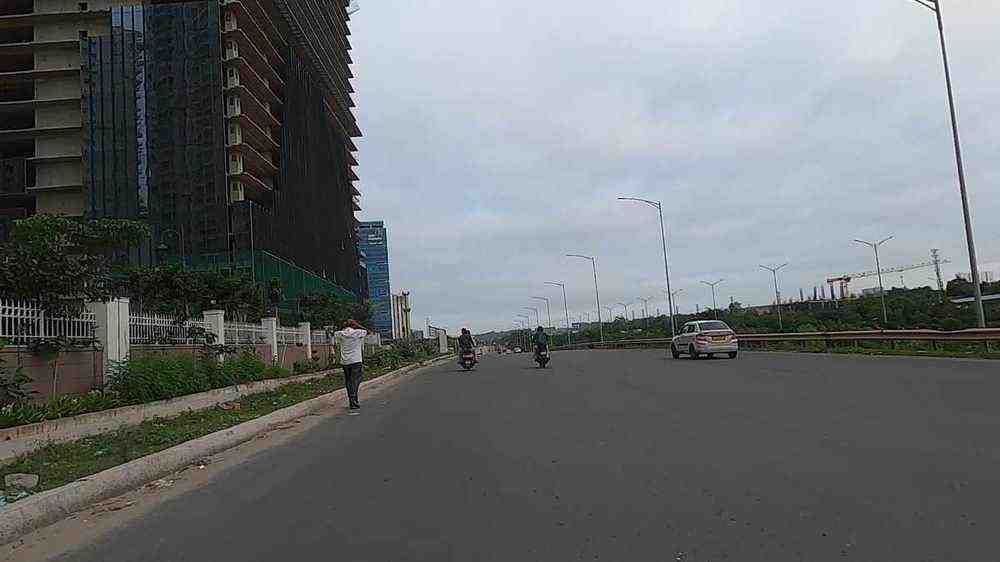} & \includegraphics[width=0.19\linewidth]{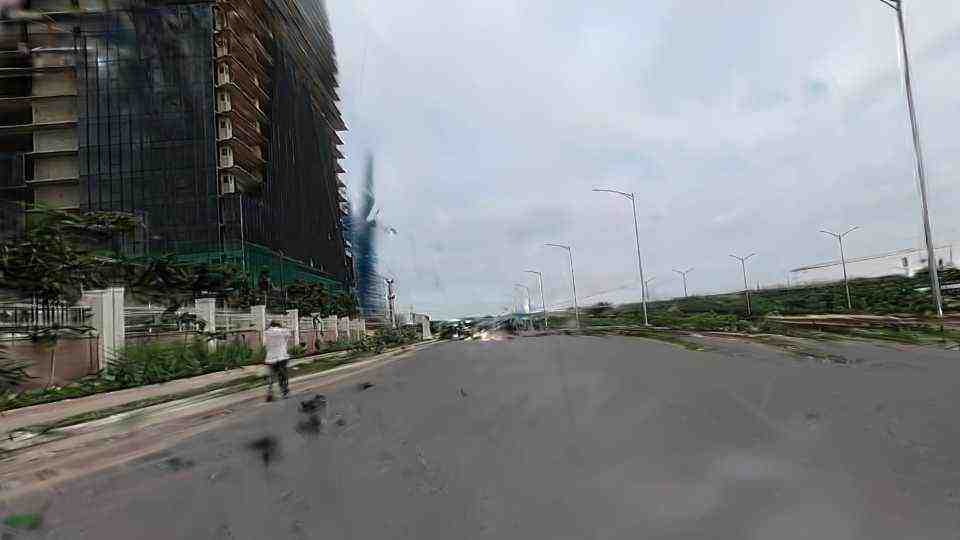} & \includegraphics[width=0.19\linewidth]{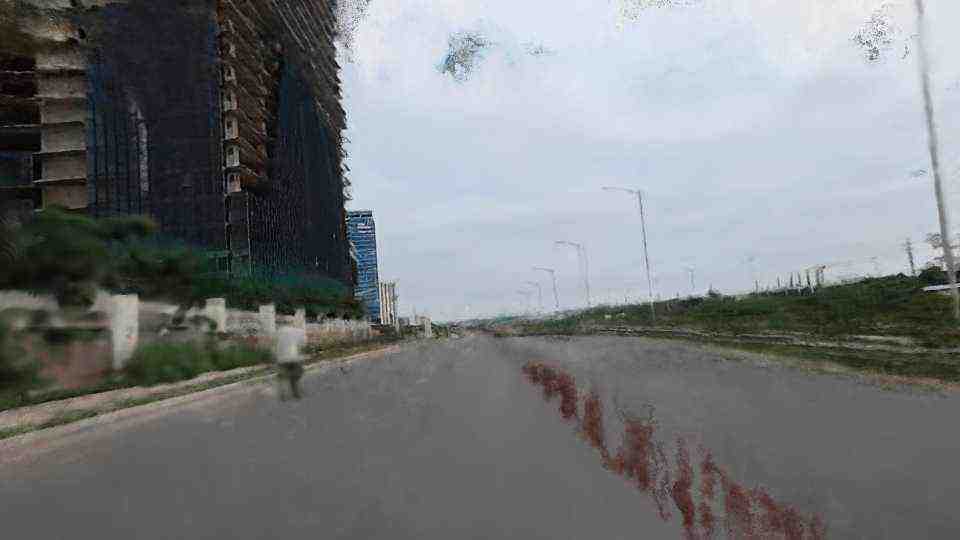} & \includegraphics[width=0.19\linewidth]{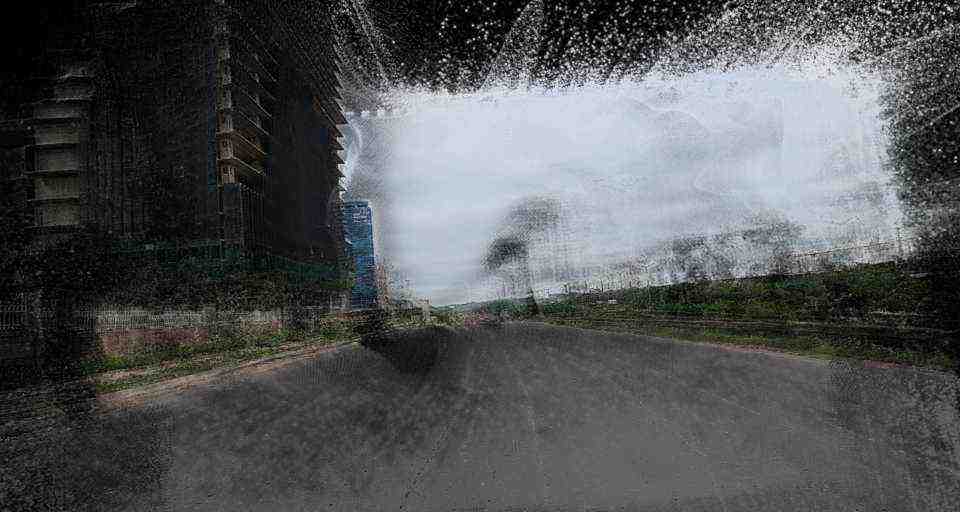} & \includegraphics[width=0.19\linewidth]{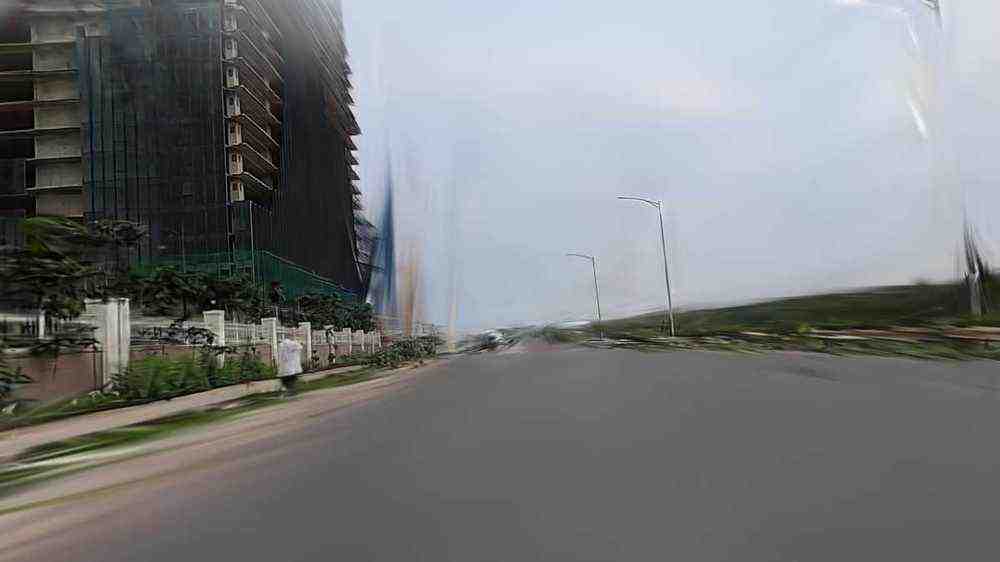} \\
& 19.16$|$0.67$|$0.50 & 19.07$|$0.73$|$0.42 & 12.63$|$0.50$|$0.65 & 19.16$|$0.72$|$0.40 \\
\includegraphics[width=0.19\linewidth]{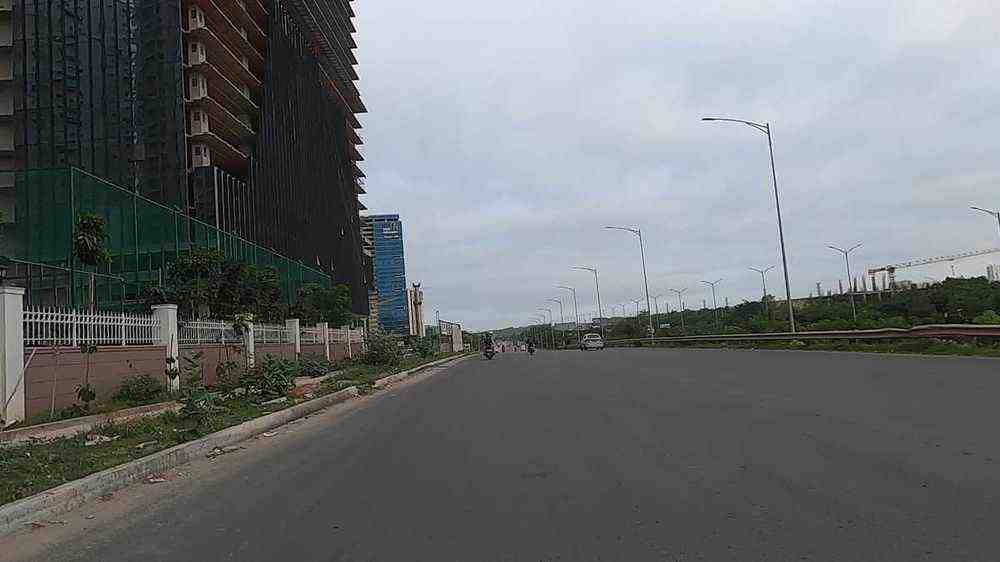} & \includegraphics[width=0.19\linewidth]{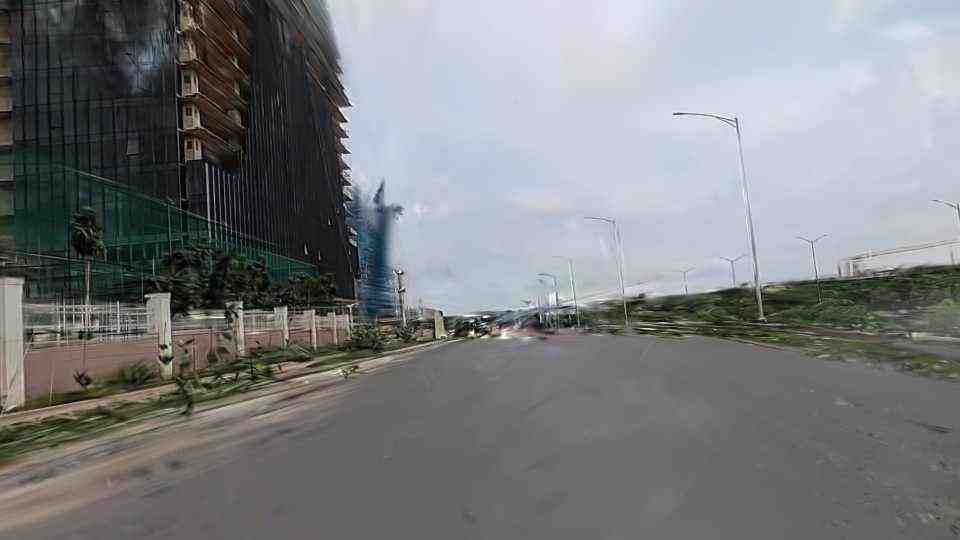} & \includegraphics[width=0.19\linewidth]{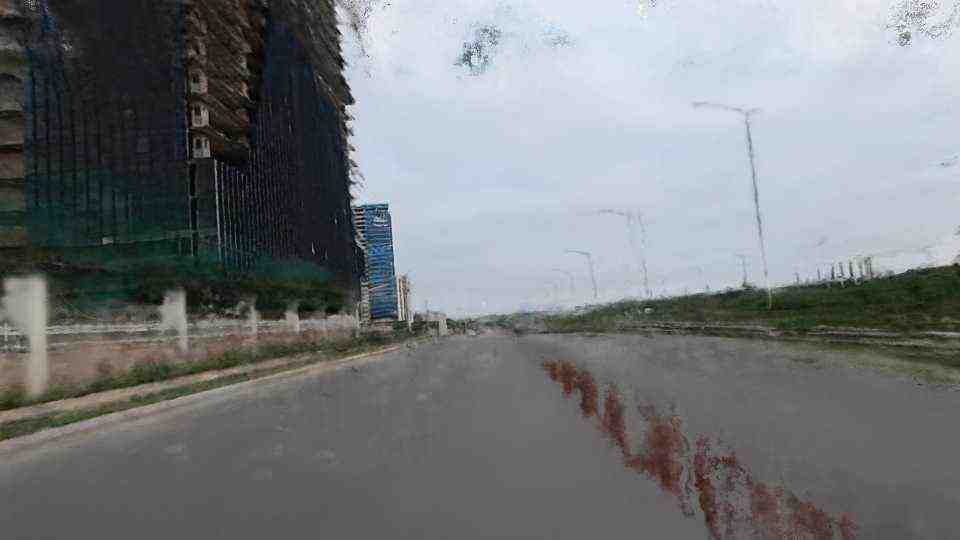} & \includegraphics[width=0.19\linewidth]{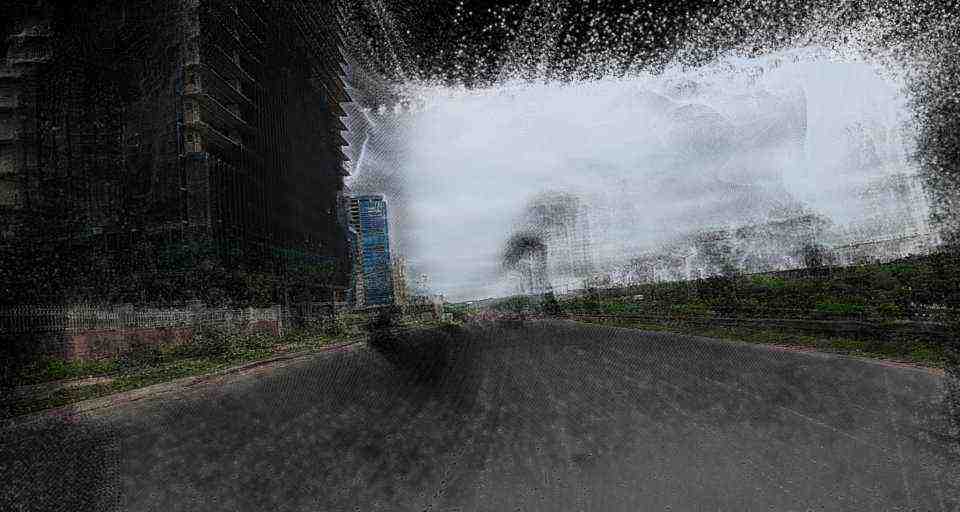} & \includegraphics[width=0.19\linewidth]{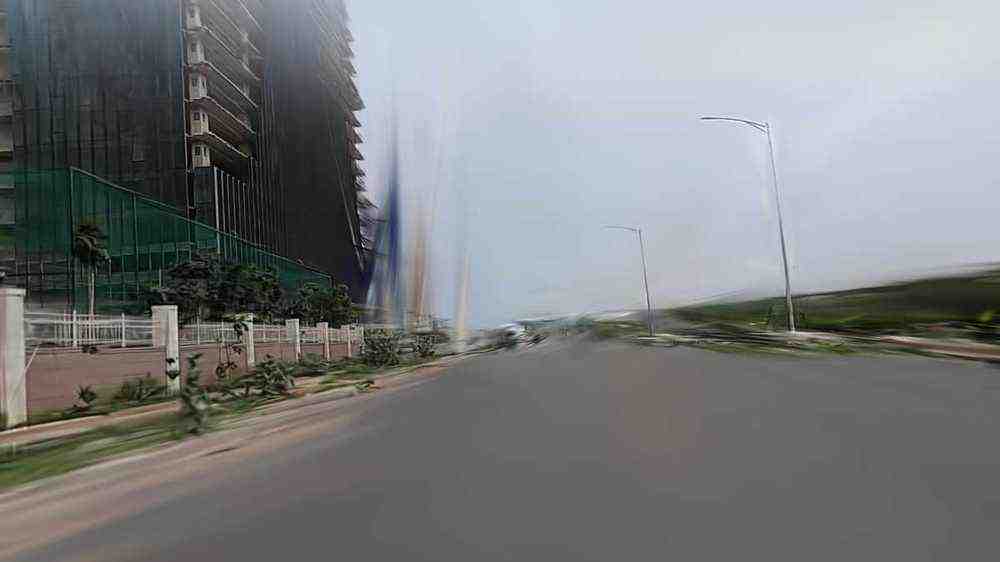} \\
& 19.19$|$0.67$|$0.51 & 18.62$|$0.73$|$0.42 & 12.93$|$0.49$|$0.67 & 19.53$|$0.72$|$0.41 \\
\includegraphics[width=0.19\linewidth]{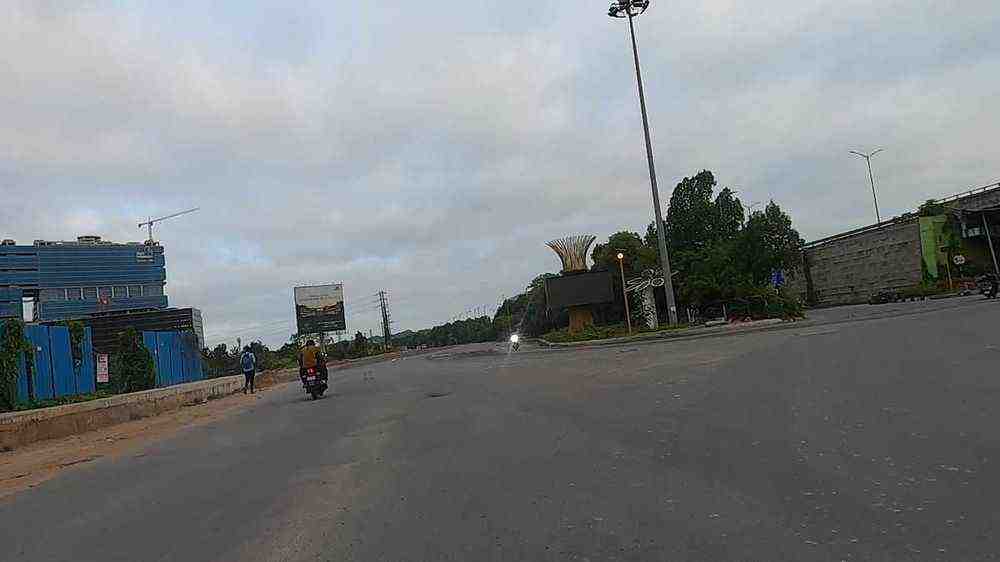} & \includegraphics[width=0.19\linewidth]{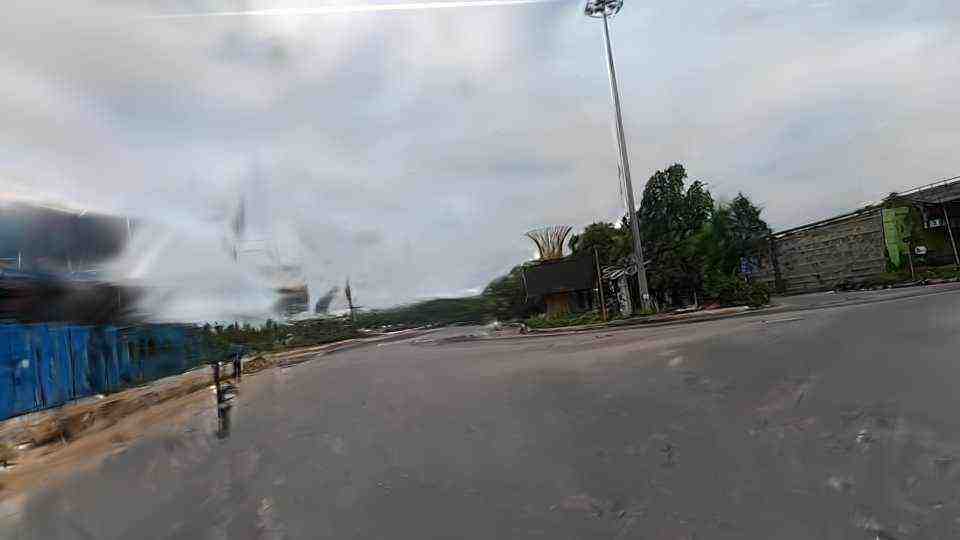} & \includegraphics[width=0.19\linewidth]{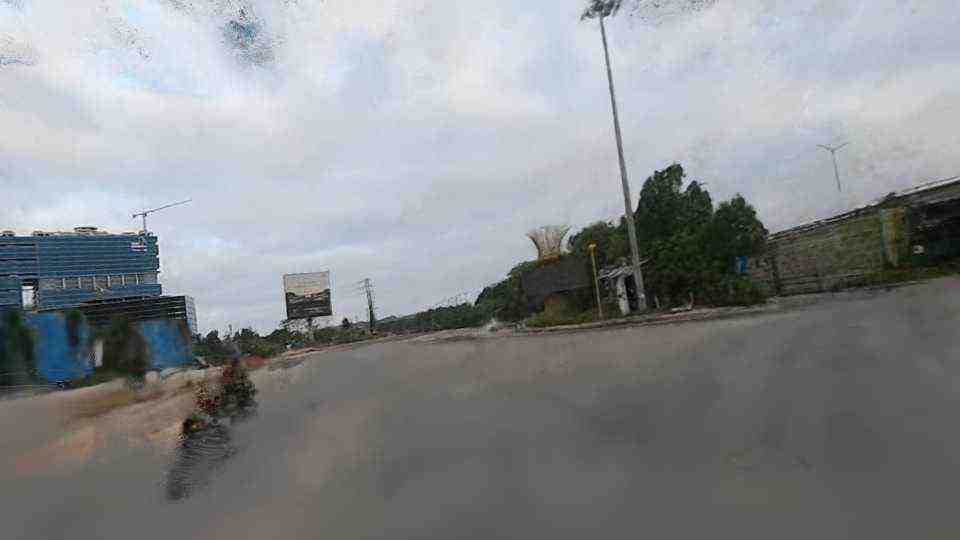} & \includegraphics[width=0.19\linewidth]{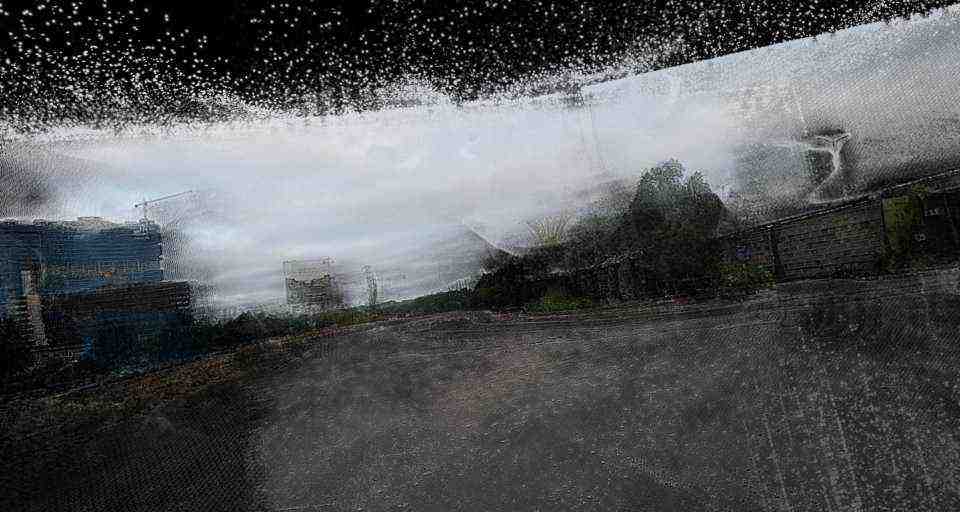} & \includegraphics[width=0.19\linewidth]{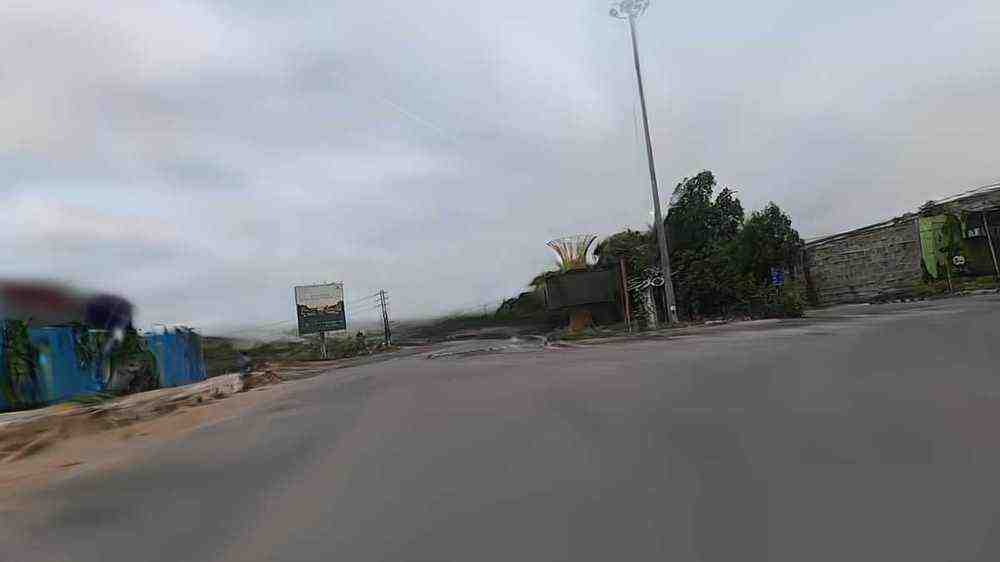} \\
& 21.40$|$0.80$|$0.40 & 20.23$|$0.82$|$0.39 & 10.37$|$0.46$|$0.70 & 20.45$|$0.80$|$0.42 \\
\includegraphics[width=0.19\linewidth]{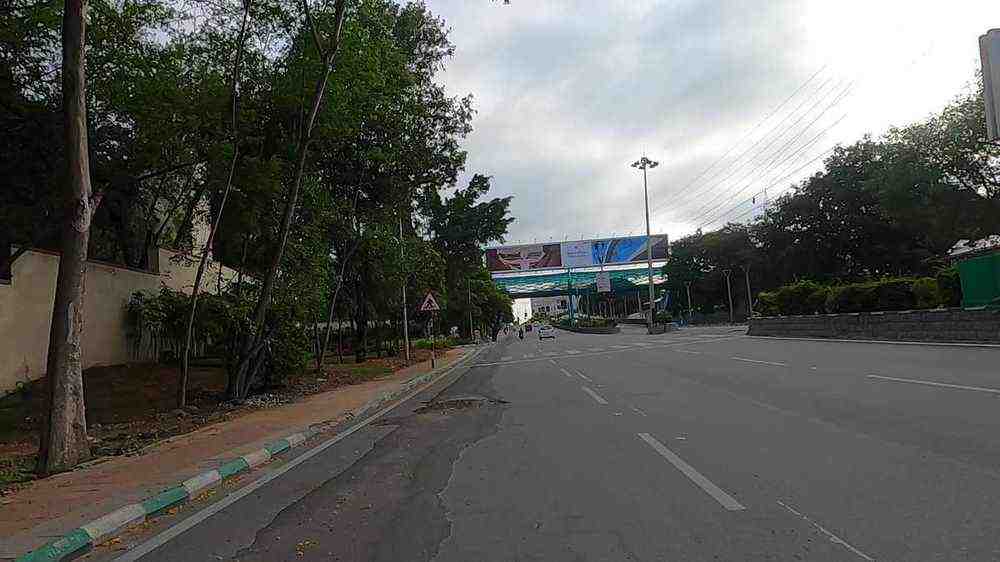} & \includegraphics[width=0.19\linewidth]{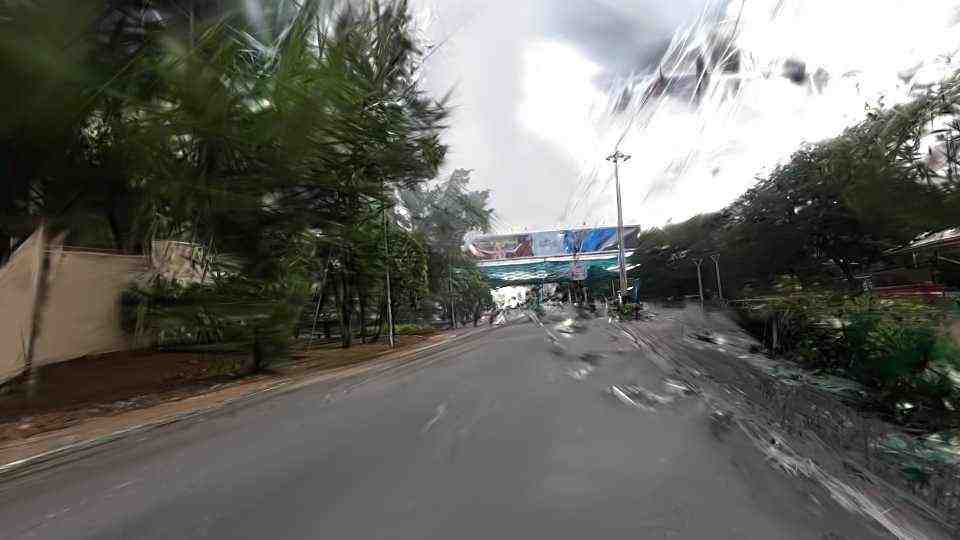} & \includegraphics[width=0.19\linewidth]{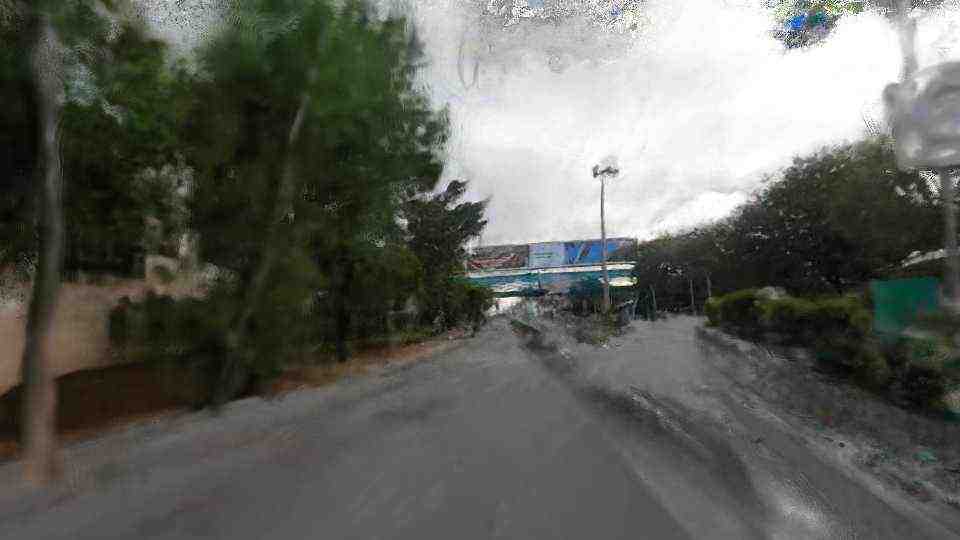} & \includegraphics[width=0.19\linewidth]{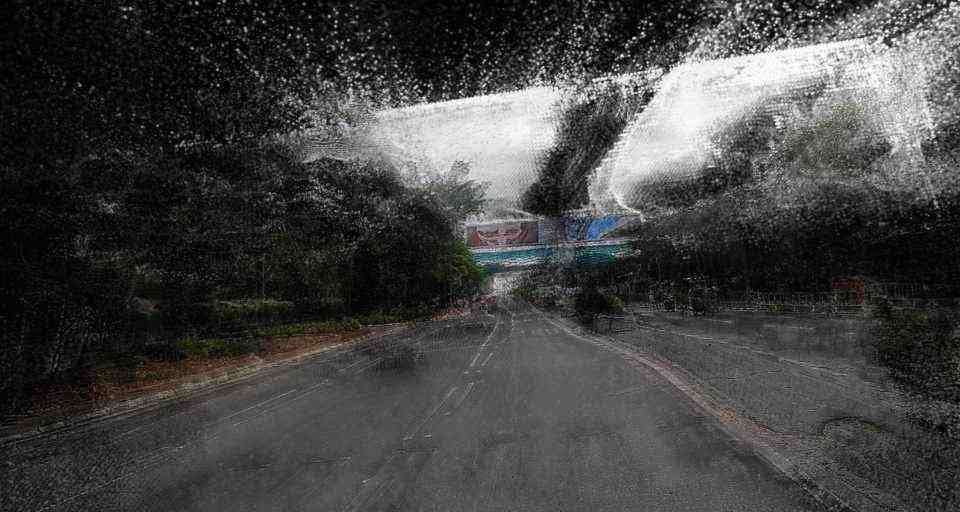} & \includegraphics[width=0.19\linewidth]{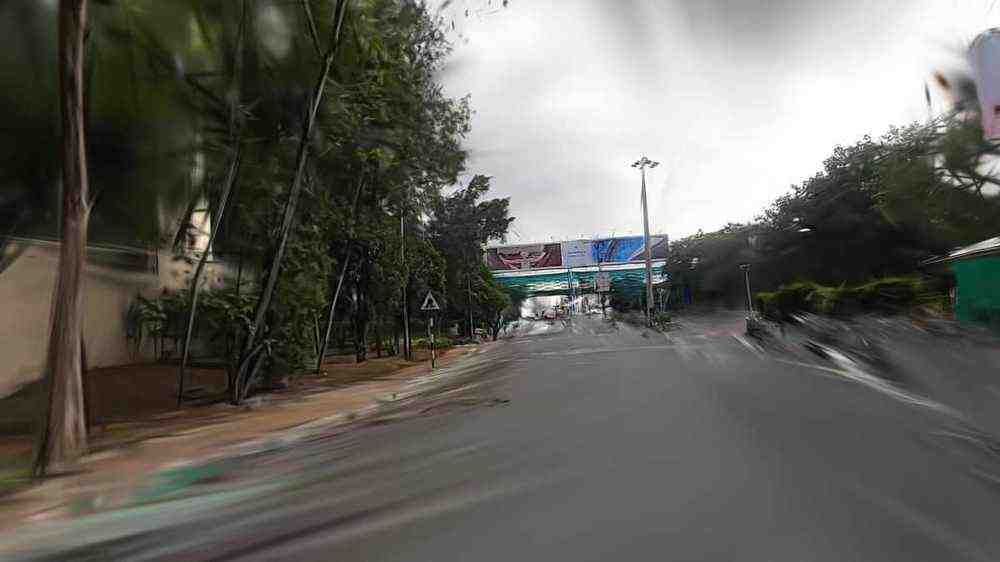} \\
& 16.74$|$0.53$|$0.69 & 18.43$|$0.61$|$0.54 & 11.38$|$0.35$|$0.72 & 16.40$|$0.53$|$0.63 \\
\includegraphics[width=0.19\linewidth]{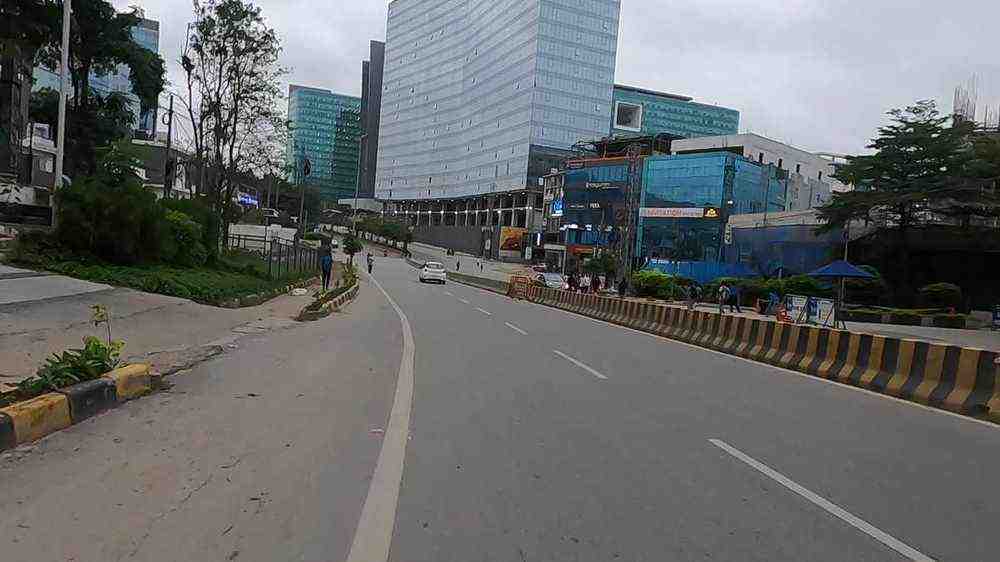} & \includegraphics[width=0.19\linewidth]{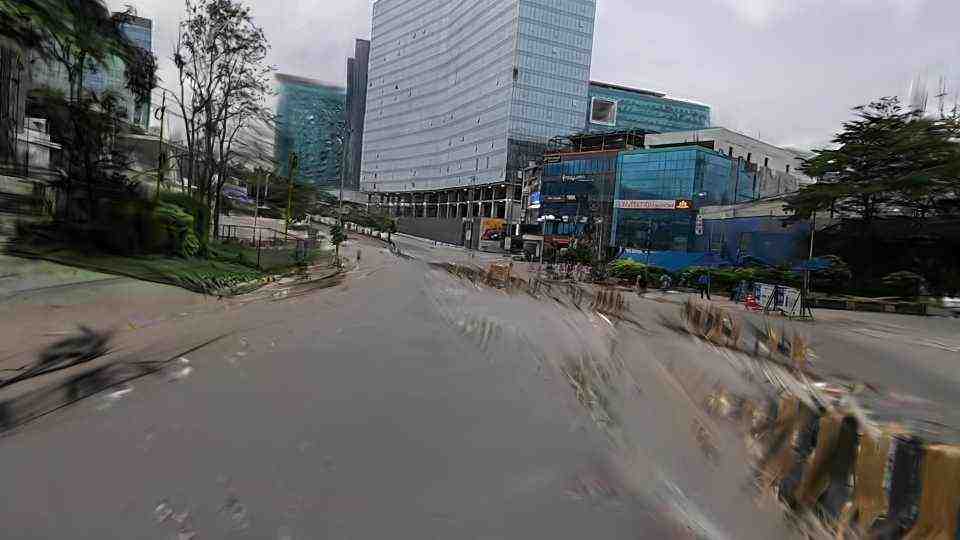} & \includegraphics[width=0.19\linewidth]{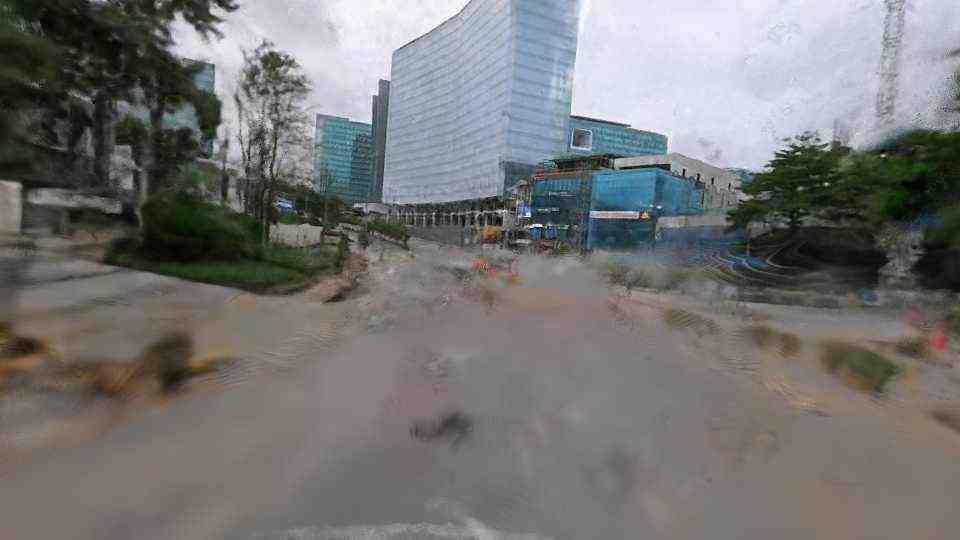} & \includegraphics[width=0.19\linewidth]{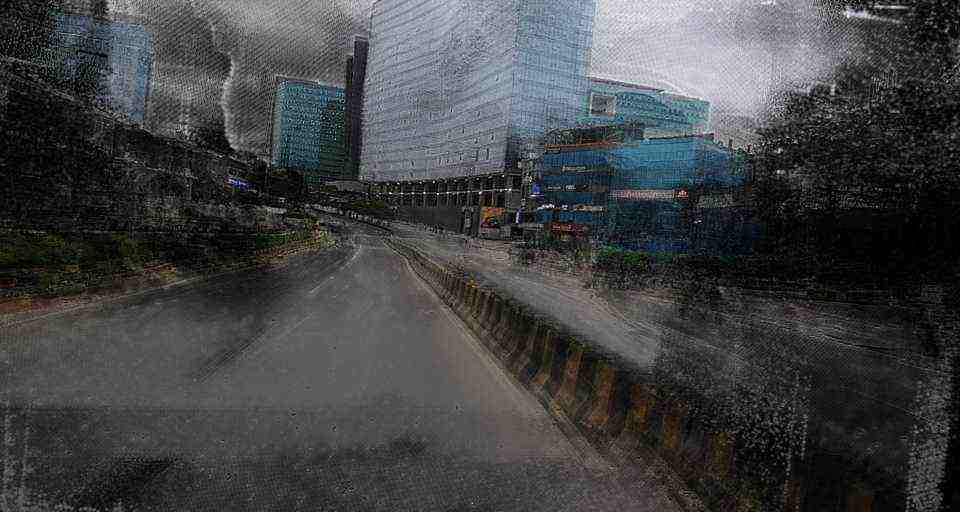} & \includegraphics[width=0.19\linewidth]{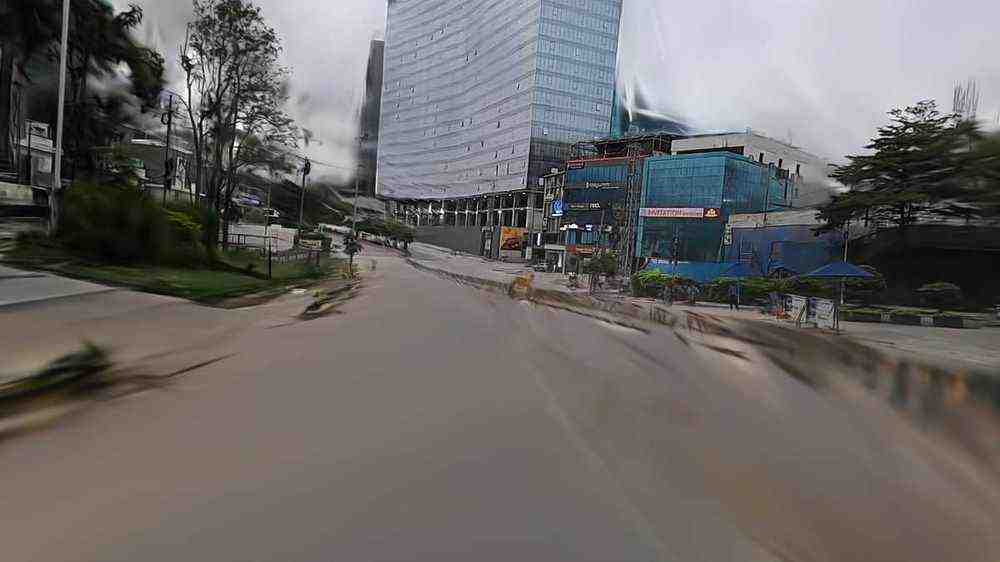} \\
& 14.94$|$0.56$|$0.63 & 19.85$|$0.72$|$0.43 & 12.93$|$0.43$|$0.67 & 20.21$|$0.69$|$0.47 \\
\includegraphics[width=0.19\linewidth]{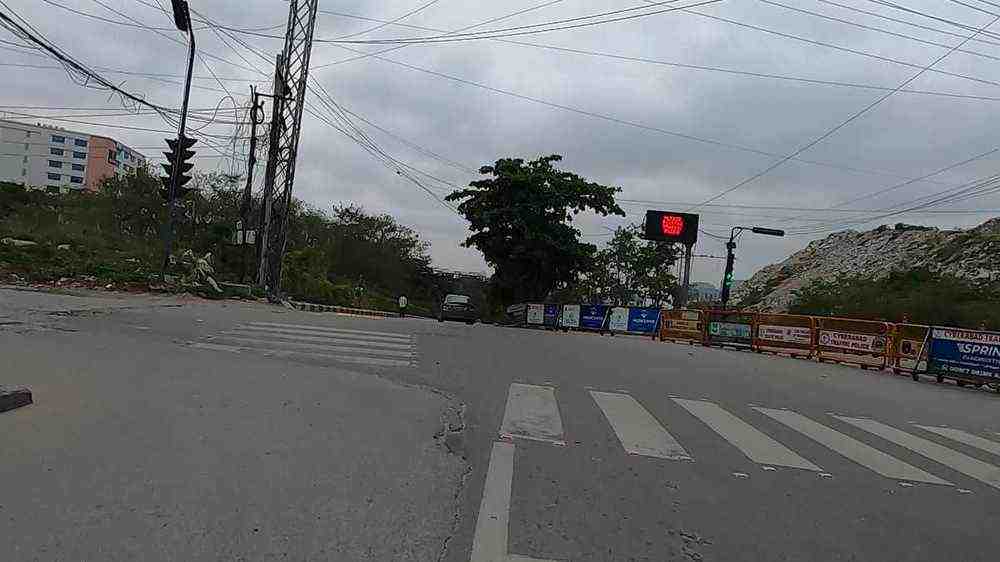} & \includegraphics[width=0.19\linewidth]{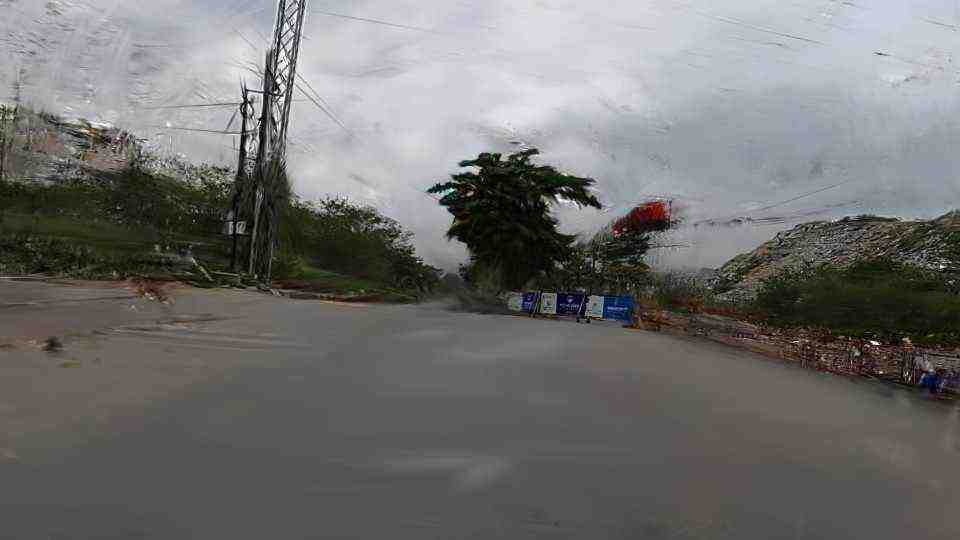} & \includegraphics[width=0.19\linewidth]{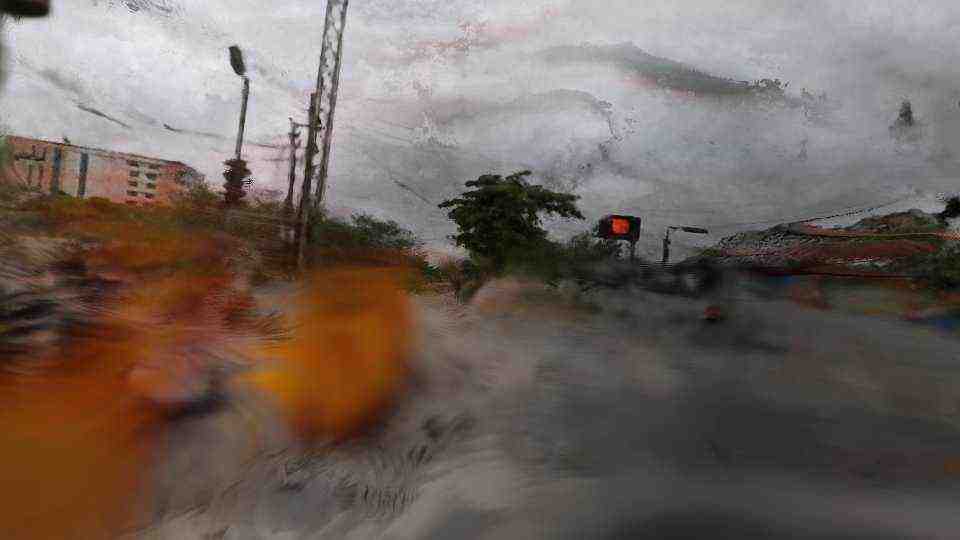} & \includegraphics[width=0.19\linewidth]{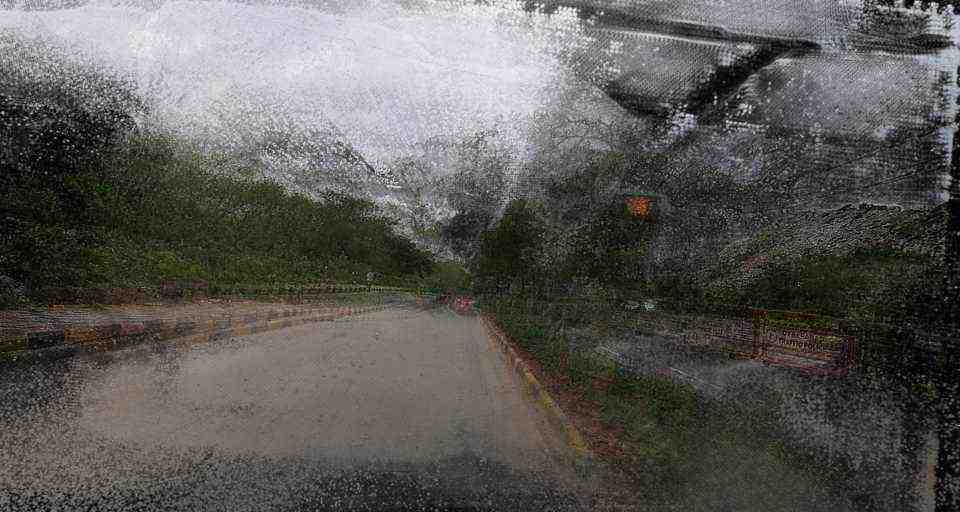} & \includegraphics[width=0.19\linewidth]{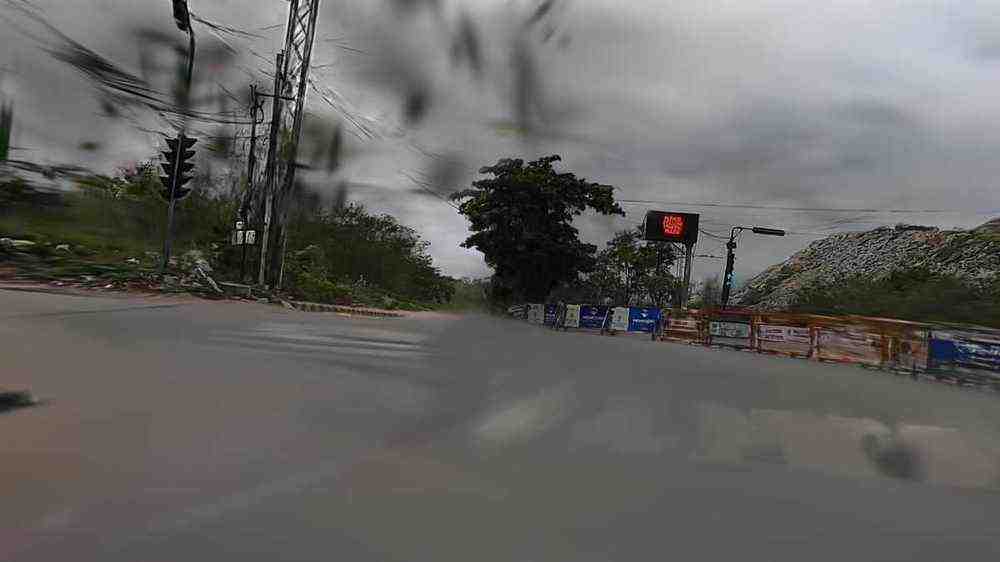} \\
& 15.20$|$0.57$|$0.78 & 16.98$|$0.71$|$0.56 & 13.29$|$0.38$|$0.77 & 20.05$|$0.68$|$0.60 \\
\includegraphics[width=0.19\linewidth]{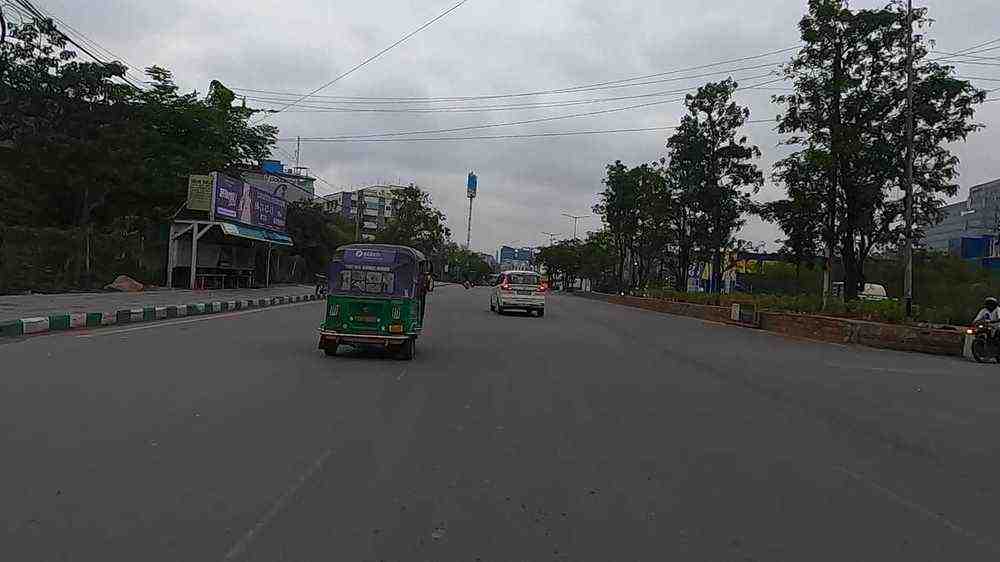} & \includegraphics[width=0.19\linewidth]{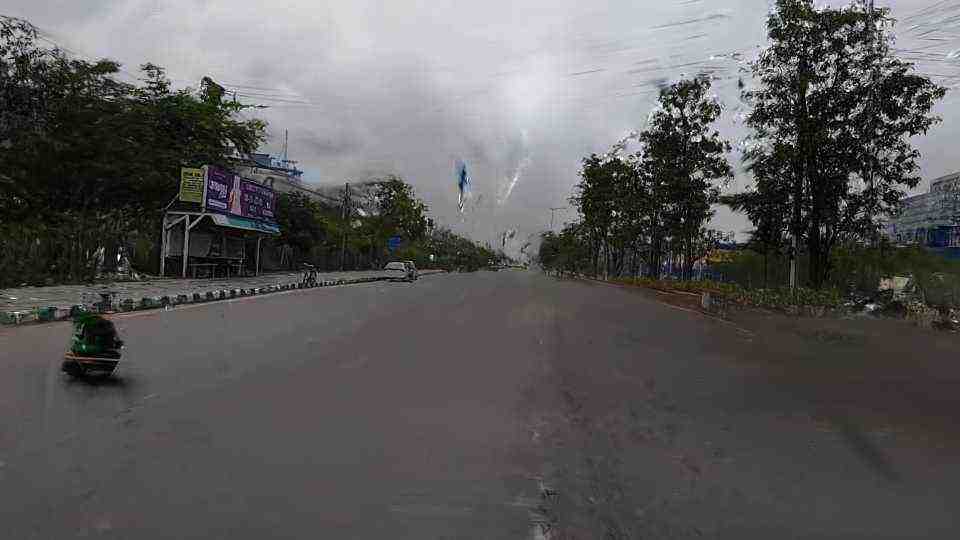} & \includegraphics[width=0.19\linewidth]{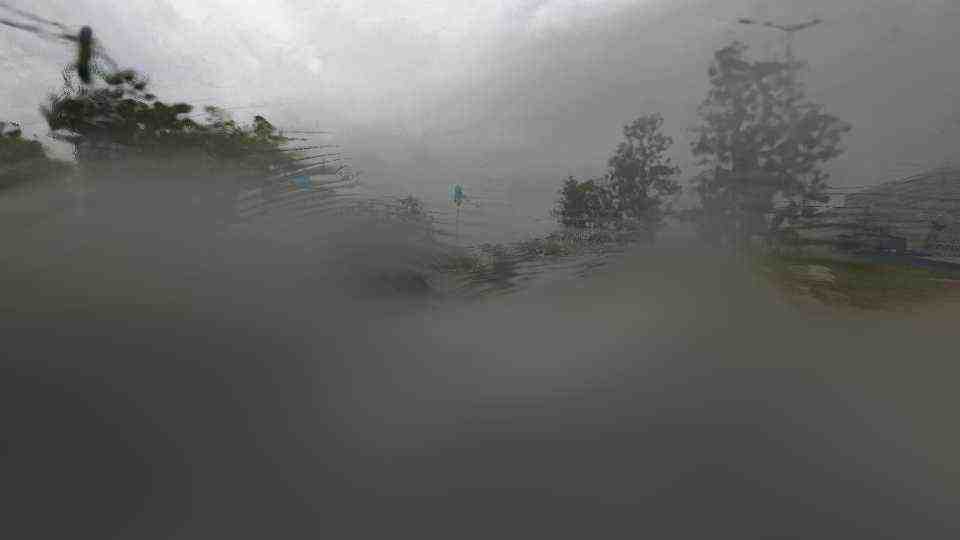} & \includegraphics[width=0.19\linewidth]{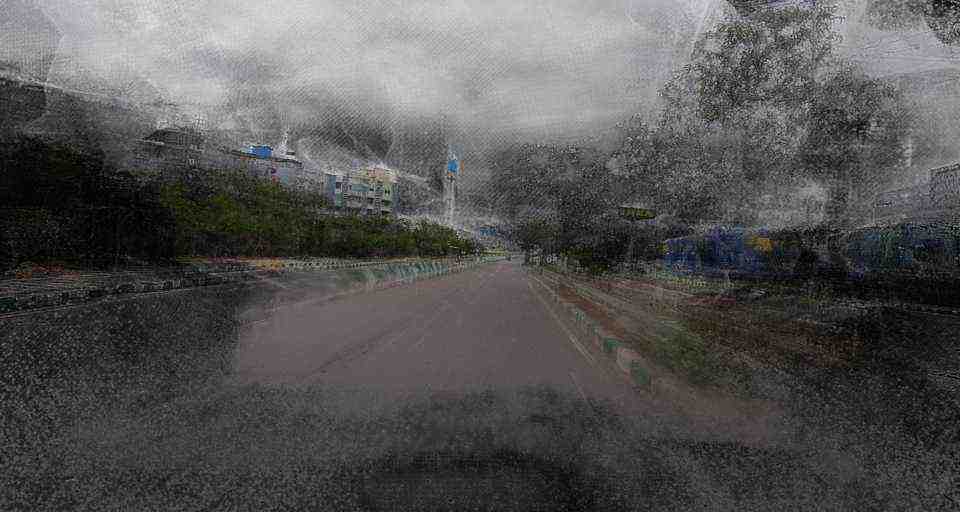} & \includegraphics[width=0.19\linewidth]{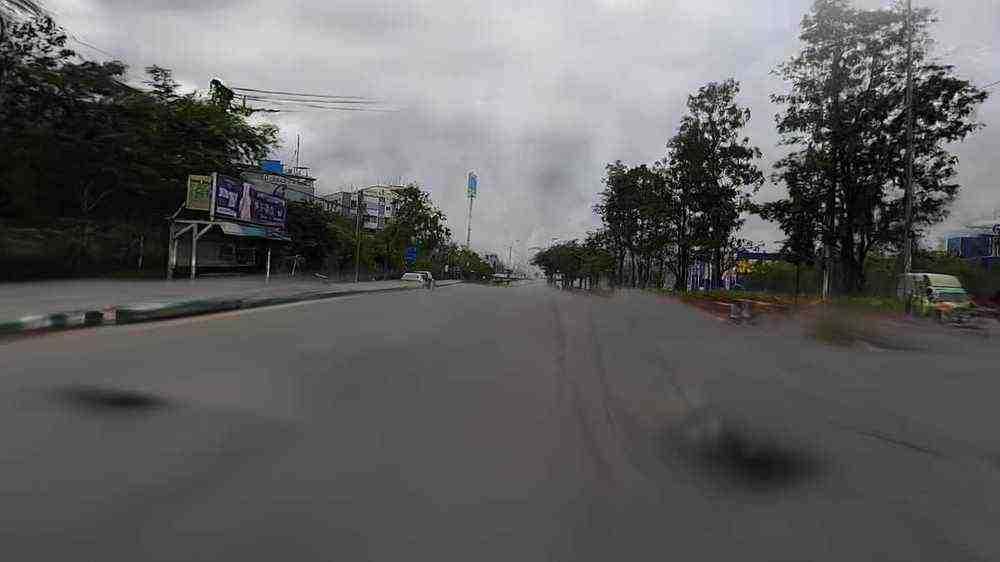} \\
& 15.12$|$0.64$|$0.61 & 19.70$|$0.73$|$0.44 & 15.21$|$0.49$|$0.74 & 20.17$|$0.71$|$0.44 \\
\includegraphics[width=0.19\linewidth]{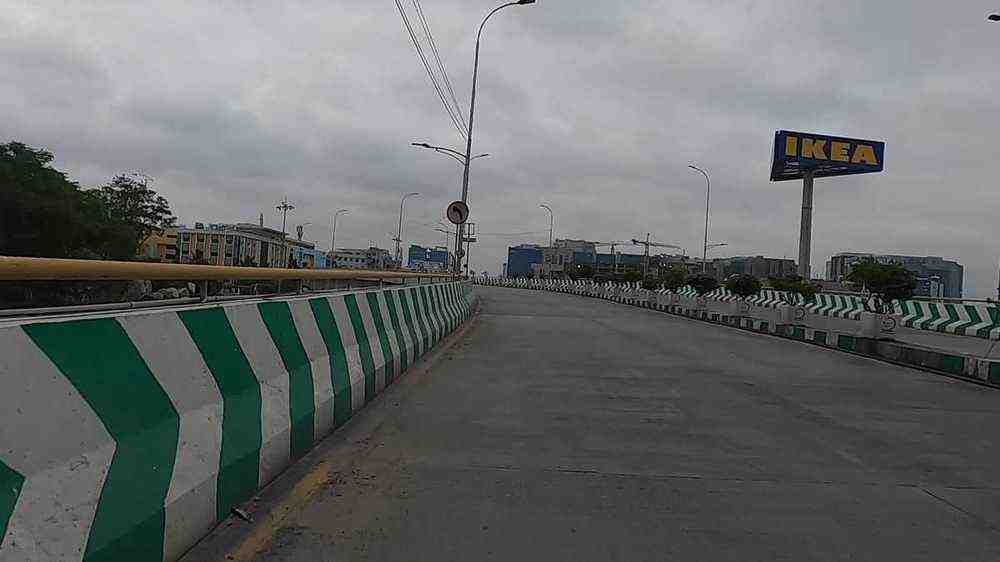} & \includegraphics[width=0.19\linewidth]{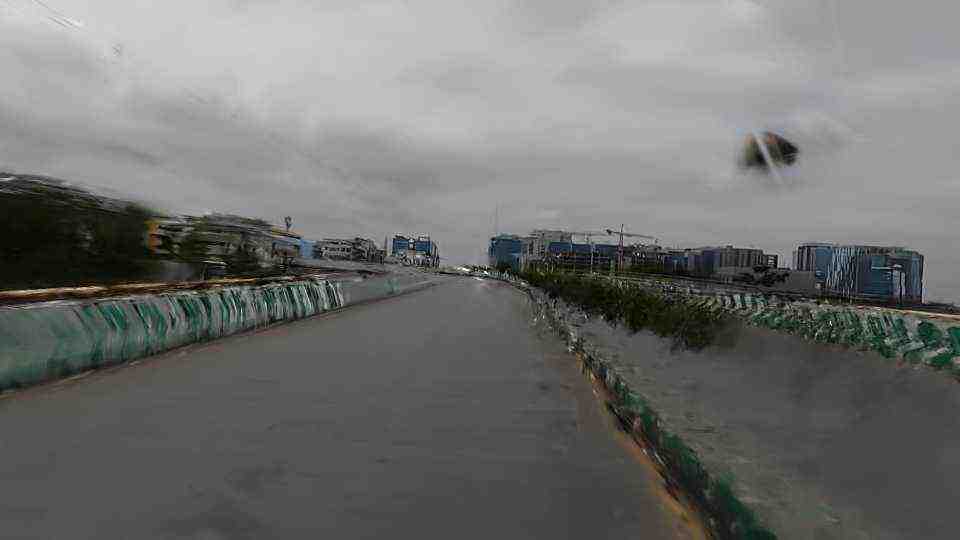} & \includegraphics[width=0.19\linewidth]{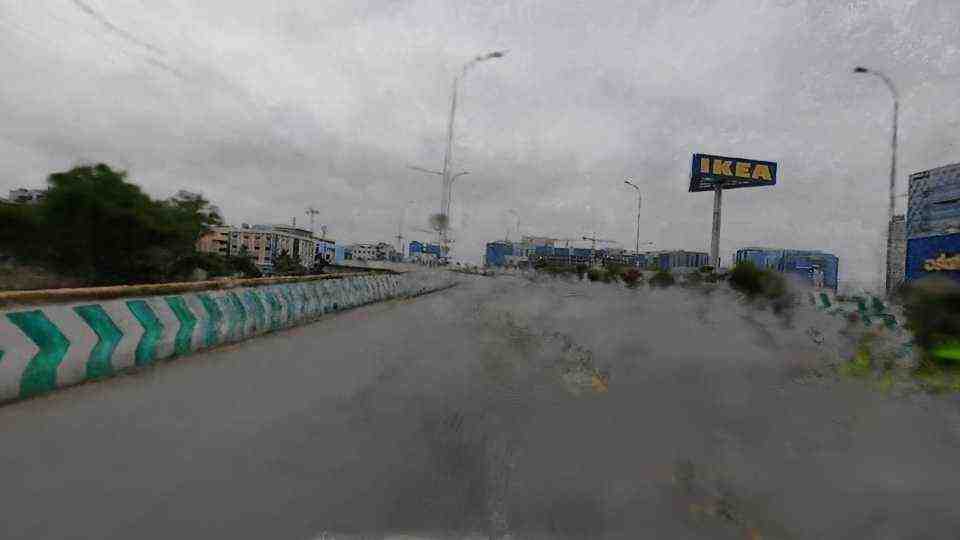} & \includegraphics[width=0.19\linewidth]{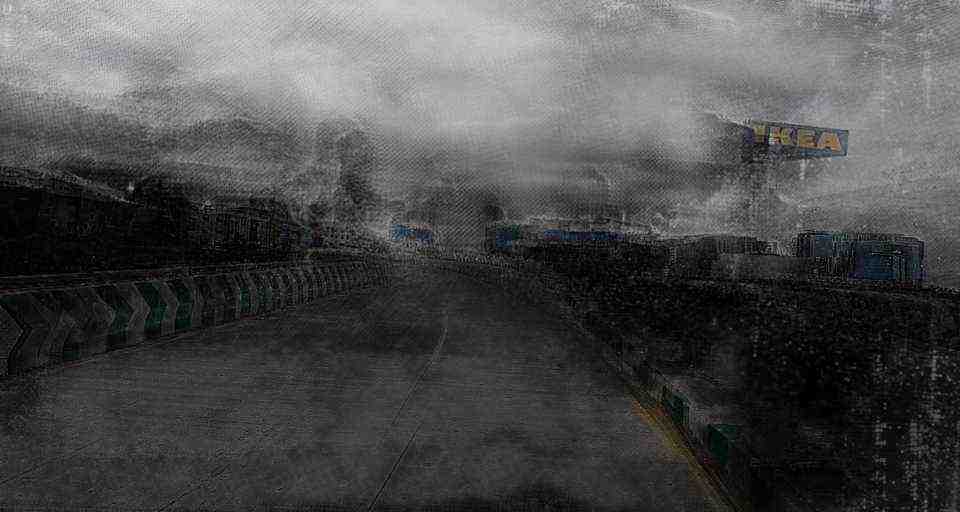} & \includegraphics[width=0.19\linewidth]{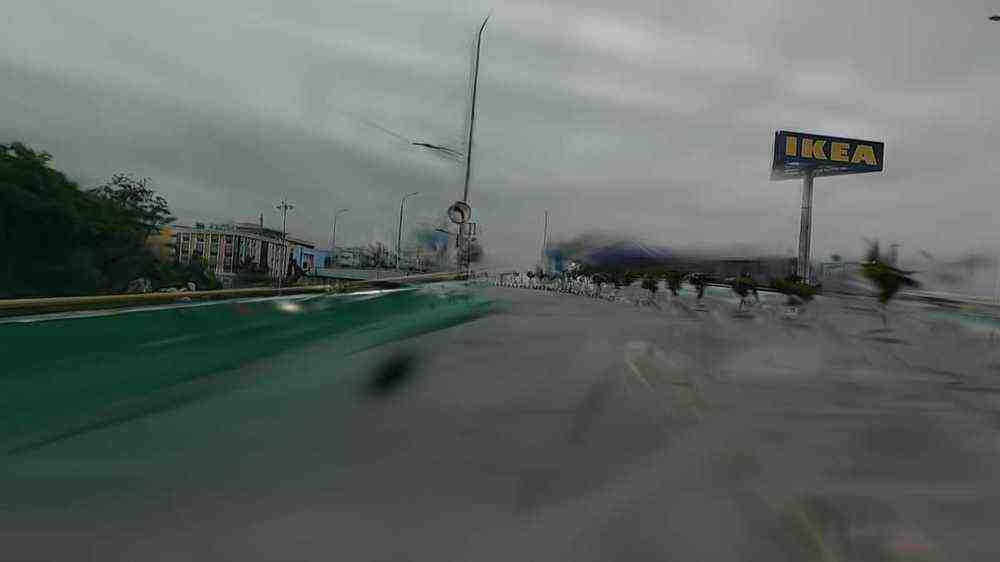} \\
& 17.07$|$0.71$|$0.57 & 19.36$|$0.77$|$0.52 & 15.45$|$0.54$|$0.72 & 19.27$|$0.74$|$0.54 \\
\includegraphics[width=0.19\linewidth]{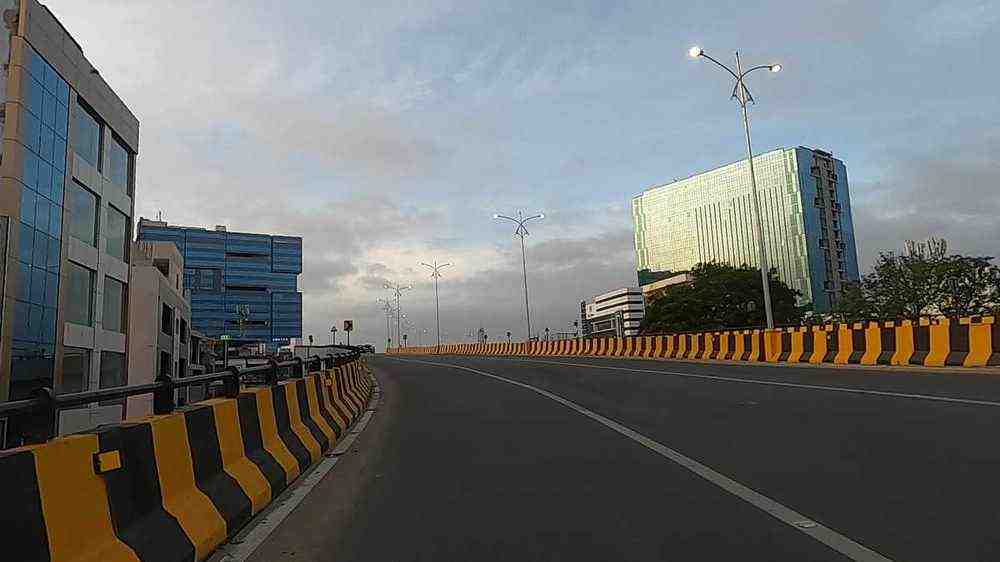} & \includegraphics[width=0.19\linewidth]{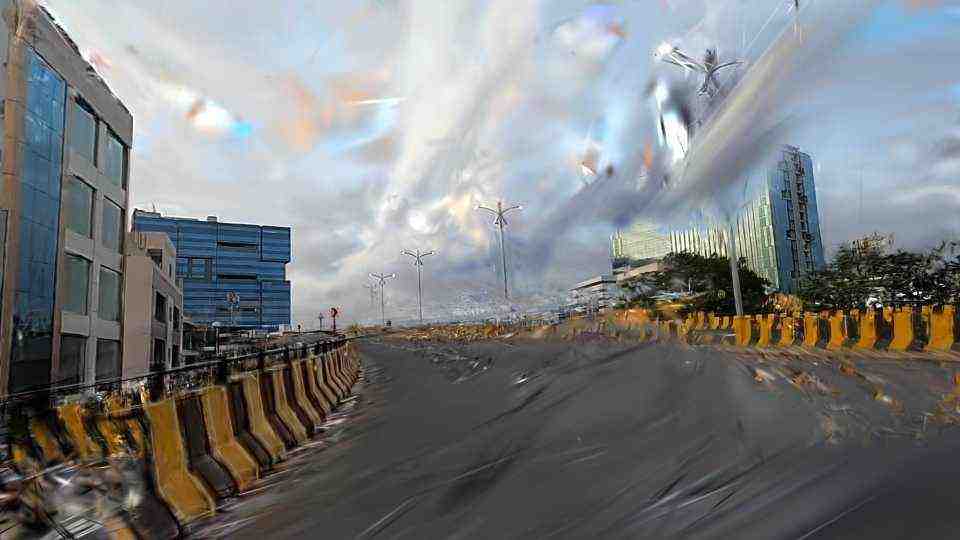} & \includegraphics[width=0.19\linewidth]{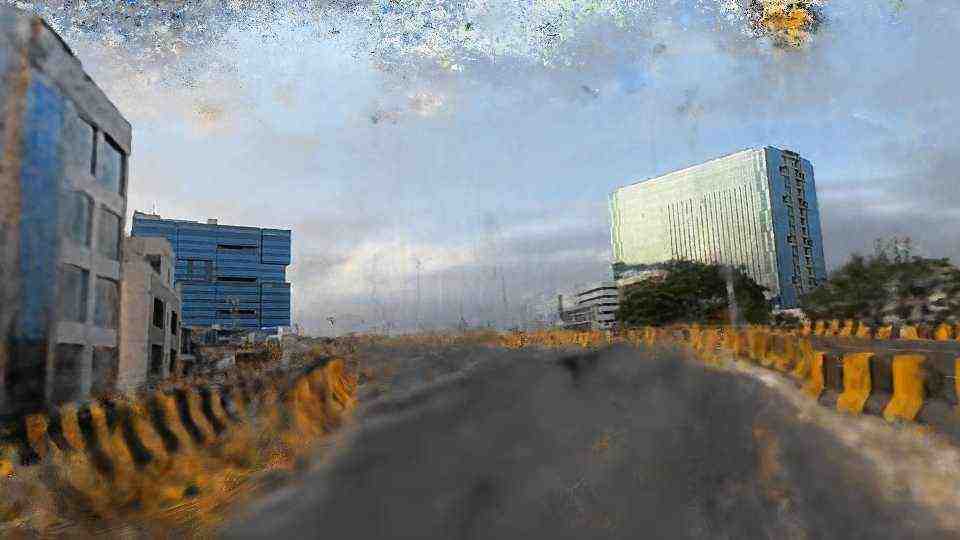} & \includegraphics[width=0.19\linewidth]{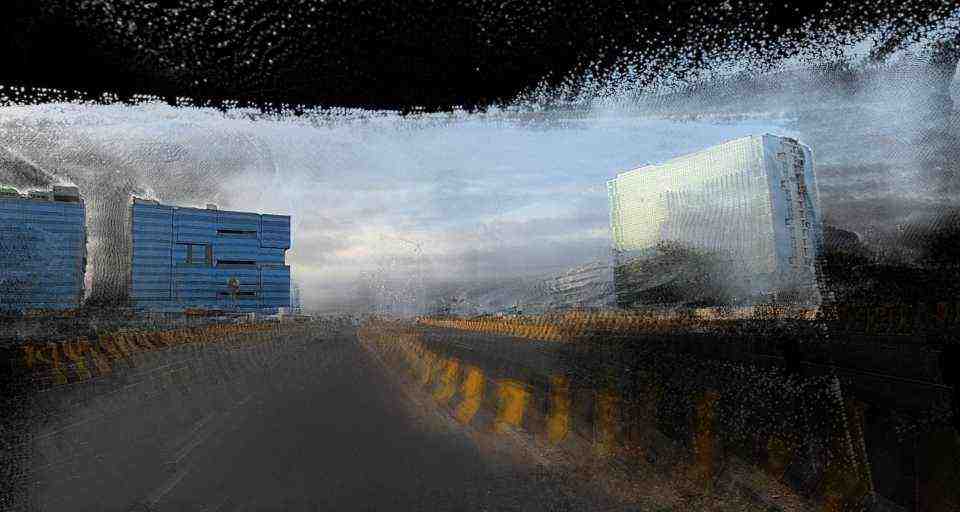} & \includegraphics[width=0.19\linewidth]{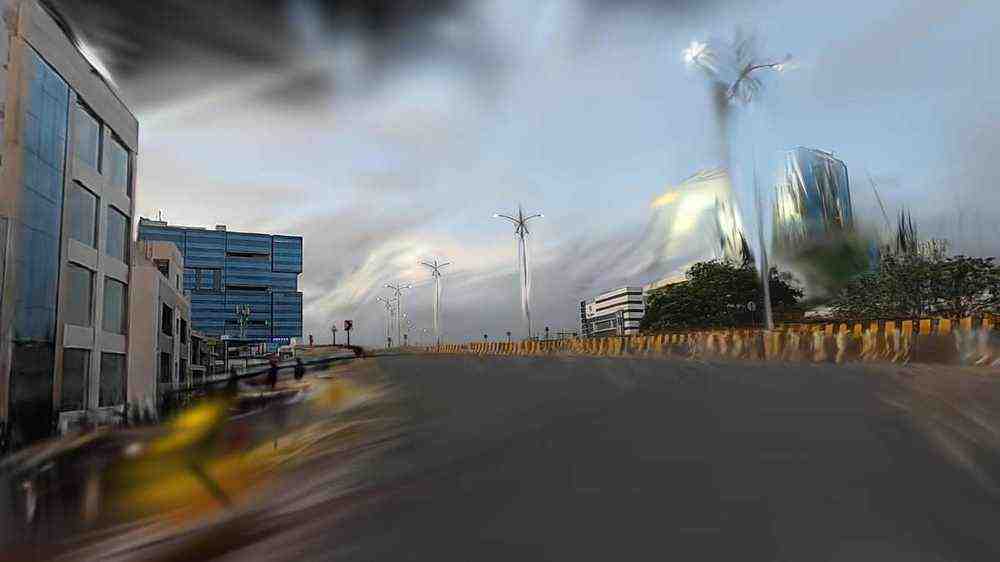} \\
& 17.01$|$0.61$|$0.56 & 16.46$|$0.72$|$0.45 & 10.37$|$0.42$|$0.73 & 18.02$|$0.68$|$0.50 \\

\end{tabular}
\end{adjustbox}

\caption{\textbf{Qualitative comparison across methods} Results from NVS methods trained on the $V^{C}$ sequences and rendered under the $T_{C\!\rightarrow S}$ viewpoint. 
Each column shows predictions from different NVS methods alongside the ground truth, illustrating the cross-domain 
generalization from car-trained models to scooty viewpoints. The values beneath each rendered image correspond 
to the NVS metrics PSNR$|$SSIM$|$LPIPS.
}
\label{fig:qual_car_scooty}
\end{figure*}

\figref{fig:qual_car_scooty} shows the $T_{C\!\rightarrow S}$ results, where models trained on $V^{C}$ must generalize to the scooty viewpoints. This cross-platform change introduces noticeable degradation due to differences in camera mounting height, field of view, and motion patterns. Methods relying heavily on learned priors exhibit stronger distortions and inconsistent geometry, while approaches with more explicit geometric reasoning demonstrate better resilience.

\begin{figure*}[ht!]
\centering
\scriptsize
\setlength{\tabcolsep}{1pt}
\renewcommand{\arraystretch}{0.8}

\begin{adjustbox}{max width=\textwidth}
\begin{tabular}{ccccc}

\textbf{GT} & \textbf{3DGS} & \textbf{NeRF} & \textbf{DepthSplat} & \textbf{PVG} \\

\includegraphics[width=0.19\linewidth]{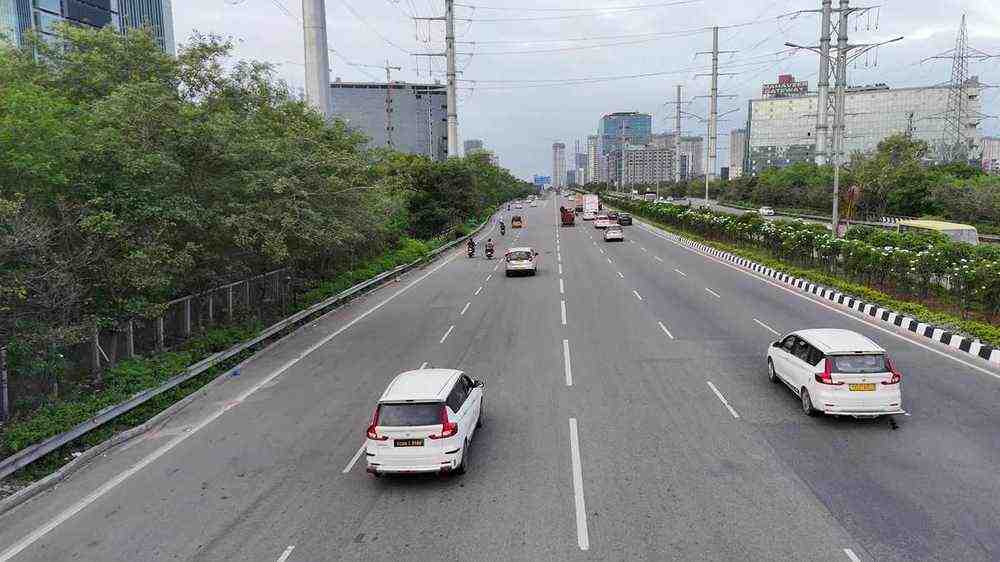} & \includegraphics[width=0.19\linewidth]{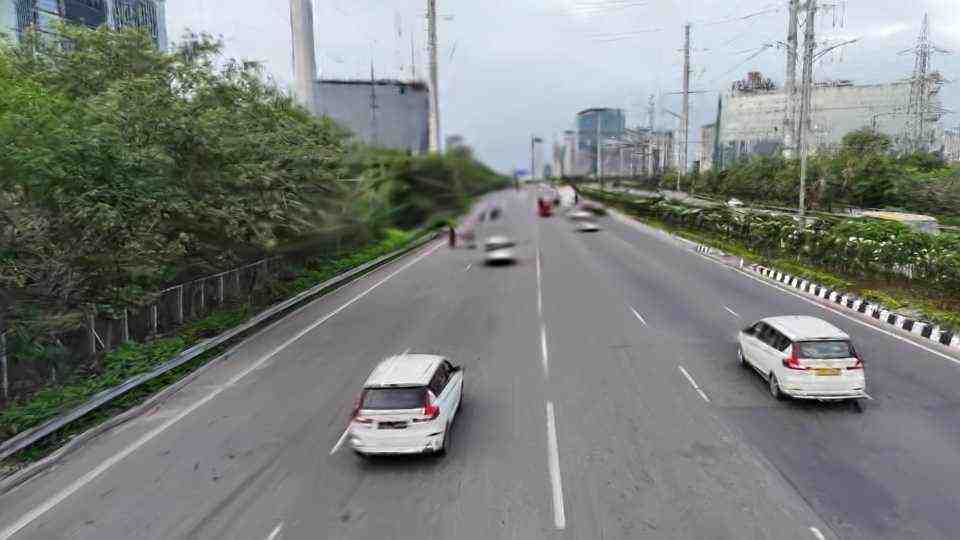} & \includegraphics[width=0.19\linewidth]{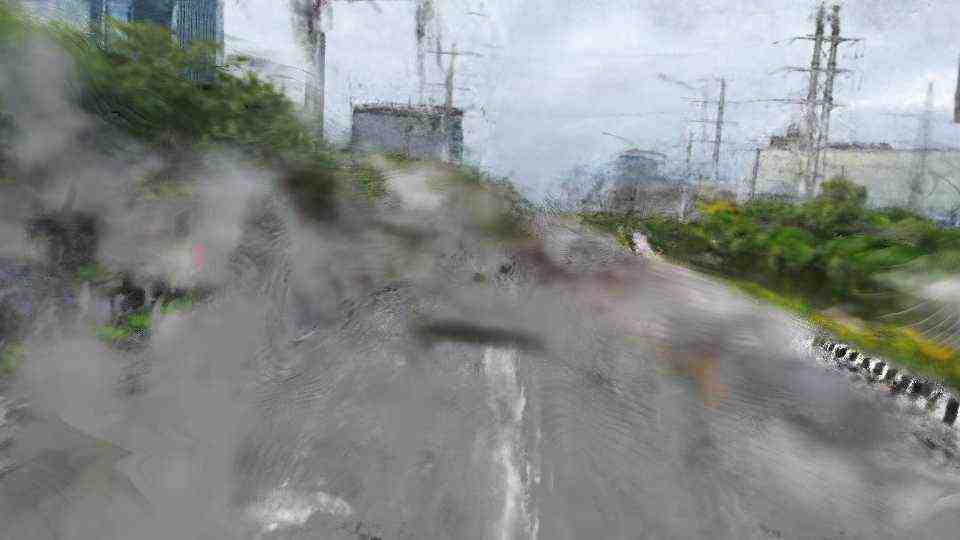} & \includegraphics[width=0.19\linewidth]{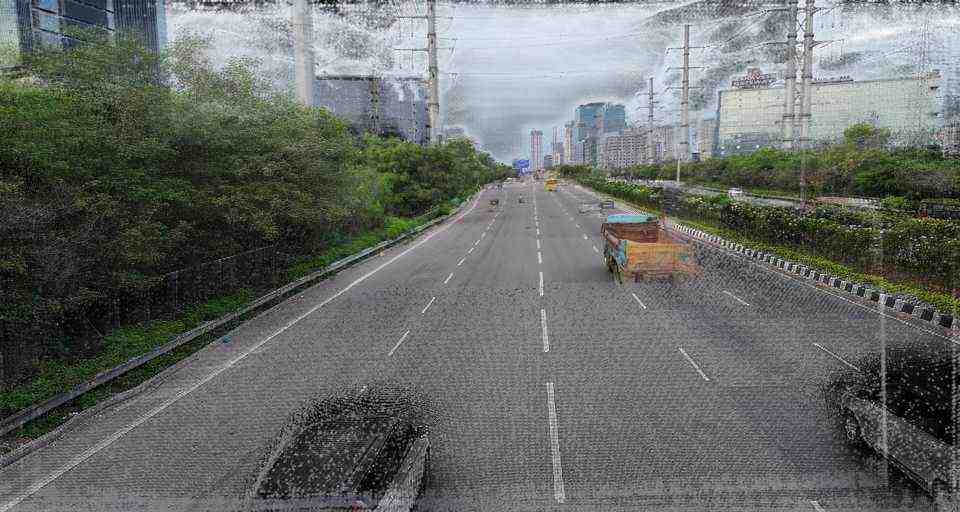} & \includegraphics[width=0.19\linewidth]{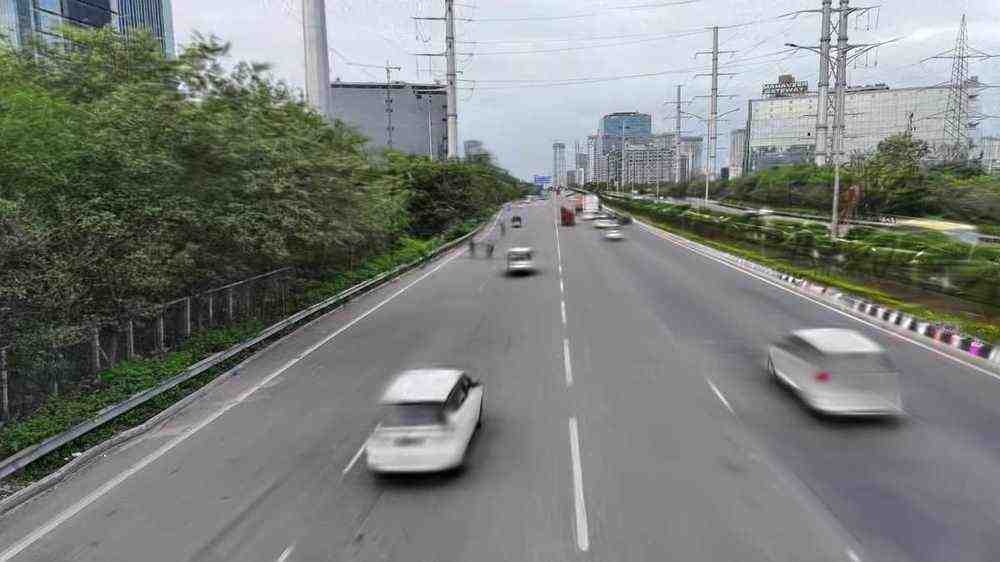} \\
& 13.96$|$0.42$|$0.79 & 24.37$|$0.73$|$0.45 & 16.18$|$0.42$|$0.58 & 24.12$|$0.64$|$0.57 \\
\includegraphics[width=0.19\linewidth]{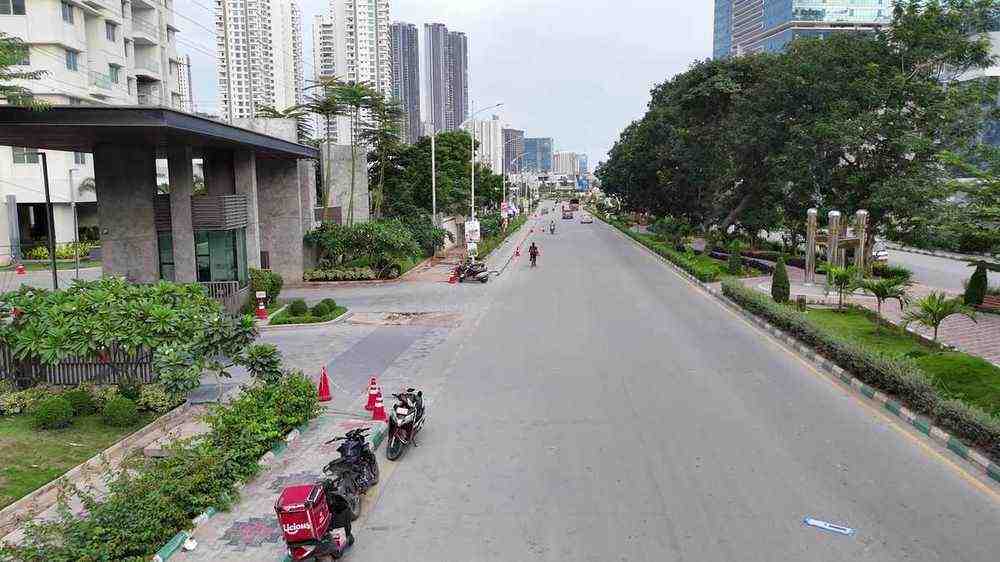} & \includegraphics[width=0.19\linewidth]{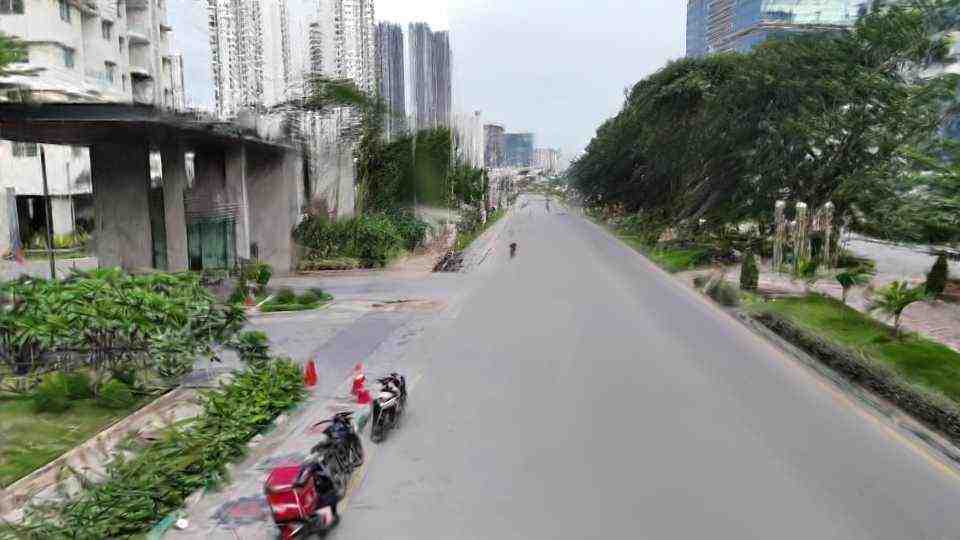} & \includegraphics[width=0.19\linewidth]{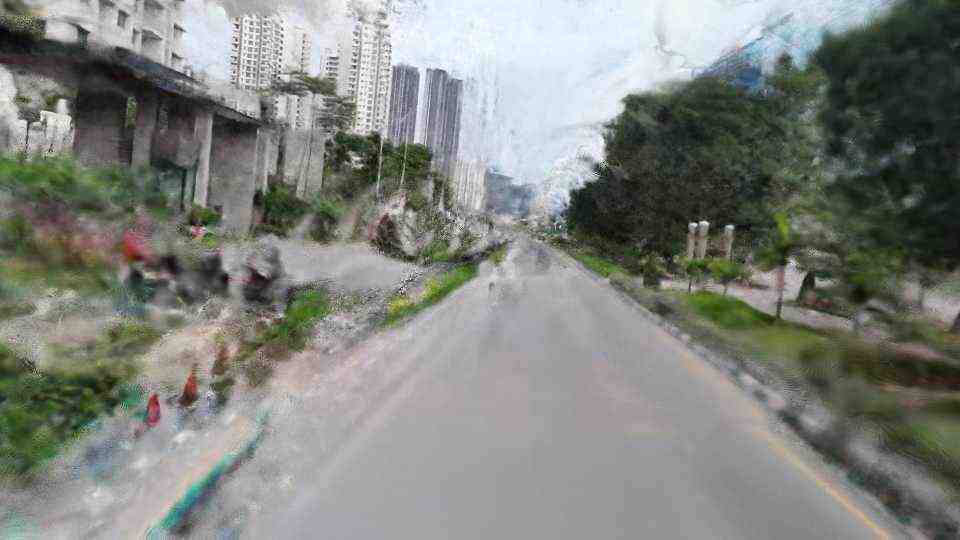} & \includegraphics[width=0.19\linewidth]{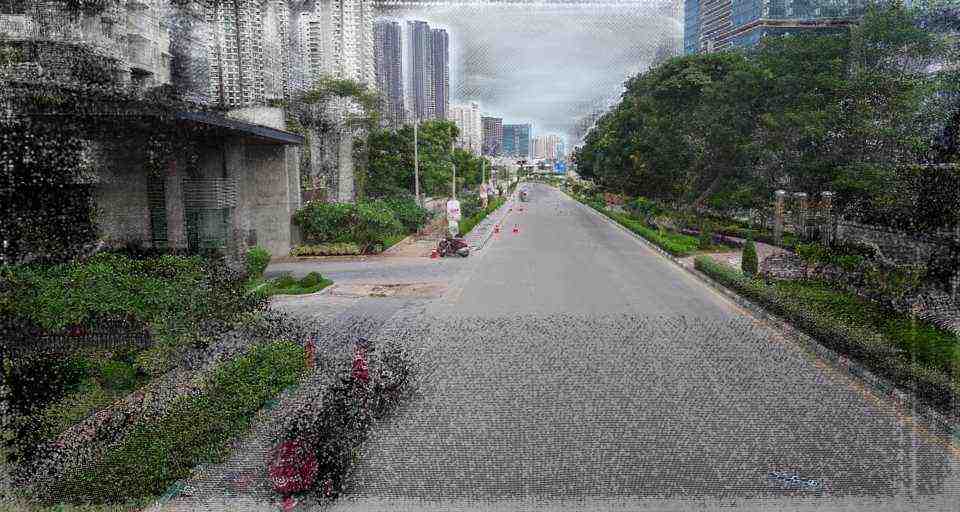} & \includegraphics[width=0.19\linewidth]{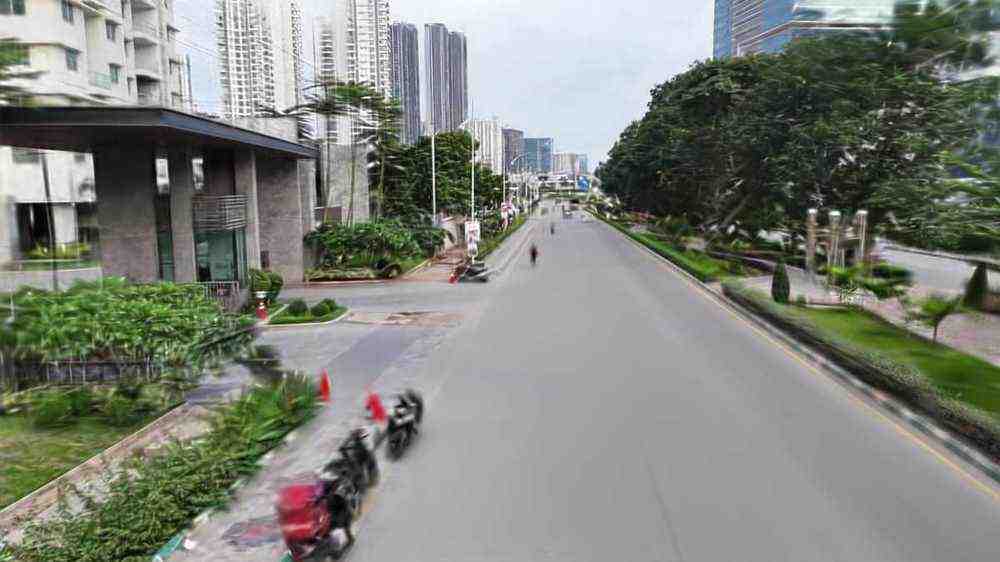} \\
& 12.18$|$0.38$|$0.75 & 23.09$|$0.70$|$0.50 & 13.59$|$0.35$|$0.63 & 21.18$|$0.61$|$0.54 \\
\includegraphics[width=0.19\linewidth]{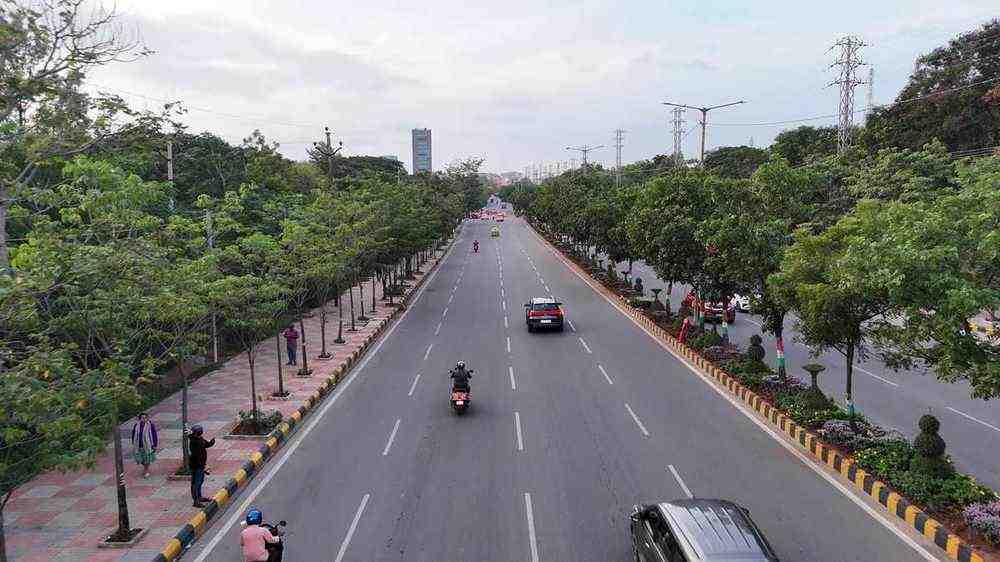} & \includegraphics[width=0.19\linewidth]{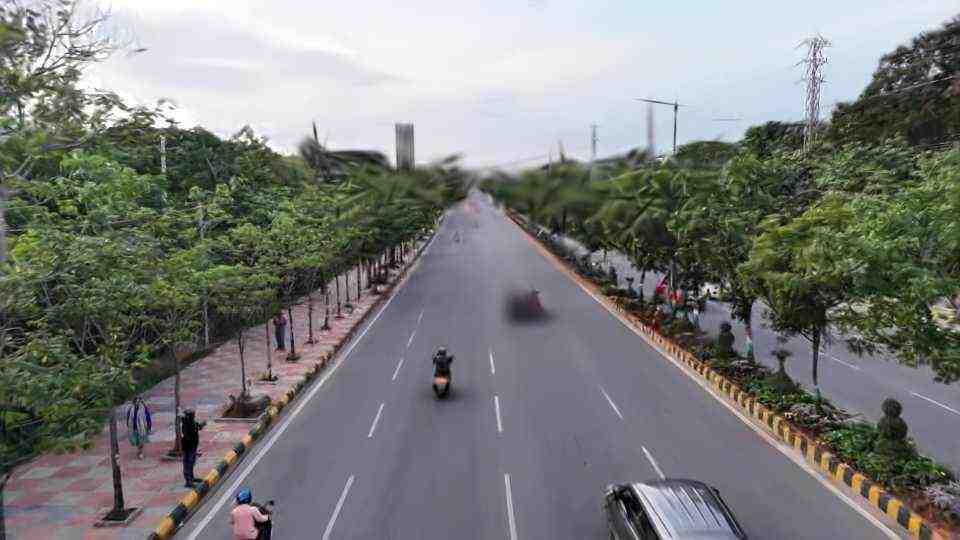} & \includegraphics[width=0.19\linewidth]{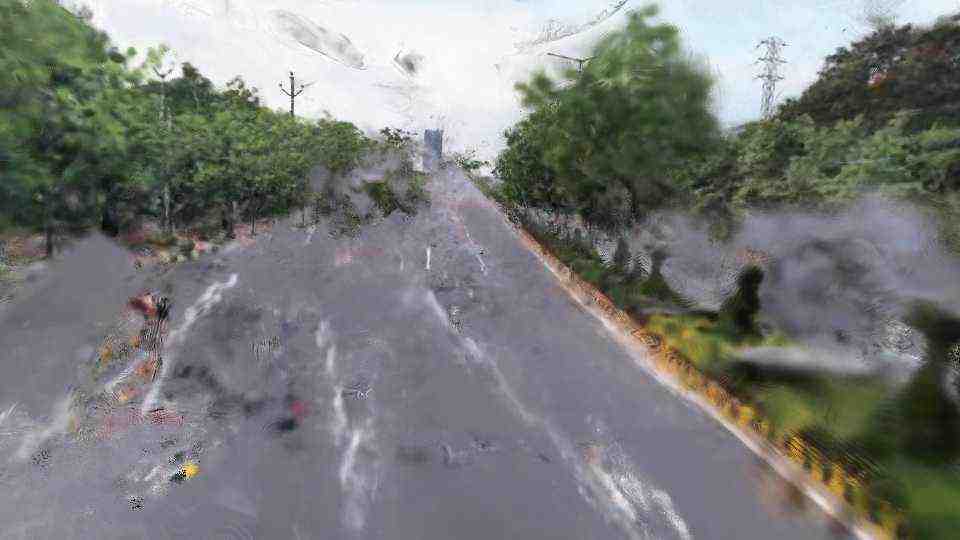} & \includegraphics[width=0.19\linewidth]{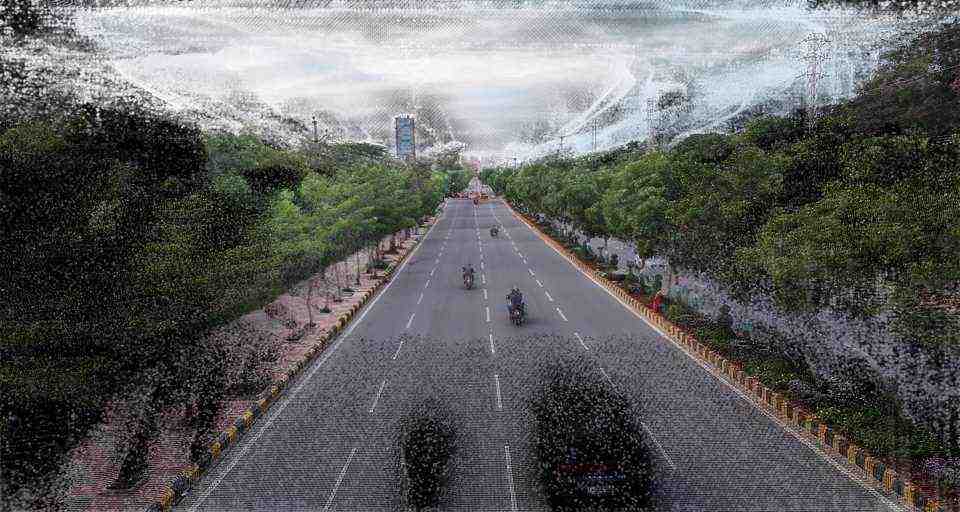} & \includegraphics[width=0.19\linewidth]{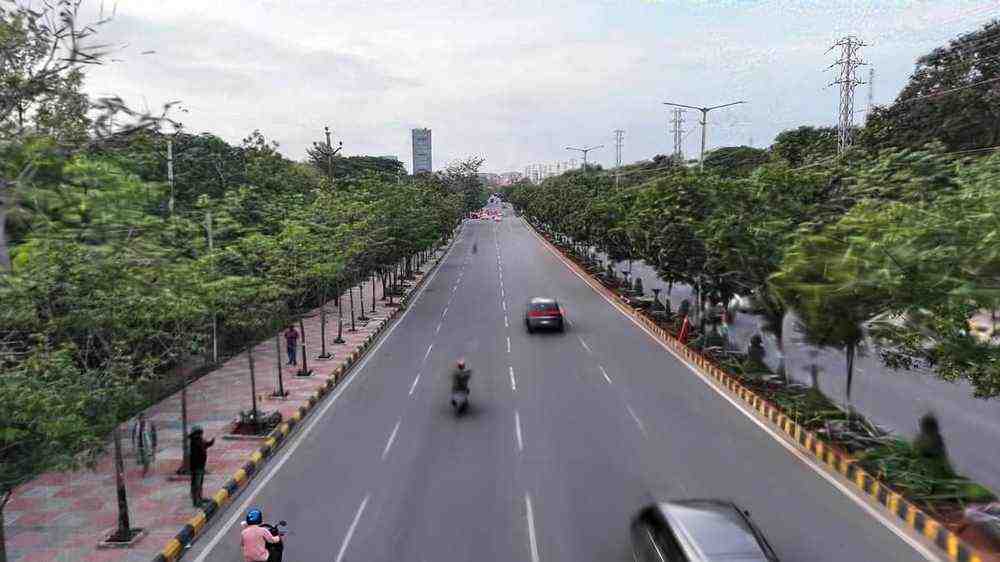} \\
& 12.69$|$0.40$|$0.77 & 24.48$|$0.73$|$0.39 & 14.28$|$0.32$|$0.64 & 24.06$|$0.70$|$0.47 \\
\includegraphics[width=0.19\linewidth]{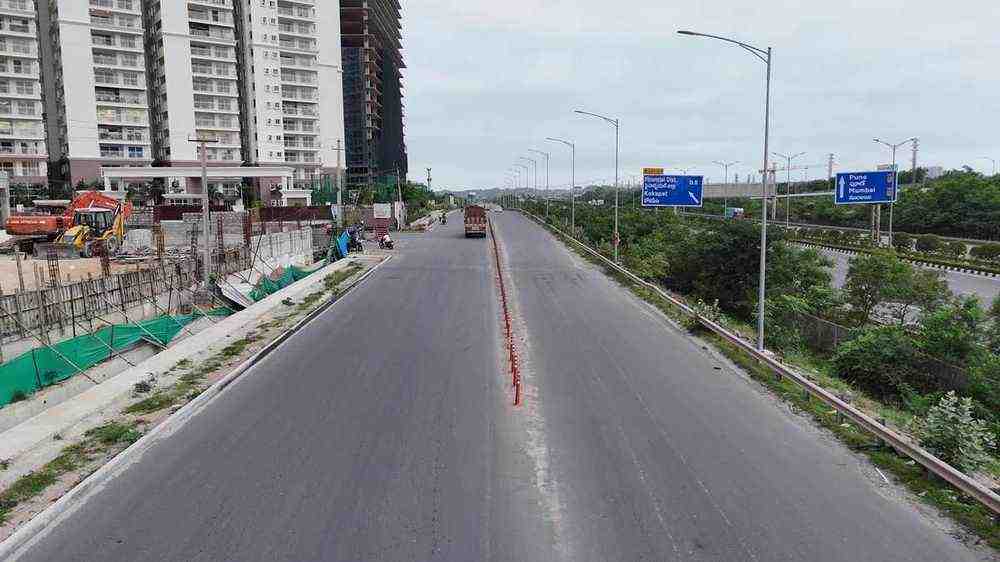} & \includegraphics[width=0.19\linewidth]{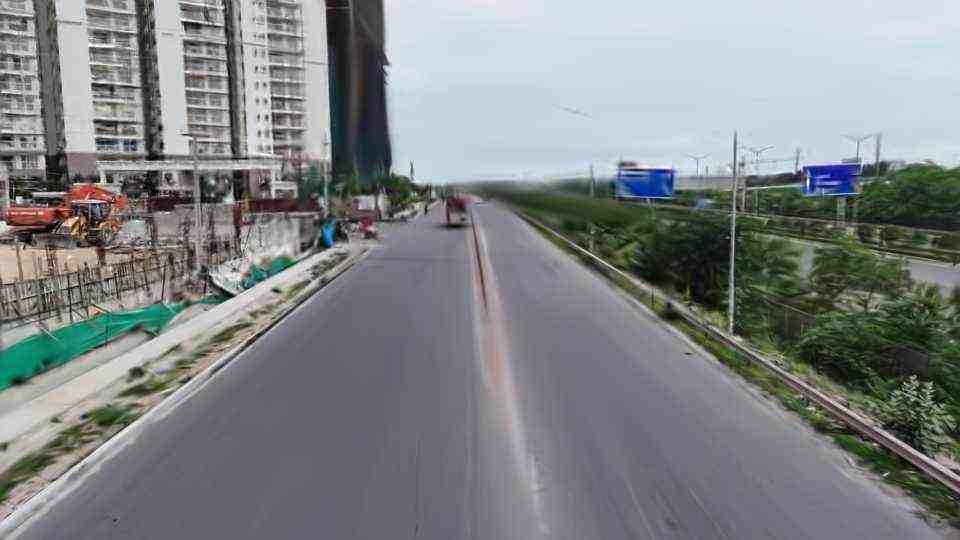} & \includegraphics[width=0.19\linewidth]{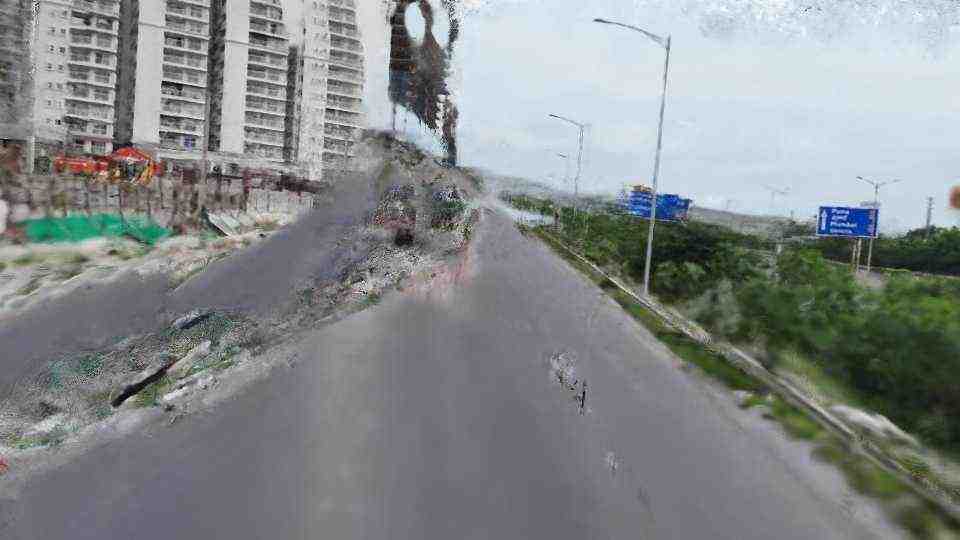} & \includegraphics[width=0.19\linewidth]{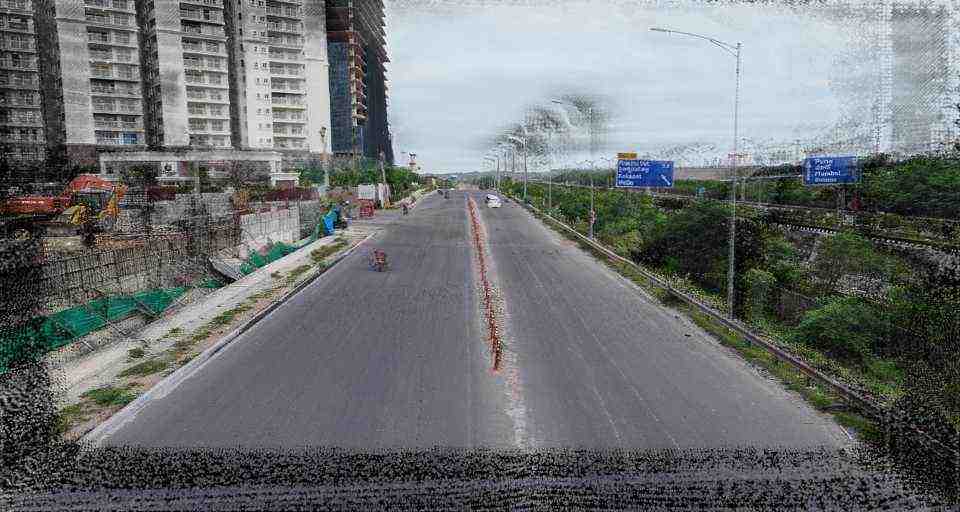} & \includegraphics[width=0.19\linewidth]{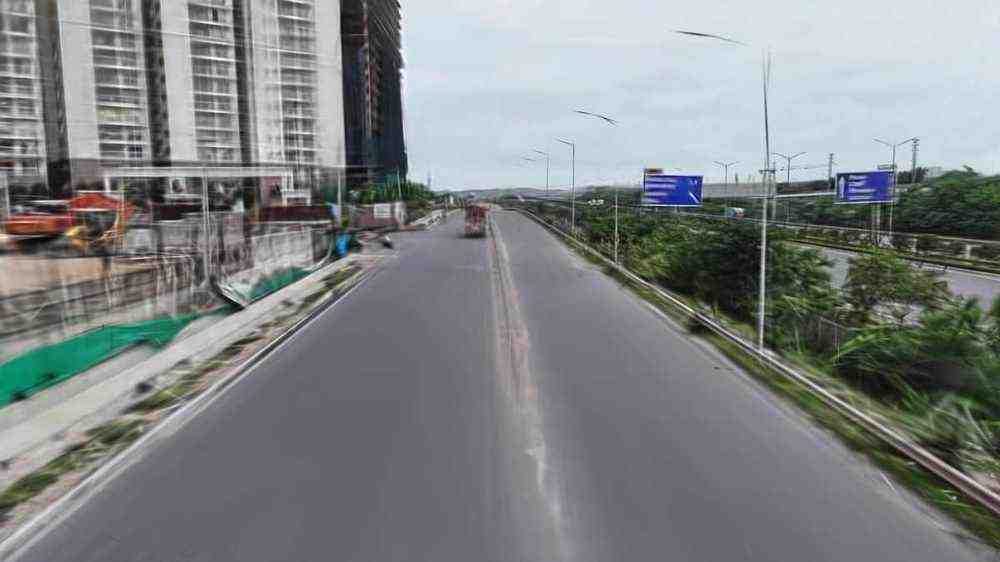} \\
& 13.28$|$0.48$|$0.68 & 23.57$|$0.74$|$0.49 & 16.10$|$0.57$|$0.48 & 23.61$|$0.73$|$0.53 \\
\includegraphics[width=0.19\linewidth]{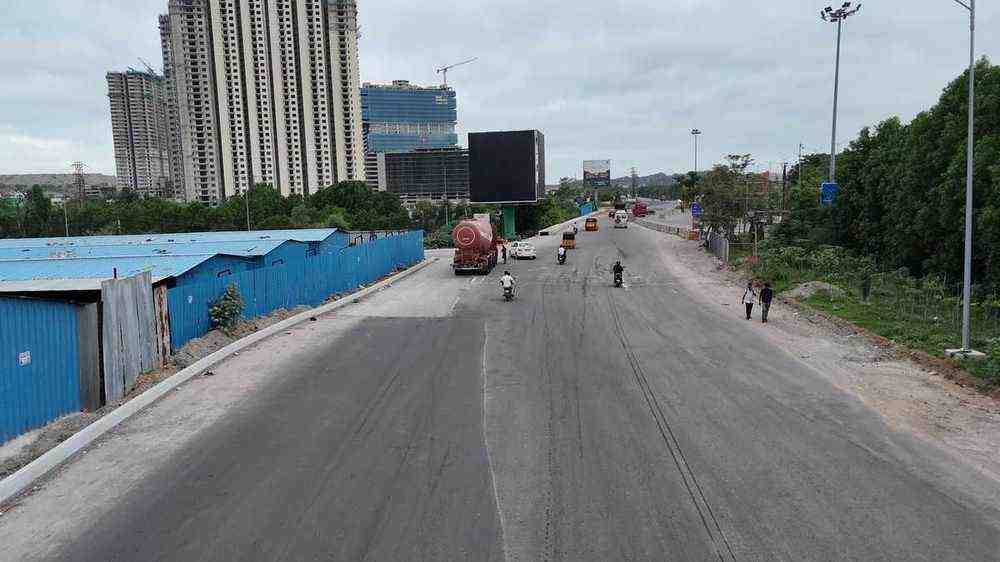} & \includegraphics[width=0.19\linewidth]{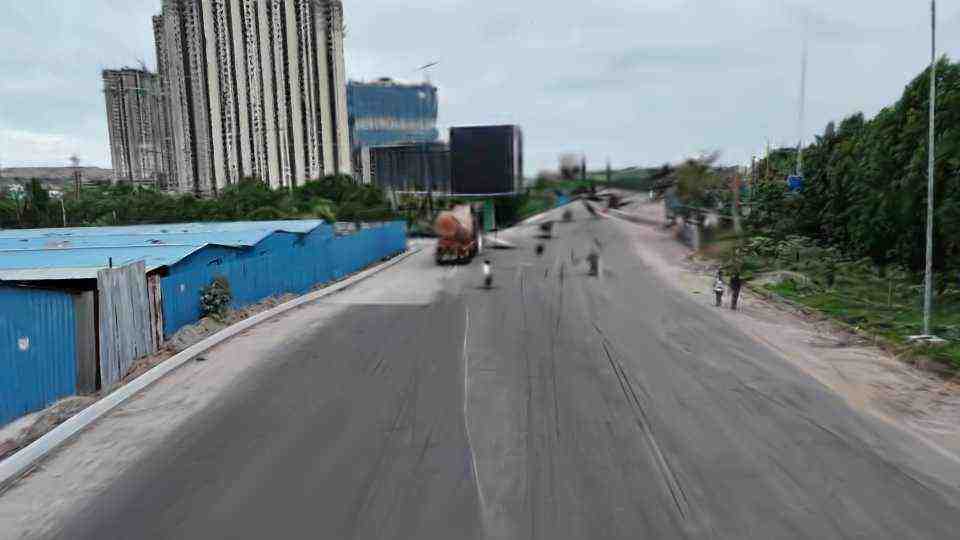} & \includegraphics[width=0.19\linewidth]{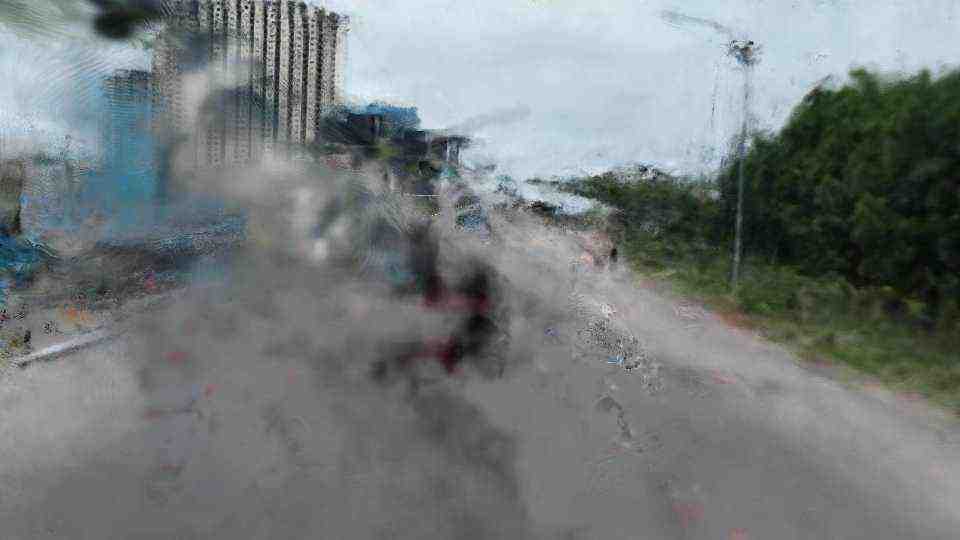} & \includegraphics[width=0.19\linewidth]{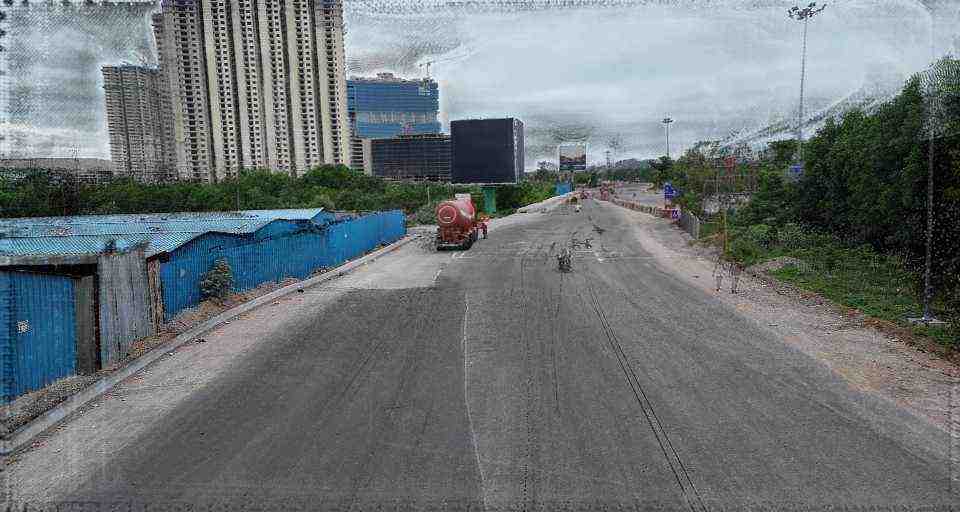} & \includegraphics[width=0.19\linewidth]{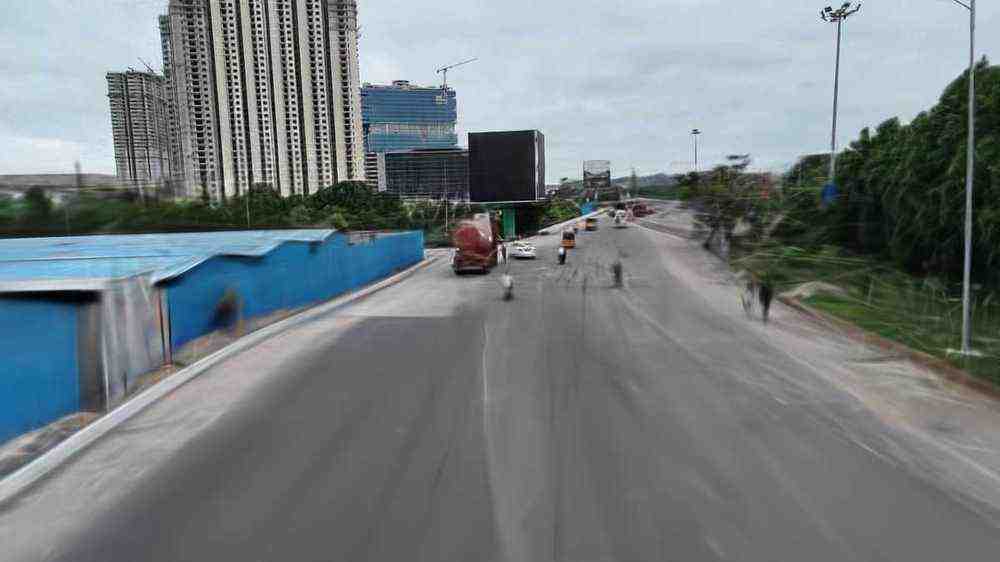} \\
& 14.06$|$0.52$|$0.74 & 25.62$|$0.74$|$0.49 & 20.21$|$0.72$|$0.38 & 24.09$|$0.72$|$0.52 \\
\includegraphics[width=0.19\linewidth]{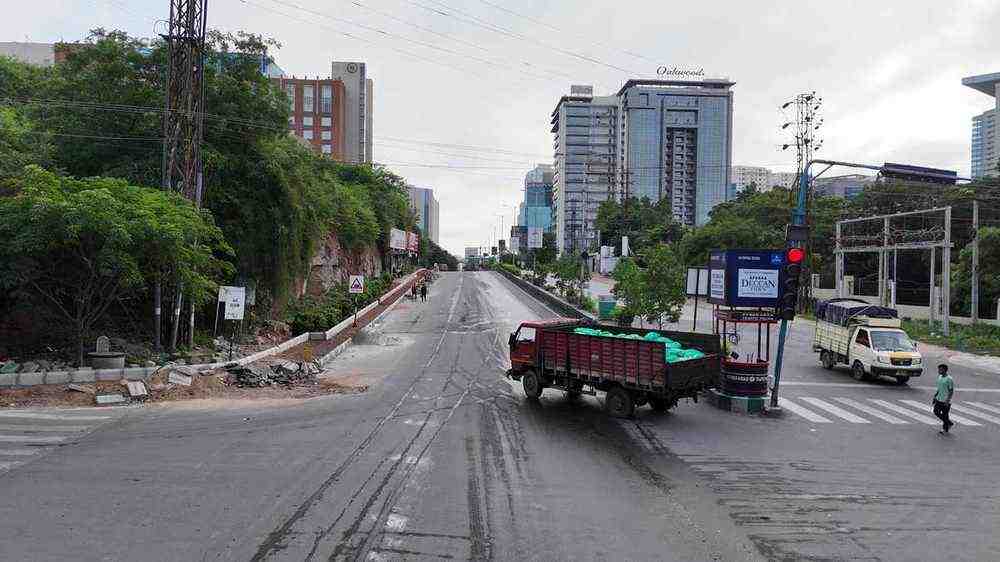} & \includegraphics[width=0.19\linewidth]{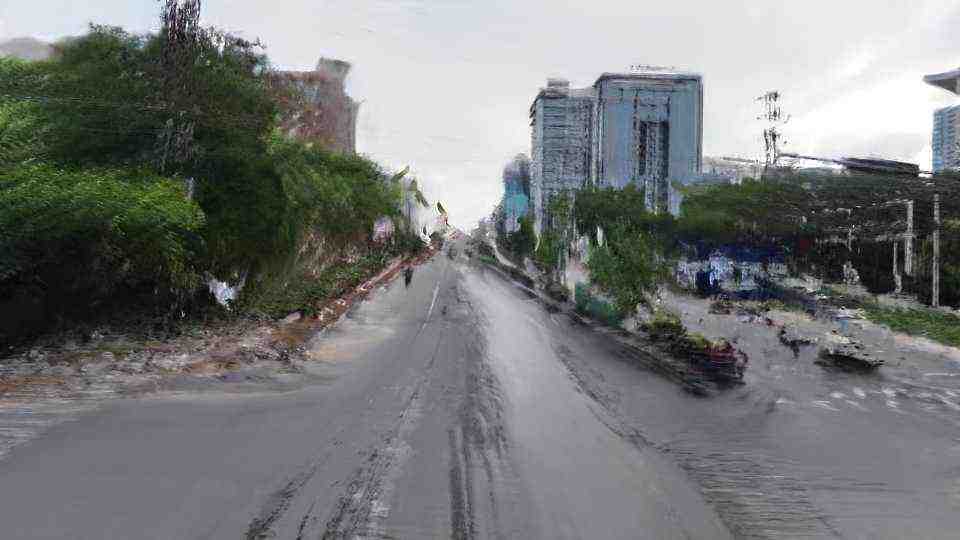} & \includegraphics[width=0.19\linewidth]{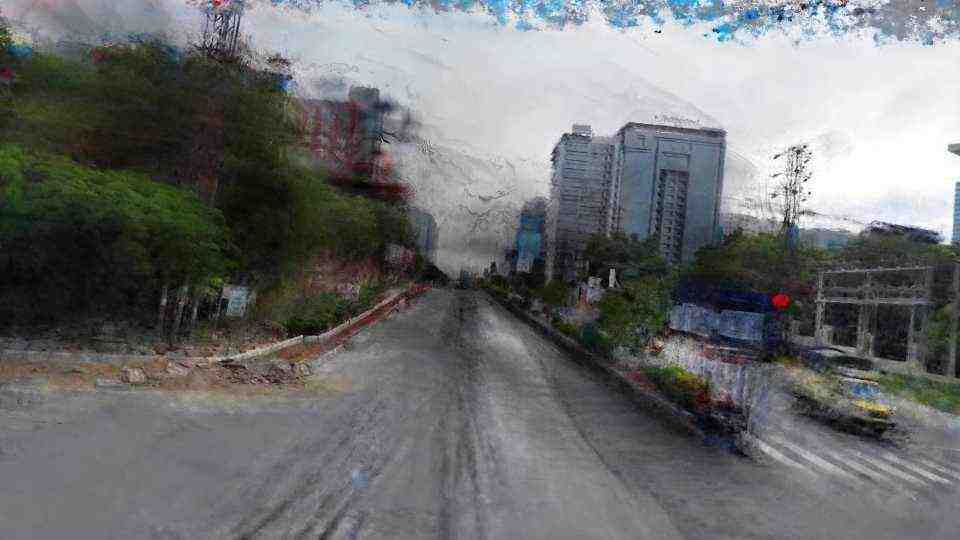} & \includegraphics[width=0.19\linewidth]{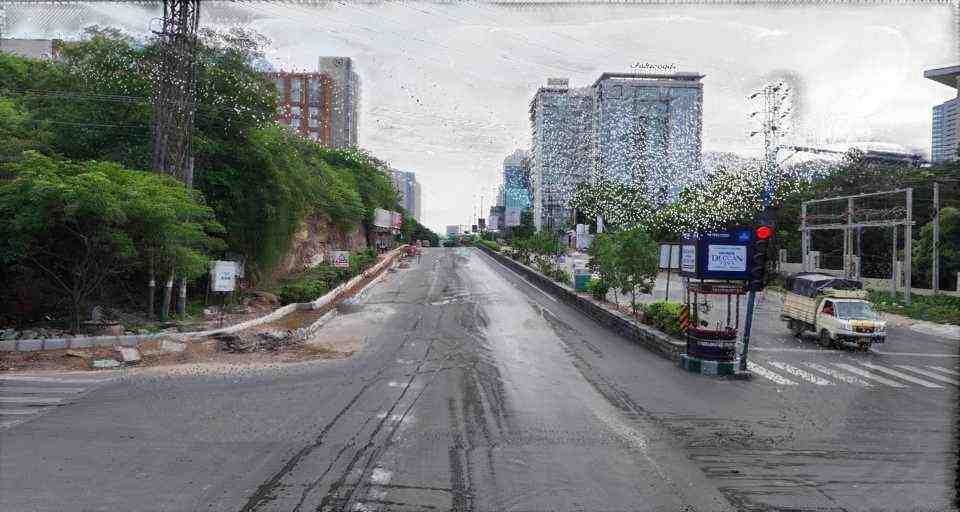} & \includegraphics[width=0.19\linewidth]{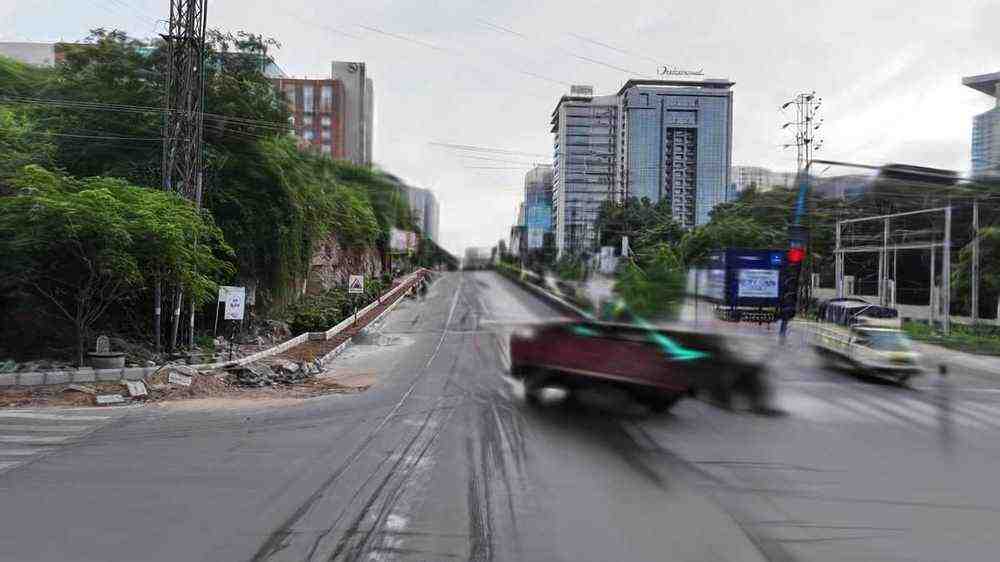} \\
& 12.99$|$0.39$|$0.77 & 24.42$|$0.72$|$0.43 & 18.27$|$0.61$|$0.47 & 17.88$|$0.52$|$0.61 \\
\includegraphics[width=0.19\linewidth]{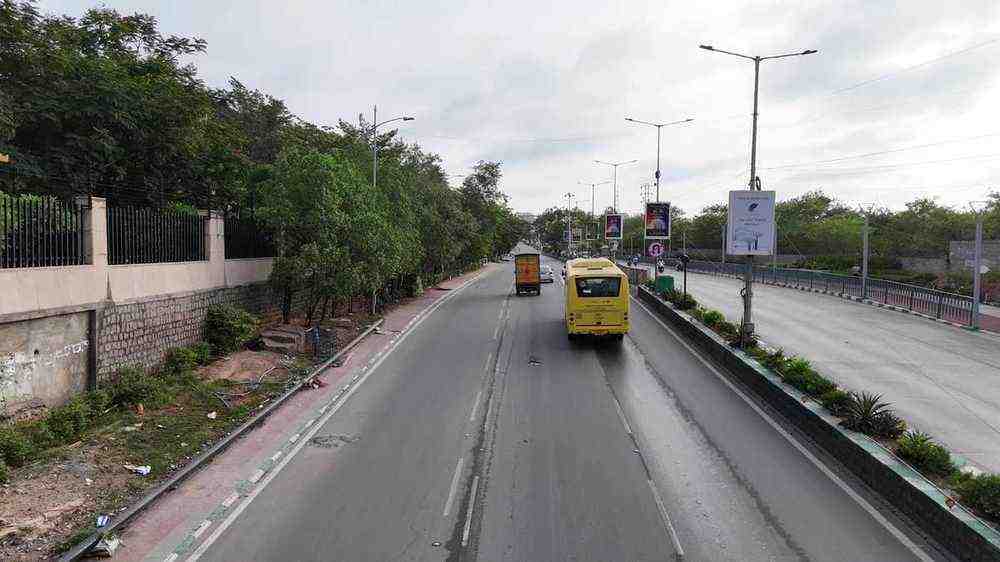} & \includegraphics[width=0.19\linewidth]{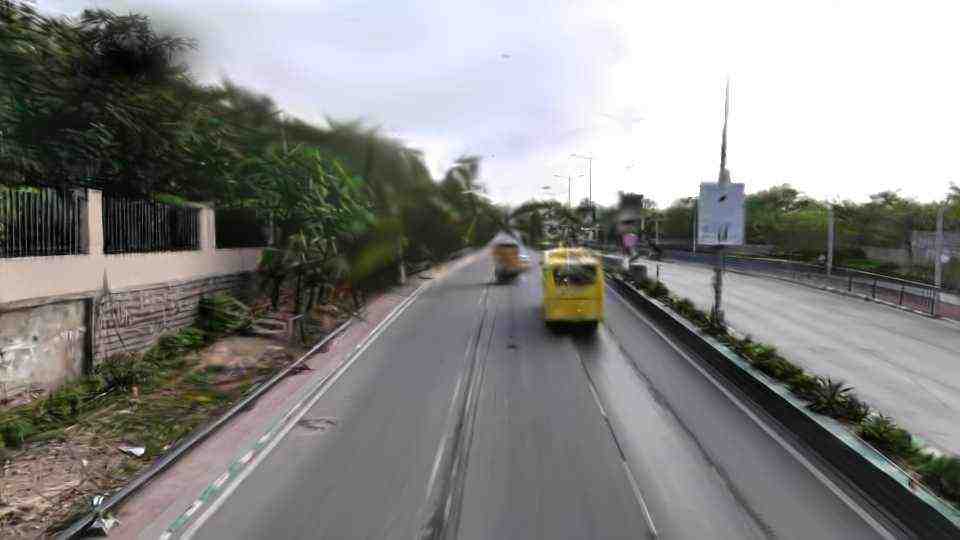} & \includegraphics[width=0.19\linewidth]{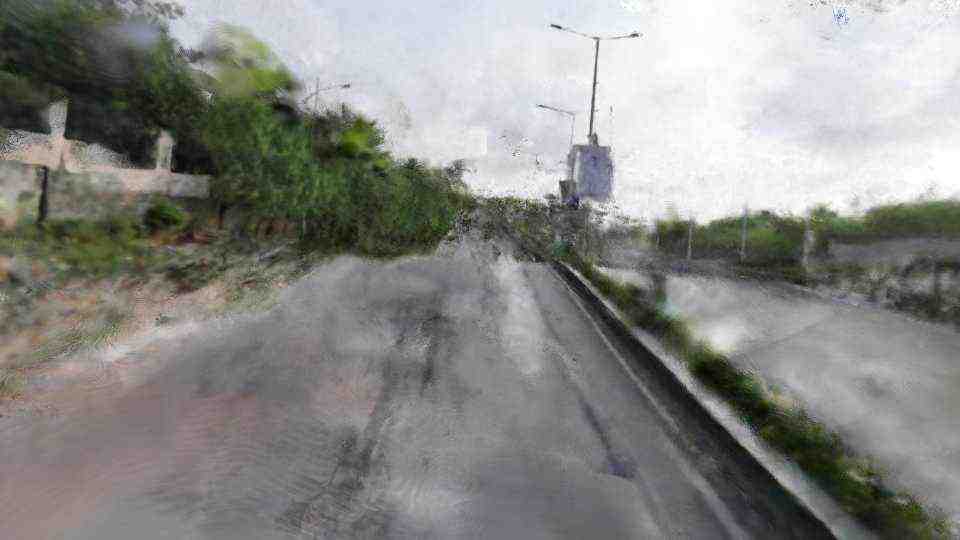} & \includegraphics[width=0.19\linewidth]{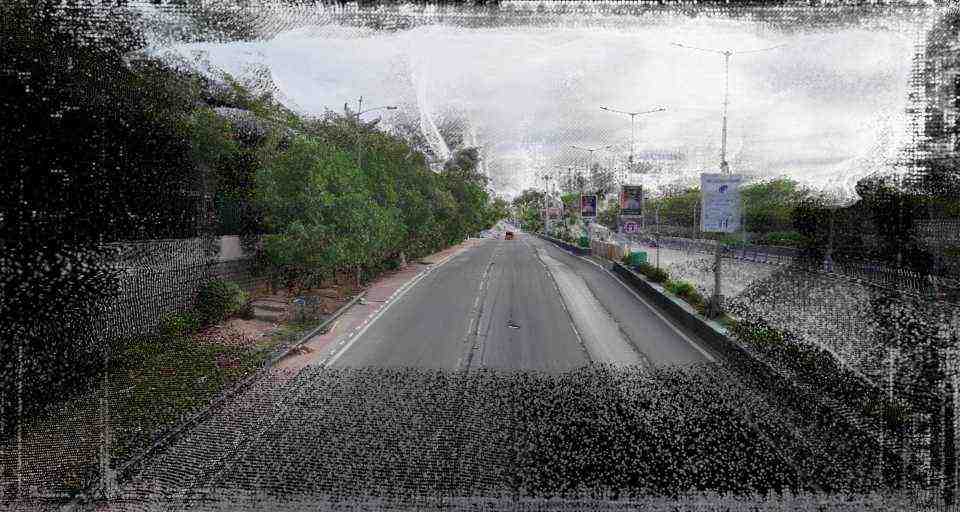} & \includegraphics[width=0.19\linewidth]{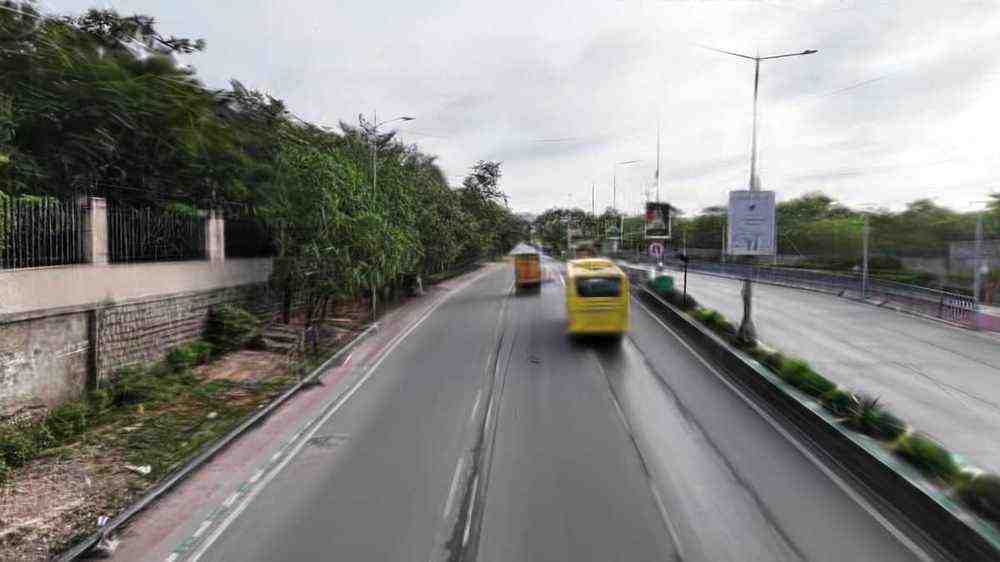} \\
& 13.63$|$0.48$|$0.73 & 24.50$|$0.73$|$0.44 & 12.30$|$0.33$|$0.69 & 22.85$|$0.70$|$0.52 \\
\includegraphics[width=0.19\linewidth]{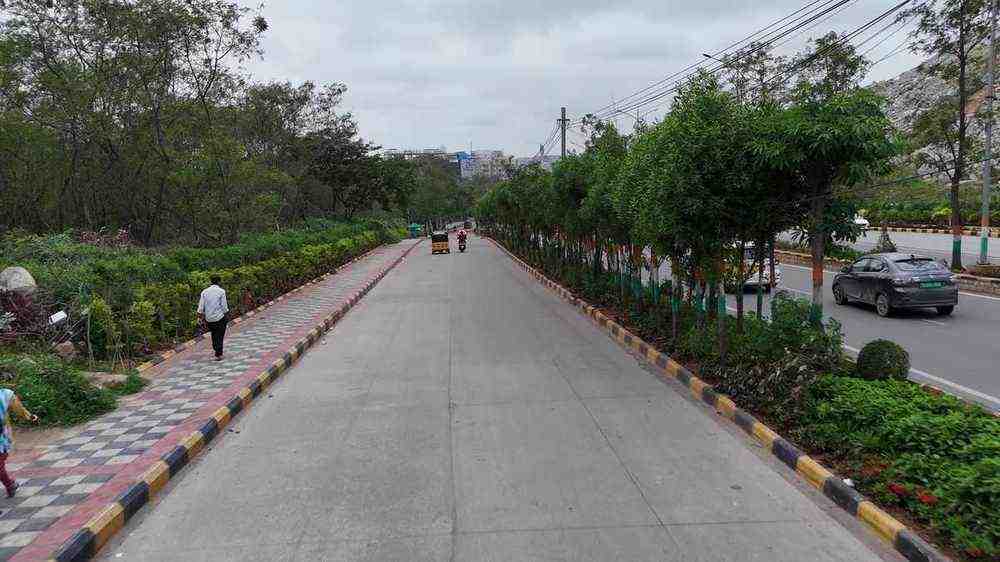} & \includegraphics[width=0.19\linewidth]{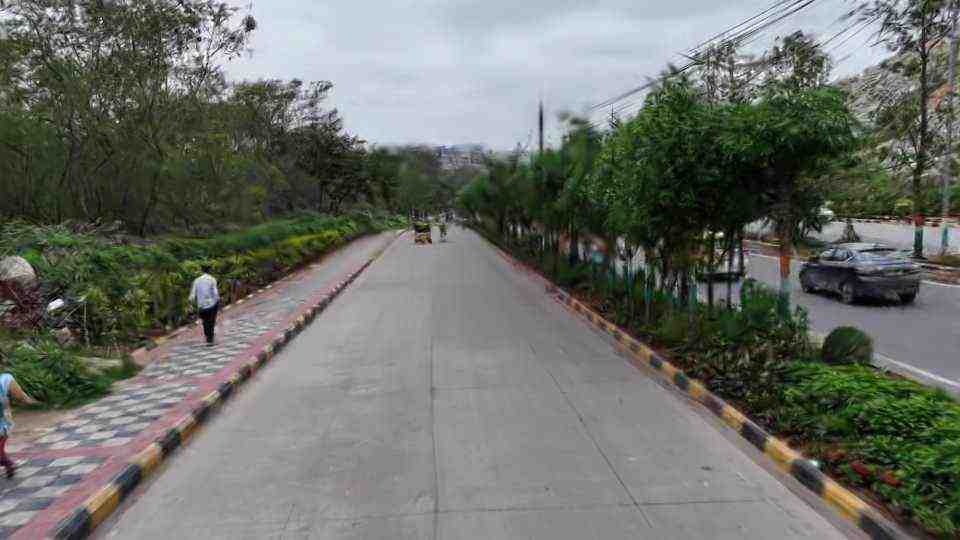} & \includegraphics[width=0.19\linewidth]{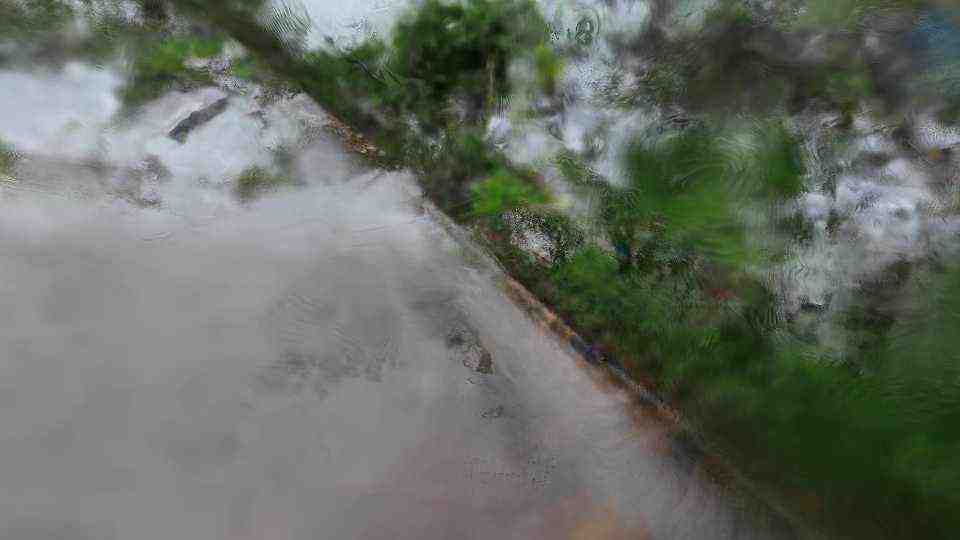} & \includegraphics[width=0.19\linewidth]{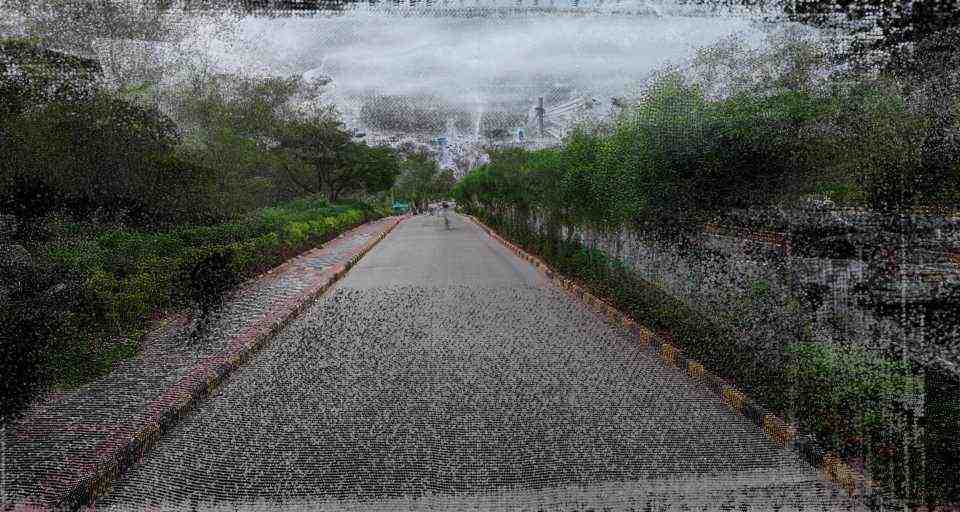} & \includegraphics[width=0.19\linewidth]{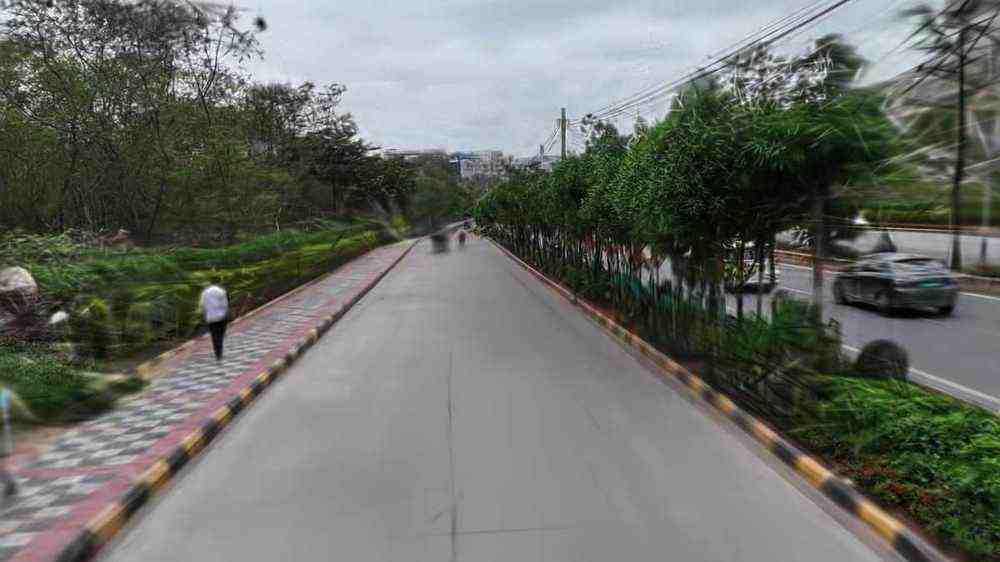} \\
& 10.43$|$0.36$|$0.86 & 23.69$|$0.68$|$0.49 & 13.69$|$0.25$|$0.70 & 23.71$|$0.66$|$0.50 \\
\includegraphics[width=0.19\linewidth]{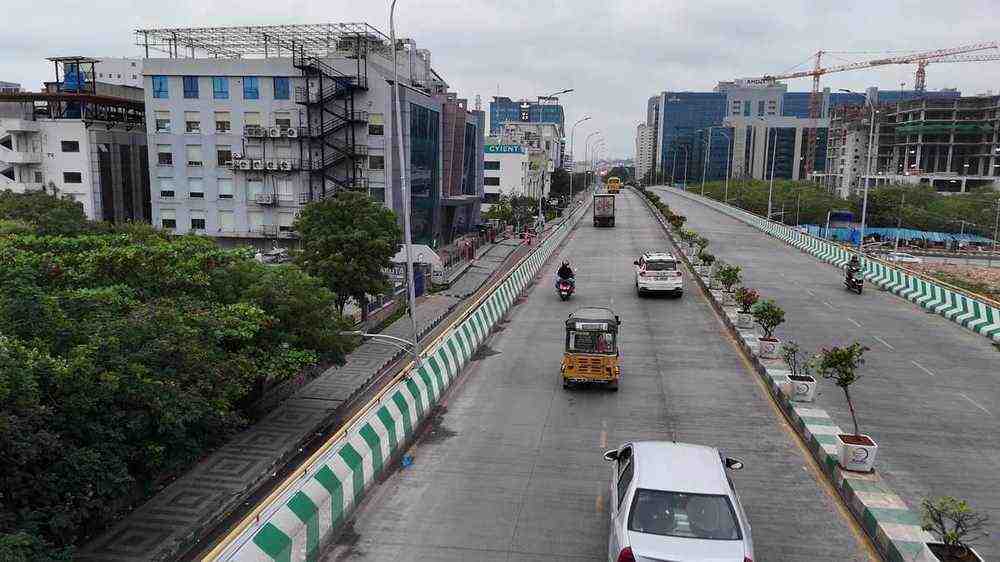} & \includegraphics[width=0.19\linewidth]{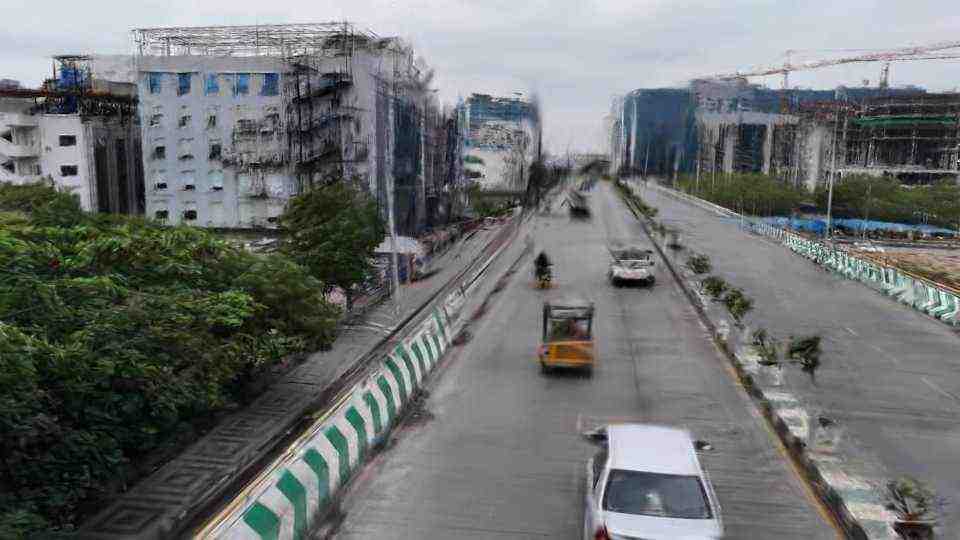} & \includegraphics[width=0.19\linewidth]{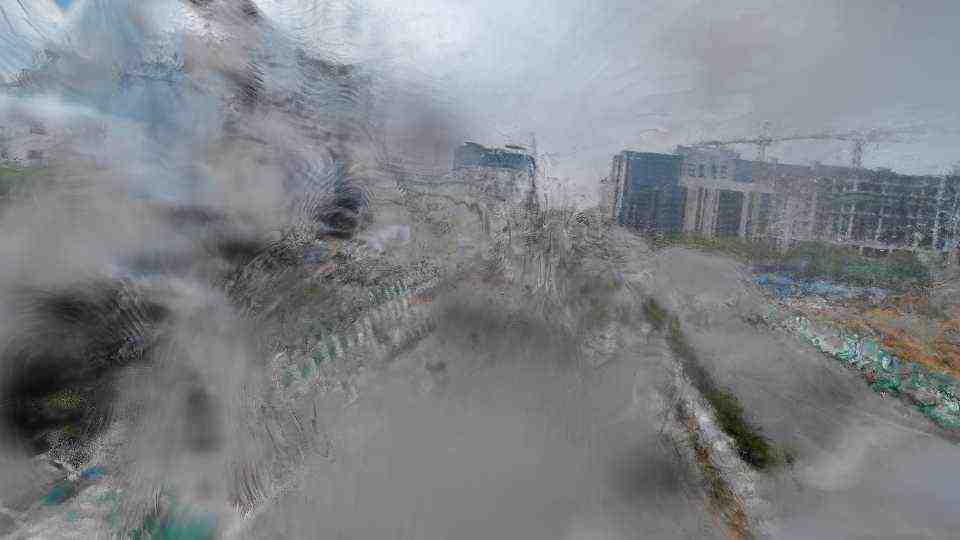} & \includegraphics[width=0.19\linewidth]{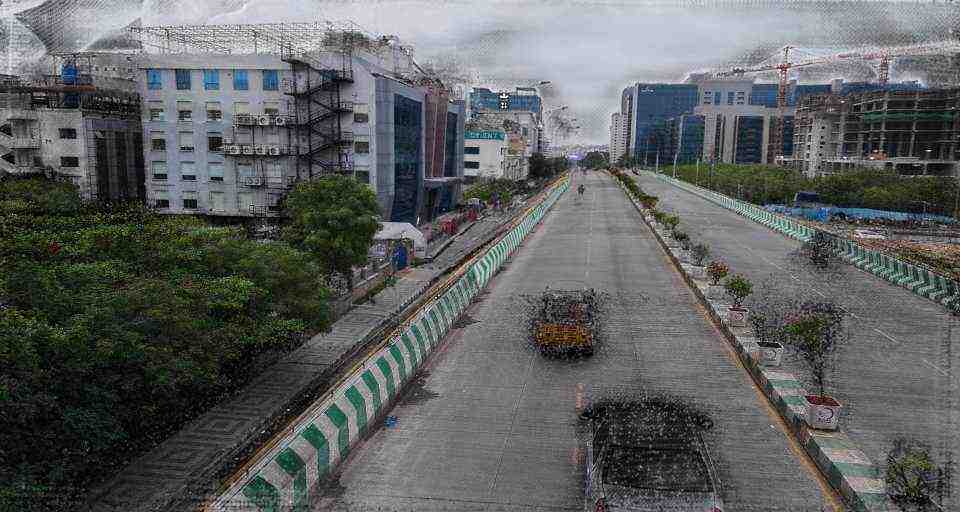} & \includegraphics[width=0.19\linewidth]{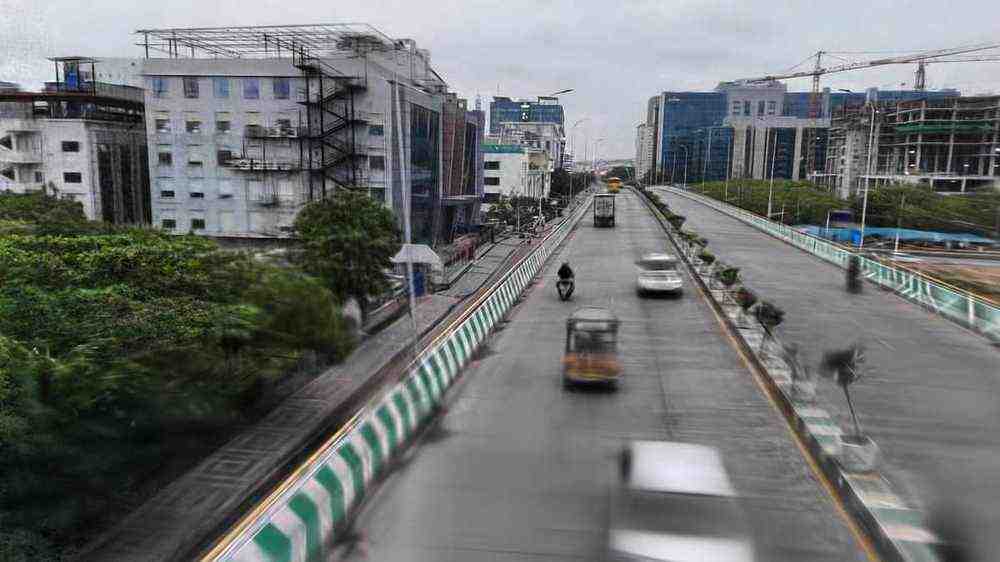} \\
& 12.57$|$0.38$|$0.81 & 23.39$|$0.70$|$0.44 & 17.28$|$0.65$|$0.41 & 21.64$|$0.62$|$0.54 \\
\includegraphics[width=0.19\linewidth]{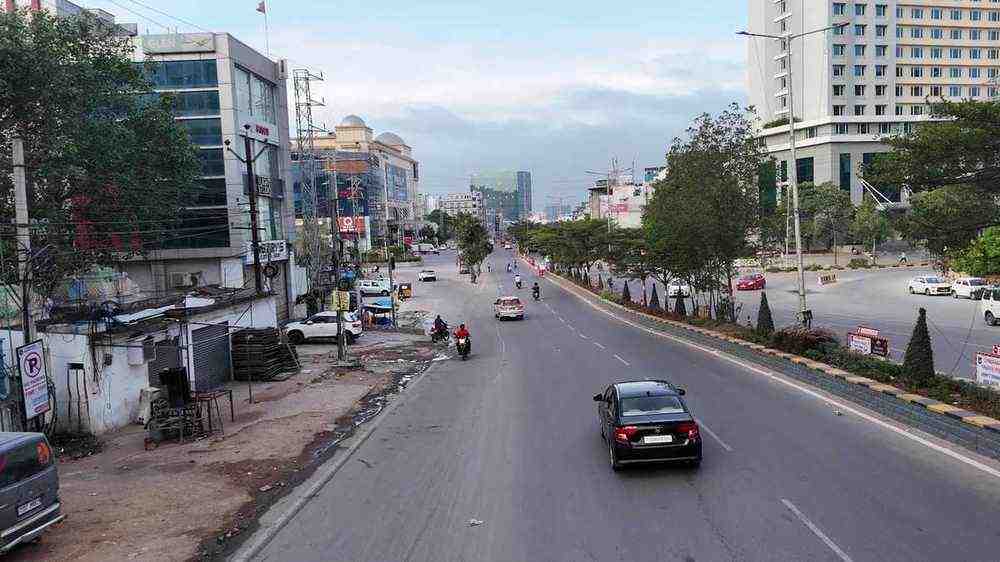} & \includegraphics[width=0.19\linewidth]{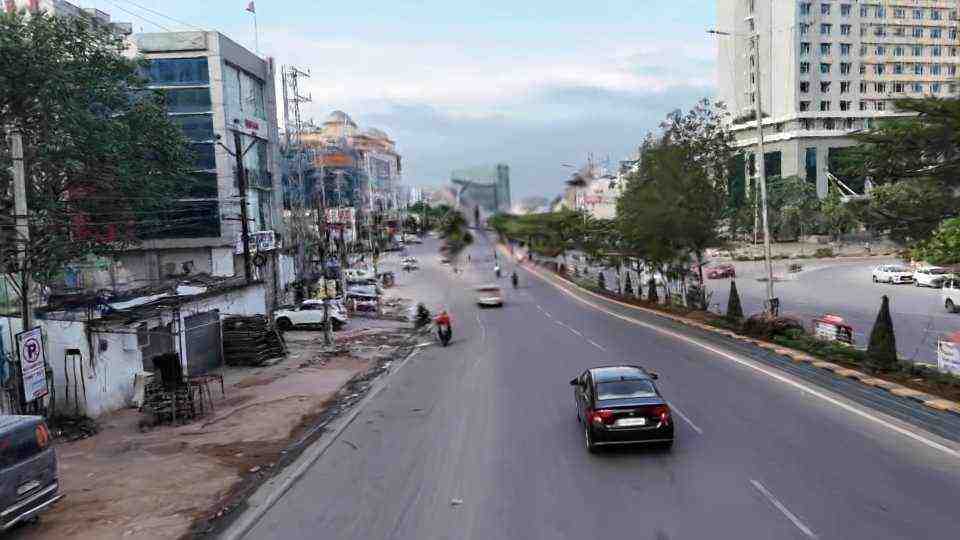} & \includegraphics[width=0.19\linewidth]{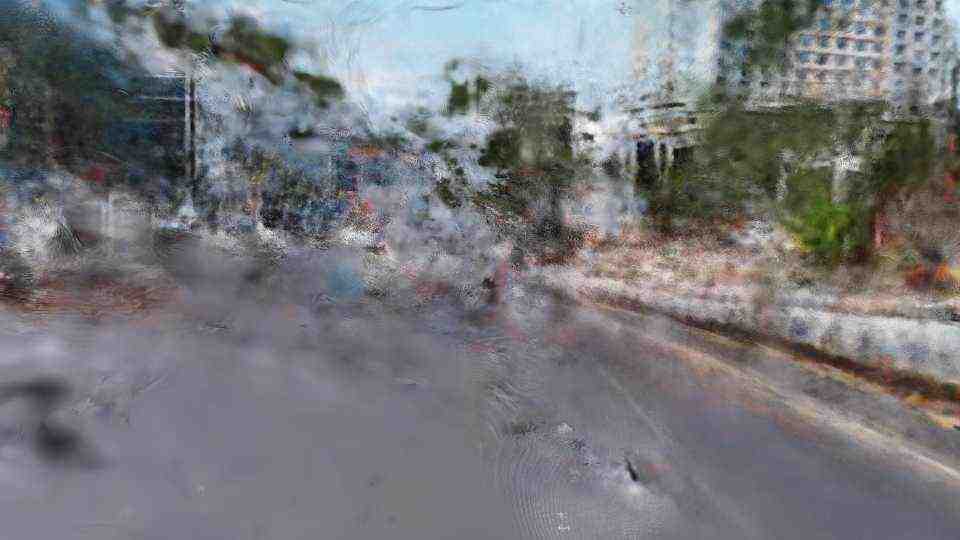} & \includegraphics[width=0.19\linewidth]{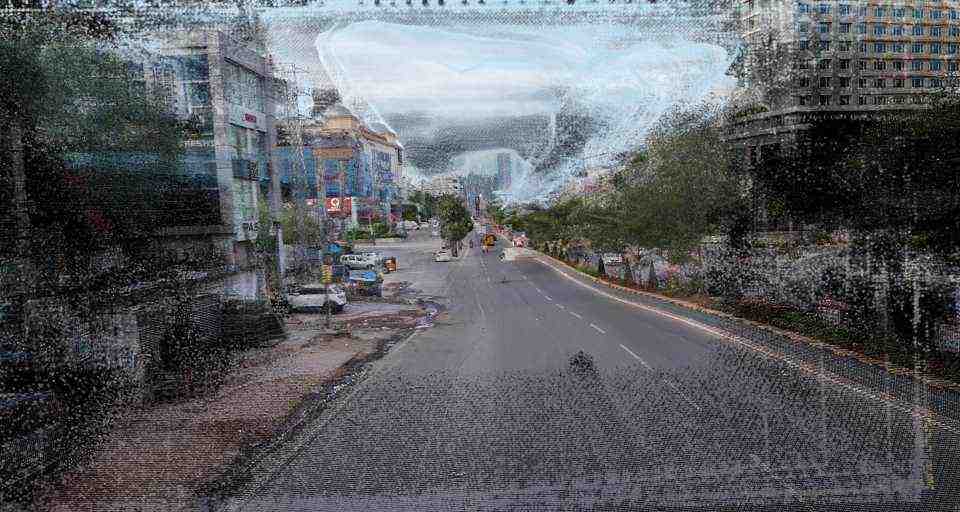} & \includegraphics[width=0.19\linewidth]{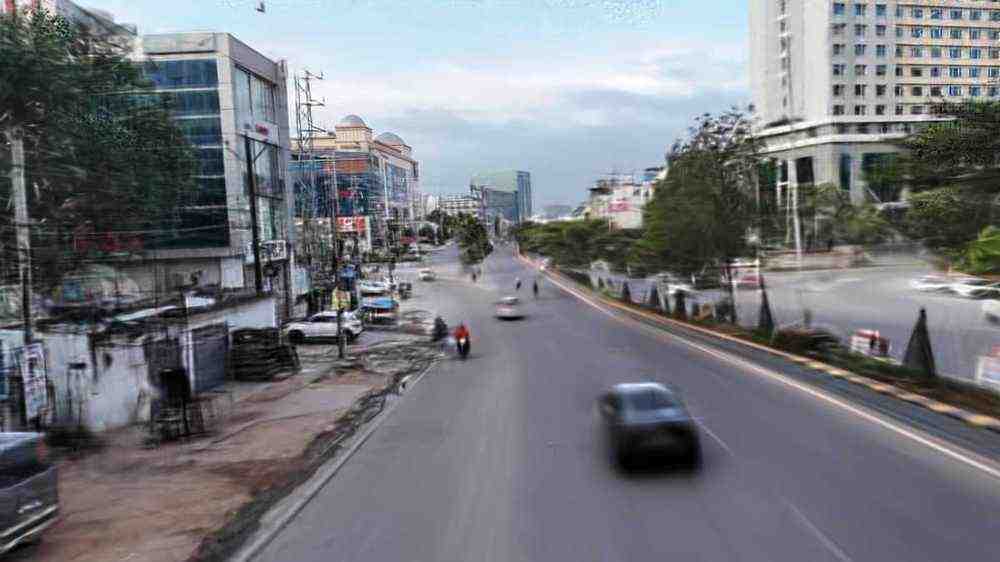} \\
& 12.78$|$0.39$|$0.80 & 22.12$|$0.65$|$0.54 & 13.85$|$0.33$|$0.68 & 23.39$|$0.70$|$0.46 \\

\end{tabular}
\end{adjustbox}

\caption{\textbf{Qualitative comparison across methods} Results from NVS methods trained on the $V^{D}$ sequences and rendered under the $T_{D\!\rightarrow D}$ viewpoint. 
Each column shows predictions from different NVS methods alongside the ground truth. 
The values beneath each rendered image correspond to the NVS metrics PSNR$|$SSIM$|$LPIPS.
}

\label{fig:qual_drone_drone}
\end{figure*}

\figref{fig:qual_drone_drone} illustrates results for models trained on the $V^{D}$ sequences and evaluated under the $T_{D\!\rightarrow D}$ setting. Within this domain, 3DGS\cite{kerbl20233d}-based methods generally preserve scene layout and structural details well. However, texture stability varies significantly across methods, with feed-forward models struggling in regions with fine-scale detail or repeated patterns.

\begin{figure*}[ht!]
\centering
\scriptsize
\setlength{\tabcolsep}{1pt}
\renewcommand{\arraystretch}{0.8}

\begin{adjustbox}{max width=\textwidth}
\begin{tabular}{ccccc}

\textbf{GT} & \textbf{3DGS} & \textbf{NeRF} & \textbf{DepthSplat} & \textbf{PVG} \\

\includegraphics[width=0.19\linewidth]{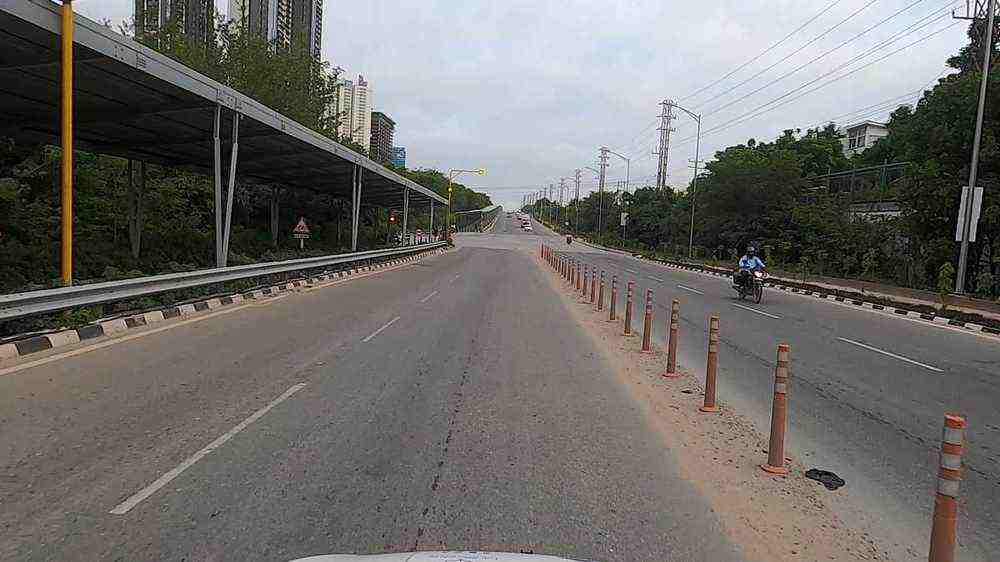} &
\includegraphics[width=0.19\linewidth]{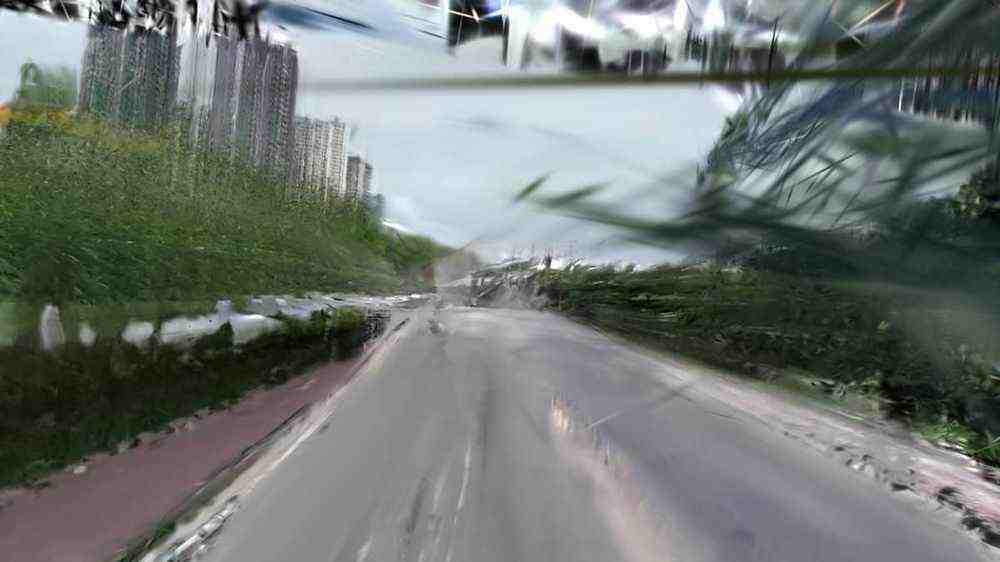} &
\includegraphics[width=0.19\linewidth]{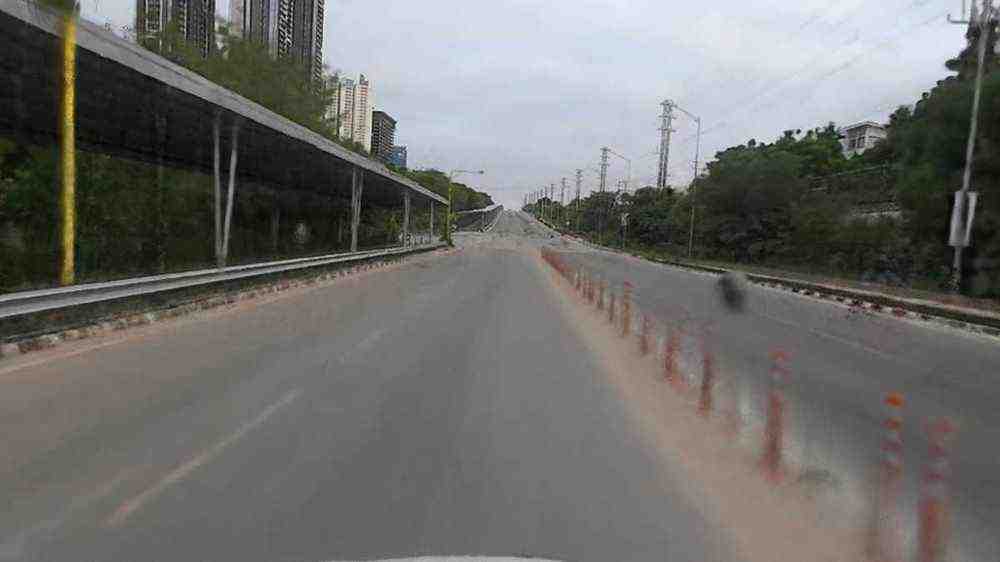} &
\includegraphics[width=0.19\linewidth]{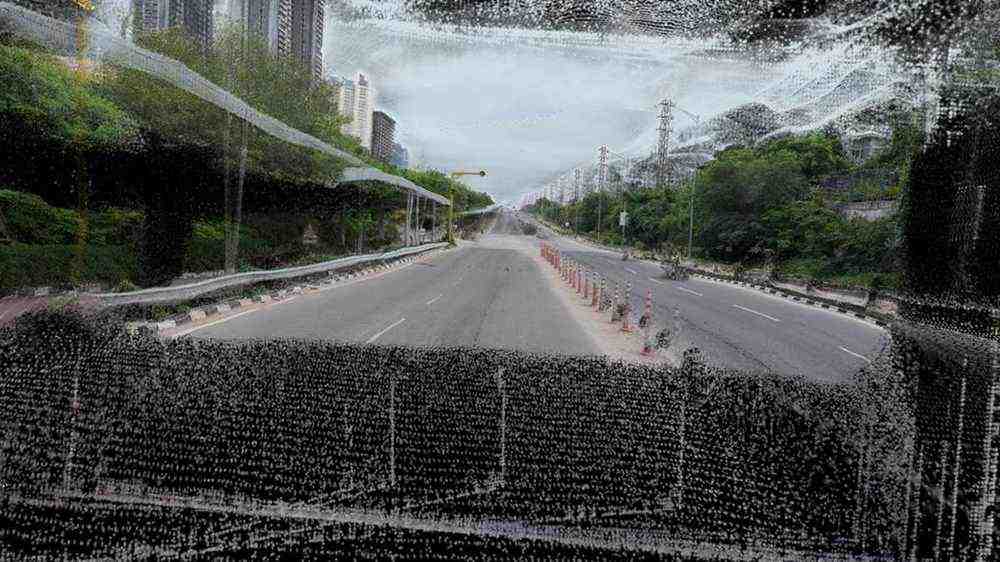} &
\includegraphics[width=0.19\linewidth]{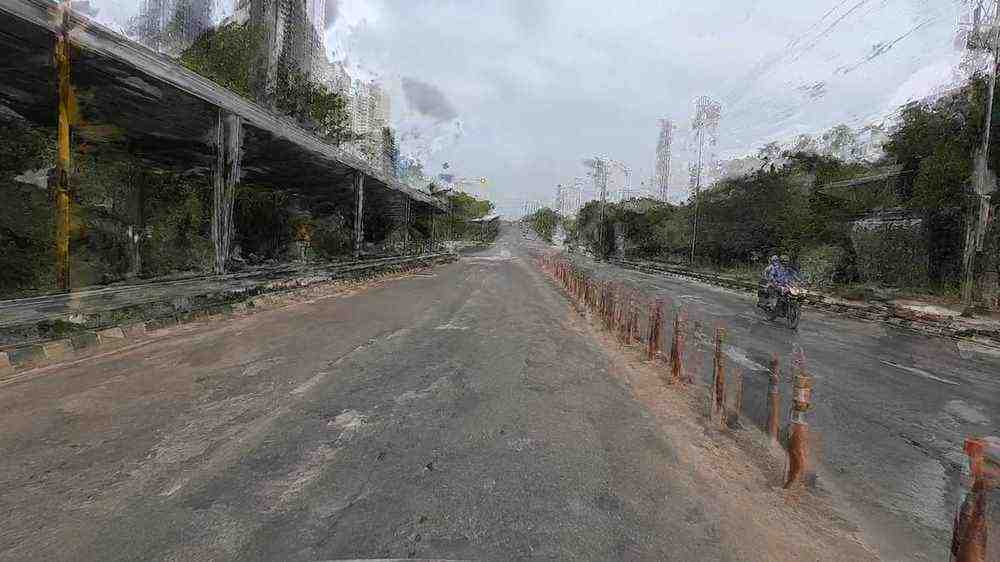} \\
 & 11.16$|$0.39$|$0.81 & 21.88$|$0.62$|$0.55 & 11.98$|$0.22$|$0.74 & 17.13$|$0.42$|$0.44 \\

\includegraphics[width=0.19\linewidth]{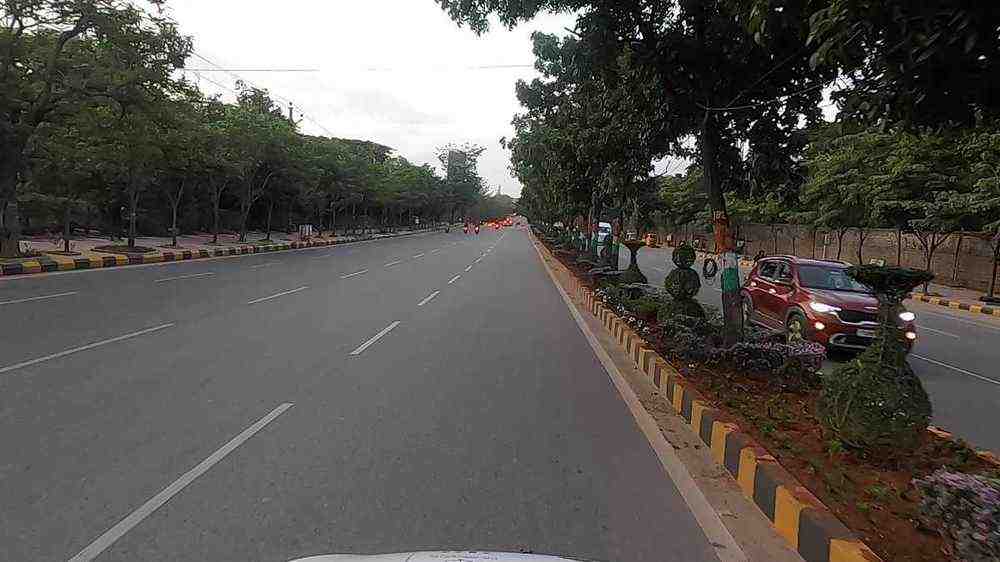} &
\includegraphics[width=0.19\linewidth]{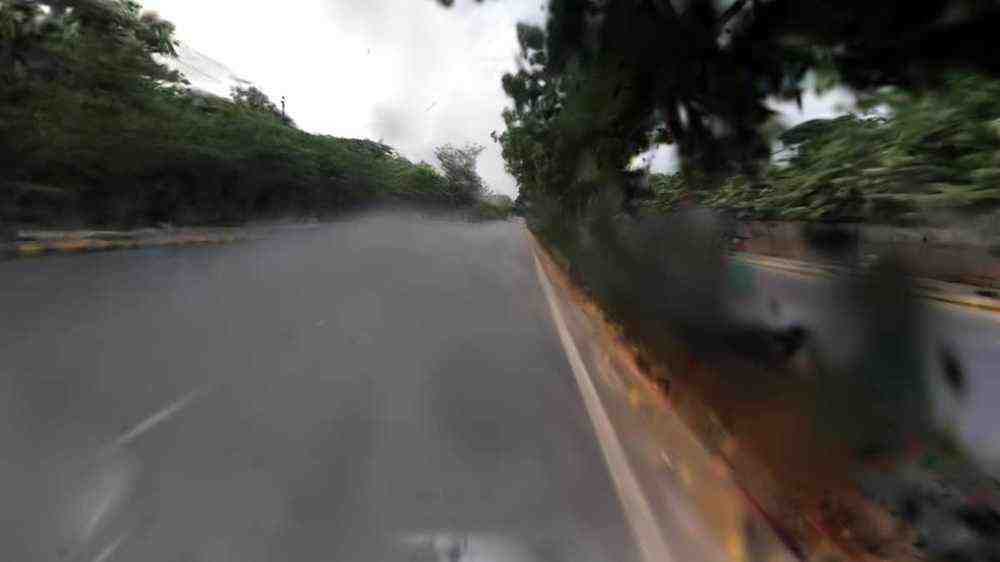} &
\includegraphics[width=0.19\linewidth]{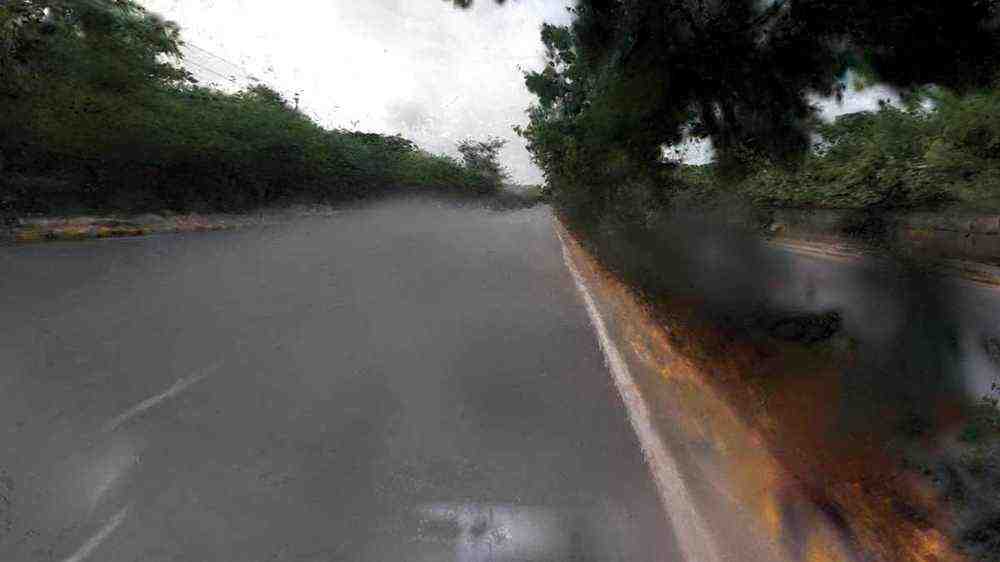} &
\includegraphics[width=0.19\linewidth]{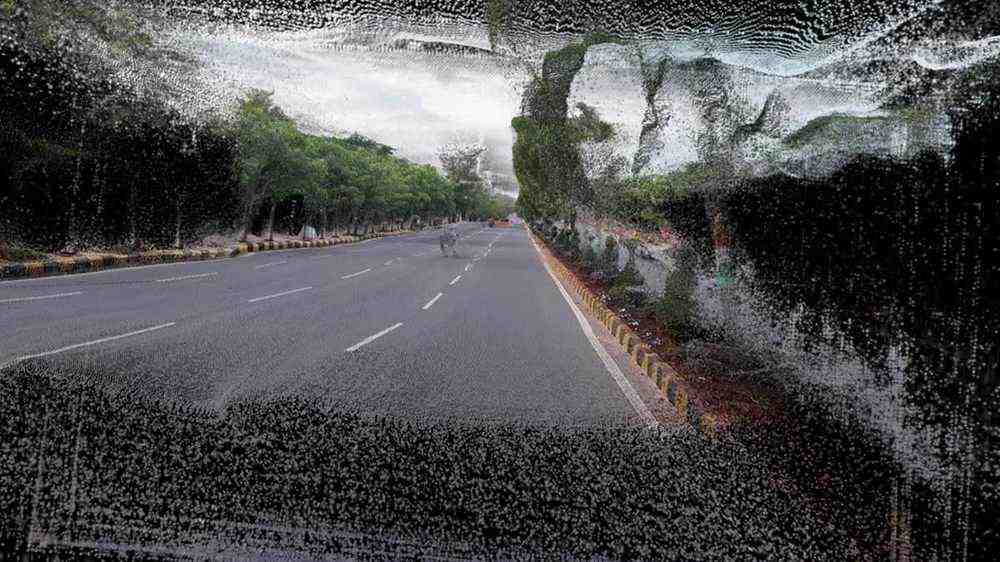} &
\includegraphics[width=0.19\linewidth]{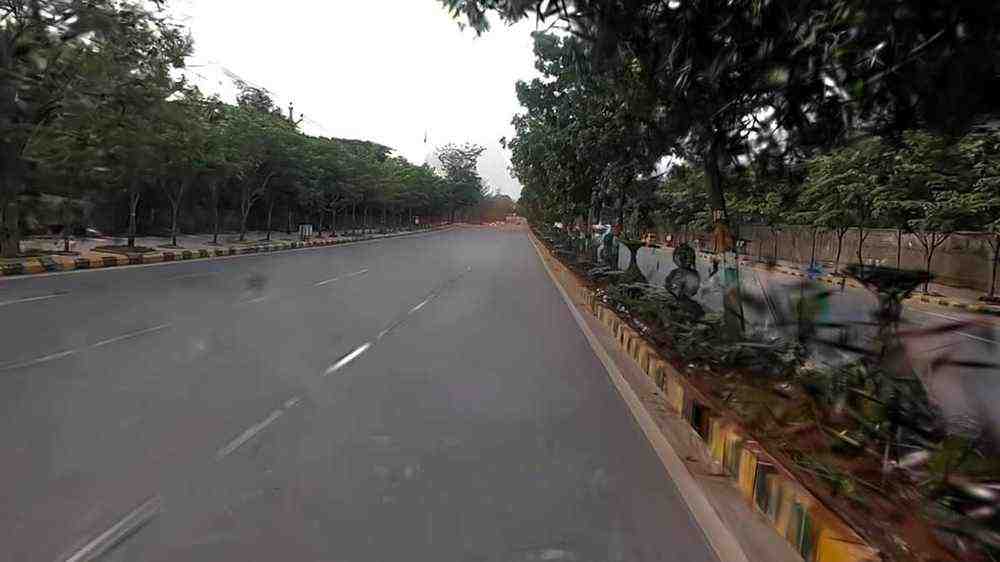} \\
 & 16.83$|$0.58$|$0.69 & 16.11$|$0.54$|$0.59 & 11.71$|$0.20$|$0.77 & 19.37$|$0.66$|$0.46 \\

\includegraphics[width=0.19\linewidth]{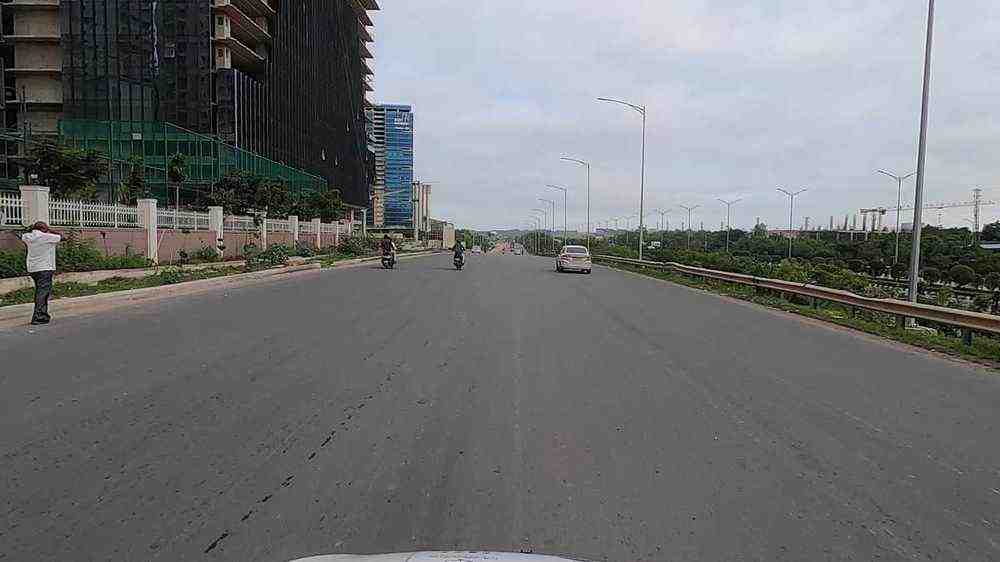} &
\includegraphics[width=0.19\linewidth]{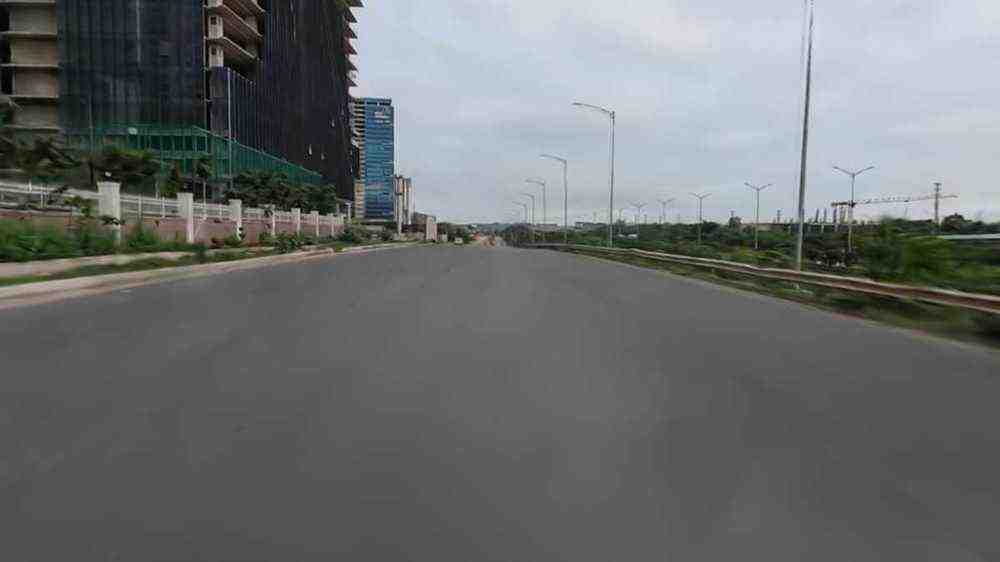} &
\includegraphics[width=0.19\linewidth]{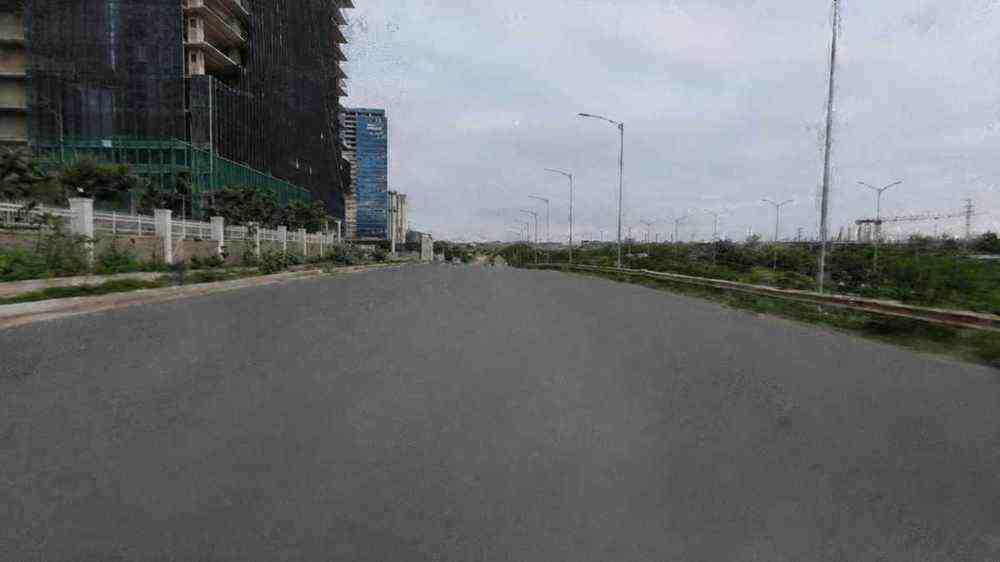} &
\includegraphics[width=0.19\linewidth]{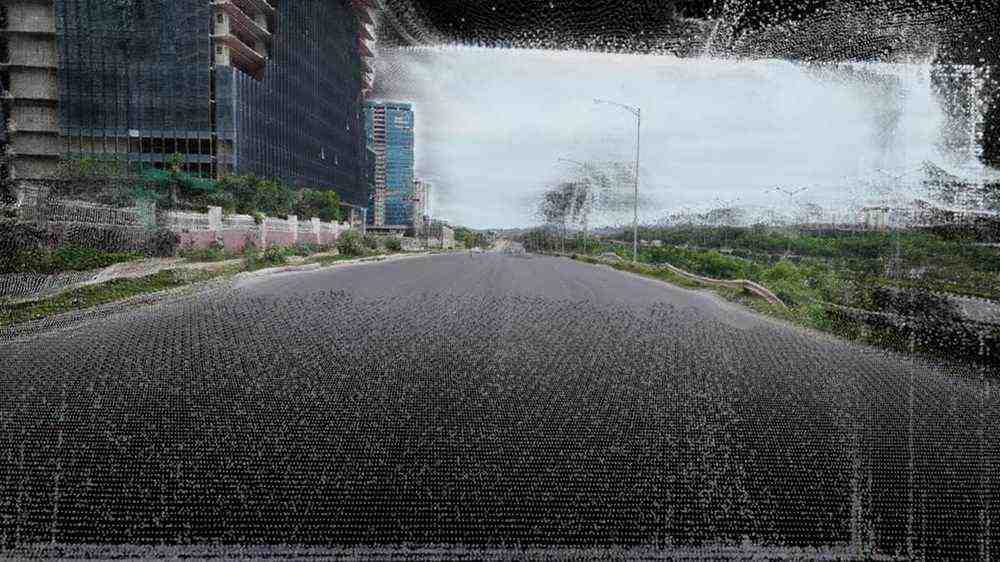} &
\includegraphics[width=0.19\linewidth]{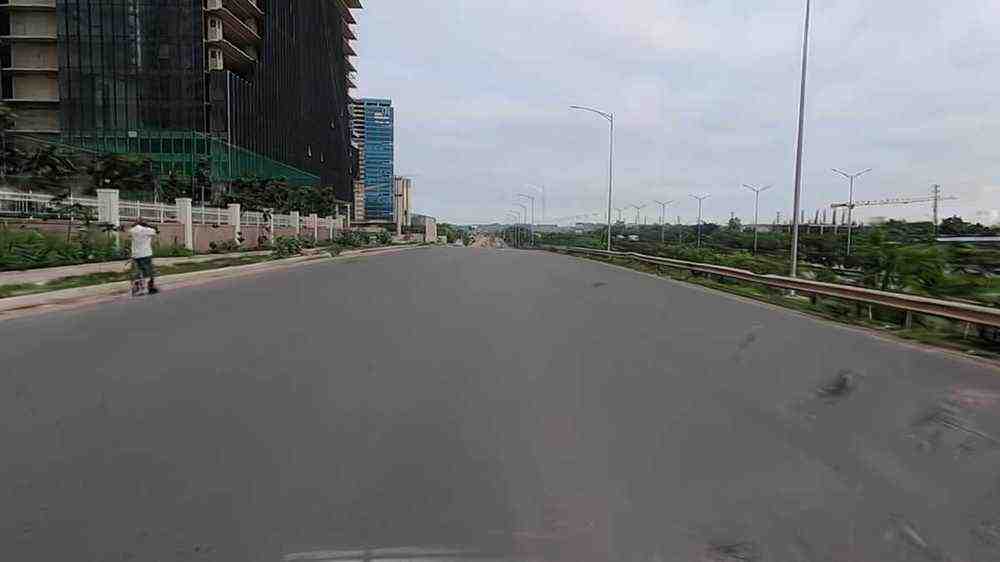} \\
 & 18.68$|$0.69$|$0.54 & 18.45$|$0.64$|$0.47 & 12.37$|$0.25$|$0.75 & 18.88$|$0.68$|$0.49 \\

\includegraphics[width=0.19\linewidth]{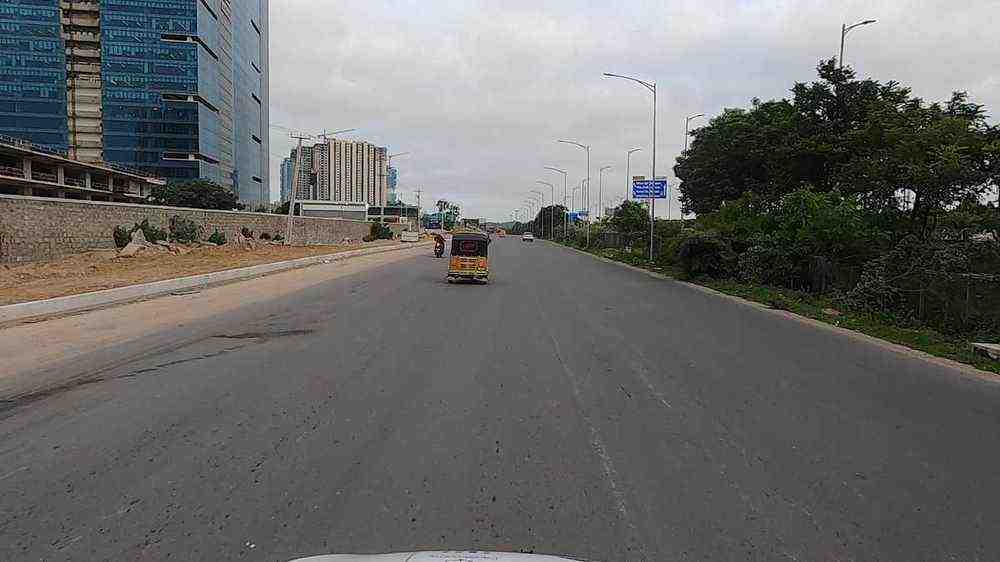} &
\includegraphics[width=0.19\linewidth]{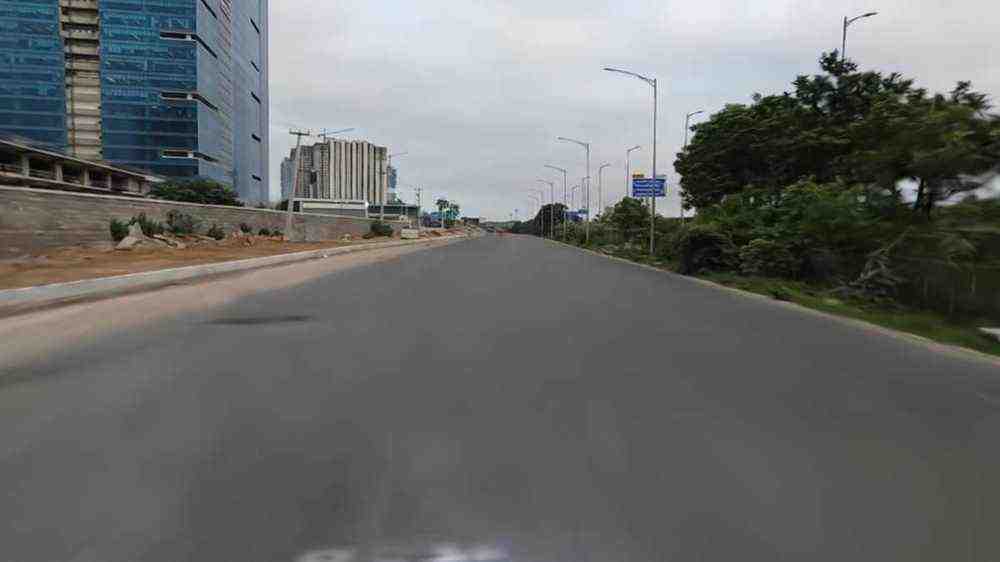} &
\includegraphics[width=0.19\linewidth]{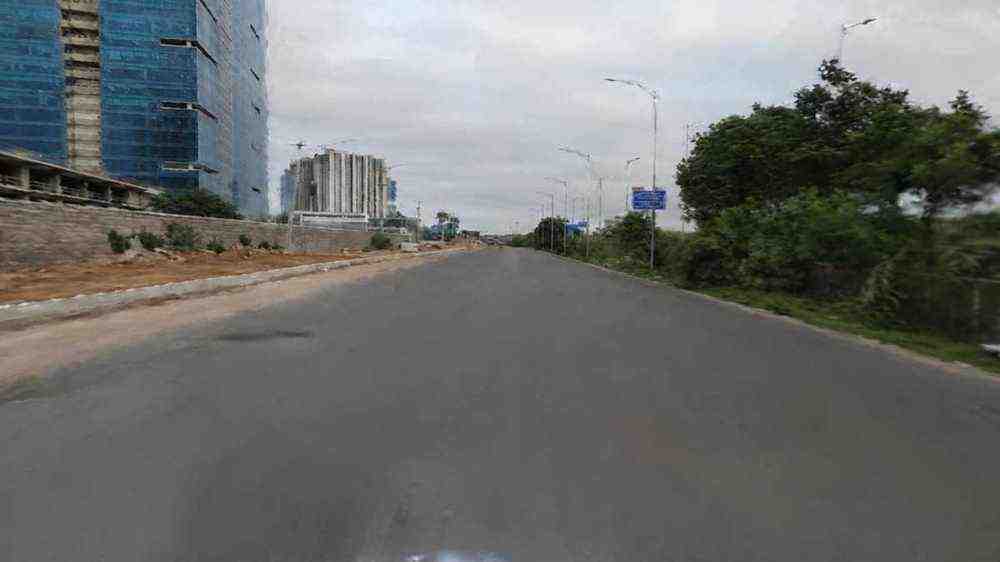} &
\includegraphics[width=0.19\linewidth]{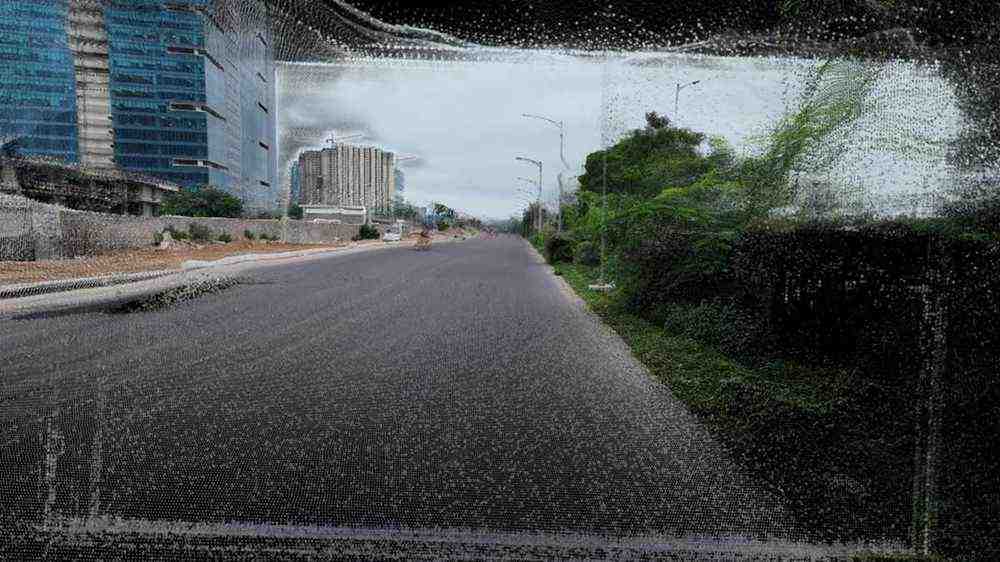} &
\includegraphics[width=0.19\linewidth]{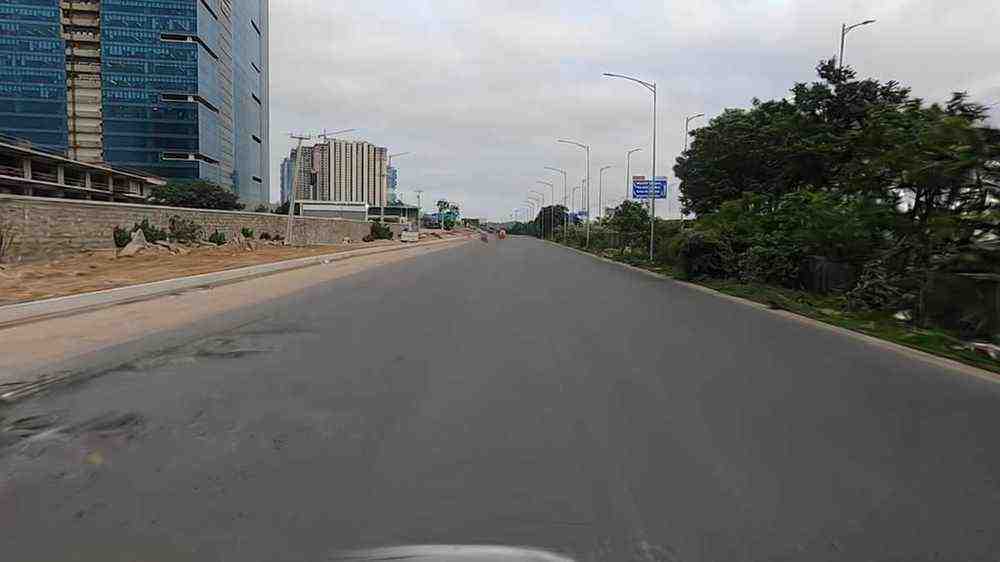} \\
 & 20.43$|$0.66$|$0.52 & 19.77$|$0.64$|$0.47 & 10.83$|$0.22$|$0.79 & 25.37$|$0.80$|$0.41 \\

\includegraphics[width=0.19\linewidth]{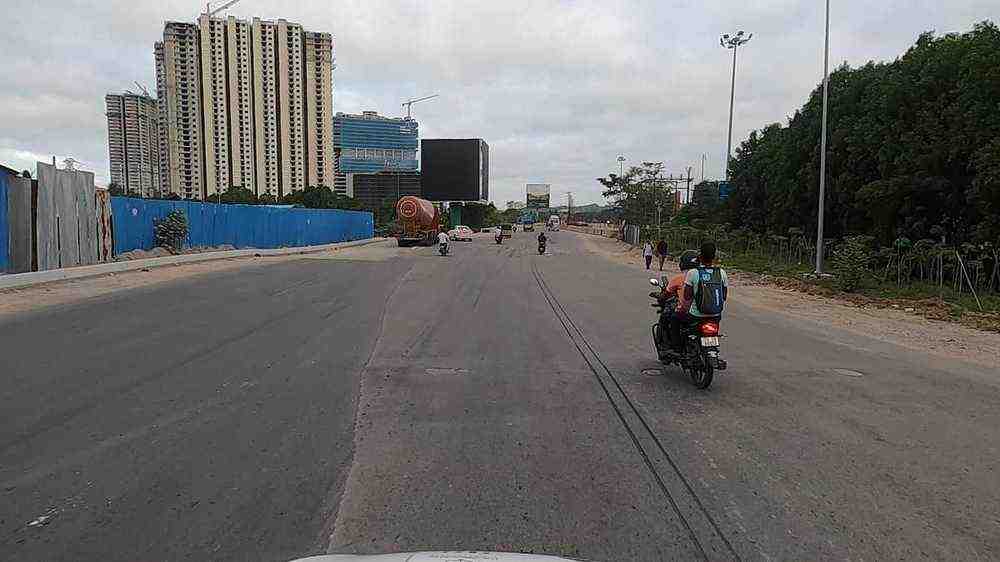} &
\includegraphics[width=0.19\linewidth]{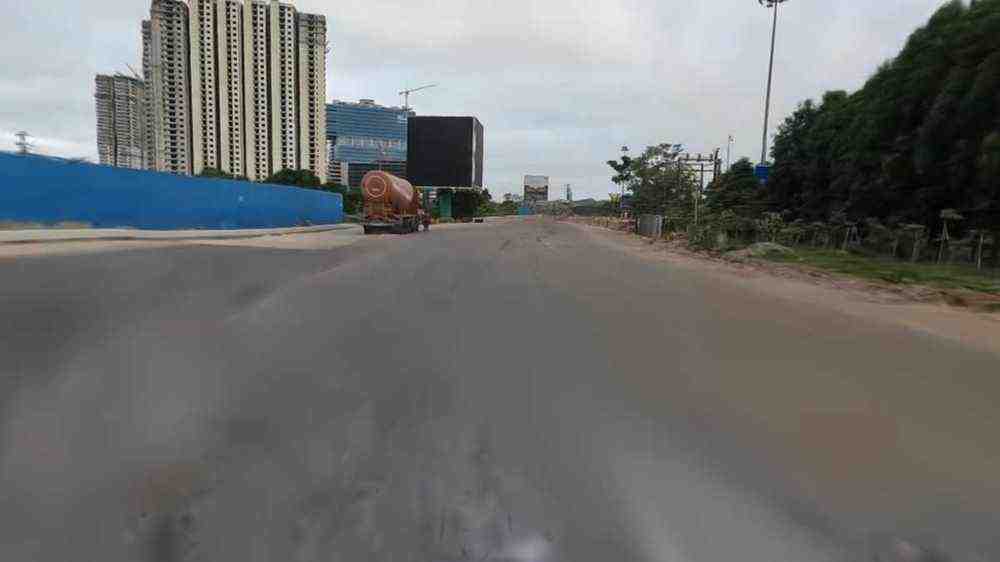} &
\includegraphics[width=0.19\linewidth]{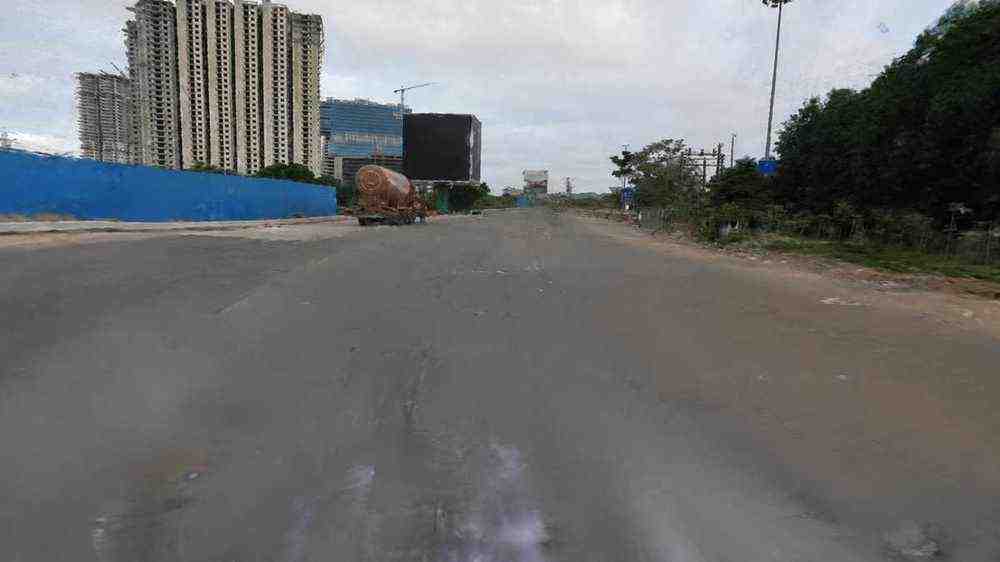} &
\includegraphics[width=0.19\linewidth]{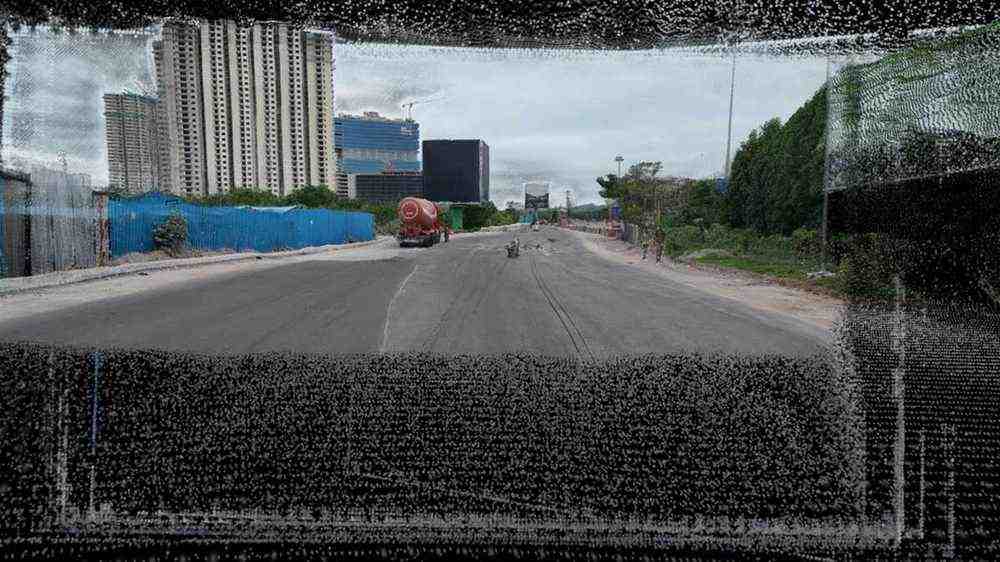} &
\includegraphics[width=0.19\linewidth]{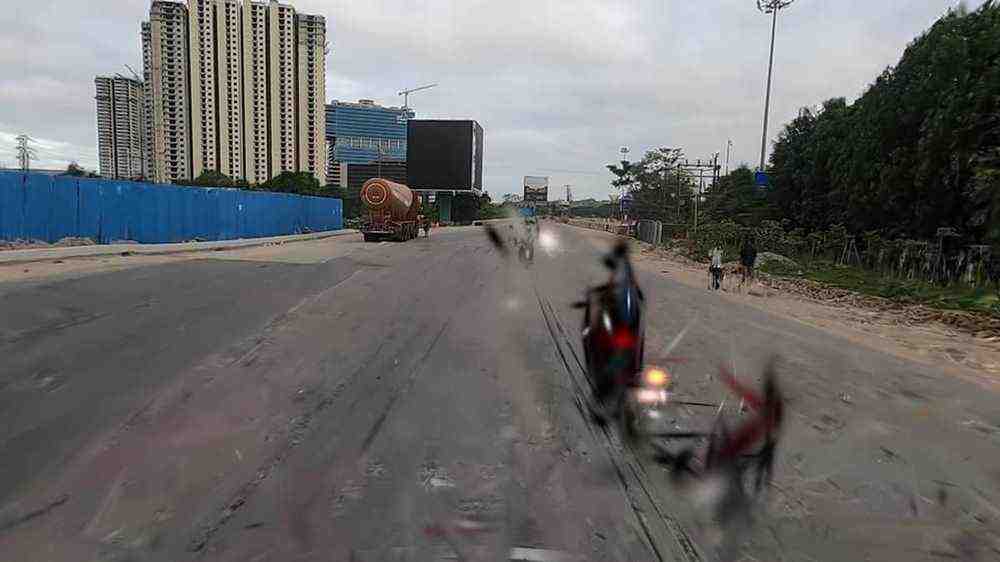} \\
 & 16.12$|$0.61$|$0.65 & 15.72$|$0.59$|$0.56 & 11.74$|$0.29$|$0.69 & 16.08$|$0.57$|$0.56 \\

\includegraphics[width=0.19\linewidth]{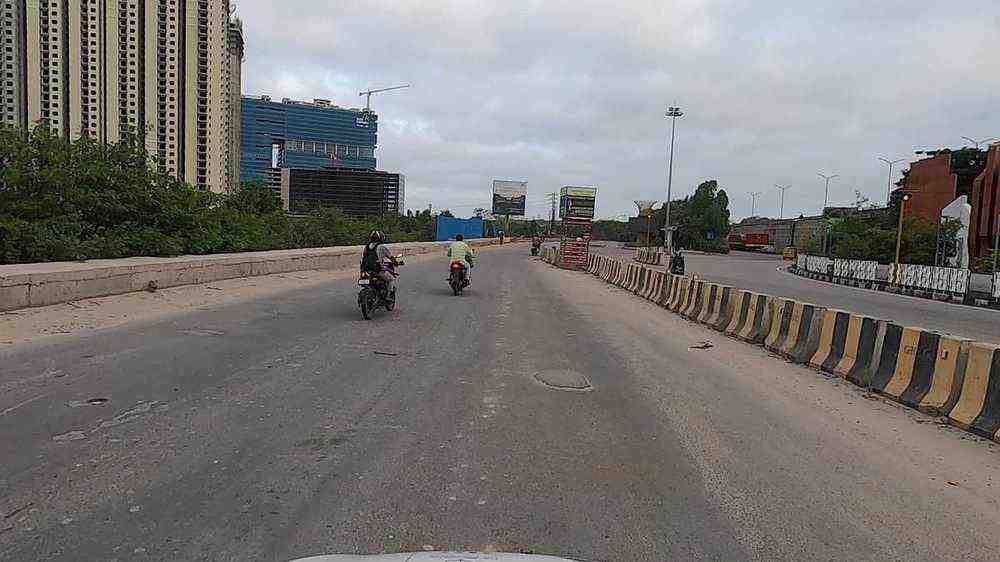} &
\includegraphics[width=0.19\linewidth]{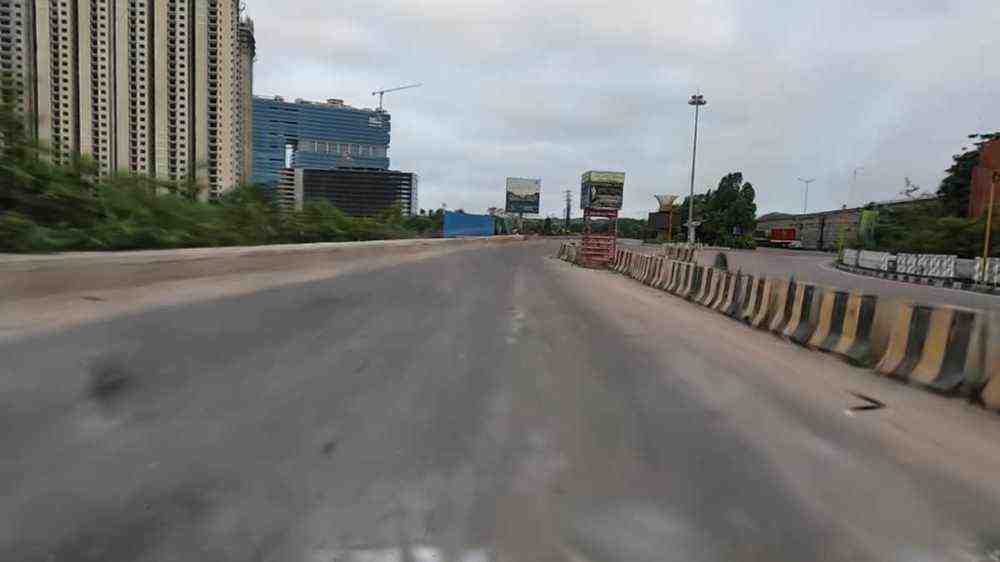} &
\includegraphics[width=0.19\linewidth]{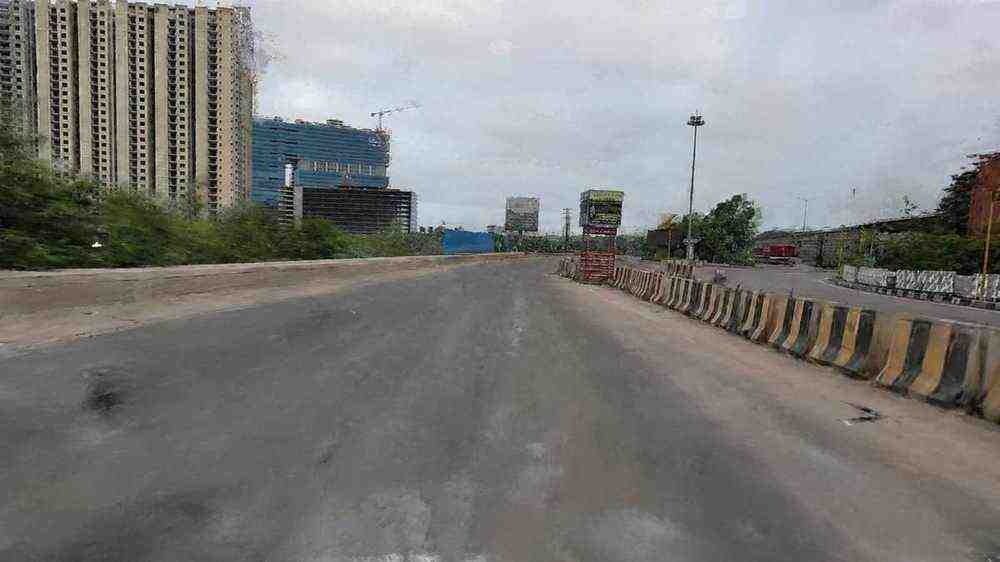} &
\includegraphics[width=0.19\linewidth]{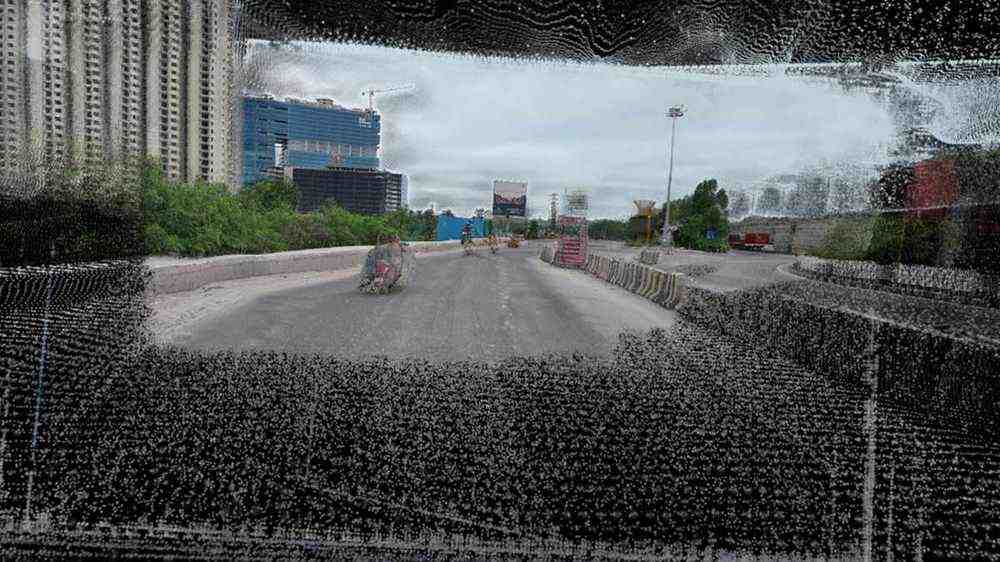} &
\includegraphics[width=0.19\linewidth]{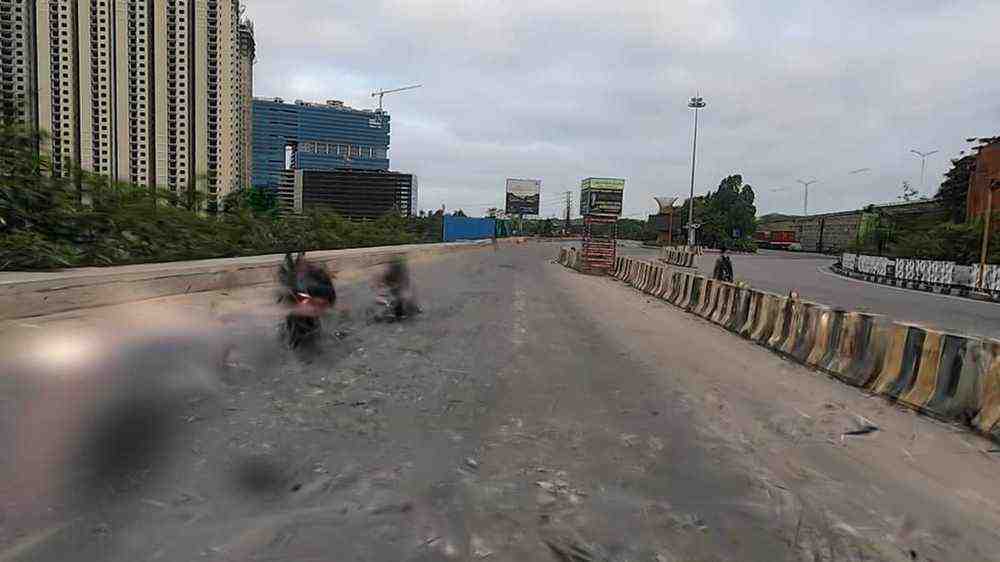} \\
 & 16.93$|$0.56$|$0.64 & 16.07$|$0.52$|$0.49 & 10.88$|$0.23$|$0.71 & 16.48$|$0.53$|$0.55 \\

\includegraphics[width=0.19\linewidth]{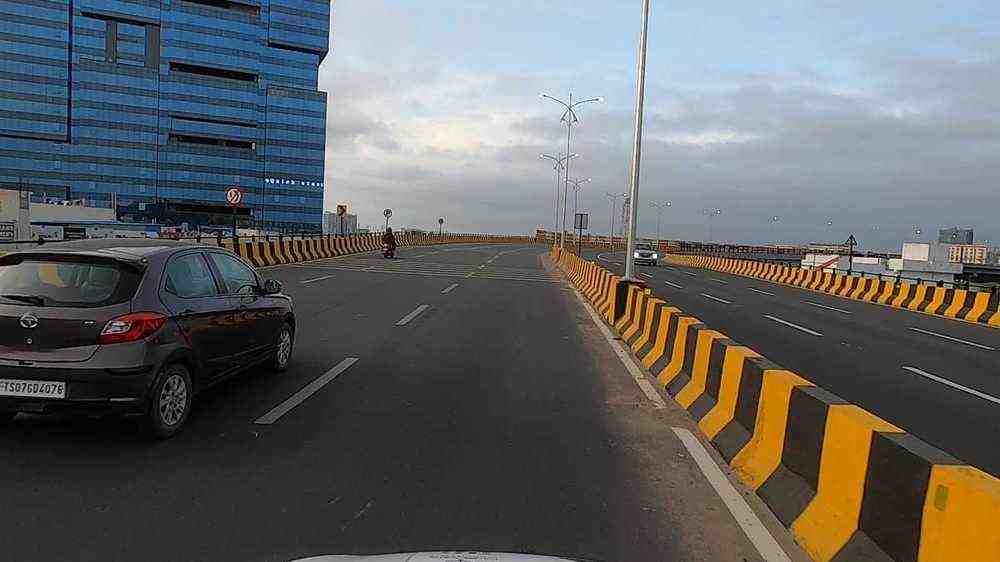} &
\includegraphics[width=0.19\linewidth]{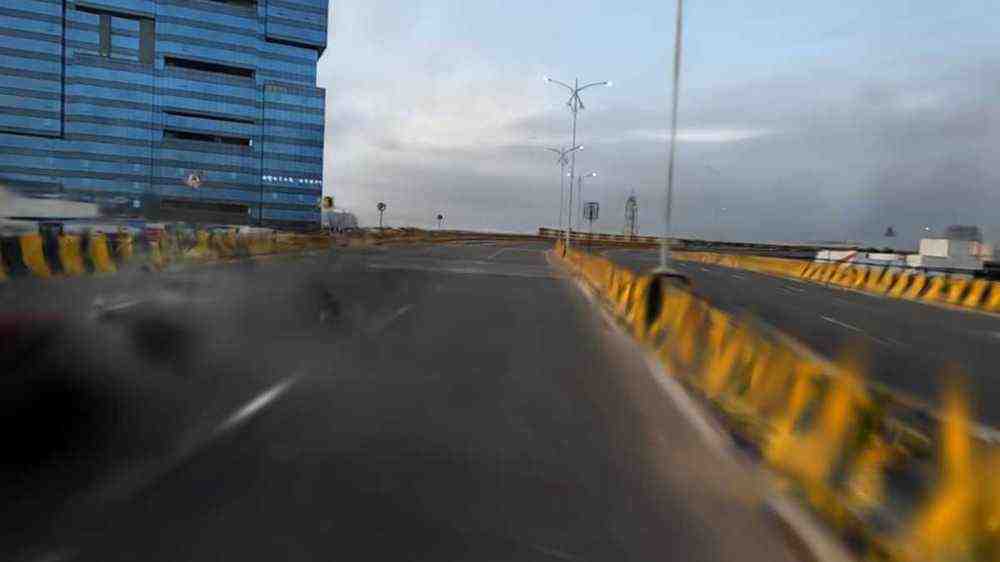} &
\includegraphics[width=0.19\linewidth]{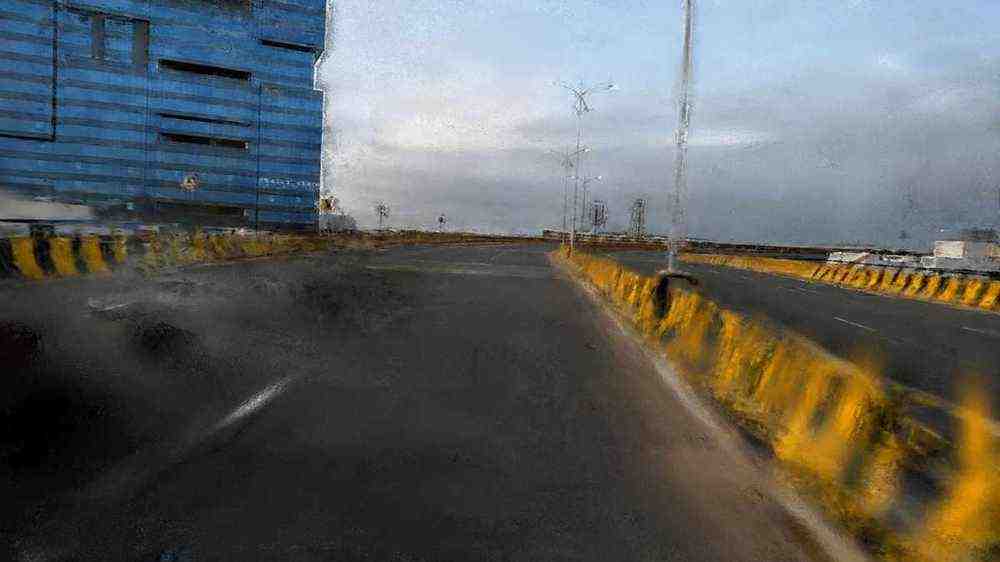} &
\includegraphics[width=0.19\linewidth]{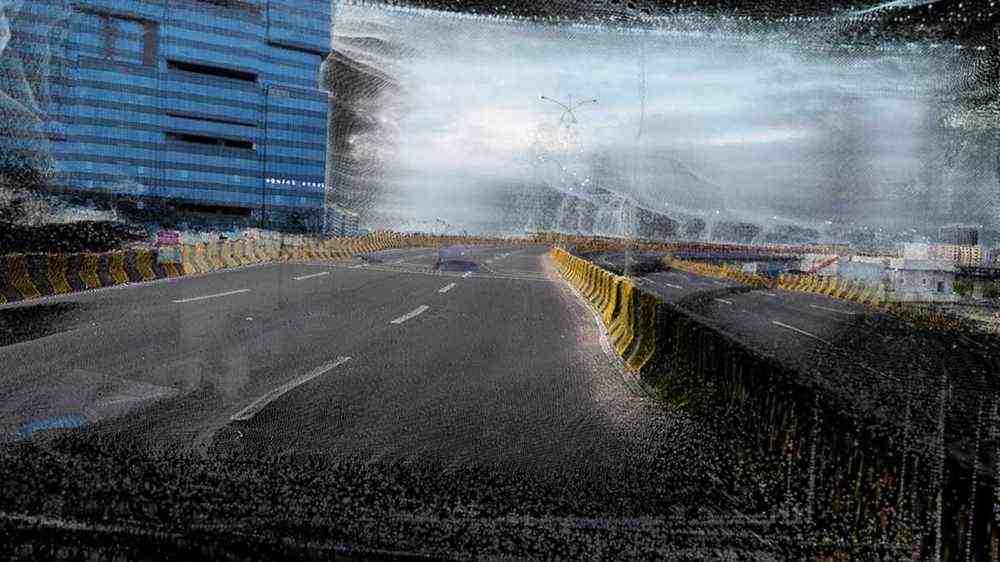} &
\includegraphics[width=0.19\linewidth]{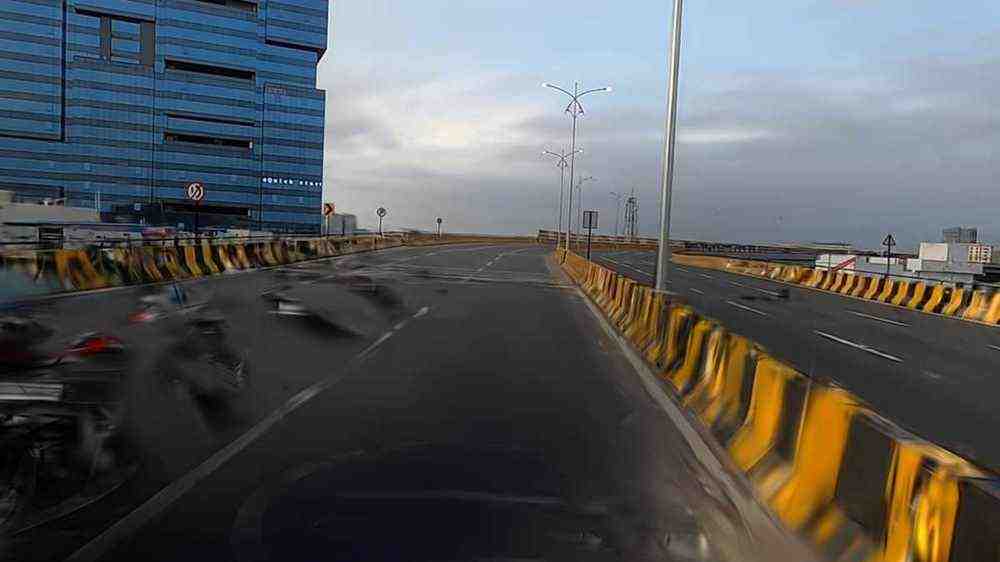} \\
 & 18.00$|$0.63$|$0.58 & 18.03$|$0.58$|$0.54 & 13.34$|$0.34$|$0.75 & 17.69$|$0.61$|$0.50 \\

\includegraphics[width=0.19\linewidth]{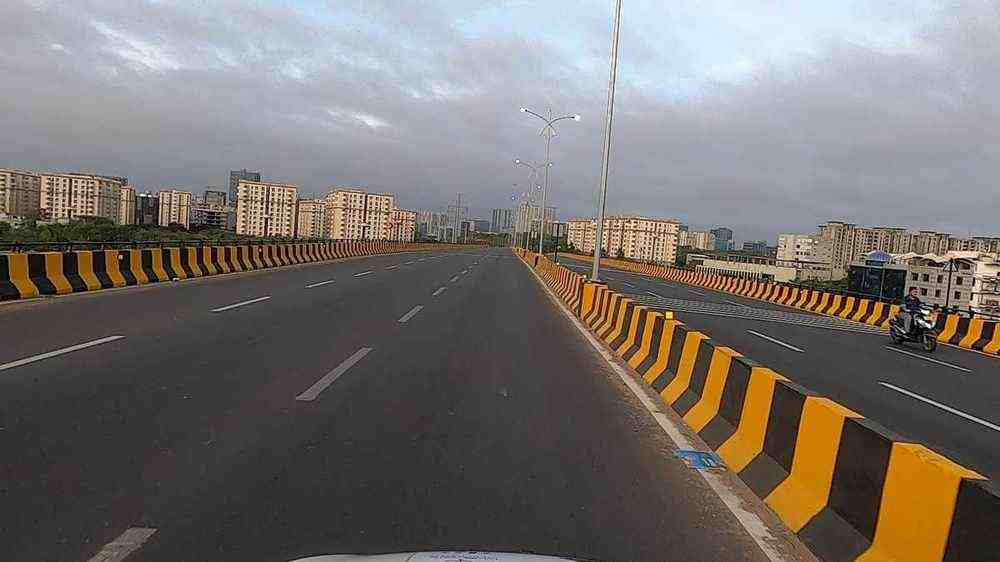} &
\includegraphics[width=0.19\linewidth]{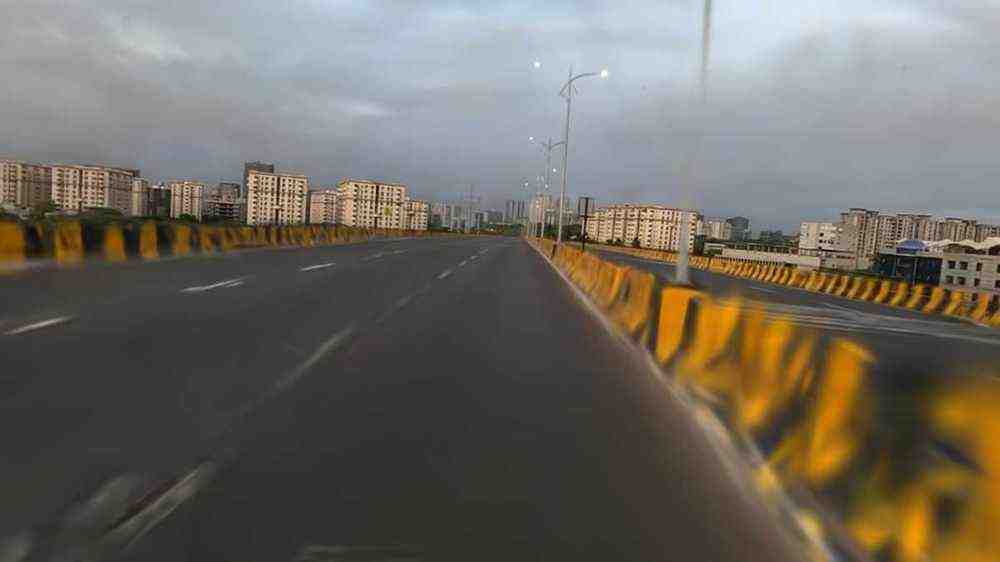} &
\includegraphics[width=0.19\linewidth]{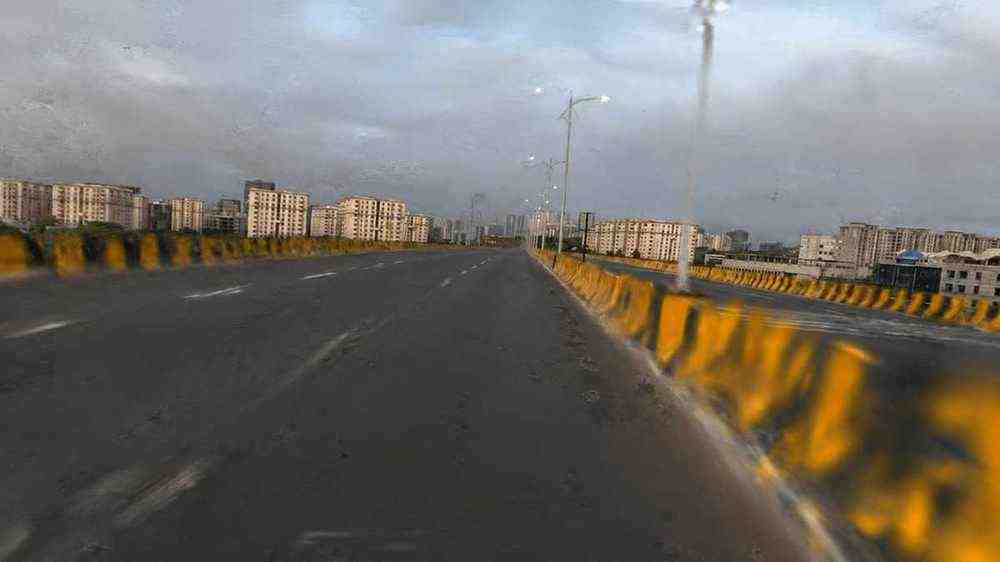} &
\includegraphics[width=0.19\linewidth]{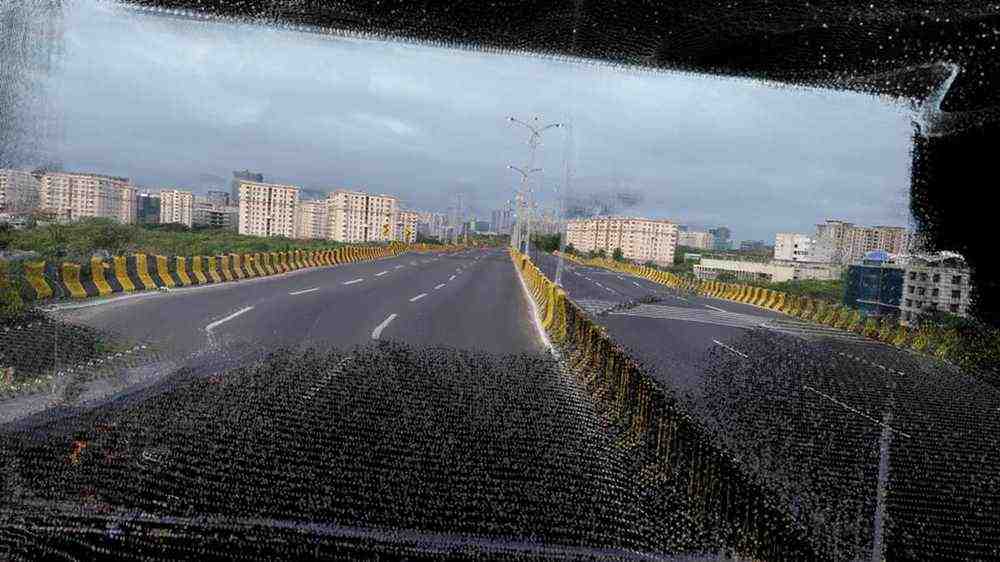} &
\includegraphics[width=0.19\linewidth]{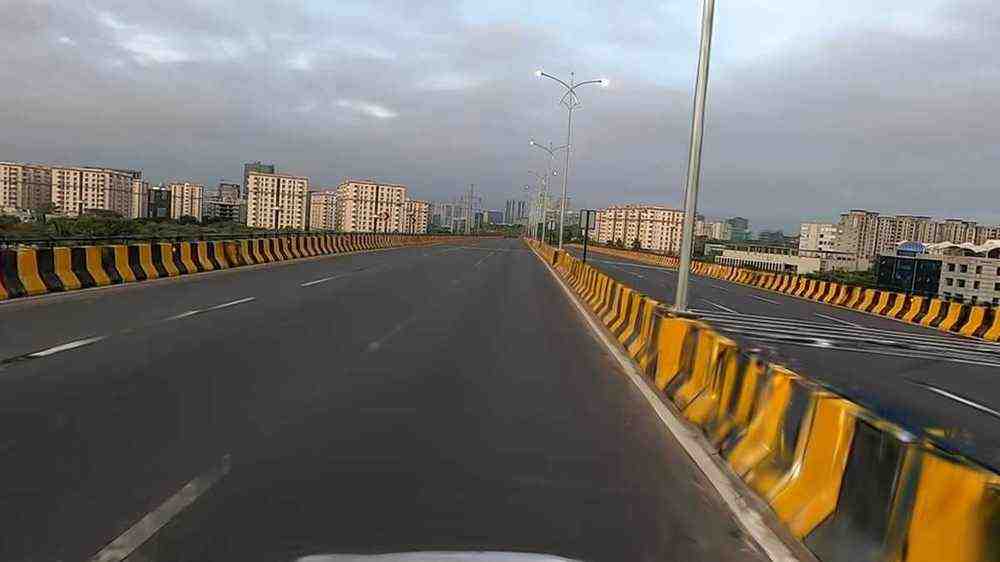} \\
 & 17.35$|$0.69$|$0.58 & 17.76$|$0.67$|$0.49 & 10.34$|$0.33$|$0.76 & 17.65$|$0.67$|$0.50 \\

\includegraphics[width=0.19\linewidth]{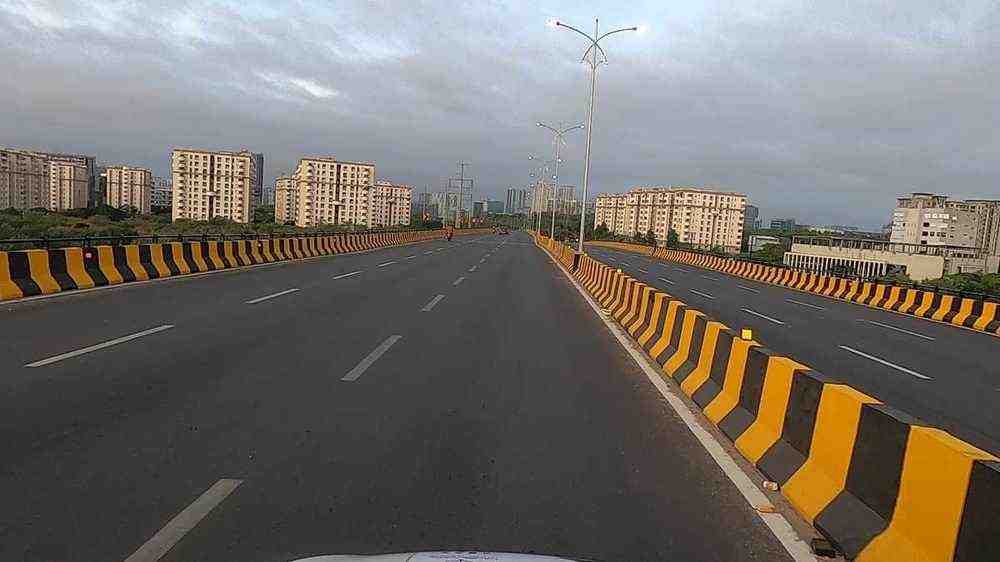} &
\includegraphics[width=0.19\linewidth]{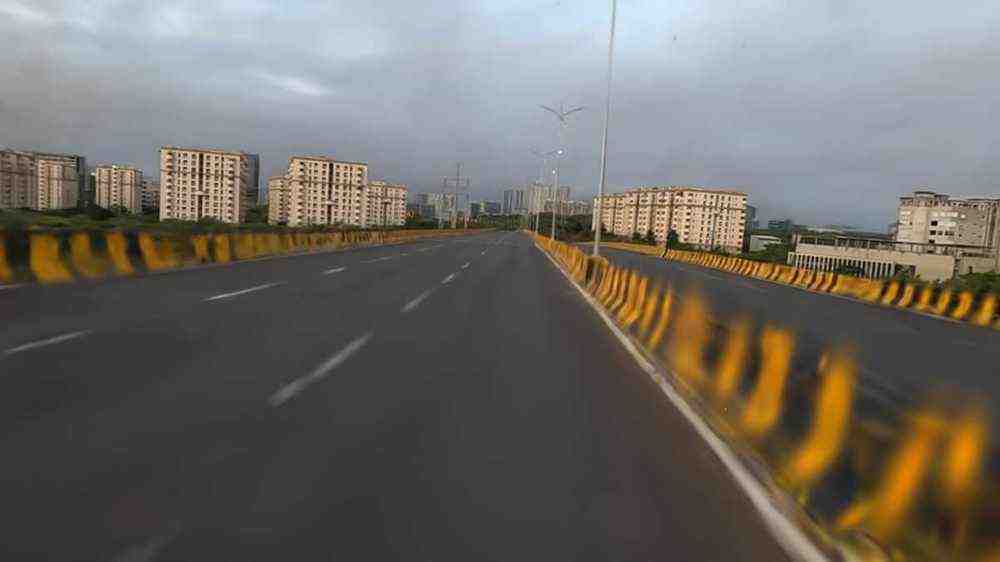} &
\includegraphics[width=0.19\linewidth]{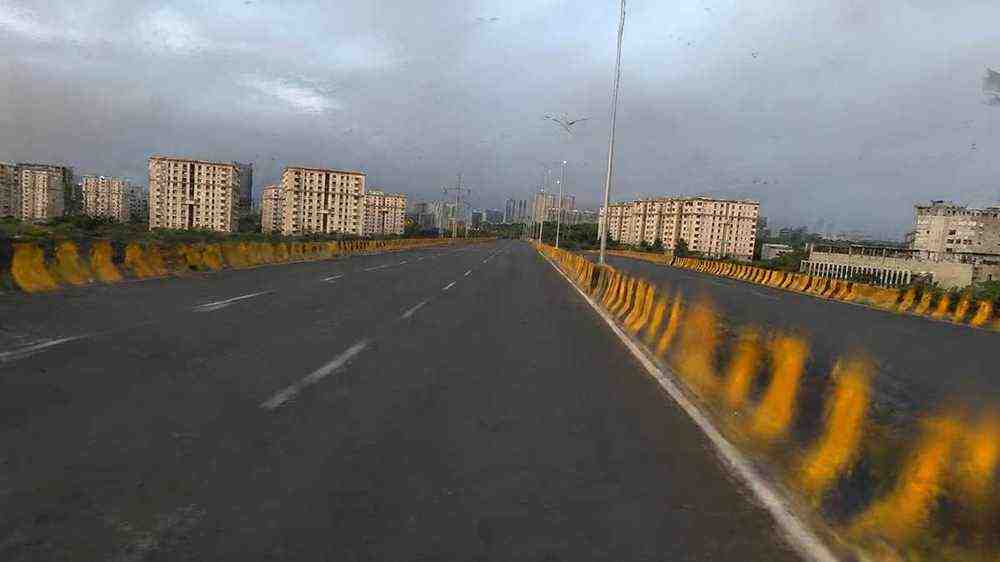} &
\includegraphics[width=0.19\linewidth]{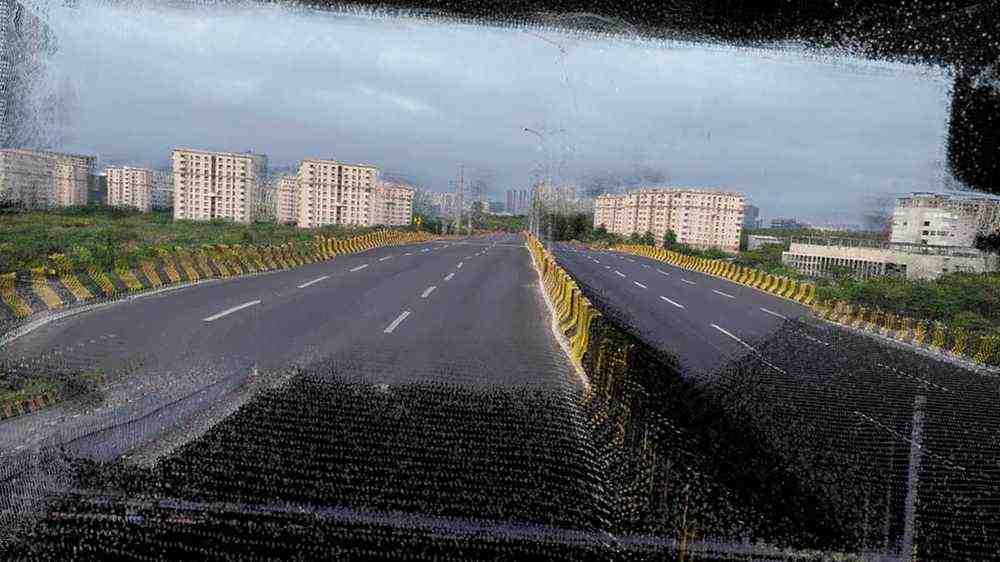} &
\includegraphics[width=0.19\linewidth]{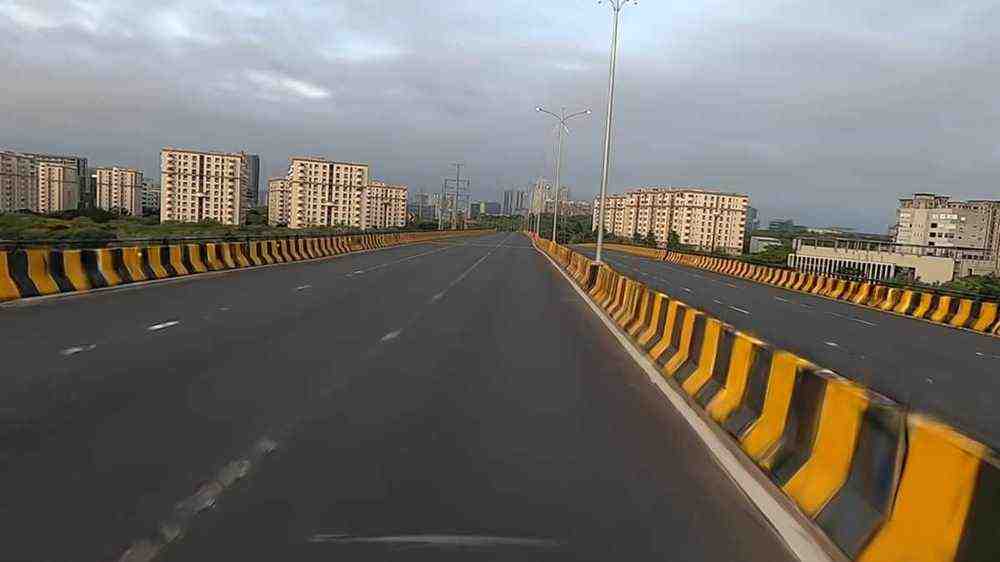} \\
 & 19.11$|$0.72$|$0.50 & 18.29$|$0.69$|$0.45 & 12.10$|$0.38$|$0.72 & 19.75$|$0.72$|$0.40 \\

\includegraphics[width=0.19\linewidth]{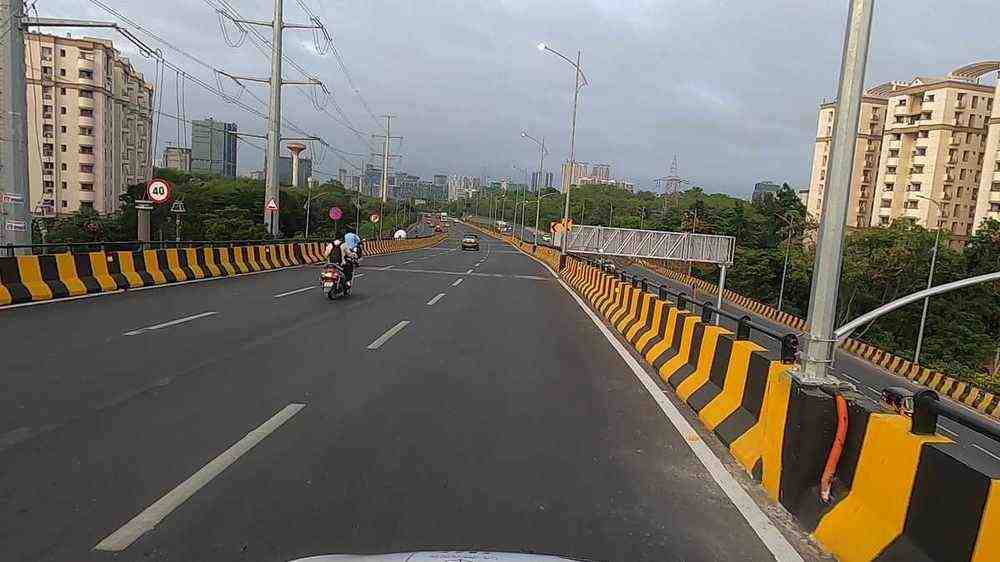} &
\includegraphics[width=0.19\linewidth]{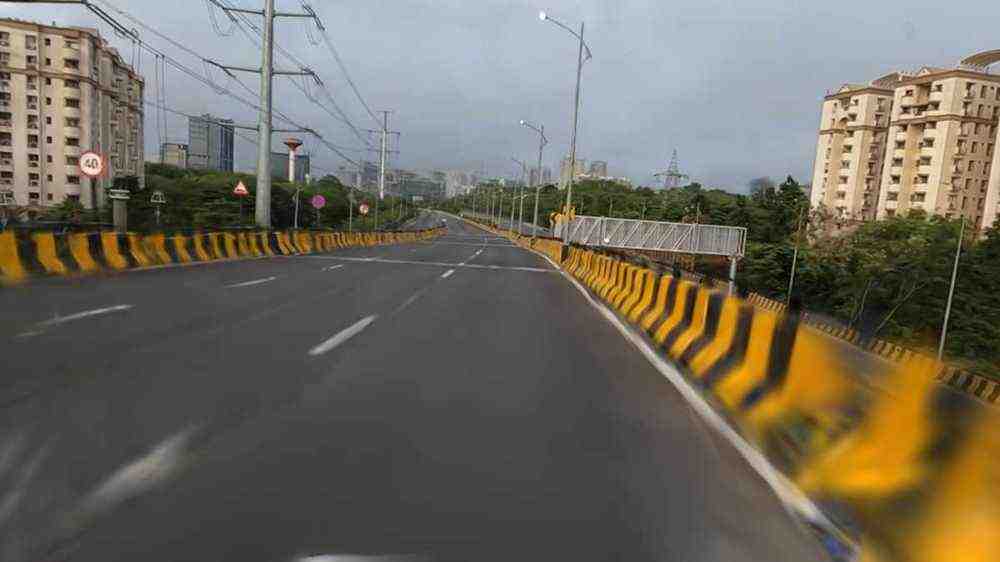} &
\includegraphics[width=0.19\linewidth]{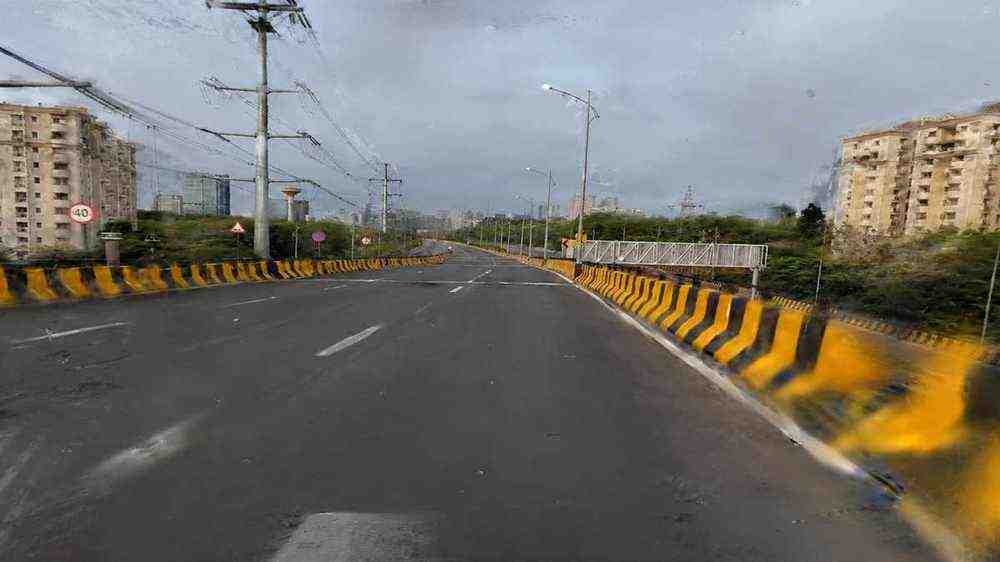} &
\includegraphics[width=0.19\linewidth]{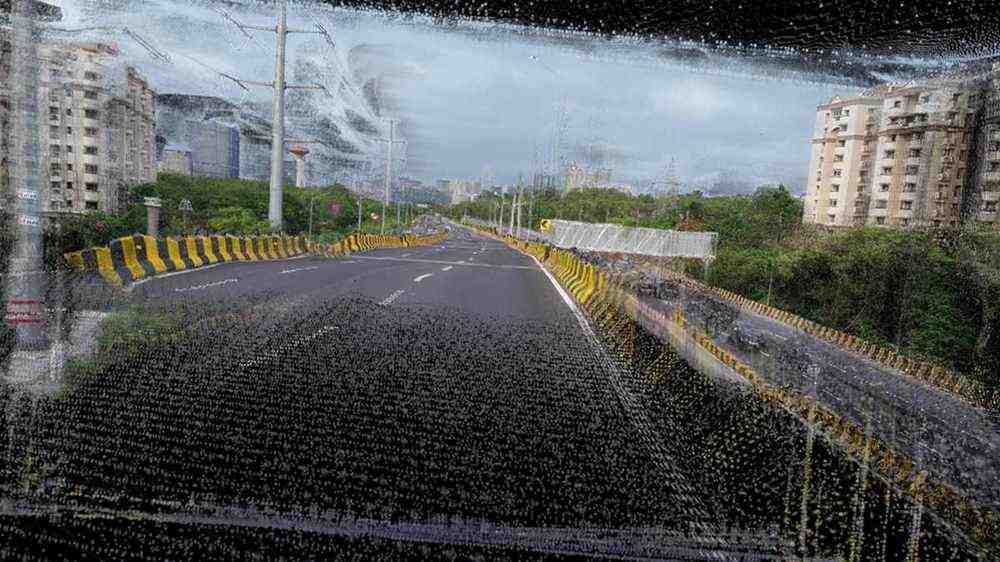} &
\includegraphics[width=0.19\linewidth]{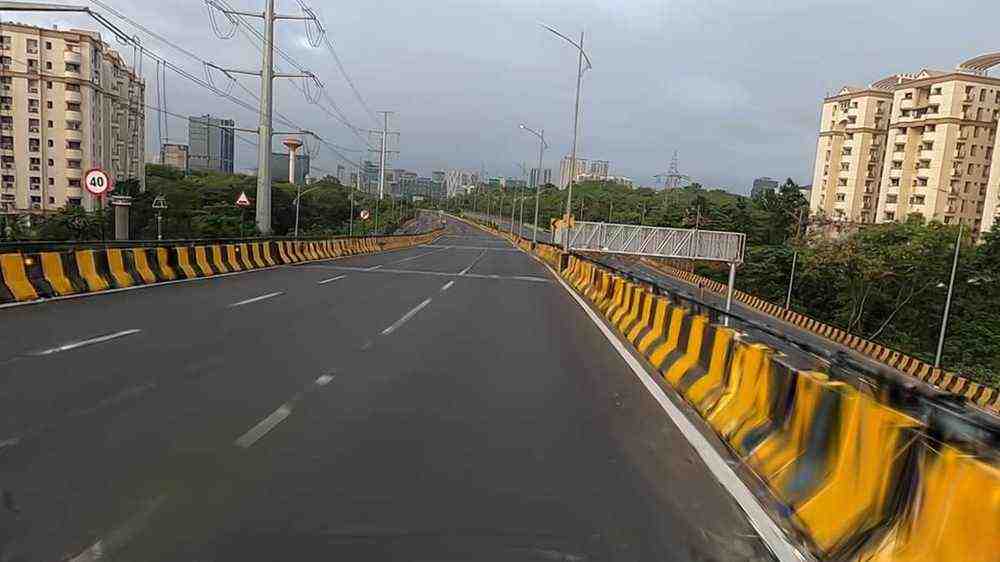} \\
 & 15.49$|$0.56$|$0.55 & 14.50$|$0.53$|$0.56 & 11.79$|$0.21$|$0.80 & 15.42$|$0.54$|$0.44 \\

\end{tabular}
\end{adjustbox}

\caption{\textbf{Qualitative comparison across methods.}
Results from NVS methods trained on the $V^{D}$ sequences and rendered under the $T_{D\!\rightarrow C}$ viewpoint.
Each column shows predictions from different NVS methods alongside the ground truth.
The values beneath each rendered image correspond to PSNR$\uparrow$$|$SSIM$\uparrow$$|$LPIPS$\downarrow$.
}
\label{fig:qual_drone_car}
\end{figure*}

\figref{fig:qual_drone_car} evaluates the cross-domain setting $T_{D\!\rightarrow C}$, where models trained on drone viewpoints must reconstruct car-mounted views. Because drone-height viewpoints differ substantially from ground-level perspectives, this change in viewpoint is the most challenging. All methods exhibit noticeable degradation, such as collapsed geometry, smeared textures, or misaligned structures. Nonetheless,  Gaussian-based methods maintain better global consistency, even if fine details remain challenging.

Overall, these qualitative comparisons highlight the challenges associated with multi-platform NVS, especially under cross-viewpoint and cross-domain transfers. They also emphasize the importance of robust geometric reasoning to handle large baseline shifts and platform-specific viewpoint changes.

\begin{figure}[t]
    \includegraphics[width=\columnwidth]{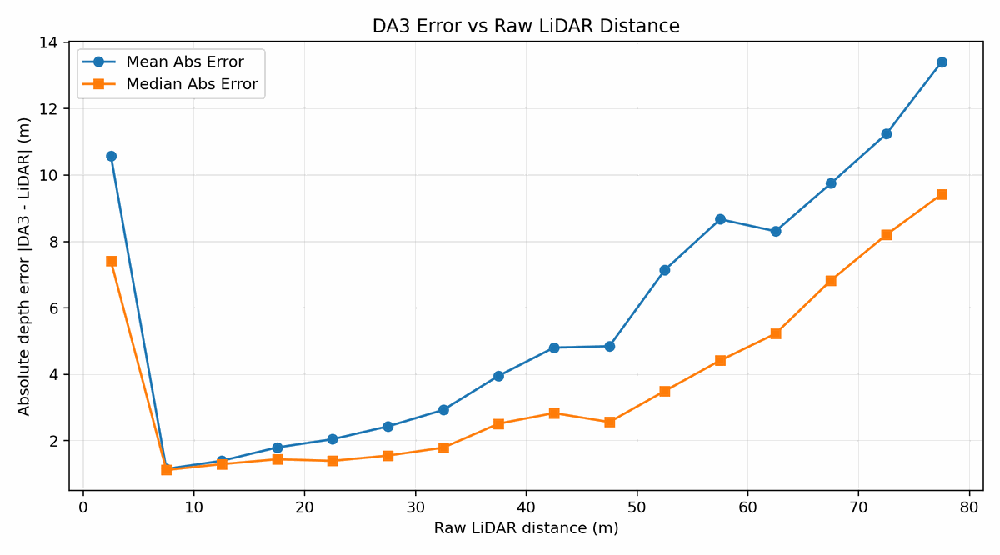}
    \caption{Analysis of monocular depth error
    \label{fig:depth_analysis}}
\end{figure}

\subsection{Depth Filtering Ablation}
\label{suppsec:depth_filter}
\textbf{Motivation for Depth Filtering.}
To analyze DaV3 reliability, we compute
$\left|D_{\mathrm{DaV3}} - D_{\mathrm{LiDAR}}\right|$ on valid LiDAR pixels
(Fig.~\ref{fig:depth_analysis}). We observe that depth errors exceed 3\,m
primarily outside the $[10,30]$\,m range and generally increase with depth.
While this error trend does not directly translate to NVS performance, it
motivates stronger outlier handling for monocular depth supervision. We
therefore compare standard IQR-95\% filtering strategy with a range-based
$[10,30]$\,m filter, which reduces the LiDAR--DaV3 gap and yields consistent
gains over IQR filtering on the WOD subset and on \MV.


The WOD sequence IDs are:
\{16, 21, 22, 25, 31, 34, 35, 49, 53, 80, 84, 86, 89, 94, 96,
102, 111, 222, 323, 382, 402, 427, 438, 546, 581, 592, 620,
640, 700, 754, 795, 796\}.

The \MV sequence IDs are:
\{1, 5, 10, 12, 17, 20, 22, 26, 30, 32, 37, 40, 42, 46, 49\}.

\subsection{Frame-wise Comparison of PVG and 3DGS}
\label{sec:framewise}

The quantitative results reported in the main paper summarize the average
PSNR, SSIM and LPIPS over all evaluation images. To better understand these
averaged results, we visualize the frame-wise metric distributions for PVG and
3DGS.

For every evaluation sequence, we compute the rendering metrics for every
fifth test image, following the evaluation protocol described in the main
paper. Each point in the scatter plots therefore corresponds to a single test
image. The horizontal axis represents the metric obtained by PVG, while the
vertical axis represents the corresponding metric obtained by 3DGS. The dashed
diagonal denotes identical performance between the two methods. Consequently,
points below the diagonal indicate better performance of PVG for PSNR and
SSIM, whereas points above the diagonal indicate better performance of PVG for
LPIPS.

\begin{figure*}[t]
\centering
\includegraphics[width=\textwidth]{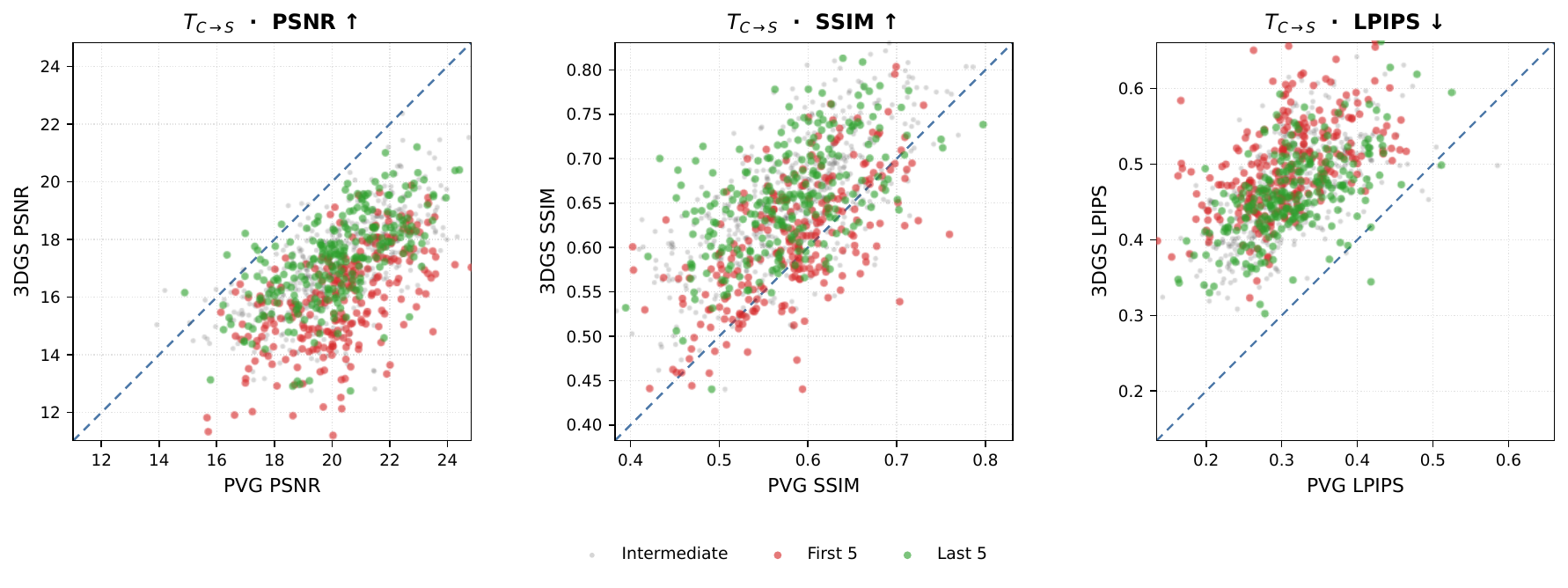}
\caption{
Frame-wise comparison between PVG and 3DGS on the
$\mathbf{T_{C\rightarrow S}}$ evaluation split.
Each point corresponds to one evaluated test image sampled every fifth frame
from all evaluation sequences.
The center of the point cloud closely matches the averaged metrics reported in
Table~1 of the main paper.
For PSNR and SSIM, the majority of samples lie below the diagonal, while for
LPIPS most samples lie above the diagonal, indicating that PVG consistently
achieves better rendering quality than 3DGS across the evaluated frames.
Rather than being driven by a few difficult examples, the improvement is
observed over the entire distribution of test images.
}
\label{fig:framewise_cs}
\end{figure*}

Figure~\ref{fig:framewise_cs} demonstrates that the improvement of PVG over
3DGS is consistently observed throughout the evaluation set. The distributions
are shifted toward the PVG-favorable side for all three metrics, indicating
that the averaged improvements reported in the main paper are representative
of the overall distribution rather than being dominated by a small number of
outlier frames.

\begin{figure*}[t]
\centering
\includegraphics[width=\textwidth]{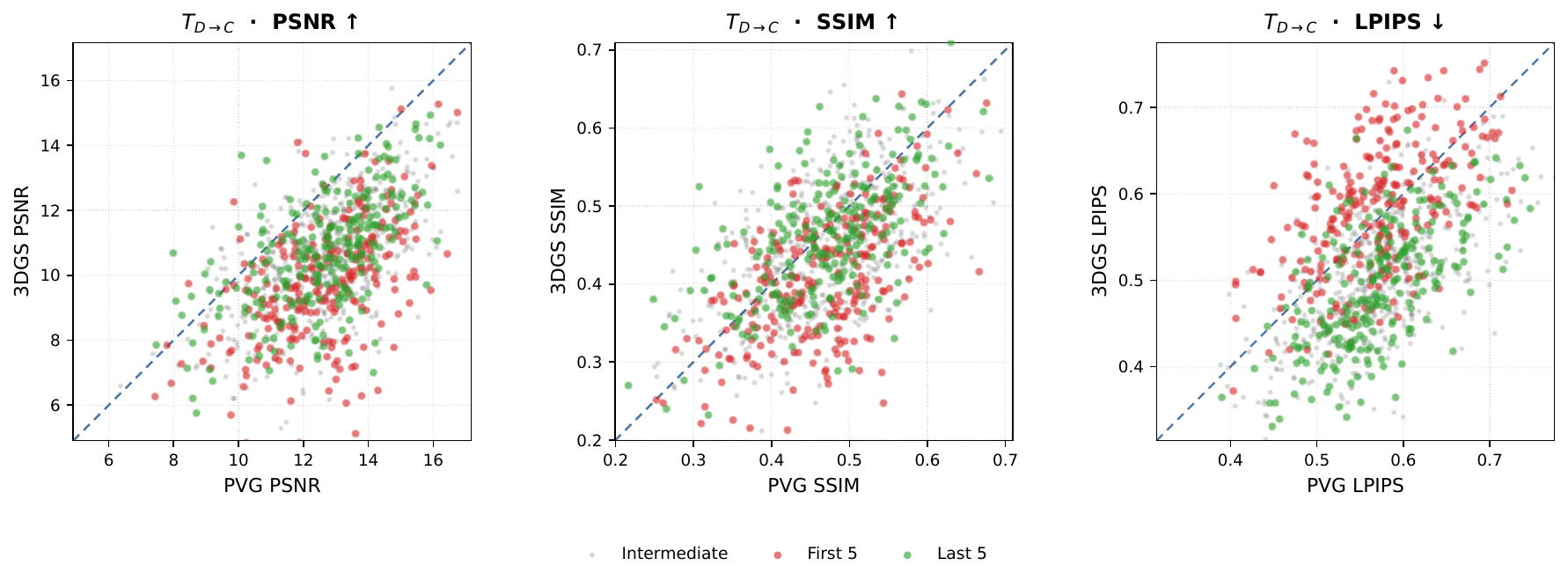}
\caption{
Frame-wise comparison between PVG and 3DGS on the
$\mathbf{T_{D\rightarrow C}}$ evaluation split.
Compared with $\mathbf{T_{C\rightarrow S}}$, the distributions exhibit
substantially larger overlap between the two methods.
The centroid of each distribution remains consistent with the averaged
performance reported in the main paper, while the overlapping point clouds
indicate that the differences between PVG and 3DGS are comparatively small for
this challenging aerial-to-ground evaluation.
}
\label{fig:framewise_dc}
\end{figure*}

Figure~\ref{fig:framewise_dc} shows that the performance gap between the two
methods becomes much smaller in the challenging aerial-to-ground setting.
Although the averaged metrics indicate slight differences between the methods,
the frame-wise distributions overlap considerably, suggesting that neither
method consistently outperforms the other across all evaluation images.
This observation is consistent with the quantitative comparison presented in
the main paper.

Overall, the frame-wise analysis complements the averaged results by showing
that the performance trends reported in Table~1 are reflected across
individual evaluation images. In particular, the improvement of PVG on the
$\mathbf{T_{C\rightarrow S}}$ split is consistently observed over the
evaluation set, whereas the $\mathbf{T_{D\rightarrow C}}$ split remains
considerably more challenging for both methods, resulting in substantially
overlapping frame-wise distributions.


\section{Additional Results on Camera Pose Estimation}
\label{suppsec:cpe}

This section expands on the camera pose estimation analysis from the main paper (\secref{subsec:camera_pe}) by providing full epipolar error distributions, per-method comparisons, and split-wise breakdowns for all benchmark directions. We include results for the two cross-view splits, $T_{C \rightarrow S}$ and $T_{D \rightarrow C}$, as well as same-view consistency analyses for $T_{C \rightarrow C}$, $T_{C \rightarrow L}$, and $T_{D \rightarrow D}$ using COLMAP\cite{colmap} (\figref{fig:supply_cam_pose_car_train}--\figref{fig:supply_cam_pose_drone_train}).

\subsection{Evaluation Protocol}
Following the protocol described in the main paper, each test image is paired with its nearest training image, and the average epipolar error $e_a$ is computed over eight manually annotated 2D correspondences (\cf~\figref{fig:epi_anno}). Feed-forward methods (MapAnything\cite{mapanything}, VGGT\cite{wang2025vggt}) take the test image and its 15 nearest training images as input and jointly regress the corresponding camera poses. COLMAP\cite{colmap} uses the same 15 neighbors for feature matching, triangulation, and pose refinement.

All methods operate on identical undistorted images, with shared intrinsics and the same correspondence annotation protocol to ensure fair comparison.

\subsection{Per-Method Error Distributions}
\label{subsec:cpe_dists}

\figref{fig:supply_cam_pose_car_train} and \figref{fig:supply_cam_pose_drone_train} show the complete error distributions for COLMAP\cite{colmap}, MapAnything\cite{mapanything}, and VGGT\cite{wang2025vggt} across the two principal evaluation splits. Following the visualization scheme from the main paper, distributions are separated into a low-error band ($e_a < 50$~px) and a high-error band ($e_a \geq 50$~px), highlighting long-tail failure modes.

\textbf{COLMAP\cite{colmap}.}
Across both splits, COLMAP\cite{colmap} produces tight, sharply peaked distributions, with a large majority of test images falling below $30$~px. Only a very small portion ($<5\%$) enters the high-error region, typically due to extreme viewpoint differences or motion blur.

\textbf{MapAnything\cite{mapanything}.}
MapAnything\cite{mapanything} exhibits much broader distributions with heavy tails. A large number of queries show errors above $80$~px, especially in cases involving large baselines or low-texture regions, where a single forward pass fails to recover accurate geometry.

\textbf{VGGT\cite{wang2025vggt}.}
VGGT\cite{wang2025vggt} performs better than MapAnything\cite{mapanything} in the low-error region, but still shows substantial high-error outliers. In the $T_{D \rightarrow C}$ split, aerial–ground viewpoint mismatches frequently cause errors exceeding $100$--$150$~px, highlighting instability under extreme viewpoint changes.

\subsection{Split-Wise Analysis}
\label{subsec:cpe_splitwise}

\textbf{$T_{C \rightarrow S}$.}
This scenario involves moderate viewpoint differences with similar elevation. COLMAP\cite{colmap} consistently maintains low errors, while feed-forward methods degrade noticeably. VGGT\cite{wang2025vggt} stays mostly below $100$~px, whereas MapAnything\cite{mapanything} produces more frequent high-error outliers.

\textbf{$T_{D \rightarrow C}$.}
This is the most challenging setting due to the wide aerial–ground viewpoint gap. All feed-forward methods show severe performance drops with long-tail distributions and many errors above $100$~px. COLMAP\cite{colmap} remains stable and significantly more accurate due to multi-view geometric optimization.




\end{document}